\documentclass{article}

\usepackage[preprint]{main}

\makeatletter
\renewcommand{\@noticestring}{Accepted at the CoRL 2026 Workshop on Continually Self-Improving Robots.}
\makeatother

\usepackage[utf8]{inputenc}
\usepackage[T1]{fontenc}
\usepackage{hyperref}
\hypersetup{pdftitle={ANCHOR: KL-Anchored Adaptation for Stable Offline-to-Online RL},pdfauthor={Geonwoo Cho, Jaegyun Im, Doyoon Kim, Lexin Li},pdfsubject={Accepted at the CoRL 2026 Workshop on Continually Self-Improving Robots}}
\usepackage{url}
\usepackage{booktabs}
\usepackage{amsfonts}
\usepackage{nicefrac}
\usepackage{microtype}
\usepackage{xcolor}

\definecolor{personalgreen}{HTML}{25643B}
\hypersetup{colorlinks=true,allcolors=personalgreen}

\usepackage{subcaption}
\usepackage{amsmath}
\usepackage{amsthm}
\usepackage{graphicx}
\usepackage{multirow}
\usepackage{algorithm}
\usepackage{algpseudocode}
\usepackage{tabularx}
\usepackage{wrapfig}

\title{ANCHOR: KL-Anchored Adaptation for Stable Offline-to-Online RL}

\author{
\textbf{Geonwoo Cho}\textsuperscript{1}\hspace{0.8em}
\textbf{Jaegyun Im}\textsuperscript{2}\hspace{0.8em}
\textbf{Doyoon Kim}\textsuperscript{3}\hspace{0.8em}
\textbf{Lexin Li}\textsuperscript{4}\\
\textsuperscript{1}University of Michigan, Ann Arbor \\
\textsuperscript{2}Gwangju Institute of Science and Technology \\
\textsuperscript{3}Pohang University of Science and Technology \\
\textsuperscript{4}University of California, Berkeley \\
\href{mailto:geonwoo@umich.edu}{\textcolor{black}{geonwoo@umich.edu}}
}

\newtheorem{lemma}{Lemma}
\newtheorem{assumption}{Assumption}

\newcommand{\figpath}{figures/low}

\begin{document}
\raggedbottom

\maketitle

\begin{abstract}
Offline-to-online reinforcement learning (RL) follows the pretrain-finetune paradigm: an
agent is first pretrained on historical data and then adapted through online interaction.
We study offline-to-online adaptation without offline replay, where the agent must improve
using only the pretrained model and newly collected experience. A key challenge is to avoid \emph{catastrophic failure}, a sharp early performance collapse during online fine-tuning, while still achieving strong \emph{asymptotic performance}. Through controlled experiments, we show that removing offline replay can induce catastrophic failure, while retaining strong conservative regularization from offline RL can suppress final online performance. We also find that common stabilization strategies do not reliably prevent early collapse, and that the critic-side mechanisms we test are insufficient on their own. Motivated by these findings, we propose ANCHOR, a simple adaptation method that anchors the actor to the offline-pretrained policy early in online fine-tuning, anneals this constraint over time, and retains only a small residual conservative critic regularizer. We provide a theoretical interpretation showing that KL anchoring helps preserve critic reliability across nearby occupancy distributions, with a two-phase regret bound explaining when anchoring should be maintained or relaxed. Across FrankaKitchen, AntMaze, and Adroit tasks, ANCHOR substantially reduces catastrophic failure while maintaining strong asymptotic performance. Ablations show that KL annealing and the residual conservative weight are important for balancing early stability and online adaptation, while additional analysis shows that ANCHOR is most beneficial when the pretrained policy contains useful behavior to preserve.
Code is available at \url{https://github.com/Cho-Geonwoo/ANCHOR}.
\end{abstract}

\section{Introduction}
\label{sec:introduction}

The pretrain-finetune paradigm has driven much of the recent success in modern machine learning across diverse domains such as natural language processing and computer vision~\citep{min2023recent, khan2024survey}. Inspired by these advances, reinforcement learning (RL) has adopted a similar paradigm:  an agent is first pretrained on historical offline data and then fine-tuned through online interaction with the target environment~\citep{agarwal2022reincarnating, luo2024optimistic}. This setting, known as offline-to-online RL, allows the agent to effectively leverage prior knowledge while adapting to the target environment with fewer costly or risky online interactions.

Prior work in this direction has shown that modifying the training objective or algorithmic structure during the online phase leads to better performance than naively deploying offline RL algorithms without modification~\citep{nakamoto2023calql,zhou2025wsrl,xiao2025porl}. However, catastrophic failure often occurs, i.e., there is an early collapse in performance when the agent first transitions from offline to online. This raises a central question: how can we adapt a pretrained agent online while avoiding early collapse and maintaining strong asymptotic performance? This question is particularly critical in safety-sensitive applications such as healthcare and robotics~\citep{singh2022reinforcement,liu2020reinforcement}. In such settings,
catastrophic failure is not merely a transient performance drop; it can lead to unsafe or
harmful decisions during deployment.

We study this question on robot-control benchmarks, including FrankaKitchen and AntMaze
from D4RL~\citep{fu2020d4rl} and Adroit dexterous manipulation
tasks~\citep{nair2020awac}. Controlled experiments show that removing offline replay can
induce catastrophic failure, while retaining the conservative regularization inherited from
offline RL can reduce final online performance. We further find that common stabilization
strategies, including UTD-ratio tuning~\citep{xiao2025porl} and warmup-based online data
collection~\citep{zhou2025wsrl}, do not reliably prevent early collapse. We also show that
the critic-side mechanisms we test can be insufficient on their own.

Motivated by these findings, we propose \textbf{ANCHOR}, a simple KL-anchored adaptation method for offline-to-online fine-tuning without offline replay. ANCHOR regularizes the online actor toward the offline-pretrained policy on online replay states during the vulnerable early phase, and gradually removes this constraint through KL annealing to enable online improvement. For the critic, ANCHOR keeps only a small residual conservative regularization term, avoiding excessive pessimism while retaining mild protection against overestimation.

We provide theoretical and empirical support for this design. Our analysis interprets KL
anchoring as a mechanism for preserving critic reliability across nearby occupancy
distributions, and gives a two-phase regret bound explaining when anchoring should be
maintained or relaxed depending on pretrained policy quality. Empirically, ANCHOR substantially reduces catastrophic failure while achieving strong
asymptotic performance, with the best average final success rate across the six tasks. Ablations show that KL annealing improves over both
keeping the KL constraint fixed and removing it from the start, that the residual
conservative regularization controls the stability-adaptation balance, and that ANCHOR is
most beneficial when the pretrained policy contains useful behavior to preserve.
\section{Preliminaries}
\label{sec:preliminaries}

\subsection{Conservative Q-Learning (CQL)}

In offline RL, an agent is trained on a fixed dataset
$\mathcal{D}$ collected by a behavior policy, without
interacting with the environment. A key challenge is that standard Q-learning objectives
can assign erroneously high values to actions not present in the dataset, leading to poor
policy performance when deployed online.  CQL~\citep{kumar2020cql} addresses this issue by
adding a regularization term to the standard temporal-difference (TD) loss,
$\mathcal{L}_{\text{TD}}$, that penalizes Q-values of actions sampled from the policy
relative to those from the dataset:
\begin{equation}
\mathcal{L}_{\text{CQL}}
=
\mathcal{L}_{\text{TD}}
+
\alpha
\left(
\mathbb{E}_{s\sim \mathcal{D},\,a \sim \pi(\cdot|s)}[Q_{\theta}(s,a)]
-
\mathbb{E}_{(s,a) \sim \mathcal{D}}[Q_{\theta}(s,a)]
\right),
\end{equation}
where $\alpha>0$ controls the penalty strength. This discourages overly optimistic
Q-values for out-of-distribution actions. We adopt CQL as the backbone of our method.

\subsection{Offline-to-Online Reinforcement Learning}

Offline-to-online (O2O) RL considers the setting in which an agent is first pretrained on
a fixed offline dataset and then further adapted through online interaction with the target
environment~\citep{nakamoto2023calql,zhou2025wsrl,xiao2025porl,Wen2024RO2O}. We focus on
the setting where the original offline trajectories are not available during online
adaptation. In this case, the agent must reuse knowledge encoded in the pretrained
model while learning from an online data distribution.

This setting presents two simultaneous requirements: the agent should avoid early
performance degradation during the transition, while still improving beyond the offline
solution through online interaction.  We evaluate these two aspects using two metrics. \emph{Catastrophic failure} measures the
drop from the initial online performance to the minimum success rate observed within the first 100K online steps; we use \emph{early stability} to refer to low catastrophic failure. \emph{Asymptotic performance} measures the average success rate
over the final 50K online steps, corresponding to the 350K-400K window, and captures the
late-stage plateau. Together, these metrics assess whether an O2O method can maintain
early stability while enabling further online improvement. Further justification for these
evaluation windows is provided in Appendix~\ref{sec:appendix:numerical_results}.
\section{Diagnosing Failure Modes in Replay-Free Offline-to-Online RL}
\label{sec:motivation}

\subsection{How Do Offline Data and Conservative Regularization Shape Online Adaptation?}
\label{sec:role_offline_data_alpha}

\begin{figure}[t]
    \centering
    \begin{minipage}[c]{0.6cm}
        \centering\textbf{(a)}
    \end{minipage}\hspace{0.1em}
    \begin{subfigure}[c]{0.23\textwidth}
        \includegraphics[width=1.0\linewidth]{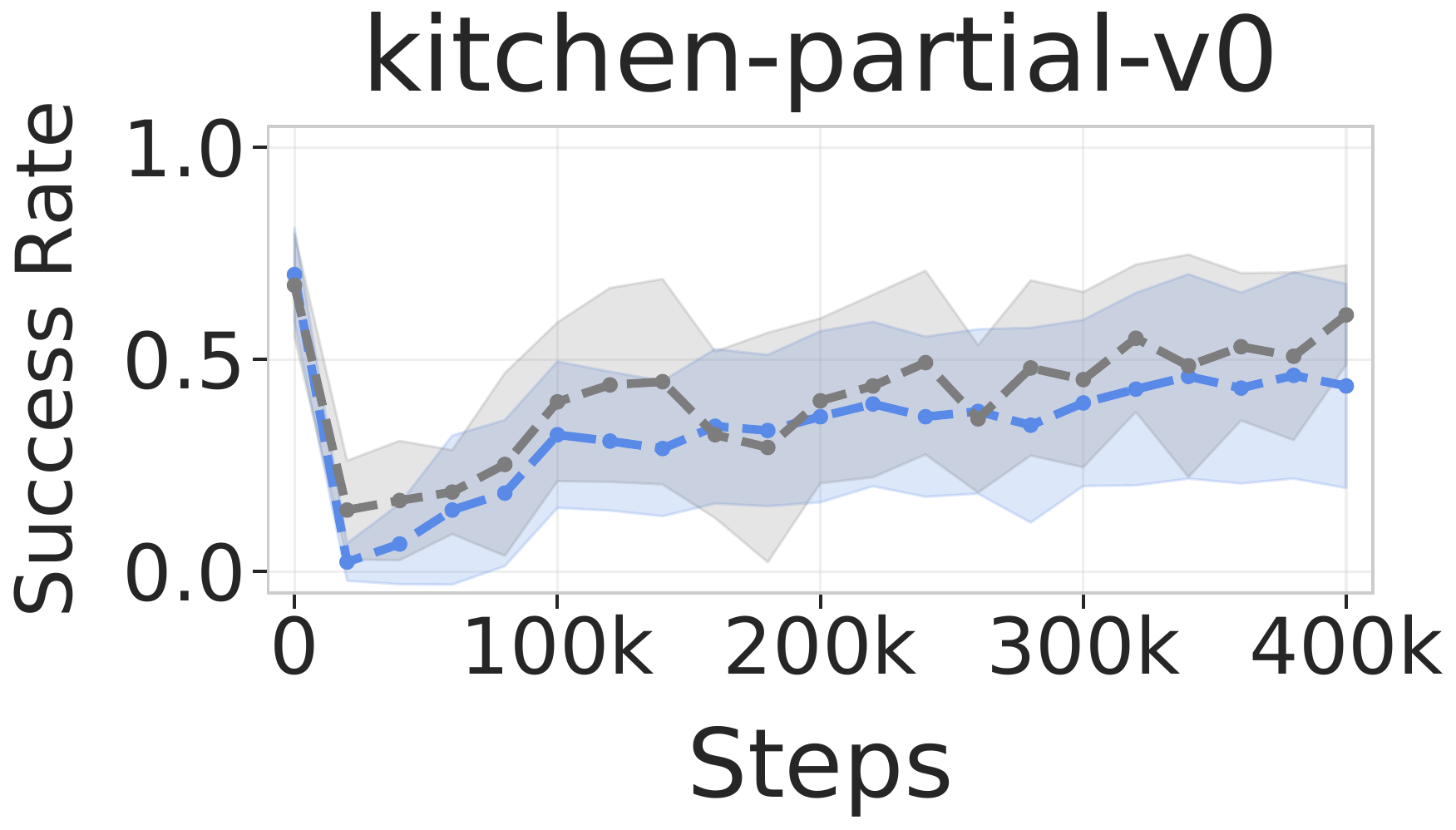}
    \end{subfigure}\hspace{0.1em}
    \begin{subfigure}[c]{0.23\textwidth}
        \includegraphics[width=1.0\linewidth]{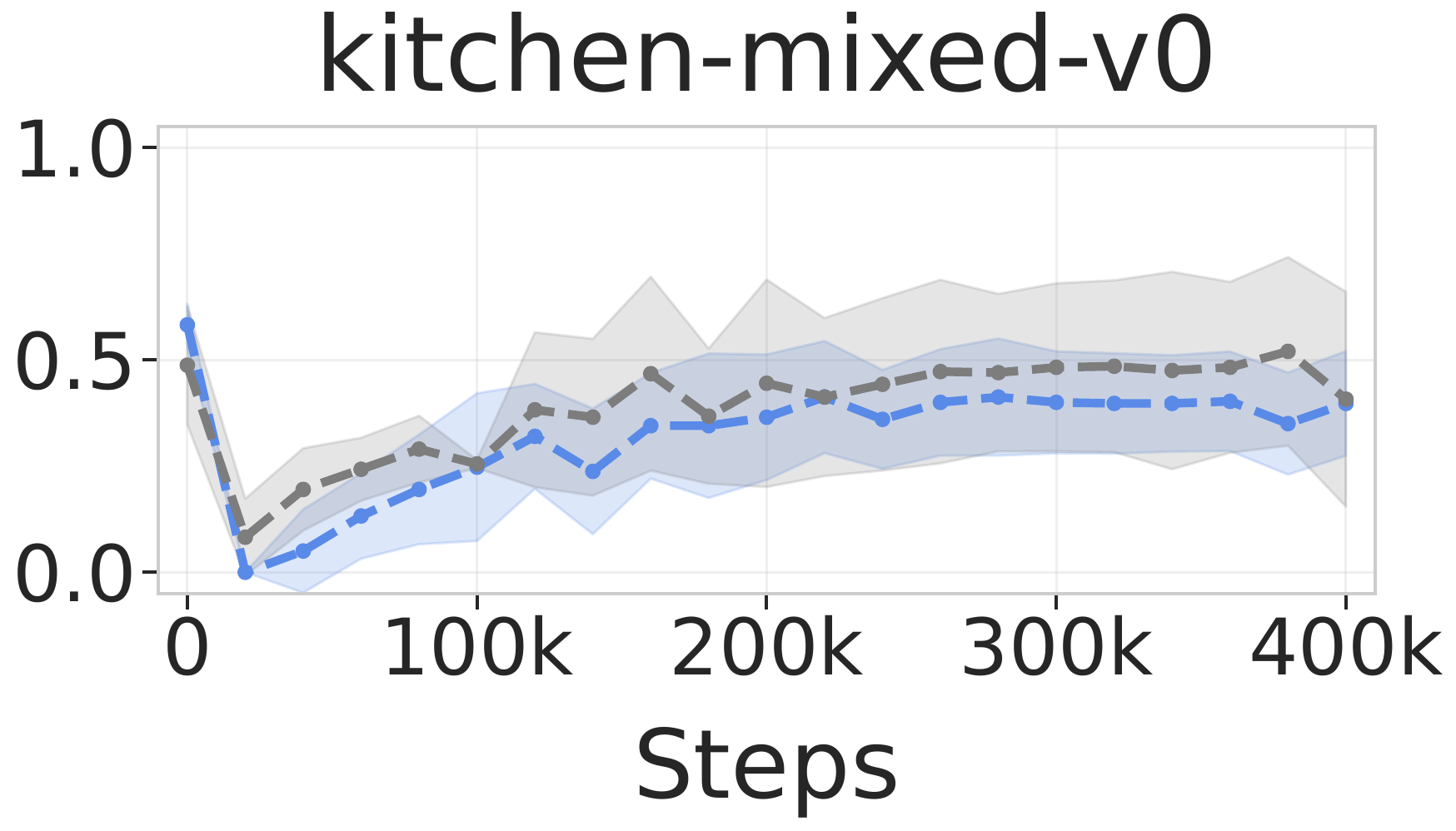}
    \end{subfigure}\hspace{0.1em}
    \begin{subfigure}[c]{0.23\textwidth}
        \includegraphics[width=1.0\linewidth]{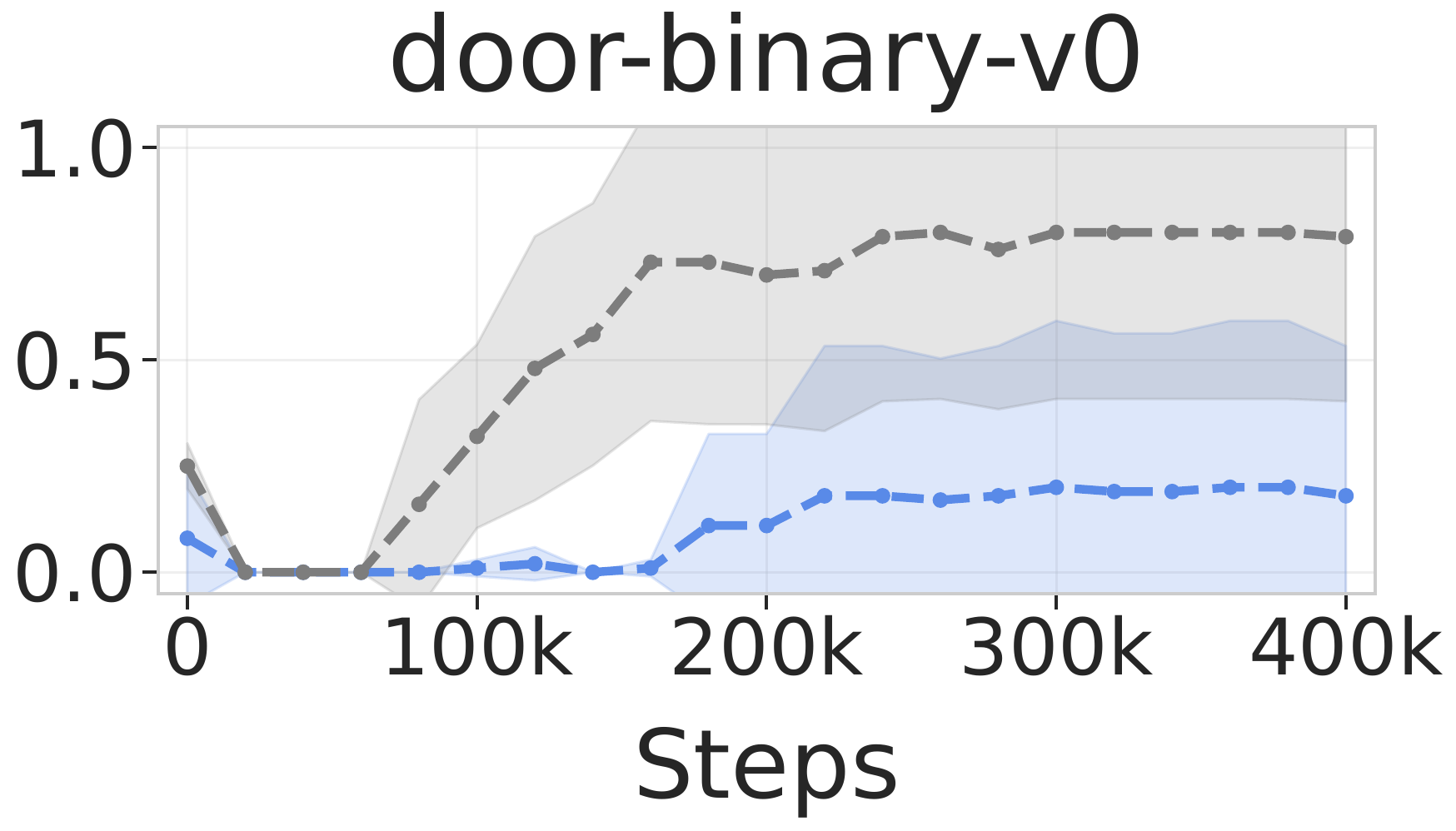}
    \end{subfigure}\hspace{0.1em}
    \begin{subfigure}[c]{0.23\textwidth}
        \includegraphics[width=1.0\linewidth]{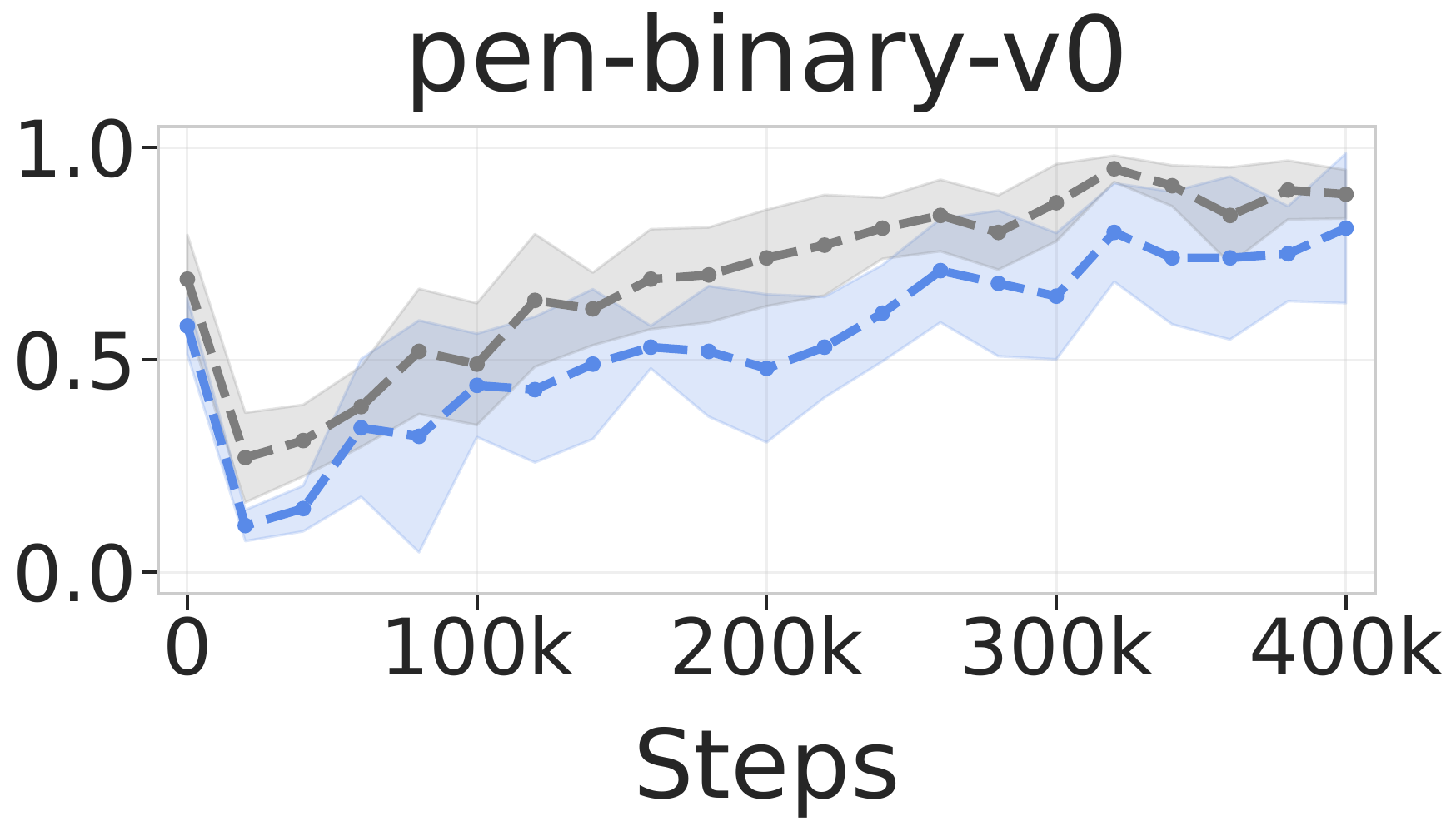}
    \end{subfigure}\\[1ex]
    \begin{minipage}[c]{0.6cm}
        \centering\textbf{(b)}
    \end{minipage}\hspace{0.1em}
    \begin{subfigure}[c]{0.23\textwidth}
        \includegraphics[width=1.0\linewidth]{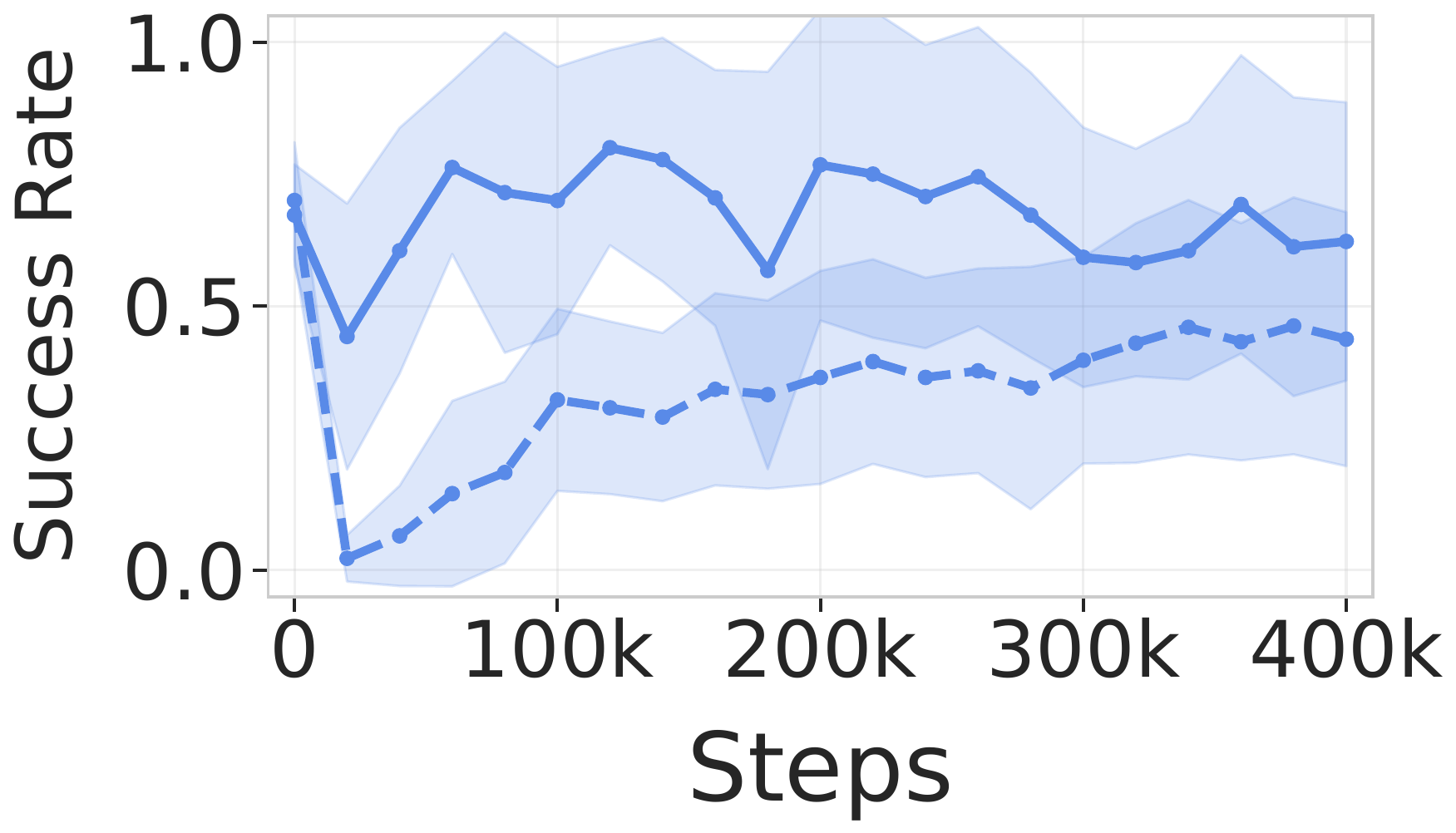}
    \end{subfigure}\hspace{0.1em}
    \begin{subfigure}[c]{0.23\textwidth}
        \includegraphics[width=1.0\linewidth]{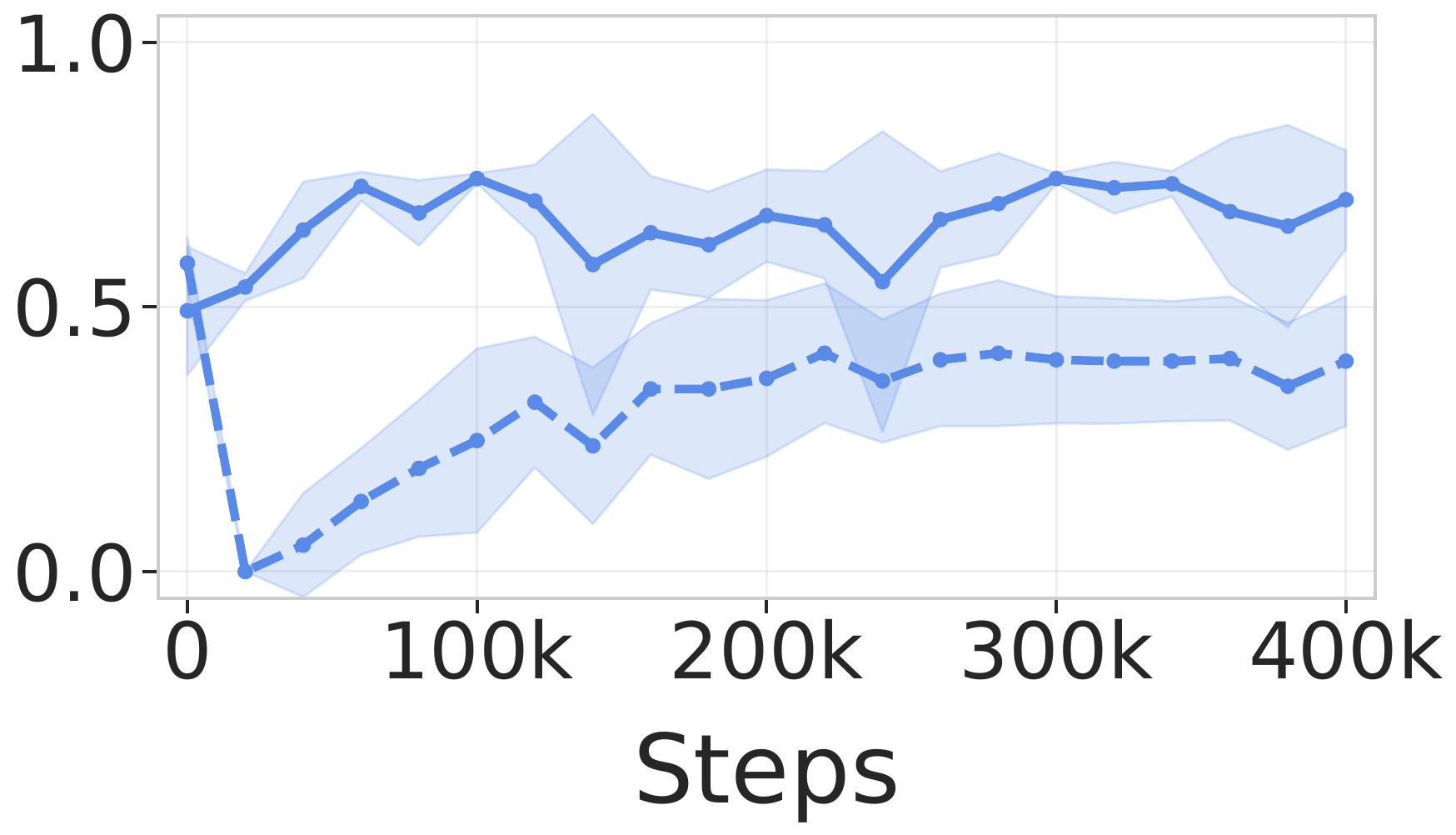}
    \end{subfigure}\hspace{0.1em}
    \begin{subfigure}[c]{0.23\textwidth}
        \includegraphics[width=1.0\linewidth]{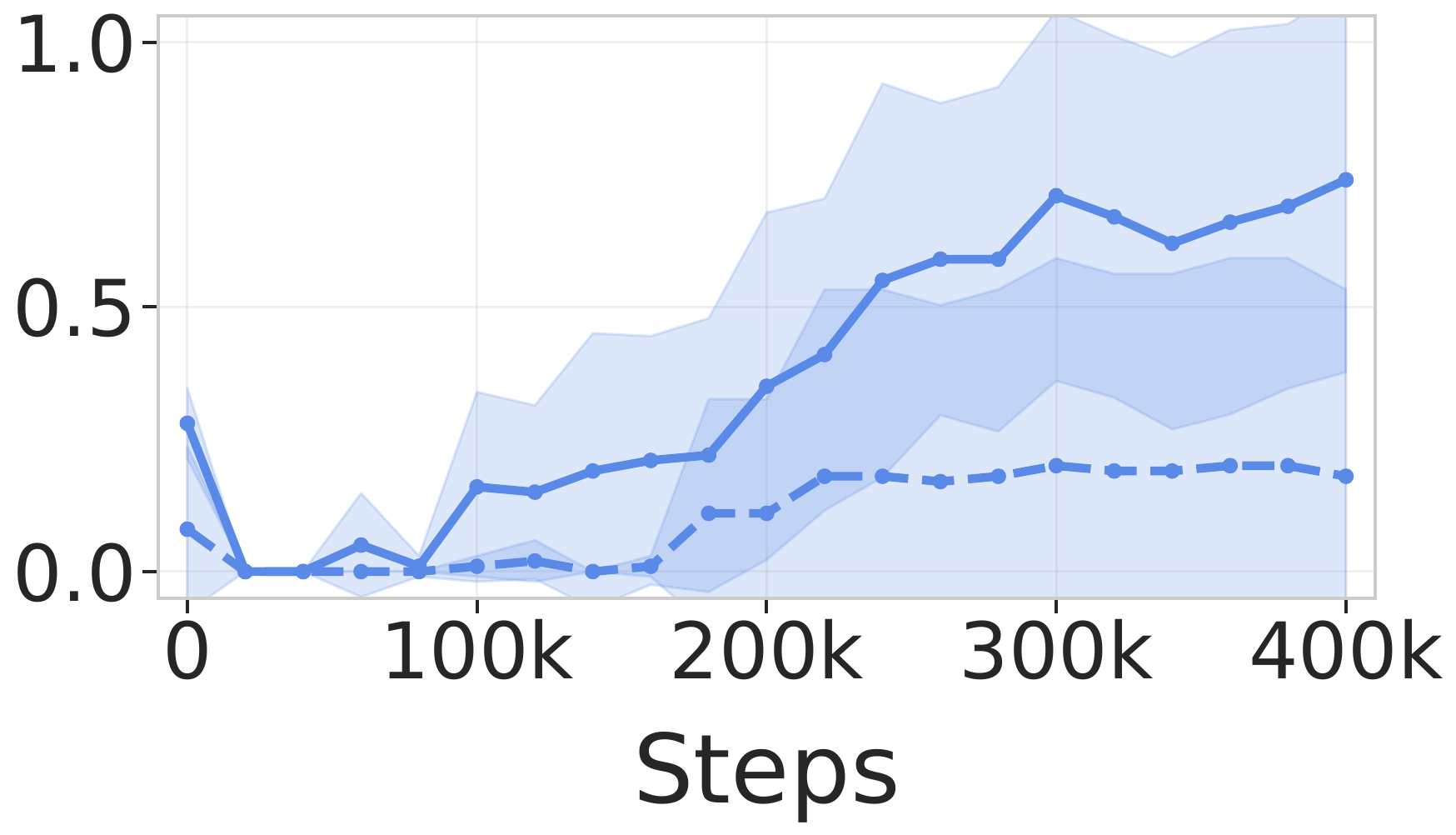}
    \end{subfigure}\hspace{0.1em}
    \begin{subfigure}[c]{0.23\textwidth}
        \includegraphics[width=1.0\linewidth]{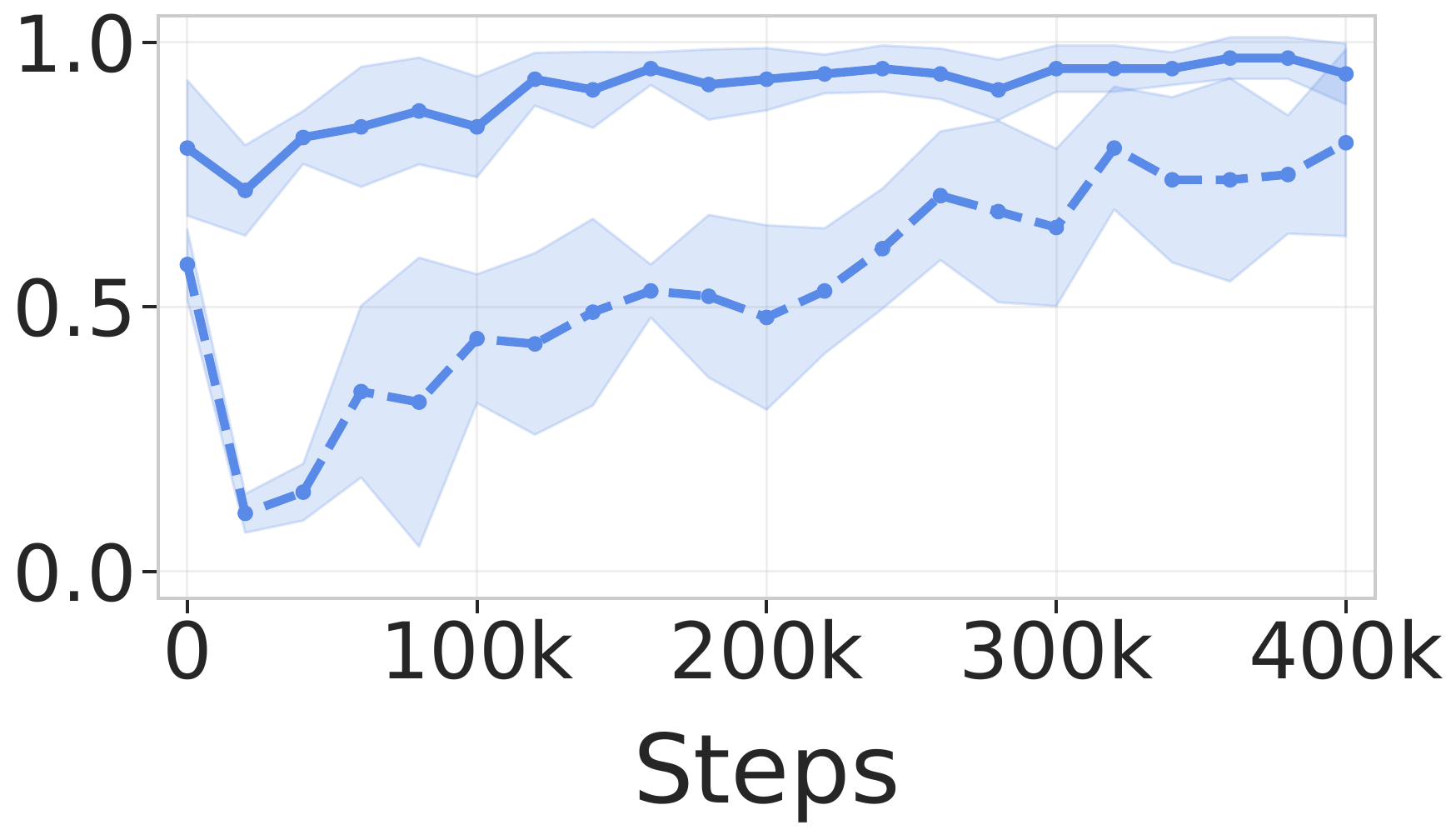}
    \end{subfigure}\\[1ex]
    \vspace{-0.5em}
    \includegraphics[width=0.7\linewidth]
    {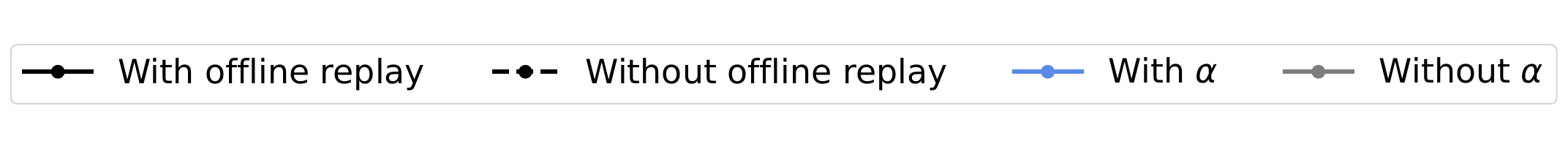}
    \vspace{-1em}
    \caption{\textbf{Controlled analysis of offline data and conservative regularization.}
    Success rates of CQL during online fine-tuning across four tasks. (a) Keeping the conservative regularizer $\alpha$ during online fine-tuning can reduce
    asymptotic performance. (b) Removing offline replay can induce early performance degradation.}
    \label{fig:offline_data_alpha_analysis}
\end{figure}

We first examine the individual roles of offline data access and conservative regularization during the offline-to-online transition. To isolate their effects, we conduct controlled CQL experiments. Unless otherwise noted, all reported experiments are run with five random seeds, and we report the mean with 95\% confidence intervals shown as shaded regions. Step 0 denotes the beginning of online fine-tuning. In Figure~\ref{fig:offline_data_alpha_analysis}, ``with offline replay'' mixes offline
data into online fine-tuning batches, while ``without offline replay'' uses only newly
collected online data. Similarly, ``with $\alpha$'' retains the CQL conservative regularizer
during online fine-tuning, whereas ``without $\alpha$'' removes it.

Our first observation is that conservative regularization can limit asymptotic online performance. To isolate the effect of the CQL regularizer, we remove offline replay in both settings and compare two variants: one that retains the conservative regularization coefficient $\alpha$, and one that removes it. As shown in Figure~\ref{fig:offline_data_alpha_analysis}(a), removing $\alpha$ achieves comparable or better final performance across all four tasks. This suggests that the pessimism induced by CQL, while useful for offline pretraining, can become restrictive once the policy must improve beyond the support of the offline dataset.

Our second observation is that removing offline data access can induce catastrophic early degradation. To isolate this effect, we compare standard CQL fine-tuning with and without offline replay. Figure~\ref{fig:offline_data_alpha_analysis}(b) shows that removing offline replay often causes a sharp performance drop at the beginning of online fine-tuning. The main exception is door-binary, where the initial success rate after offline pretraining is already low, making the drop less pronounced. This indicates that offline replay serves as an important stabilizing mechanism during the early online phase.

Together, these results identify two distinct factors that shape offline-to-online
adaptation: offline replay stabilizes the early online phase, while strong conservative
regularization can restrict asymptotic improvement. When offline replay is unavailable,
we therefore need an adaptation mechanism that prevents early catastrophic failure without
maintaining overly conservative regularization throughout online fine-tuning.

\subsection{Can Existing Stabilization Strategies Prevent Catastrophic Failure?}
\label{sec:analysis_catastrophic_failure}

Section~\ref{sec:role_offline_data_alpha} shows that offline-to-online fine-tuning can
become unstable without offline replay. We therefore test whether existing stabilization
strategies can mitigate catastrophic failure in this setting. The diagnostic plots are provided in Appendix~\ref{sec:appendix:diagnostic_figures}.

We first show that standard stabilization strategies fail to consistently resolve early
collapse across tasks. We consider the update-to-data (UTD) ratio, which increases the number of critic
updates per online interaction and has been suggested to alleviate catastrophic
failure~\citep{xiao2025porl}. As shown in the top row of Figure~\ref{fig:hyperparams},
varying the UTD ratio does not consistently reduce early collapse across tasks. We next
evaluate a warmup phase, where the agent collects online trajectories with the
offline-pretrained policy before parameter updates, following~\citet{zhou2025wsrl}. As
shown in the middle row of Figure~\ref{fig:hyperparams}, varying the warmup length also
fails to reliably prevent catastrophic failure. Additional tuning attempts, including
changes to batch size, learning rate, network width, hidden dimension, and LayerNorm, are
reported in Appendix~\ref{sec:appendix:other_attempts}. None of these modifications
consistently prevent catastrophic failure across tasks.

We next test whether online calibration of the conservatively pretrained Q-function can
stabilize fine-tuning. Cal-QL provides a natural testbed for this question, as it modifies
the CQL conservative penalty to calibrate the Q-function using an online evidence-based
reference value~\citep{nakamoto2023calql}. The coefficient
$\alpha_{\mathrm{Cal\text{-}QL}}$ controls the strength of this calibrated conservative
regularization. As shown in the bottom row of Figure~\ref{fig:hyperparams}, tuning
$\alpha_{\mathrm{Cal\text{-}QL}}$ can reduce early collapse in antmaze-large-diverse and
antmaze-large-play, but the effect is not consistent across tasks or hyperparameter values.
Moreover, Appendix~\ref{sec:appendix:numerical_results} shows that ANCHOR achieves lower
catastrophic failure than Cal-QL on average, suggesting that there remains room for
improvement.

The limited effectiveness of Cal-QL raises the question of whether critic-side
stabilization alone is sufficient. Motivated by the hypothesis of~\citet{zhou2025wsrl}
that rapid critic shift contributes to early collapse, we test this question by explicitly
penalizing deviation from the offline-pretrained critic:
\begin{equation*}
    \min_{\theta}\ %
    \mathcal{L}_{\mathrm{TD}}(\theta)
    +
    \alpha_{\mathrm{KL}}\,
    \mathbb{E}_{(s,a)\sim \mathcal{B}}\left[
    D_{\mathrm{KL}}\left(
    \mathrm{softmax}\big(Q_{\mathrm{offline}}(s,\cdot)\big)
    \,\Vert\,
    \mathrm{softmax}\big(Q_{\mathrm{online}}(s,\cdot)\big)
    \right)
    \right],
\end{equation*}
where $\mathcal{B}$ is the online replay buffer, $D_{\mathrm{KL}}$ denotes the KL
divergence, $Q_{\mathrm{offline}}$ is the offline-pretrained critic, and
$Q_{\mathrm{online}}$ is the critic being updated during online fine-tuning.
The coefficient $\alpha_{\mathrm{KL}}>0$ controls the strength of this critic-shift
penalty. As shown in Figure~\ref{fig:critic_shift}, increasing $\alpha_{\mathrm{KL}}$ slows critic distribution shift, yet the corresponding success curves show that catastrophic failure persists. Appendix~\ref{sec:appendix:q_catastropic_failure_analysis} further shows that
Q-driven metrics do not reliably predict early collapse. These findings do not rule out critic instability as a contributing factor, but suggest that the critic-side stabilization strategies tested here are insufficient on their own, motivating policy-side stabilization.
\section{Method}
\label{sec:method}

Building on Sections~\ref{sec:role_offline_data_alpha} and~\ref{sec:analysis_catastrophic_failure}, we propose ANCHOR, a KL-anchored adaptation scheme that mitigates catastrophic failure while preserving high asymptotic performance during offline-to-online adaptation. We also provide a theoretical interpretation of ANCHOR.

\subsection{ANCHOR: KL-anchored checkpoint adaptation}
\label{sec:method:anchor}

Let $\pi_{\mathrm{off}}$ denote the actor obtained after offline pretraining, and let $\pi_t$ and $Q_t$ denote the online actor and critic at online step $t$. ANCHOR regularizes the actor toward $\pi_{\mathrm{off}}$ during the early adaptation phase and gradually relaxes this constraint. Concretely, we use the actor objective
\begin{equation}
\label{eq:anchor-actor-loss}
\mathcal{J}^{\text{ANCHOR}}_{\pi,t}(\phi)
=
\mathbb{E}_{s\sim \mathcal{D}_{\mathrm{on}},\, a\sim \pi_{\phi}(\cdot|s)}\left[Q_t(s,a)\right]
- \beta_t\, \mathbb{E}_{s\sim \mathcal{D}_{\mathrm{on}}}\left[D_{\mathrm{KL}}\big(\pi_{\phi}(\cdot|s)\,\Vert\,\pi_{\mathrm{off}}(\cdot|s)\big)\right],
\end{equation}
where $\beta_t \ge 0$ is the KL regularization weight. The standard entropy term is omitted here for clarity; implementation details are in Appendix~\ref{sec:appendix:objective_details}.

The critic is updated with the usual online TD loss together with a conservative CQL term whose coefficient is fixed to $\alpha_{\mathrm{on}}$ during online fine-tuning:
\begin{equation}
\label{eq:anchor-critic-loss}
\mathcal{L}^{\text{ANCHOR}}_{Q}(\theta)
=
\mathcal{L}_{\mathrm{TD}}(\theta;\mathcal{D}_{\mathrm{on}})
+
\alpha_{\mathrm{on}} \Big(
\mathbb{E}_{s\sim \mathcal{D}_{\mathrm{on}},\, a\sim \pi_t(\cdot|s)}[Q_{\theta}(s,a)]
-
\mathbb{E}_{(s,a)\sim \mathcal{D}_{\mathrm{on}}}[Q_{\theta}(s,a)]
\Big).
\end{equation}
Here, $\alpha_{\mathrm{on}} \ge 0$ denotes the residual conservative regularization strength used
throughout online fine-tuning, which is set smaller than the coefficient used during
offline pretraining. This design avoids excessive conservatism that may limit online improvement,
while retaining a small amount of critic regularization to prevent abrupt critic distribution
shift and stabilize adaptation.

For intuition, consider a fixed state $s$ and an unconstrained policy class over a
continuous action space $\mathcal A$. Assuming policies are absolutely continuous with
respect to $\pi_{\mathrm{off}}$ and the normalizer is finite, the state-wise objective in
Equation~\ref{eq:anchor-actor-loss} admits the closed-form solution~\citep{foster2025goodfoundation}
\begin{equation}
\label{eq:anchor-actor-opt}
\pi_t^{\star}(a|s)
=
\frac{
\pi_{\mathrm{off}}(a|s)\exp\!\left(Q_t(s,a)/\beta_t\right)
}{
Z_t(s)
},
\qquad
Z_t(s)
=
\int_{\mathcal A}
\pi_{\mathrm{off}}(a'|s)
\exp\!\left(Q_t(s,a')/\beta_t\right)
\,da' .
\end{equation}
Thus, $\beta_t$ controls the strength of anchoring: as $\beta_t\to 0$, the ideal policy
concentrates on high-$Q_t$ actions, while large $\beta_t$ keeps it close to
$\pi_{\mathrm{off}}$. In our continuous-control experiments, we do not compute this
nonparametric solution explicitly; instead, we optimize the actor using Equation~\ref{eq:anchor-actor-loss}.

In practice, ANCHOR uses a linear KL annealing schedule: \[ \beta_t = \beta_0 \cdot \max\left(1 - \frac{t}{T_{\mathrm{KL}}},\, 0\right), \] where $\beta_0$ is the initial KL regularization weight and $T_{\mathrm{KL}}$ is the KL annealing interval. Thus, $\beta_t$ decreases linearly from $\beta_0$ to $0$ during the first $T_{\mathrm{KL}}$ online steps. After $t \ge T_{\mathrm{KL}}$, the KL anchoring term is completely removed and the actor is optimized using the standard online objective. Exact hyperparameter settings are reported in Appendix~\ref{sec:appendix:implementation_details}. We detail the difference from trust-region methods such as TRPO~\citep{schulman2015trust} in Appendix~\ref{sec:appendix:related_works}.

\subsection{Theoretical Interpretation}

We provide a formal interpretation of ANCHOR through two supporting lemmas. The first lemma shows that KL anchoring can control the distribution-shift component of critic instability. The second lemma gives a two-phase regret bound that quantifies the cost of using KL anchoring during the early online phase before switching to the unanchored online learner. Detailed assumptions and proofs are deferred to Appendix~\ref{sec:appendix:lemma}.

\begin{lemma}[Critic error difference under KL-occupancy control]\label{thm:1}
Let $\widehat Q_t$ denote the learned critic at online step $t$, and let $Q^\pi$ denote the true action-value function of an arbitrary target policy $\pi$. Define
\[
e_t(s,a)=|\widehat Q_t(s,a)-Q^\pi(s,a)|.
\]
If $|\widehat Q_t(s,a)|\le Q_{\max}$ and $|Q^\pi(s,a)|\le Q_{\max}$, then for any two occupancy measures $d$ and $d'$,
\[
\left|
\mathbb E_{d}[e_t]
-
\mathbb E_{d'}[e_t]
\right|
\le
2Q_{\max}\|d-d'\|_1.
\]
If the online actor additionally satisfies a KL constraint around the offline policy,
\[
D_{\rho^{\pi_{\mathrm{off}}}}(\pi_t \| \pi_{\mathrm{off}}) \le \delta_t,
\]
where $\rho^{\pi_{\mathrm{off}}}$ is the offline policy's state marginal, and the occupancy shift is locally controlled by policy divergence, then
\[
\left|
\mathbb E_{d^{\pi_t}}[e_t]
-
\mathbb E_{d^{\pi_{\mathrm{off}}}}[e_t]
\right|
=
O(\sqrt{\delta_t}).
\]
\end{lemma}

Lemma~\ref{thm:1} is a distribution-transfer bound for critic error. It does not assume that the critic error is small; instead, it shows that if the critic is accurate under the offline-policy occupancy, then its error under the current online-policy occupancy remains controlled when $\pi_t$ stays close to $\pi_{\mathrm{off}}$. Thus, if the pretrained critic is reliable under the offline-policy occupancy and this reliability is preserved during early online updates, KL anchoring prevents the online policy from moving too quickly into regions where critic error may be much larger.

In practice, ANCHOR computes the KL penalty on online replay states rather than under the offline-policy occupancy used in the lemma. Although these distributions differ, the replay-buffer KL serves the same local role: it constrains policy updates on states encountered during fine-tuning and thereby limits abrupt policy drift.

\begin{lemma}[Two-phase regret decomposition]\label{thm:2}
Over $T = T_1 + T_2$ steps, suppose phase I uses KL anchoring around
$\pi_{\mathrm{off}}$ with budgets $\delta_t$. In phase II, suppose the actor-side KL
anchor is removed, and the resulting online learner has
regret $\mathrm{Reg}_{\mathrm{base}}(T_2)$. Then under bounded advantage and local
occupancy control,
\[
\mathrm{Reg}_T
\le
T_1 \Delta_{\mathrm{off}}
+
C \sum_{t=1}^{T_1}\sqrt{\delta_t}
+
\mathrm{Reg}_{\mathrm{base}}(T_2),
\]
where $\Delta_{\mathrm{off}} = J(\pi^*) - J(\pi_{\mathrm{off}})$ and $C$ is a problem-dependent constant.
\end{lemma}

Lemma~\ref{thm:2} separates the regret into the cost of staying near the offline policy, the cost of controlled policy deviation during the anchored phase, and the regret after the anchor is removed. This decomposition clarifies how the usefulness of KL anchoring depends on the quality of the pretrained policy. When $\pi_{\mathrm{off}}$ is weak, $\Delta_{\mathrm{off}}$ is large, so remaining close to the offline policy for too long can incur substantial regret; in this case, annealing the KL coefficient is important for allowing the actor to move beyond the pretrained solution. When $\pi_{\mathrm{off}}$ is strong, $\Delta_{\mathrm{off}}$ is small, so the cost of staying near the offline policy is limited, and stronger or longer anchoring can be justified if it prevents unstable online updates.

Overall, the analysis supports the central design of ANCHOR: KL anchoring limits early distribution shift and helps preserve critic reliability near the pretrained policy, while annealing removes this constraint when online improvement beyond the pretrained solution is needed. The results should be viewed as an interpretation of the stabilizing role of anchoring, rather than a convergence guarantee.
\section{Experiments}
\label{sec:experiments}

\subsection{Experimental Setup}
\label{sec:experiments:setup}


\textbf{Environments and tasks.}
We evaluate ANCHOR on three widely used offline-to-online RL benchmarks: FrankaKitchen and AntMaze from D4RL~\citep{fu2020d4rl}, and Adroit dexterous manipulation tasks~\citep{nair2020awac}. Our main benchmark consists of six tasks: kitchen-mixed, kitchen-partial, antmaze-large-diverse, antmaze-large-play, door-binary, and pen-binary. All offline pretraining datasets match those used in WSRL~\citep{zhou2025wsrl}. Further environment details and task visualizations are provided in Appendices~\ref{sec:appendix:implementation_details} and~\ref{sec:appendix:visualizations}.

\textbf{Training procedure.}
We first perform offline pretraining for 1M steps in AntMaze, 250K steps in FrankaKitchen,
and 40K steps in Adroit, followed by 400K steps of online fine-tuning. All methods are
trained without offline data replay and use the same online interaction budget.

\textbf{Baselines.}
We compare ANCHOR against WSRL~\citep{zhou2025wsrl}, PORL~\citep{xiao2025porl},
Cal-QL~\citep{nakamoto2023calql}, CQL~\citep{kumar2020cql},
IQL~\citep{kostrikov2022iql}, and SAC~\citep{haarnoja2018sac}. Detailed descriptions of
the baselines are provided in Appendix~\ref{sec:appendix:details_on_baseline_algorithms}.

For baseline reproduction, we largely follow the experimental setup of
WSRL~\citep{zhou2025wsrl}. Any modifications made to stabilize training are described in
Appendix~\ref{sec:appendix:reproducing_baselines}. ANCHOR is implemented on top of CQL,
with modifications to the online training objective. Details of
the ANCHOR implementation and hyperparameters are provided in
Appendix~\ref{sec:appendix:implementation_details}.

\subsection{ANCHOR Improves Both Early Stability and Final Performance}
\label{sec:experiments:main_results}

\begin{figure}[t]
    \centering
    \begin{subfigure}[c]{0.24\textwidth}
        \includegraphics[width=1.0\linewidth]{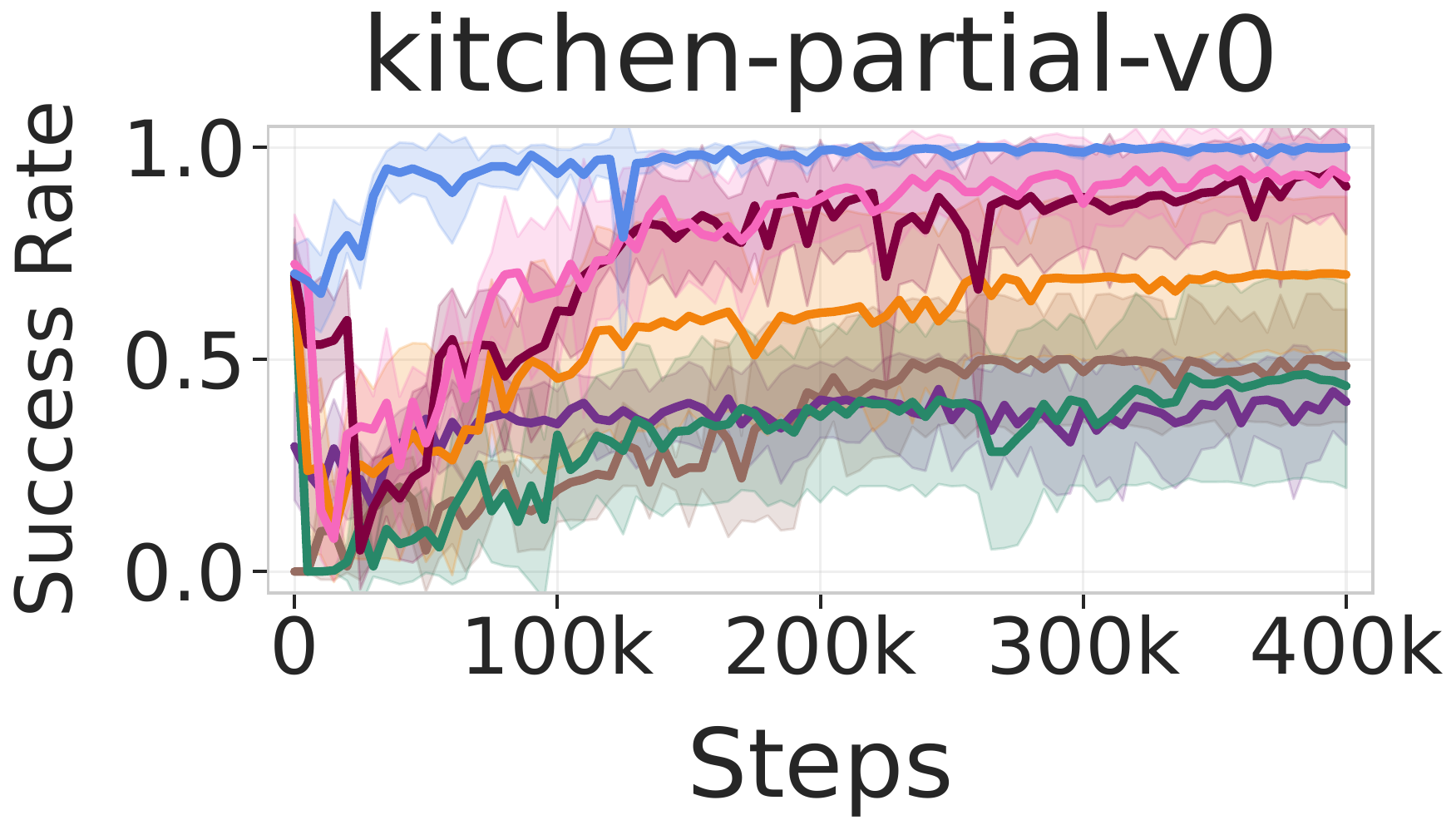}
    \end{subfigure}\hspace{0.1em}
    \begin{subfigure}[c]{0.24\textwidth}
        \includegraphics[width=1.0\linewidth]{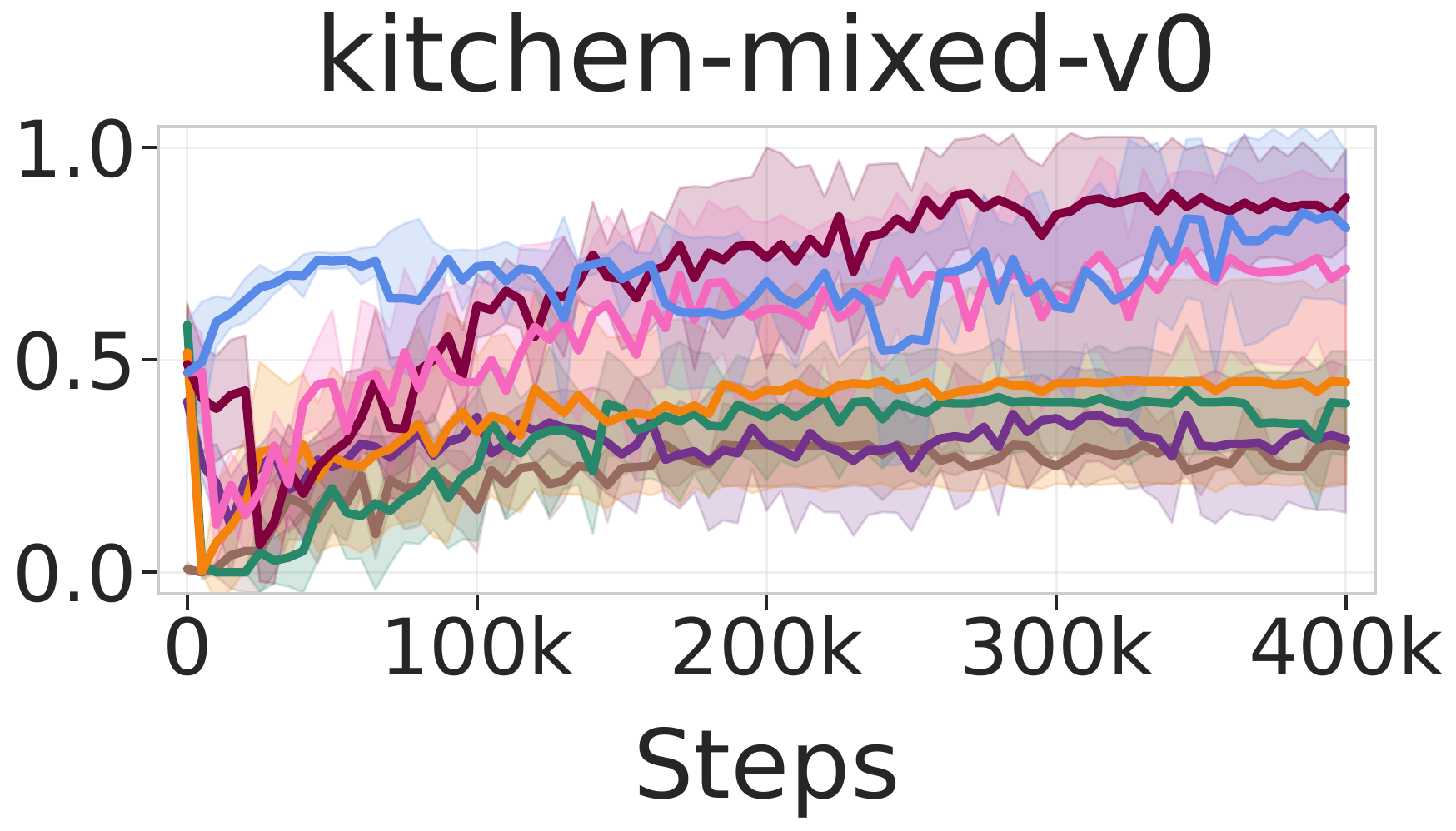}
    \end{subfigure}\hspace{0.1em}
    \begin{subfigure}[c]{0.24\textwidth}
        \includegraphics[width=1.0\linewidth]{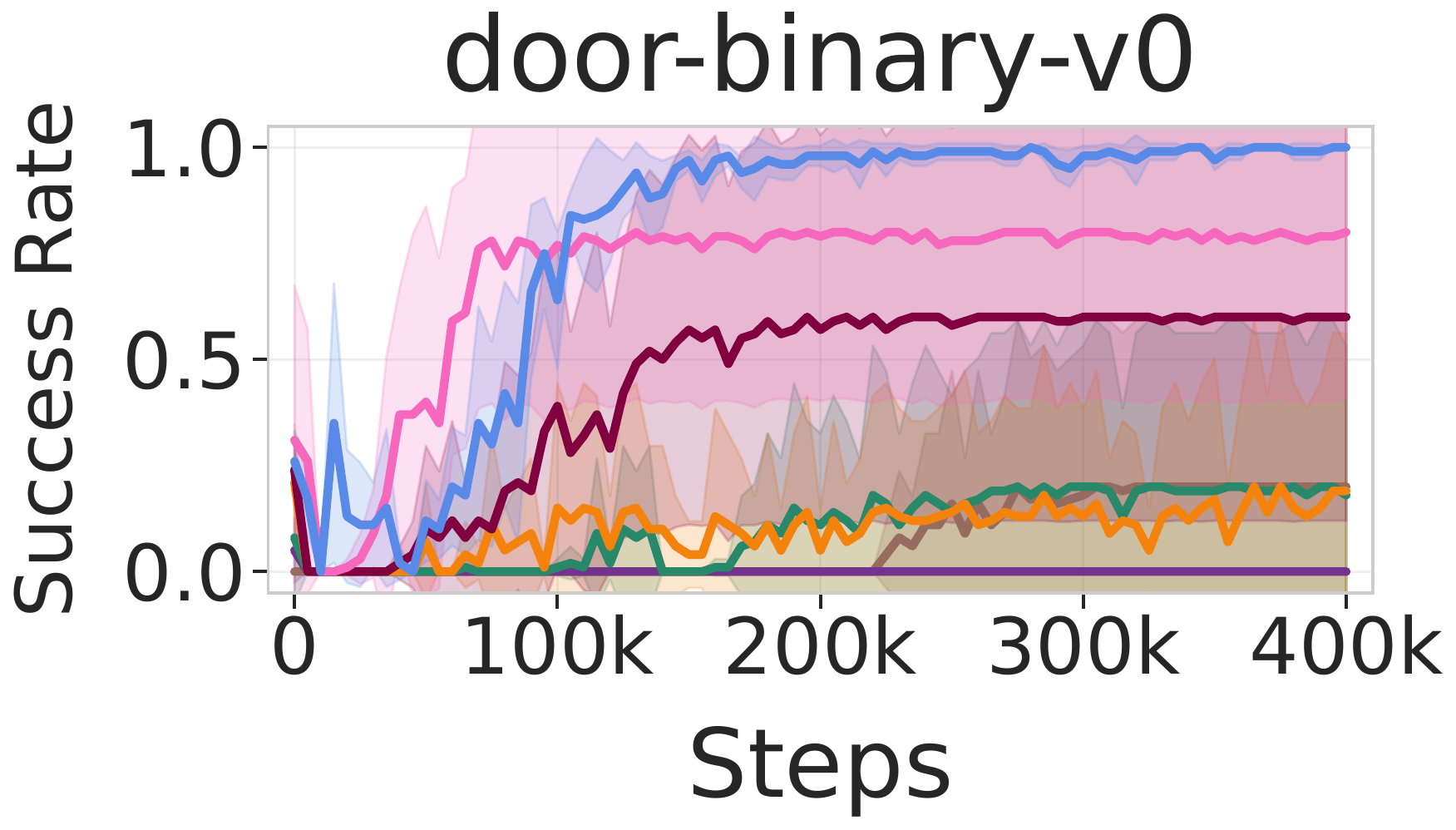}
    \end{subfigure}\hspace{0.1em}
    \begin{subfigure}[c]{0.24\textwidth}
        \includegraphics[width=1.0\linewidth]{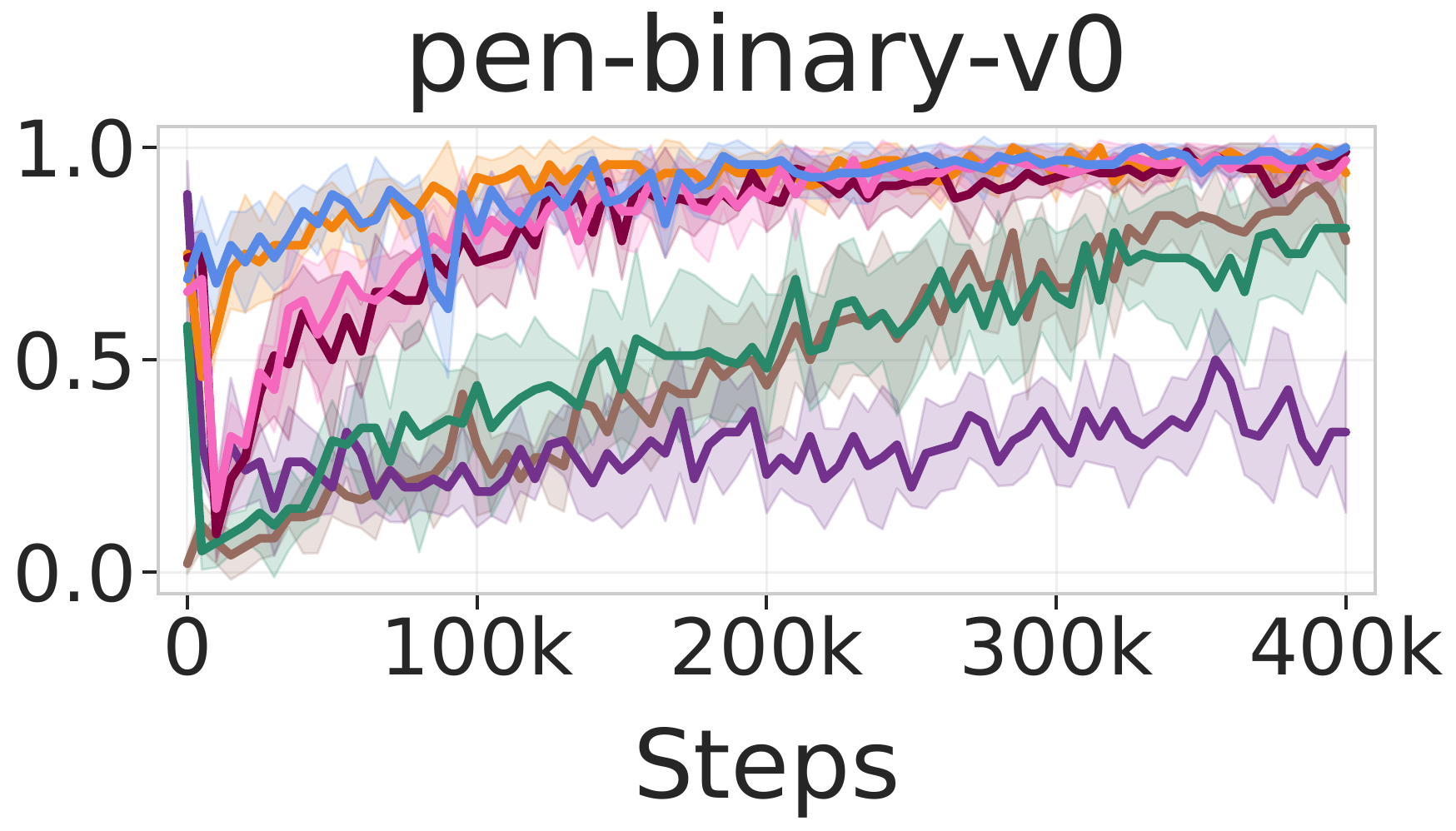}
    \end{subfigure}\hspace{0.1em} \\
    \begin{subfigure}[c]{0.24\textwidth}
        \includegraphics[width=1.0\linewidth]{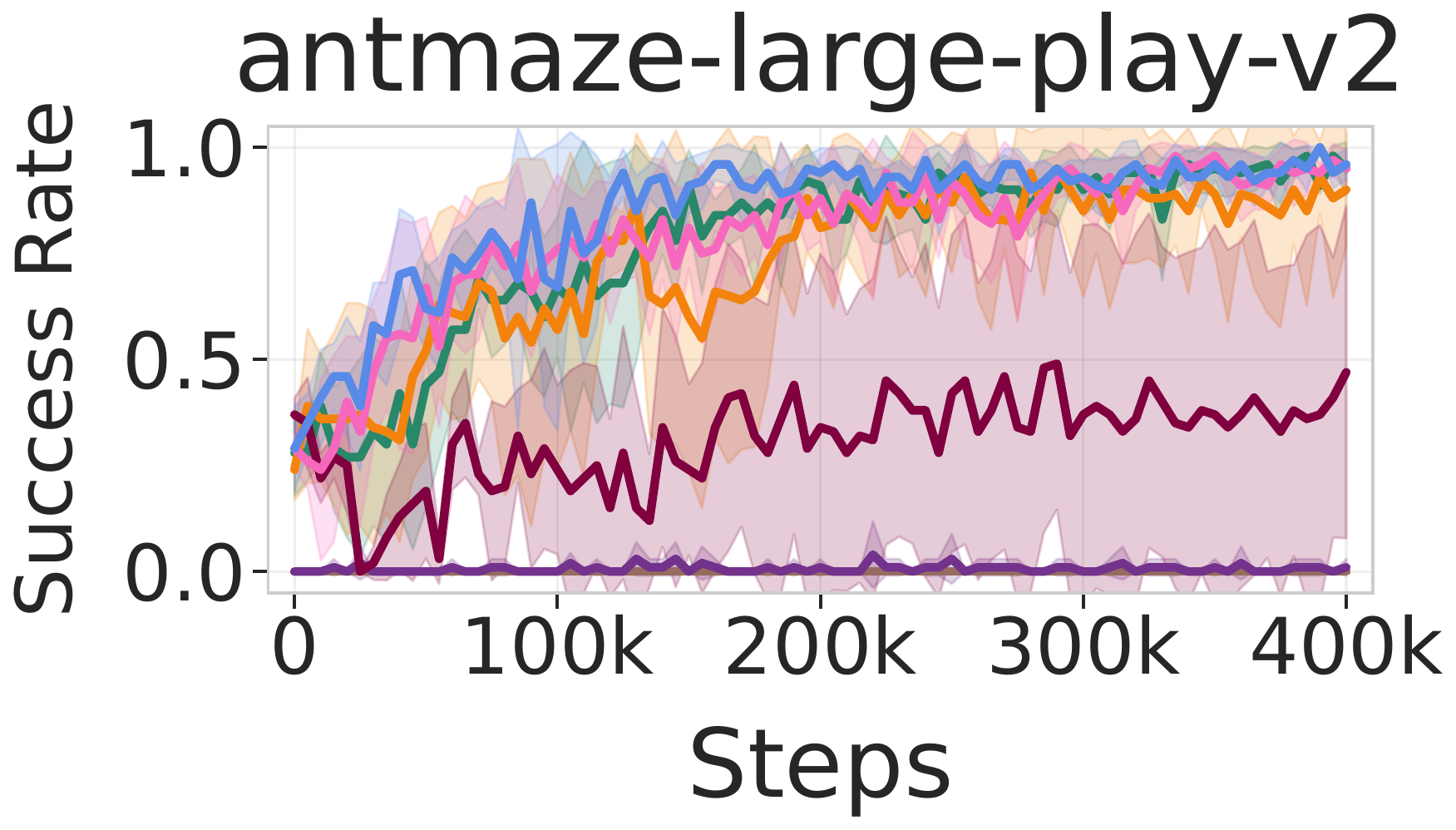}
    \end{subfigure}\hspace{0.1em}
    \begin{subfigure}[c]{0.24\textwidth}
        \includegraphics[width=1.0\linewidth]{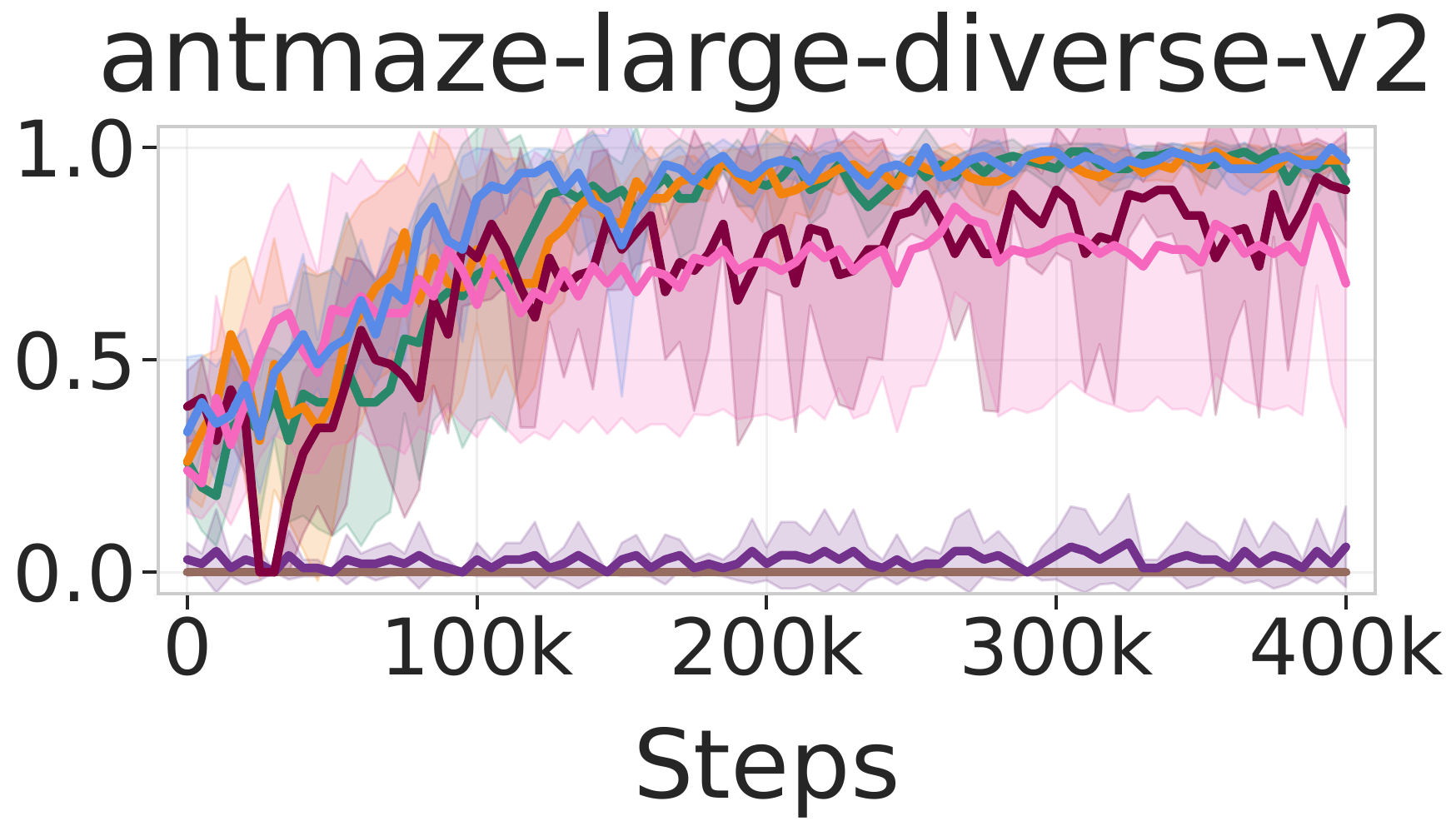}
    \end{subfigure}\hspace{0.1em}
    \begin{subfigure}[c]{0.24\textwidth}
        \includegraphics[width=1.0\linewidth]{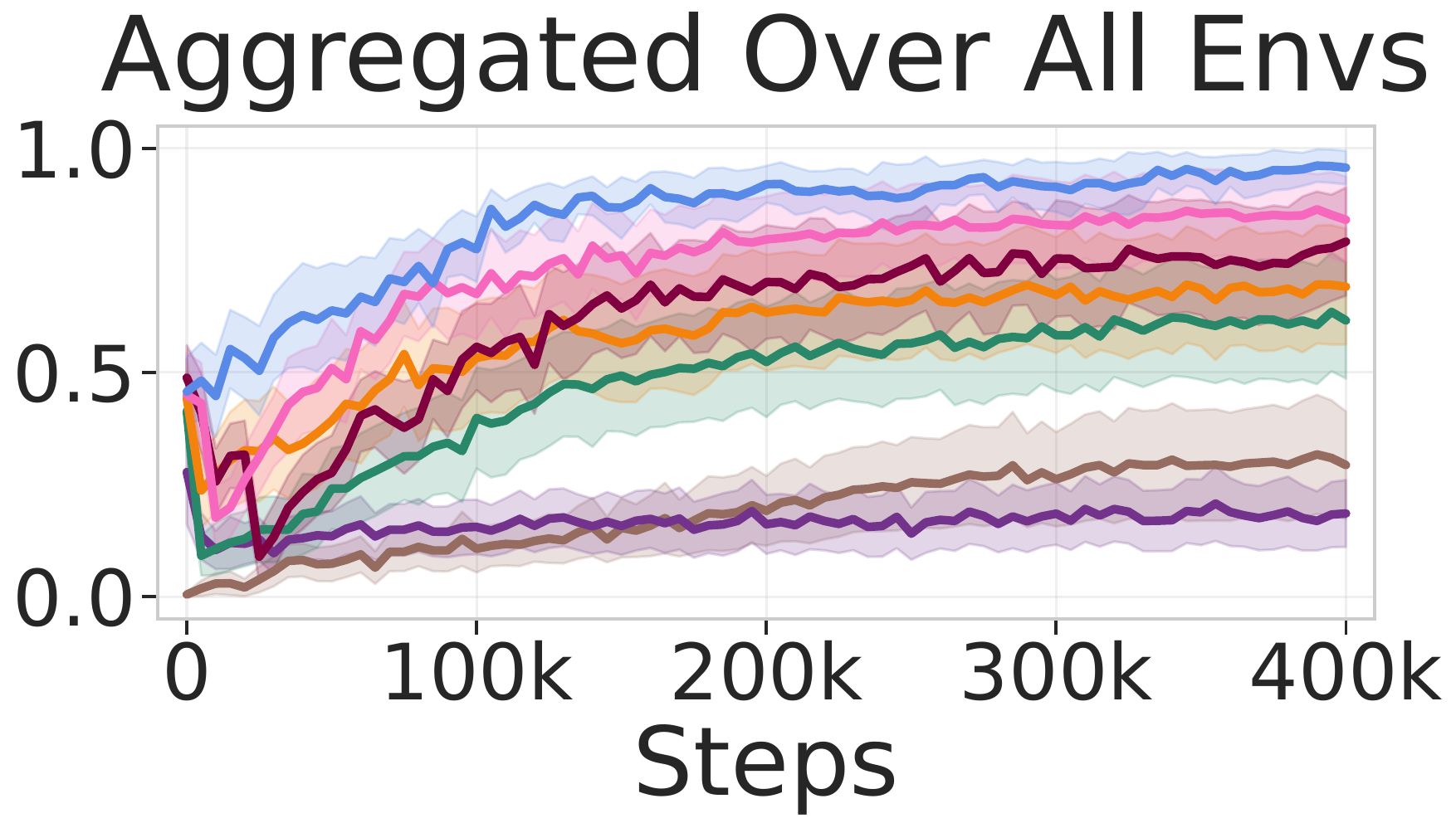}
    \end{subfigure}\hspace{0.1em}
    \includegraphics[width=0.6\linewidth]{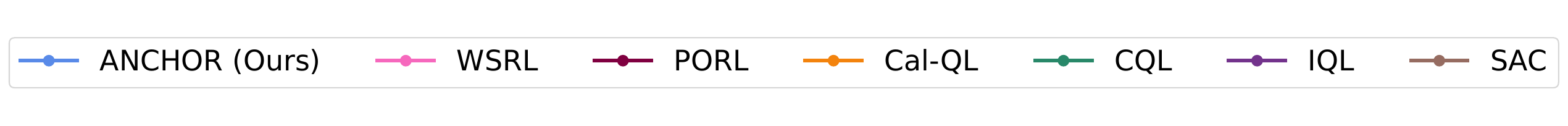}
    \caption{\textbf{ANCHOR improves both early stability and final performance.}
    Existing methods do not consistently achieve both low catastrophic failure and strong asymptotic performance across tasks. In contrast, ANCHOR reduces early performance collapse while attaining strong final performance across tasks. The aggregate curve reports the mean success rate averaged over the six tasks.}
    \label{fig:baseline_performance}
    \vspace{-1em}
\end{figure}

Figure~\ref{fig:baseline_performance} summarizes the main benchmark results. Across the six-task suite, ANCHOR achieves the best average final performance while
substantially reducing catastrophic failure. In door-binary, where the offline-pretrained policy starts from relatively low performance, the benefit of preventing catastrophic failure is less pronounced; nevertheless, ANCHOR achieves the best asymptotic performance.

Existing baselines often fail to achieve both early stability and strong final
performance. Cal-QL, which retains stronger conservative regularization, can reduce early
collapse but often adapts more slowly, whereas more aggressive online adaptation methods,
including WSRL and PORL, can suffer larger early performance drops. Full numerical results, reported as mean $\pm$ standard error, are provided in Appendix~\ref{sec:appendix:numerical_results}.

We further evaluate ANCHOR in additional environments that stress different failure modes:
near-zero pretrained performance in kitchen-complete and relocate-binary, and harder
sparse-reward exploration in antmaze-ultra-diverse
(Appendix~\ref{sec:appendix:additional_environments}). In kitchen-complete, ANCHOR remains competitive with the KL coefficient set to zero, but, like other baselines, fails to improve
asymptotic performance in relocate-binary. In antmaze-ultra-diverse, ANCHOR also fails to
improve over the pretrained policy. These results suggest that KL anchoring is useful when
there is pretrained behavior to preserve, but may need to be combined with stronger
exploration or adaptation mechanisms in harder failure cases.

We also analyze whether KL-anchored adaptation can be integrated with another offline RL
backbone, IQL. Appendix~\ref{sec:appendix:ours_with_other_offline_rl} shows that adding KL
anchoring to IQL reduces early catastrophic failure compared to naive online fine-tuning,
although it does not improve asymptotic performance.

Additional comparisons with policy-side adaptations and BC controls, aggregate statistical analysis, RND diagnostics, and FQL/IFQL extensions are reported in Appendix~\ref{sec:appendix:additional_evidence}.

\subsection{Analysis of ANCHOR}
\label{sec:experiments:kl_anchoring}

\textbf{Annealing the KL constraint balances early anchoring and later adaptation.}
We first ask whether the KL constraint should be kept throughout online fine-tuning,
removed from the beginning, or annealed over time. To answer this, we compare ANCHOR
against two alternatives: fixed-KL, which keeps the initial KL weight throughout online
fine-tuning without annealing, and no-KL, which does not use KL regularization.

\begin{figure}[ht]
    \centering
    \hspace{-1.0em}
    \begin{subfigure}[c]{0.15\textwidth}
        \hfill
        \includegraphics[width=1.0\linewidth]{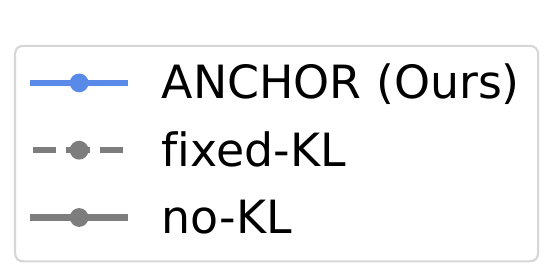}
    \end{subfigure}\hspace{1.0em}
    \begin{subfigure}[c]{0.24\textwidth}
        \includegraphics[width=1.0\linewidth]{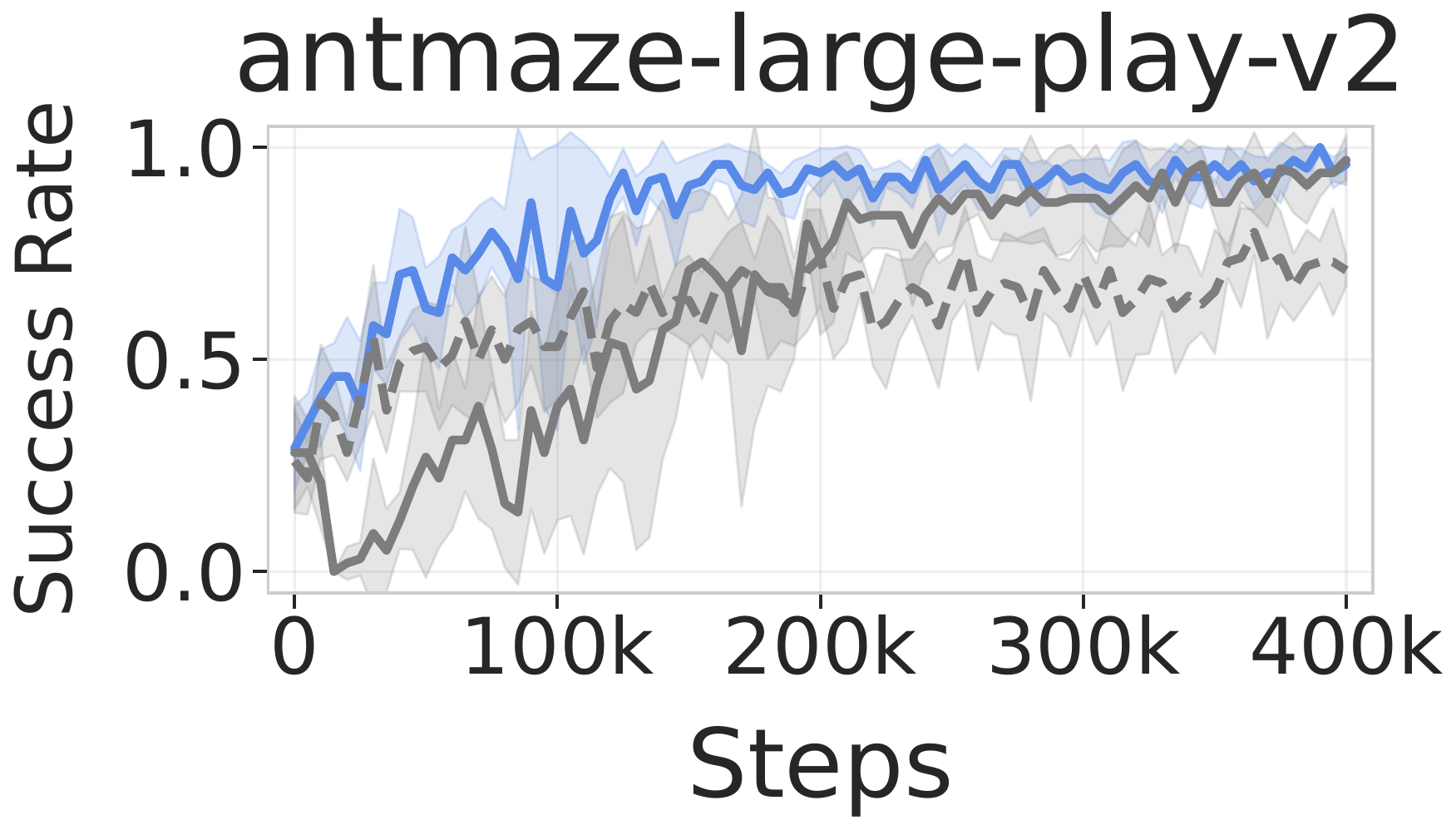}
    \end{subfigure}\hspace{0.1em}
    \begin{subfigure}[c]{0.24\textwidth}
        \includegraphics[width=1.0\linewidth]{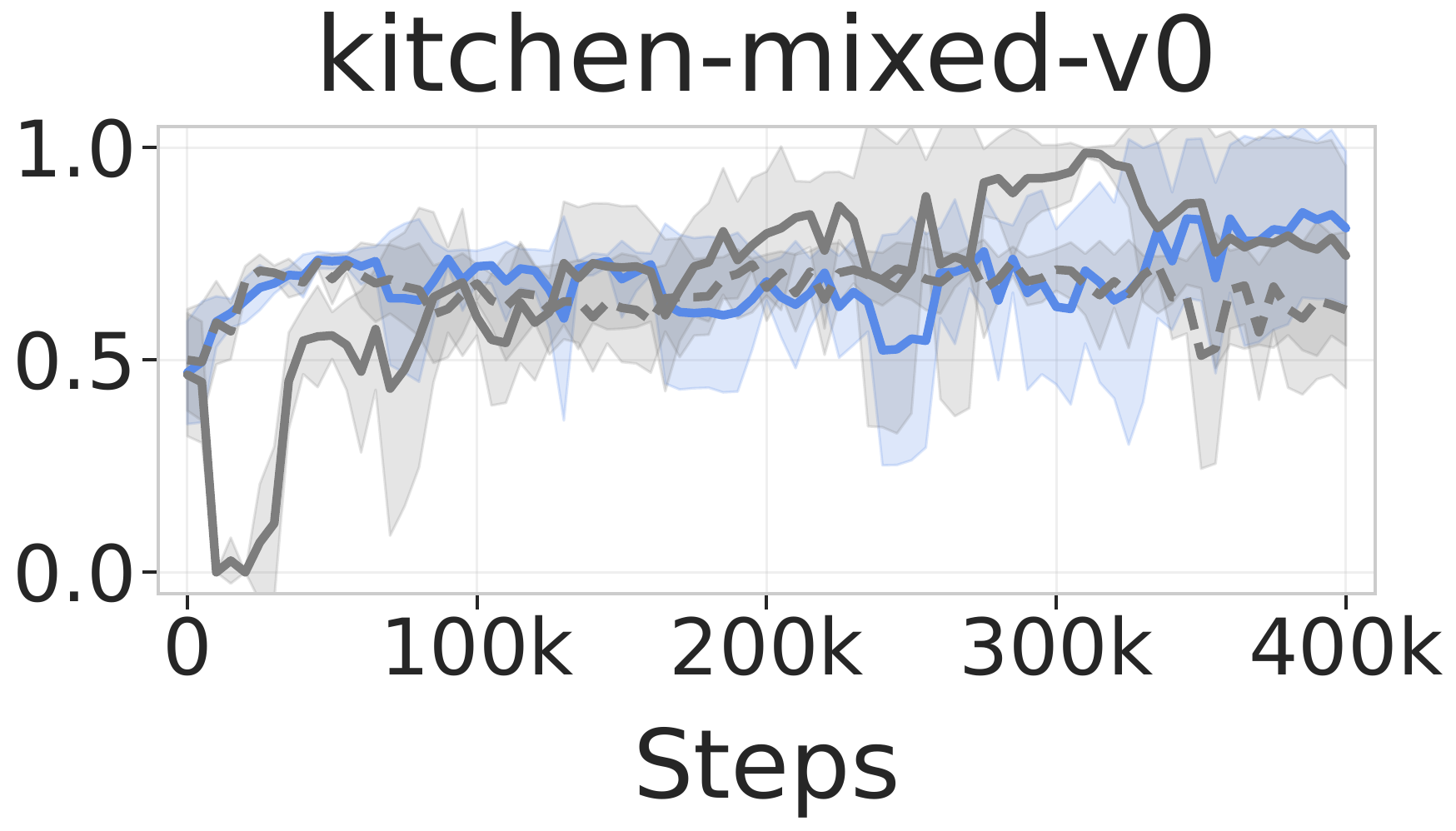}
    \end{subfigure}\hspace{0.1em}
    \begin{subfigure}[c]{0.24\textwidth}
        \includegraphics[width=1.0\linewidth]{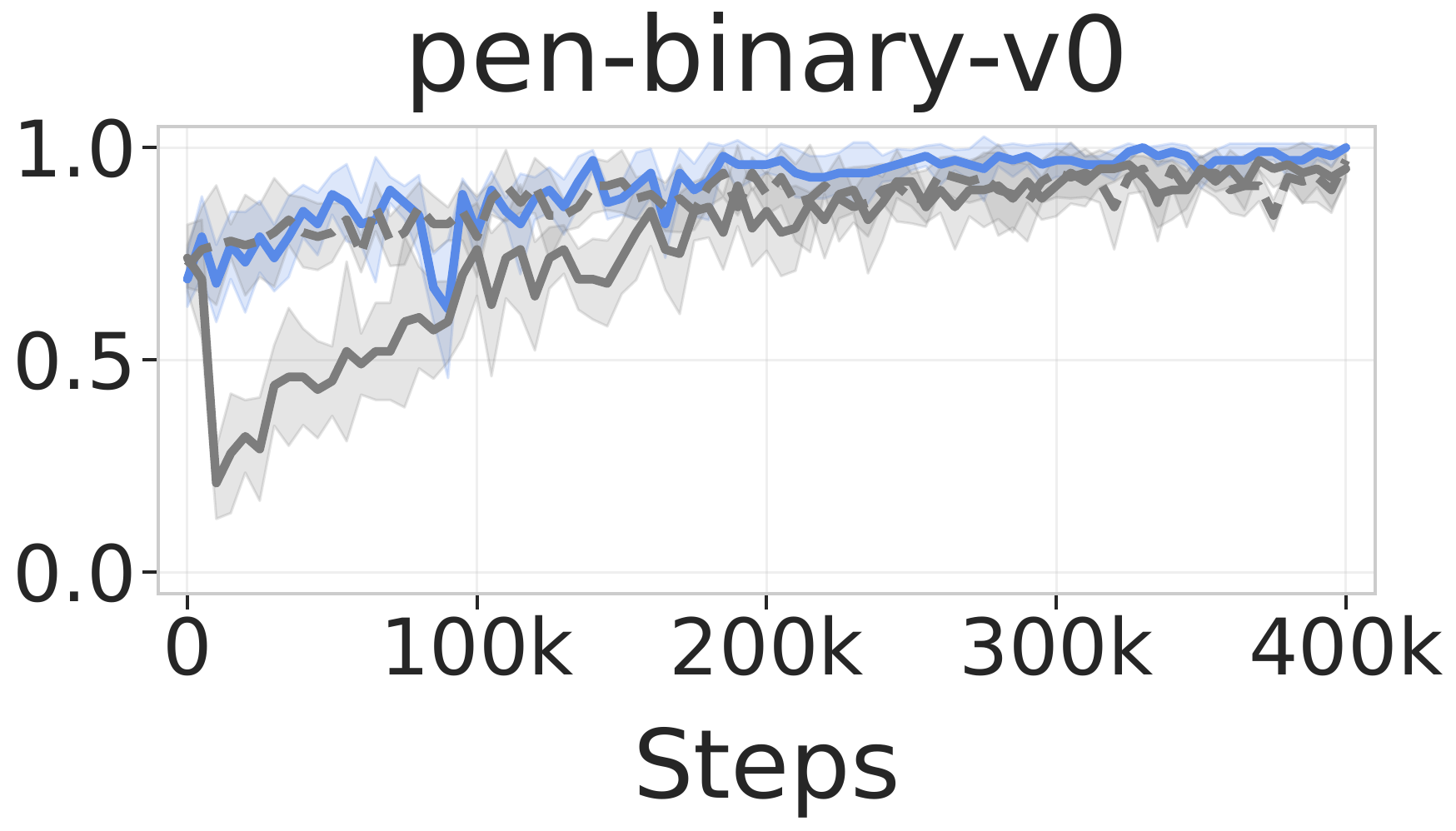}
    \end{subfigure}\hspace{0.1em}
    \caption{\textbf{KL annealing improves early stability while preserving final performance.}
    ANCHOR is compared with the fixed-KL and no-KL variants. The fixed-KL variant improves
    early stability but can limit final improvement, while the no-KL variant adapts
    aggressively but causes early collapse.}
    \label{fig:fixed_kl_vs_annealed_kl}
\end{figure}

As shown in Figure~\ref{fig:fixed_kl_vs_annealed_kl}, ANCHOR most consistently achieves
both low catastrophic failure and strong final performance across the three tasks. The fixed-KL variant demonstrates the benefit of anchoring: keeping the actor close to the pretrained policy can suppress early collapse. However,
because this constraint is never annealed, it can remain overly restrictive and reduce
asymptotic performance, especially when the offline-pretrained policy is not already
strong, as in antmaze-large-play and kitchen-mixed. Conversely, the no-KL variant removes
this restriction entirely and allows more aggressive online adaptation, but it exhibits
large catastrophic failure across all three tasks because the policy can drift away from
the pretrained behavior too abruptly. These patterns justify the annealing design of
ANCHOR: KL annealing provides a middle path by using strong anchoring when online
fine-tuning is most vulnerable, and then relaxing the constraint as the online policy and
critic adapt.

\textbf{The benefit of ANCHOR depends on the quality of the pretrained policy.}
We next examine when ANCHOR is most beneficial by varying the quality of the
offline-pretrained policy. We compare ANCHOR with the no-KL variant across pretrained
checkpoints with different initial success rates.

\begin{figure}[t]
    \centering
    \begin{subfigure}[c]{0.27\textwidth}
        \includegraphics[width=1.0\linewidth]{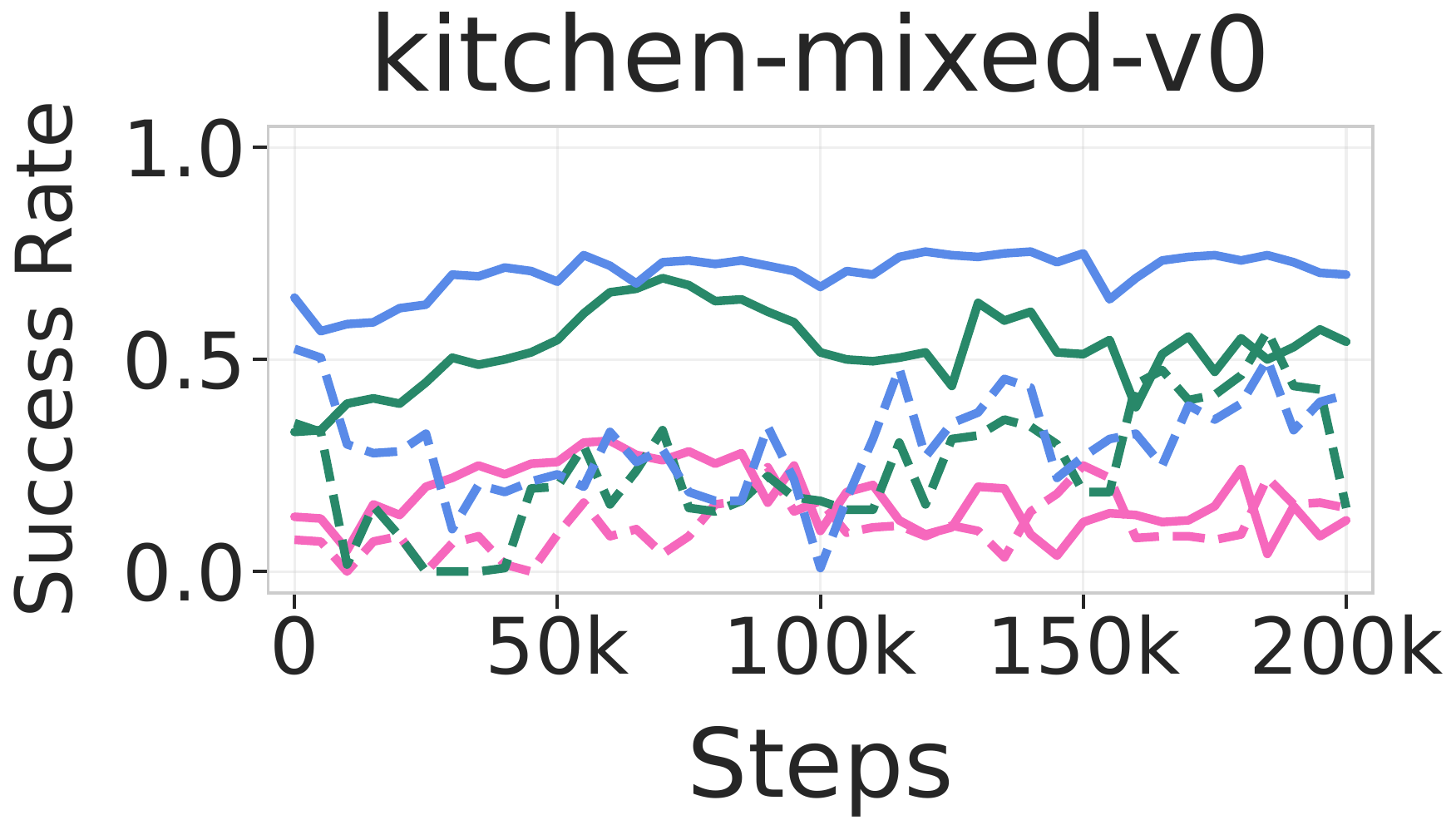}
    \end{subfigure}\hspace{0.02\linewidth}
    \begin{subfigure}[c]{0.27\textwidth}
        \includegraphics[width=1.0\linewidth]{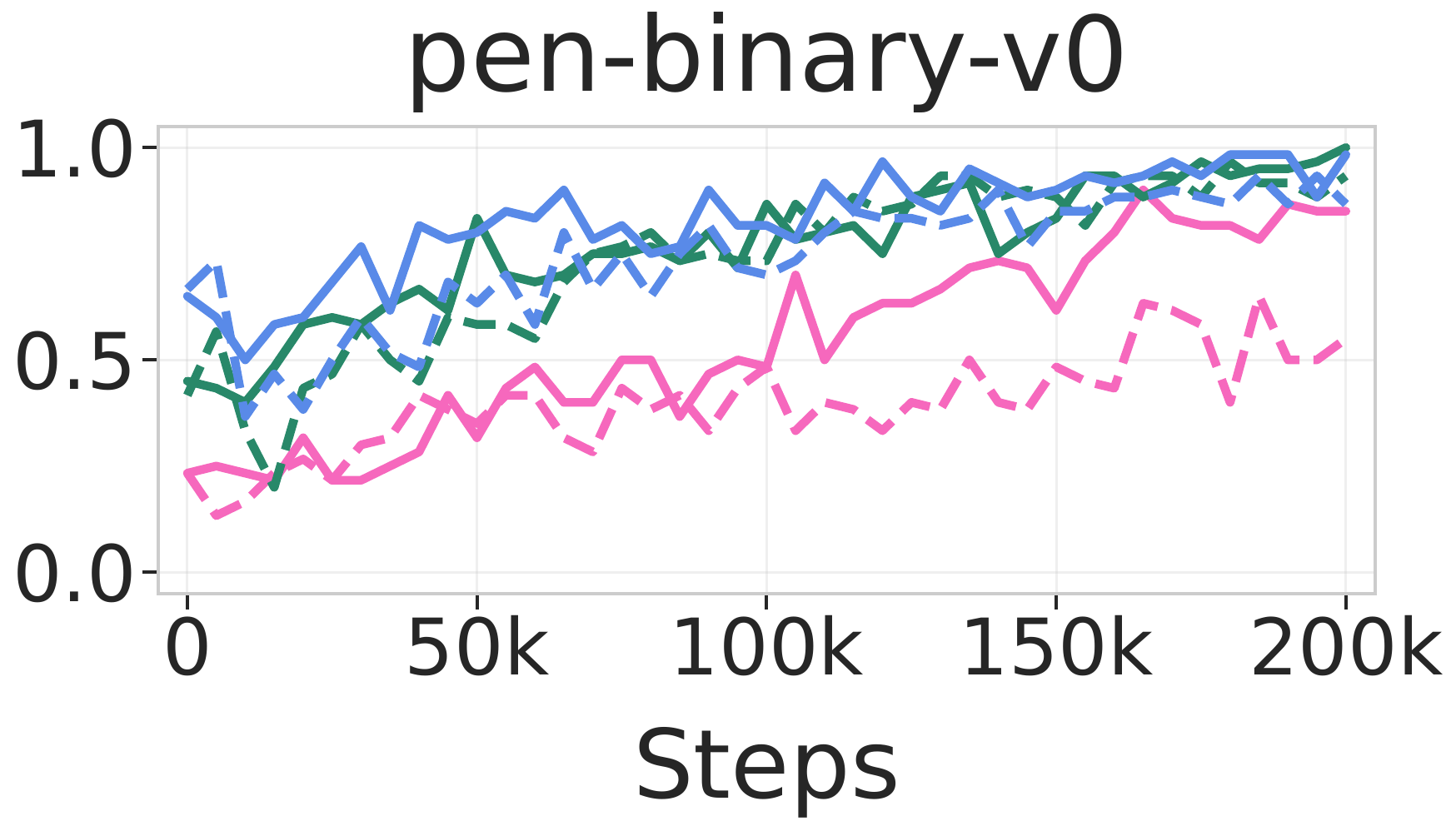}
    \end{subfigure}

    \includegraphics[width=0.5\linewidth]{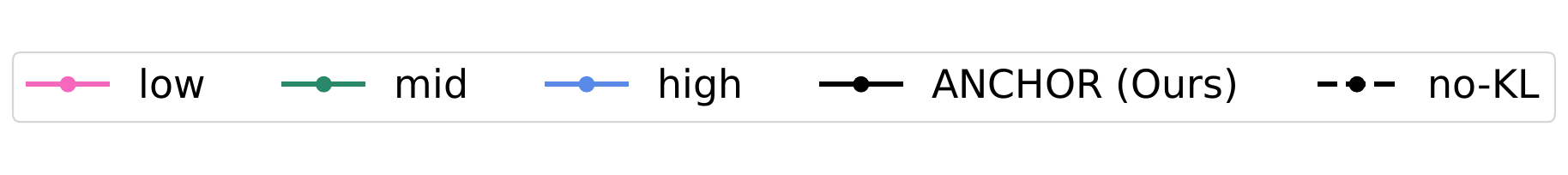}
    \vspace{-0.6em}

    \caption{\textbf{Benefit of ANCHOR depends on pretrained policy quality.}
    ANCHOR reduces catastrophic failure more clearly when the offline-pretrained policy
    starts from a higher initial success rate. Curves with the same color correspond to
    the same pretrained policy. Results are averaged over three seeds.}
    \label{fig:kl_anchoring_importance}
\end{figure}

Figure~\ref{fig:kl_anchoring_importance} shows that the effect of ANCHOR depends on the
quality of the offline-pretrained policy. In both pen-binary and kitchen-mixed, when the
pretrained policy starts with a low initial success rate, the difference in catastrophic
failure between ANCHOR and the no-KL variant is less pronounced. In contrast, when the
pretrained policy starts from a stronger initial success rate, the gap becomes much
clearer: ANCHOR substantially reduces catastrophic failure relative to the no-KL variant.

Importantly, even in the low-initial-success regime, ANCHOR remains a reasonable design
choice. Compared to the no-KL variant, it still achieves strong asymptotic performance in
pen-binary and comparable asymptotic performance in kitchen-mixed. Taken together, these
results indicate that the main value of ANCHOR lies in protecting useful pretrained
behavior when it exists, while not substantially hindering final online performance even
when the pretrained policy is less reliable. Results with full confidence intervals are
provided in Appendix~\ref{sec:appendix:numerical_results}.

\textbf{How sensitive is ANCHOR to its hyperparameters?} We finally study the sensitivity of ANCHOR to three hyperparameters: the initial KL weight
$\beta_0$, the KL annealing interval $T_{\mathrm{KL}}$, and the residual online
conservative regularization strength $\alpha_{\mathrm{on}}$. The full analysis and Figure~\ref{fig:anchor_ablation} are provided in Appendix~\ref{sec:appendix:diagnostic_figures}.

\section{Conclusion}
\label{sec:conclusion}

Offline-to-online adaptation without offline replay requires balancing early stability with
online improvement. ANCHOR addresses this by anchoring the actor to the offline-pretrained
policy during the vulnerable early phase, annealing this constraint as online learning
progresses, and retaining only a small residual conservative penalty for the critic. Across
FrankaKitchen, AntMaze, and Adroit tasks, this simple design reduces catastrophic failure
while maintaining strong asymptotic performance.

Our analysis and ablations clarify the operating conditions of ANCHOR. The theoretical
results suggest that anchoring is most useful when the pretrained policy induces nearby
occupancy distributions under which the critic remains reliable, while the two-phase regret bound explains why the anchor should eventually be relaxed when further online improvement is needed. Empirically, the ablations show that ANCHOR is not simply a fixed constraint: its benefit comes from using the pretrained policy as a temporary behavioral prior, while allowing both the actor and critic to adapt as online data accumulates. They also show that the residual conservative penalty should be kept small enough to avoid limiting final performance, but can still help stabilize the early online phase in some tasks.

These findings suggest that offline-to-online RL methods should control how strongly and how long they rely on the pretrained policy during deployment-time adaptation. Future work could develop adaptive anchoring schedules that adjust the KL constraint based on online signals. Another important direction is to better disentangle the roles of actor drift and critic error in catastrophic failure, which may lead to more targeted stabilization mechanisms. Finally,
settings where the pretrained policy provides little useful guidance may require combining
anchoring with stronger exploration mechanisms that enable online improvement without
reintroducing early collapse. Additional limitations are discussed in
Appendix~\ref{sec:appendix:limitations}.

\label{main-text-end}
\clearpage

\bibliography{main}
\bibliographystyle{plainnat}

\clearpage
\appendix

\section{Theoretical Discussion}
\label{sec:appendix:lemma}

\setcounter{lemma}{0}
\renewcommand{\theHlemma}{appendix.\arabic{lemma}}

This appendix provides a formal discussion of ANCHOR in the offline-to-online setting.
For simplicity, we present the formal argument for finite action spaces. The same
interpretation extends to continuous actions by replacing sums with integrals under
standard absolute-continuity and boundedness assumptions. Our goal is to formalize the
local occupancy-control intuition rather than to provide a full convergence guarantee for
the continuous-control implementation.

\paragraph{Notation.}
We consider an infinite-horizon discounted MDP with discount factor $\gamma \in (0,1)$,
finite action set $\mathcal A$, reward function $r(s,a)$, and normalized discounted
state-action occupancy
\[
d^\pi(s,a)
:=
(1-\gamma)\sum_{k=0}^\infty \gamma^k \Pr_\pi(s_k=s,a_k=a).
\]
Write $\rho^\pi(s)=\sum_a d^\pi(s,a)$ for the state marginal. With this normalization, $d^\pi$ is a probability distribution over $(s,a)$ and
\[
J(\pi)
=
\frac{1}{1-\gamma}\,
\mathbb E_{(s,a)\sim d^\pi}[r(s,a)].
\]

For two probability distributions $P$ and $Q$ over a finite set, we define
\[
D_{\mathrm{KL}}(P\|Q)
:=
\sum_x P(x)\log\frac{P(x)}{Q(x)},
\qquad
\mathrm{TV}(P,Q)
:=
\frac12\sum_x |P(x)-Q(x)|
=
\frac12\|P-Q\|_1.
\]
Under this convention, $\mathrm{TV}(P,Q)\in[0,1]$ and Pinsker's inequality gives
\[
\mathrm{TV}(P,Q)
\le
\sqrt{\frac12 D_{\mathrm{KL}}(P\|Q)}.
\]

For two policies $\pi,\mu$ and a reference state distribution $\rho$, define the
state-averaged KL and TV distances as
\[
D_{\rho}(\pi\|\mu)
:=
\mathbb E_{s\sim \rho}\!\left[
D_{\mathrm{KL}}(\pi(\cdot|s)\|\mu(\cdot|s))
\right],
\qquad
\mathrm{TV}_{\rho}(\pi,\mu)
:=
\mathbb E_{s\sim \rho}\!\left[
\mathrm{TV}(\pi(\cdot|s),\mu(\cdot|s))
\right].
\]
We also write
\[
A^\mu(s,a):=Q^\mu(s,a)-V^\mu(s).
\]

\paragraph{Performance difference lemma.}
We will repeatedly use the standard identity
\begin{equation}
\label{eq:pdl}
J(\pi)-J(\mu)
=
\frac{1}{1-\gamma}\,
\mathbb E_{(s,a)\sim d^\pi}\!\left[A^\mu(s,a)\right].
\end{equation}
This is the classical performance difference lemma.

\begin{assumption}[Local occupancy control around the offline policy]
\label{ass:occ}
There exists a constant $C_{\mathrm{occ}}>0$ such that, for every policy $\pi$ in the anchored phase,
\[
\|d^\pi-d^{\pi_{\mathrm{off}}}\|_1
\le
C_{\mathrm{occ}}\,
\mathrm{TV}_{\rho^{\pi_{\mathrm{off}}}}(\pi,\pi_{\mathrm{off}}).
\]
\end{assumption}

Assumption~\ref{ass:occ} is a local smoothness condition on the policy-to-occupancy map
around $\pi_{\mathrm{off}}$: within the anchored region, small policy deviations lead to
proportionally controlled occupancy shifts. It excludes pathological cases in which a
change on states with negligible offline visitation produces a large downstream change in the trajectory distribution. Since ANCHOR applies KL anchoring precisely during the early phase to keep the policy close to $\pi_{\mathrm{off}}$ on sampled states, our lemma is intended to characterize this local anchored regime rather than arbitrary policies far from the offline reference.

\subsection{Lemma 1: Critic error under occupancy shift}
\label{sec:lemma:value_divergence}

\begin{lemma}[Critic error difference under occupancy shift]
\label{thm:app:1}
Let $\widehat Q_t$ be a critic trained on online data collected from occupancy $d_t$.
Fix any target policy $\pi$, and define
\[
e_t(s,a):=\big|\widehat Q_t(s,a)-Q^\pi(s,a)\big|.
\]
Assume that
\[
|\widehat Q_t(s,a)|\le Q_{\max},
\qquad
|Q^\pi(s,a)|\le Q_{\max}
\qquad
\text{for all }(s,a).
\]
Then, for any two occupancy measures $d$ and $d'$,
\begin{equation}
\label{eq:thm1_main}
\left|
\mathbb E_{(s,a)\sim d}[e_t(s,a)]
-
\mathbb E_{(s,a)\sim d'}[e_t(s,a)]
\right|
\le
2Q_{\max}\|d-d'\|_1.
\end{equation}

In particular, if the online actor satisfies
\[
D_{\rho^{\pi_{\mathrm{off}}}}(\pi_t\|\pi_{\mathrm{off}})\le \delta_t,
\]
then under Assumption~\ref{ass:occ},
\begin{equation}
\label{eq:thm1_occ}
\|d^{\pi_t}-d^{\pi_{\mathrm{off}}}\|_1
\le
C_{\mathrm{occ}}\sqrt{\delta_t/2},
\end{equation}
and therefore
\begin{equation}
\label{eq:thm1_offsupport}
\left|
\mathbb E_{d^{\pi_t}}[e_t]
-
\mathbb E_{d^{\pi_{\mathrm{off}}}}[e_t]
\right|
\le
2Q_{\max}C_{\mathrm{occ}}\sqrt{\delta_t/2}.
\end{equation}
\end{lemma}

\begin{proof}
By the boundedness assumptions and the triangle inequality,
\[
e_t(s,a)
=
|\widehat Q_t(s,a)-Q^\pi(s,a)|
\le
|\widehat Q_t(s,a)|+|Q^\pi(s,a)|
\le
2Q_{\max}.
\]
Thus, for any two occupancy measures $d$ and $d'$,
\[
\mathbb E_d[e_t]-\mathbb E_{d'}[e_t]
=
\sum_{s,a}\big(d(s,a)-d'(s,a)\big)e_t(s,a).
\]
Taking absolute values and applying Hölder's inequality,
\[
\left|
\mathbb E_d[e_t]-\mathbb E_{d'}[e_t]
\right|
\le
\|e_t\|_\infty \|d-d'\|_1
\le
2Q_{\max}\|d-d'\|_1.
\]
This proves~\eqref{eq:thm1_main}.

Now suppose
\[
D_{\rho^{\pi_{\mathrm{off}}}}(\pi_t\|\pi_{\mathrm{off}})\le \delta_t.
\]
By Pinsker's inequality, for each state $s$,
\[
\mathrm{TV}(\pi_t(\cdot|s),\pi_{\mathrm{off}}(\cdot|s))
\le
\sqrt{
\frac12
D_{\mathrm{KL}}(\pi_t(\cdot|s)\|\pi_{\mathrm{off}}(\cdot|s))
}.
\]
Taking expectation over $s\sim \rho^{\pi_{\mathrm{off}}}$ and applying Jensen's inequality,
\[
\mathrm{TV}_{\rho^{\pi_{\mathrm{off}}}}(\pi_t,\pi_{\mathrm{off}})
\le
\sqrt{
\frac12
D_{\rho^{\pi_{\mathrm{off}}}}(\pi_t\|\pi_{\mathrm{off}})
}
\le
\sqrt{\delta_t/2}.
\]
Combining this with Assumption~\ref{ass:occ} gives
\[
\|d^{\pi_t}-d^{\pi_{\mathrm{off}}}\|_1
\le
C_{\mathrm{occ}}\sqrt{\delta_t/2}.
\]
Finally, applying~\eqref{eq:thm1_main} with
$d=d^{\pi_t}$ and $d'=d^{\pi_{\mathrm{off}}}$ yields
\[
\left|
\mathbb E_{d^{\pi_t}}[e_t]
-
\mathbb E_{d^{\pi_{\mathrm{off}}}}[e_t]
\right|
\le
2Q_{\max}C_{\mathrm{occ}}\sqrt{\delta_t/2}.
\]
\end{proof}

\subsection{Lemma 2: Two-phase regret decomposition}
\label{sec:lemma:two_phase_regret}

\begin{lemma}[Two-phase regret decomposition]
Let $T=T_1+T_2$, and define the policy-sequence regret as
\[
\mathrm{Reg}_T
:=
\sum_{t=1}^{T}\left(J(\pi^*)-J(\pi_t)\right),
\]
where $\pi^*$ is an optimal policy. In phase I ($1\le t\le T_1$), suppose ANCHOR uses KL anchoring satisfying
\[
D_{\rho^{\pi_{\mathrm{off}}}}(\pi_t\|\pi_{\mathrm{off}})\le \delta_t.
\]
In phase II ($T_1+1\le t\le T$), suppose the base online learner satisfies
\[
\sum_{t=T_1+1}^{T}
\left(J(\pi^*)-J(\pi_t)\right)
\le
\mathrm{Reg}_{\mathrm{base}}(T_2).
\]
Assume that
\[
|A^{\pi_{\mathrm{off}}}(s,a)|\le A_{\max}
\qquad\text{for all }(s,a),
\]
and define
\[
\Delta_{\mathrm{off}}:=J(\pi^*)-J(\pi_{\mathrm{off}}).
\]
Then under Assumption~\ref{ass:occ},

\begin{equation}
\label{eq:thm2_main}
\mathrm{Reg}_T
\le
T_1\Delta_{\mathrm{off}}
+
\frac{A_{\max}C_{\mathrm{occ}}}{1-\gamma}
\sum_{t=1}^{T_1}\sqrt{\delta_t/2}
+\mathrm{Reg}_{\mathrm{base}}(T_2).
\end{equation}
\end{lemma}

\begin{proof}
By definition,
\[
\mathrm{Reg}_T
=
\sum_{t=1}^T \big(J(\pi^*)-J(\pi_t)\big)
=
\sum_{t=1}^{T_1}\big(J(\pi^*)-J(\pi_t)\big)
+
\sum_{t=T_1+1}^{T}\big(J(\pi^*)-J(\pi_t)\big).
\]

For phase I, add and subtract $J(\pi_{\mathrm{off}})$:
\[
J(\pi^*)-J(\pi_t)
=
\Delta_{\mathrm{off}}+\big(J(\pi_{\mathrm{off}})-J(\pi_t)\big).
\]
Thus it remains to bound $|J(\pi_t)-J(\pi_{\mathrm{off}})|$.

By the performance difference lemma~\eqref{eq:pdl},
\[
J(\pi_t)-J(\pi_{\mathrm{off}})
=
\frac{1}{1-\gamma}
\mathbb E_{(s,a)\sim d^{\pi_t}}\!\left[A^{\pi_{\mathrm{off}}}(s,a)\right].
\]
Also,
\[
\mathbb E_{(s,a)\sim d^{\pi_{\mathrm{off}}}}\!\left[A^{\pi_{\mathrm{off}}}(s,a)\right]=0,
\]
because
\[
\sum_a \pi_{\mathrm{off}}(a|s)A^{\pi_{\mathrm{off}}}(s,a)=0
\qquad\text{for every state }s.
\]
Therefore,
\[
J(\pi_t)-J(\pi_{\mathrm{off}})
=
\frac{1}{1-\gamma}
\sum_{s,a}\big(d^{\pi_t}(s,a)-d^{\pi_{\mathrm{off}}}(s,a)\big)
A^{\pi_{\mathrm{off}}}(s,a).
\]
Using $|A^{\pi_{\mathrm{off}}}(s,a)|\le A_{\max}$,
\[
|J(\pi_t)-J(\pi_{\mathrm{off}})|
\le
\frac{A_{\max}}{1-\gamma}\,
\|d^{\pi_t}-d^{\pi_{\mathrm{off}}}\|_1.
\]
By Assumption~\ref{ass:occ} and Pinsker,
\[
\|d^{\pi_t}-d^{\pi_{\mathrm{off}}}\|_1
\le
C_{\mathrm{occ}}\,
\mathrm{TV}_{\rho^{\pi_{\mathrm{off}}}}(\pi_t,\pi_{\mathrm{off}})
\le
C_{\mathrm{occ}}\sqrt{\delta_t/2}.
\]
Hence
\[
J(\pi^*)-J(\pi_t)
\le
\Delta_{\mathrm{off}}
+
\frac{A_{\max}C_{\mathrm{occ}}}{1-\gamma}\sqrt{\delta_t/2}.
\]
Summing over $t=1,\dots,T_1$ yields
\[
\sum_{t=1}^{T_1}\big(J(\pi^*)-J(\pi_t)\big)
\le
T_1\Delta_{\mathrm{off}}
+
\frac{A_{\max}C_{\mathrm{occ}}}{1-\gamma}
\sum_{t=1}^{T_1}\sqrt{\delta_t/2}.
\]

For phase II, apply the assumed regret guarantee of the base learner:
\[
\sum_{t=T_1+1}^{T}\big(J(\pi^*)-J(\pi_t)\big)
\le
\mathrm{Reg}_{\mathrm{base}}(T_2).
\]
Combining the two parts proves~\eqref{eq:thm2_main}.
\end{proof}

\section{Other Attempts to Mitigate Catastrophic Failure}
\label{sec:appendix:other_attempts}

\begin{figure}[H]
    \centering
    \begin{subfigure}[c]{0.1\textwidth}
        \includegraphics[width=1.0\textwidth]{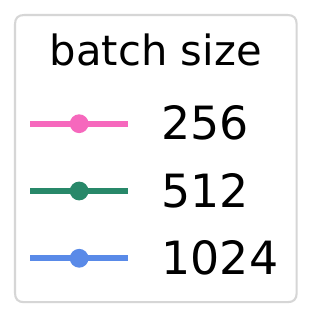}
    \end{subfigure}\hspace{0.1em}
    \begin{subfigure}[c]{0.21\textwidth}
        \includegraphics[width=1.0\linewidth]{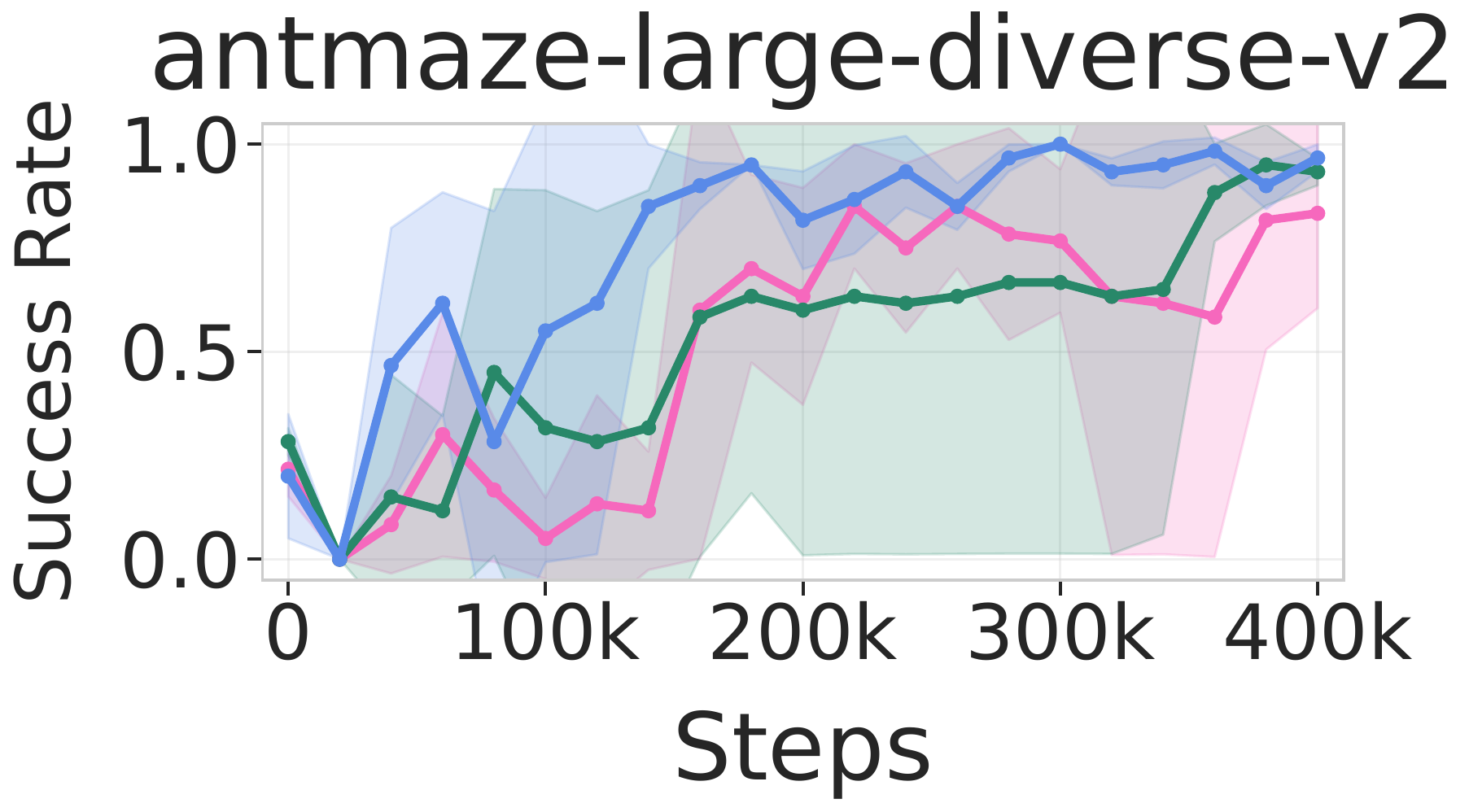}
    \end{subfigure}\hspace{0.1em}
    \begin{subfigure}[c]{0.21\textwidth}
        \includegraphics[width=1.0\linewidth]{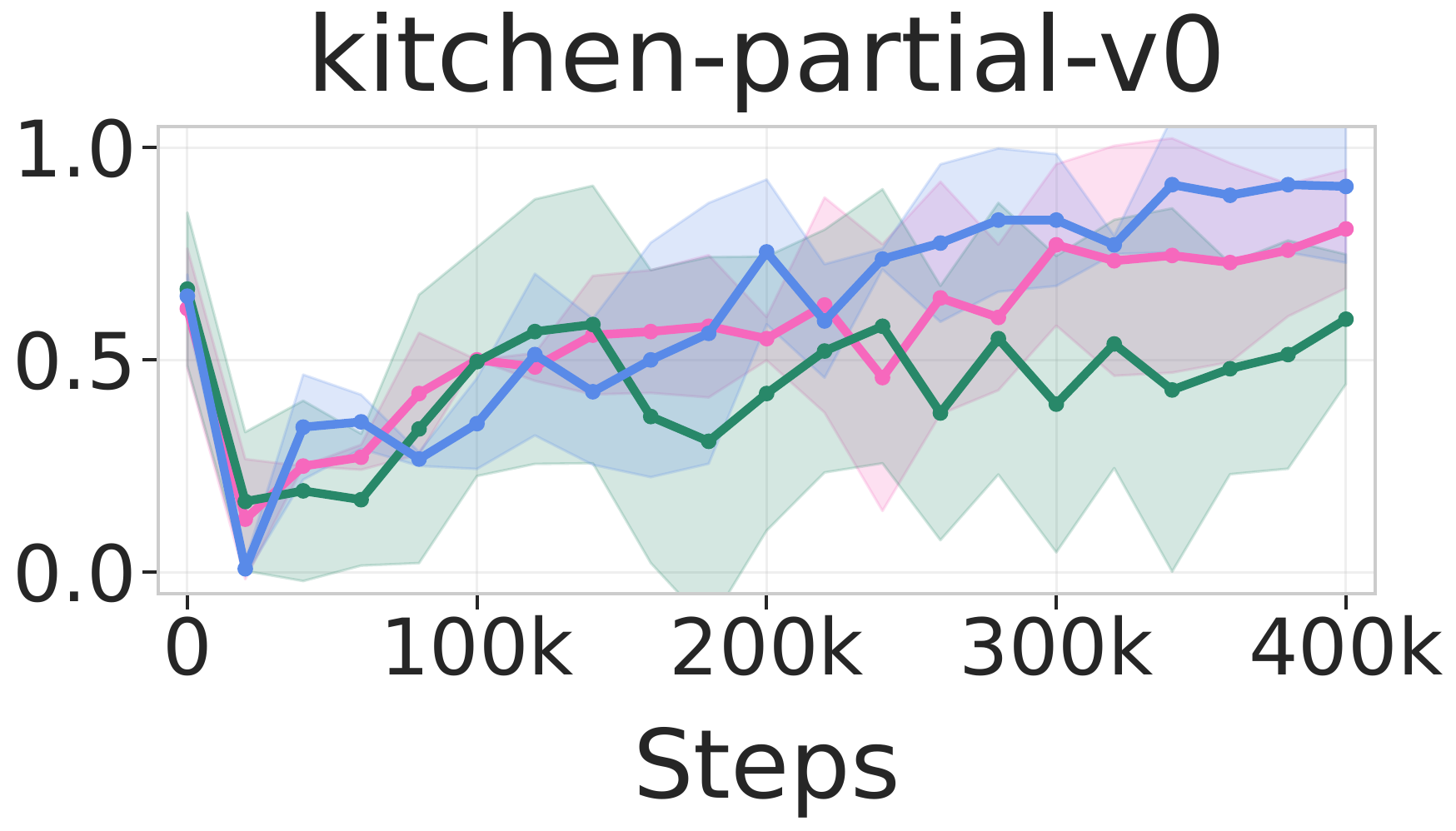}
    \end{subfigure}\hspace{0.1em}
    \begin{subfigure}[c]{0.21\textwidth}
        \includegraphics[width=1.0\linewidth]{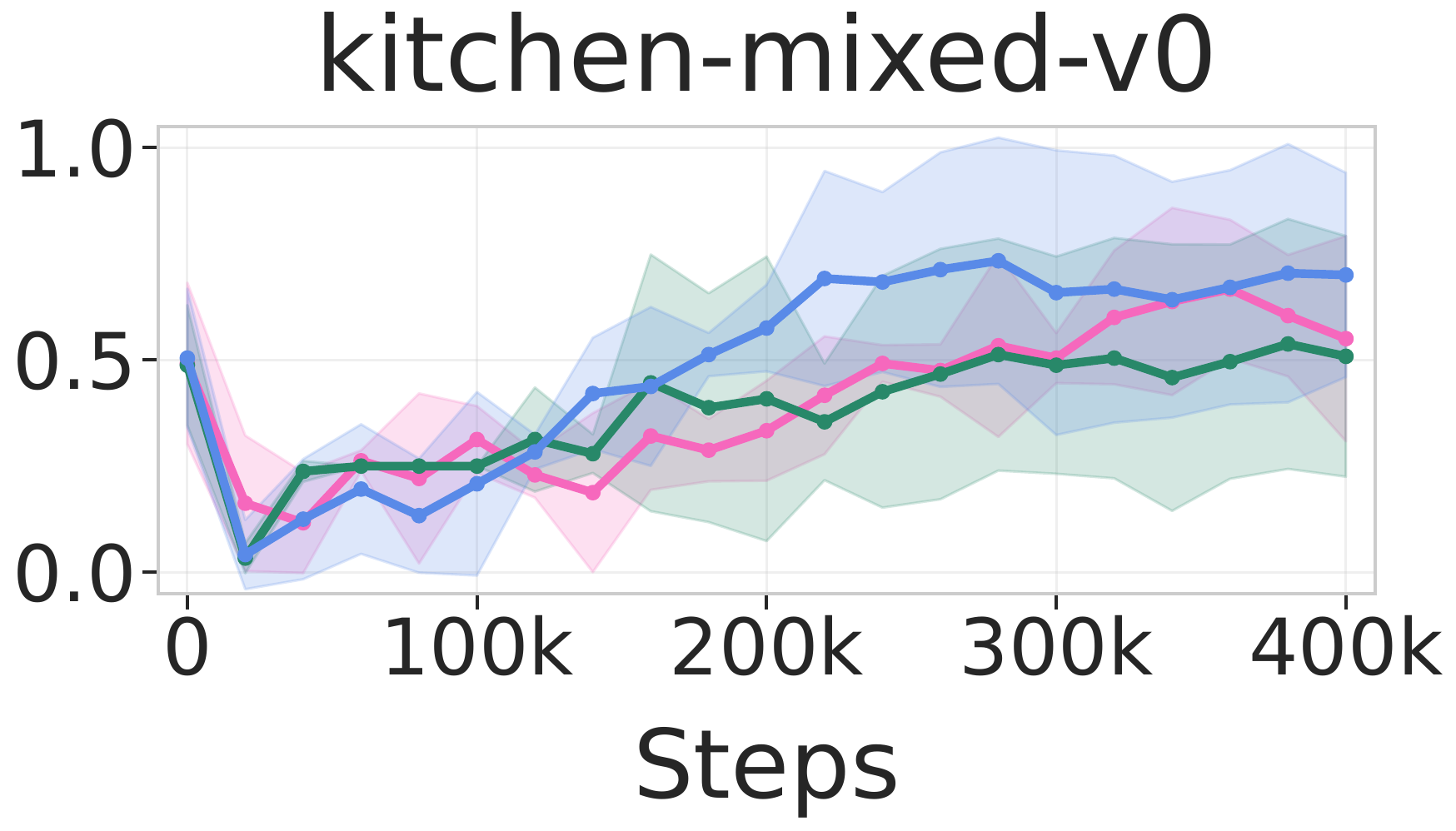}
    \end{subfigure}\hspace{0.1em}
    \begin{subfigure}[c]{0.21\textwidth}
        \includegraphics[width=1.0\linewidth]{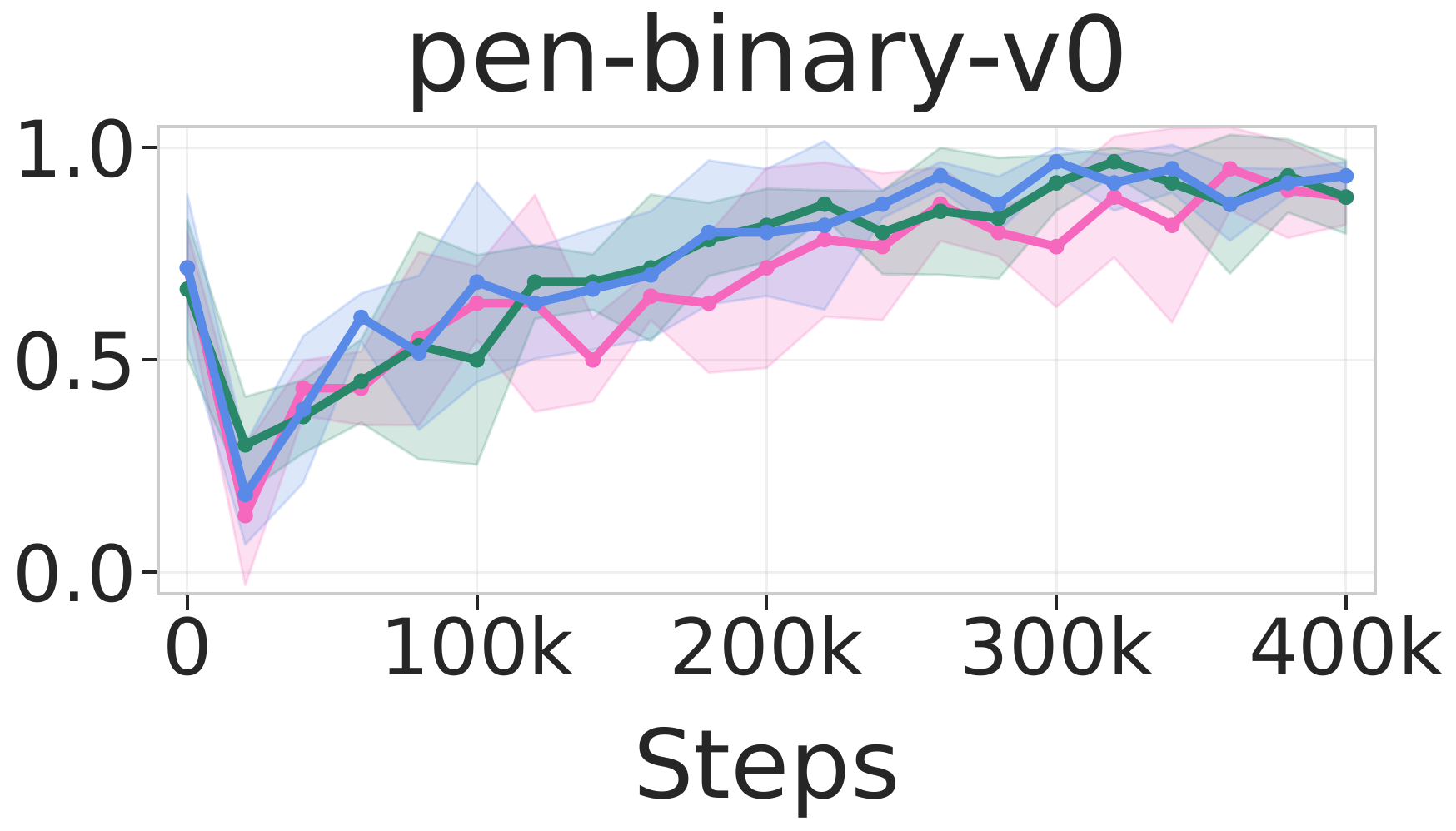}
    \end{subfigure}\\[1ex]
    \begin{subfigure}[c]{0.1\textwidth}
        \includegraphics[width=1.0\textwidth]{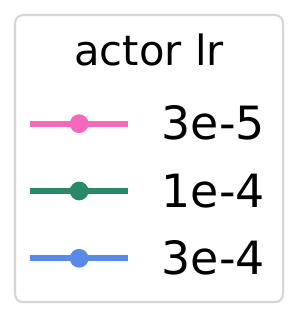}
    \end{subfigure}\hspace{0.1em}
    \begin{subfigure}[c]{0.21\textwidth}
        \includegraphics[width=1.0\linewidth]{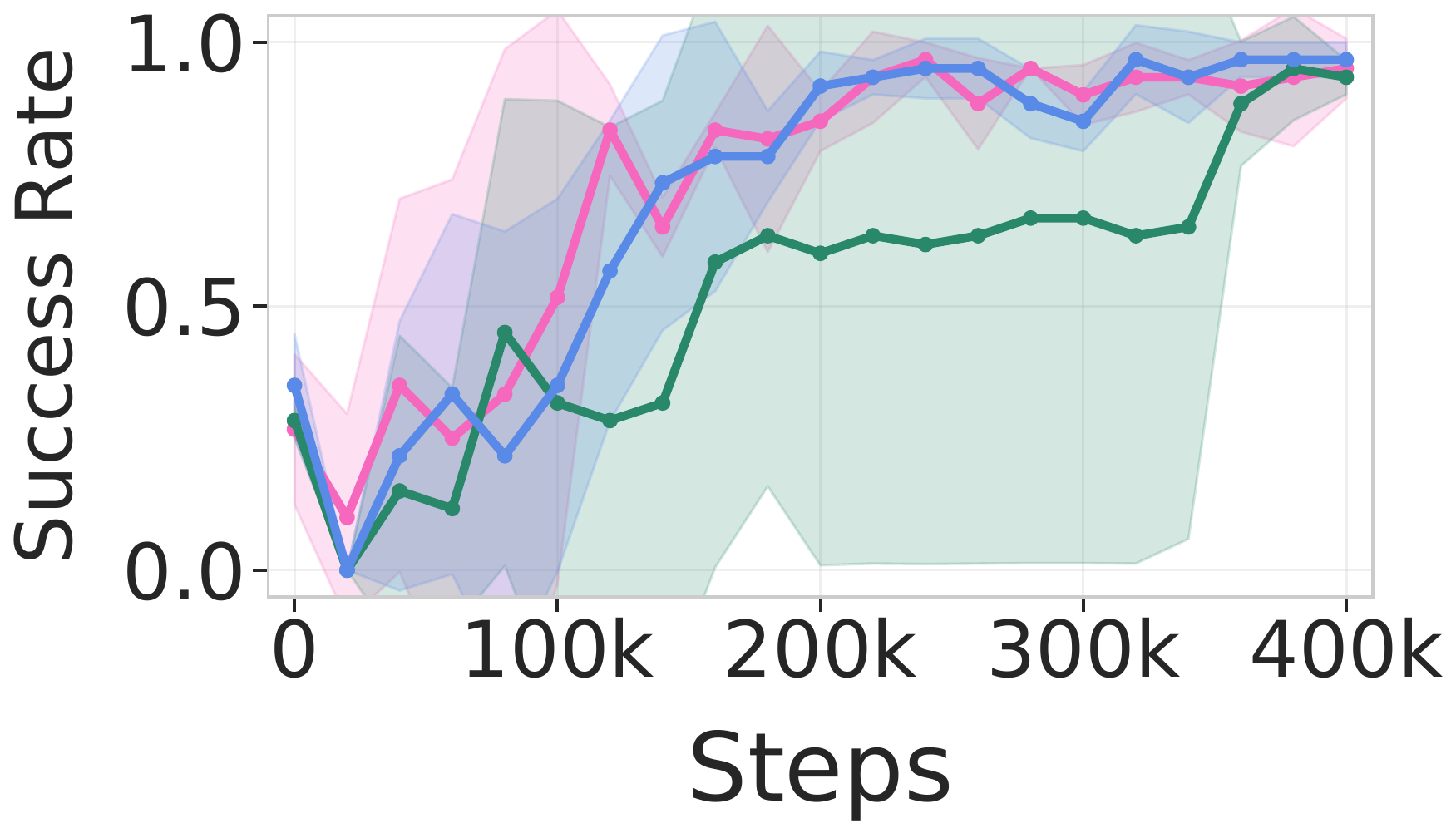}
    \end{subfigure}\hspace{0.1em}
    \begin{subfigure}[c]{0.21\textwidth}
        \includegraphics[width=1.0\linewidth]{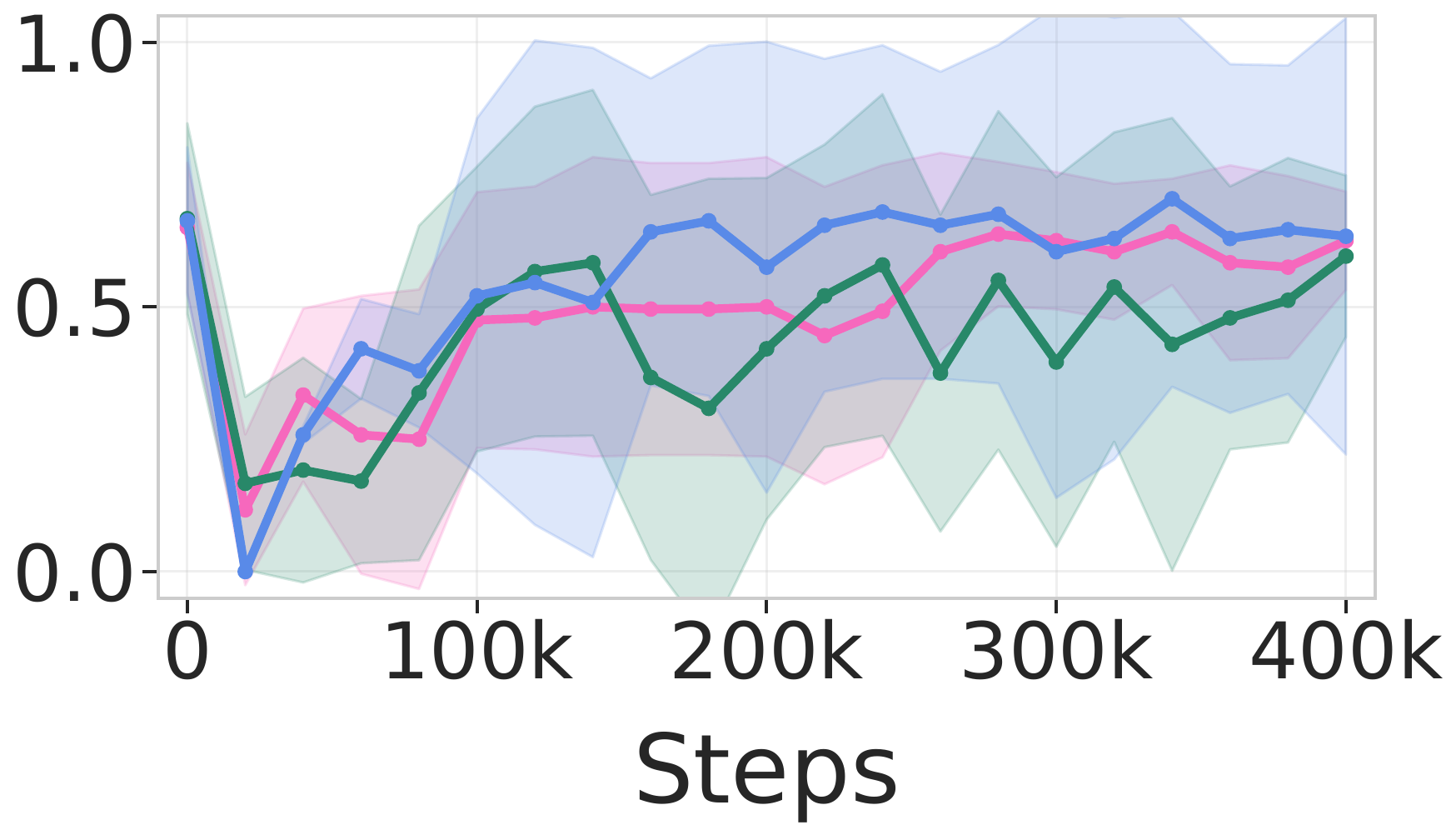}
    \end{subfigure}\hspace{0.1em}
    \begin{subfigure}[c]{0.21\textwidth}
        \includegraphics[width=1.0\linewidth]{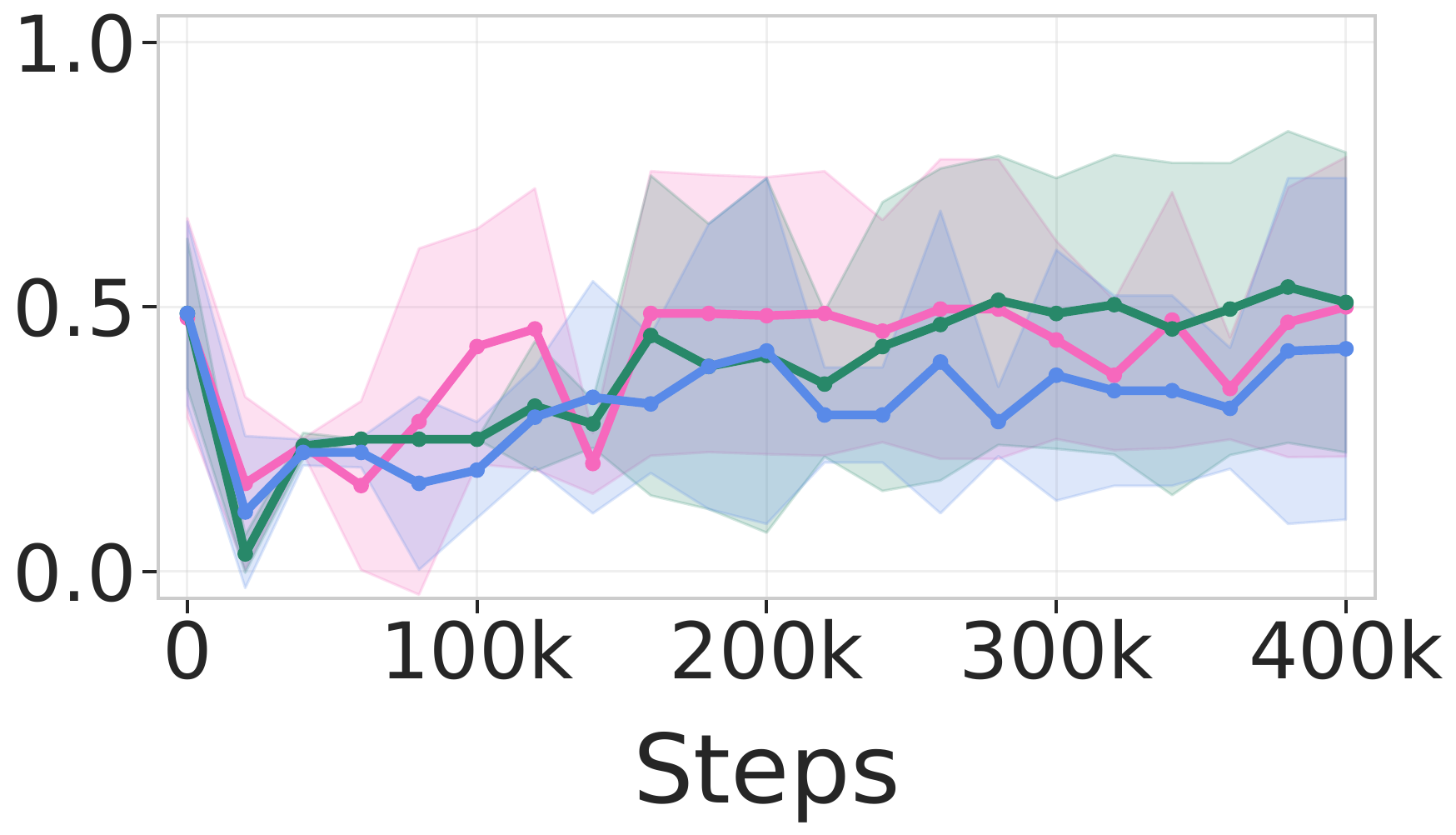}
    \end{subfigure}\hspace{0.1em}
    \begin{subfigure}[c]{0.21\textwidth}
        \includegraphics[width=1.0\linewidth]{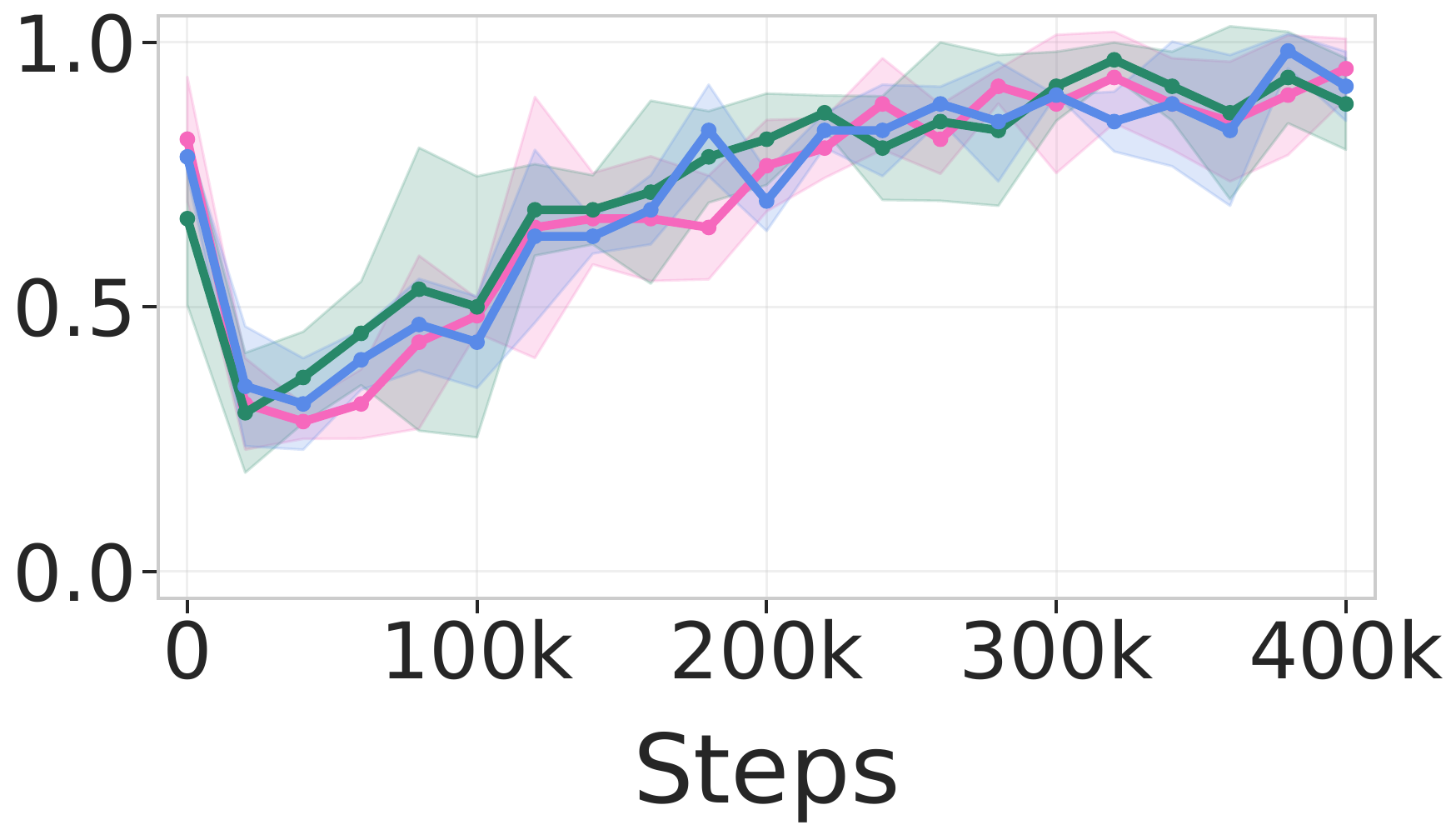}
    \end{subfigure}\\[1ex]
    \begin{subfigure}[c]{0.1\textwidth}
        \includegraphics[width=1.0\textwidth]{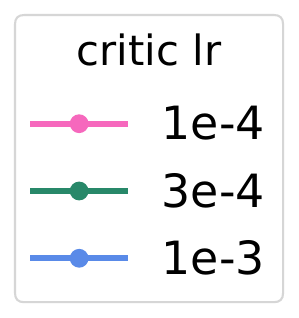}
    \end{subfigure}\hspace{0.1em}
    \begin{subfigure}[c]{0.21\textwidth}
        \includegraphics[width=1.0\linewidth]{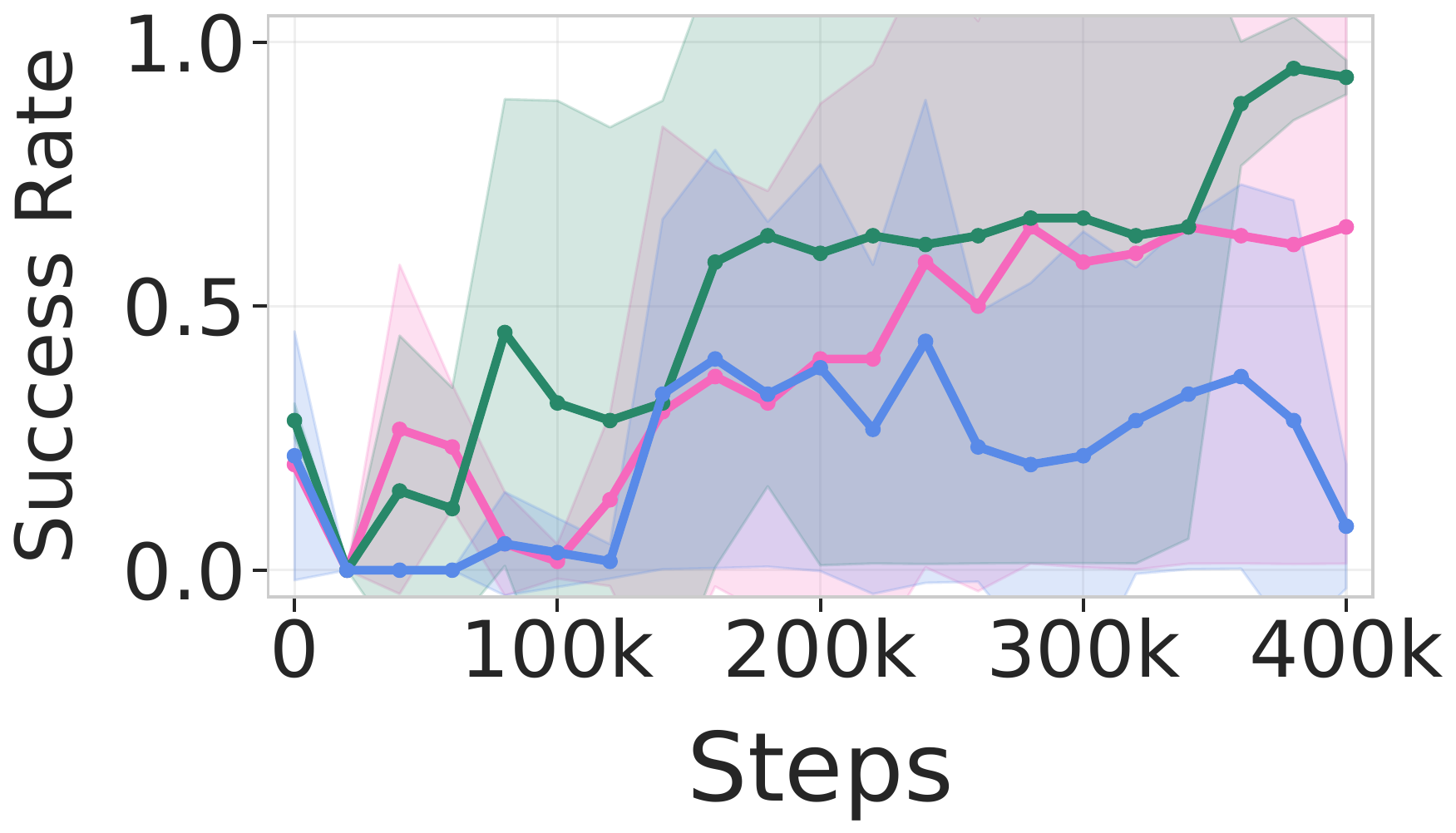}
    \end{subfigure}\hspace{0.1em}
    \begin{subfigure}[c]{0.21\textwidth}
        \includegraphics[width=1.0\linewidth]{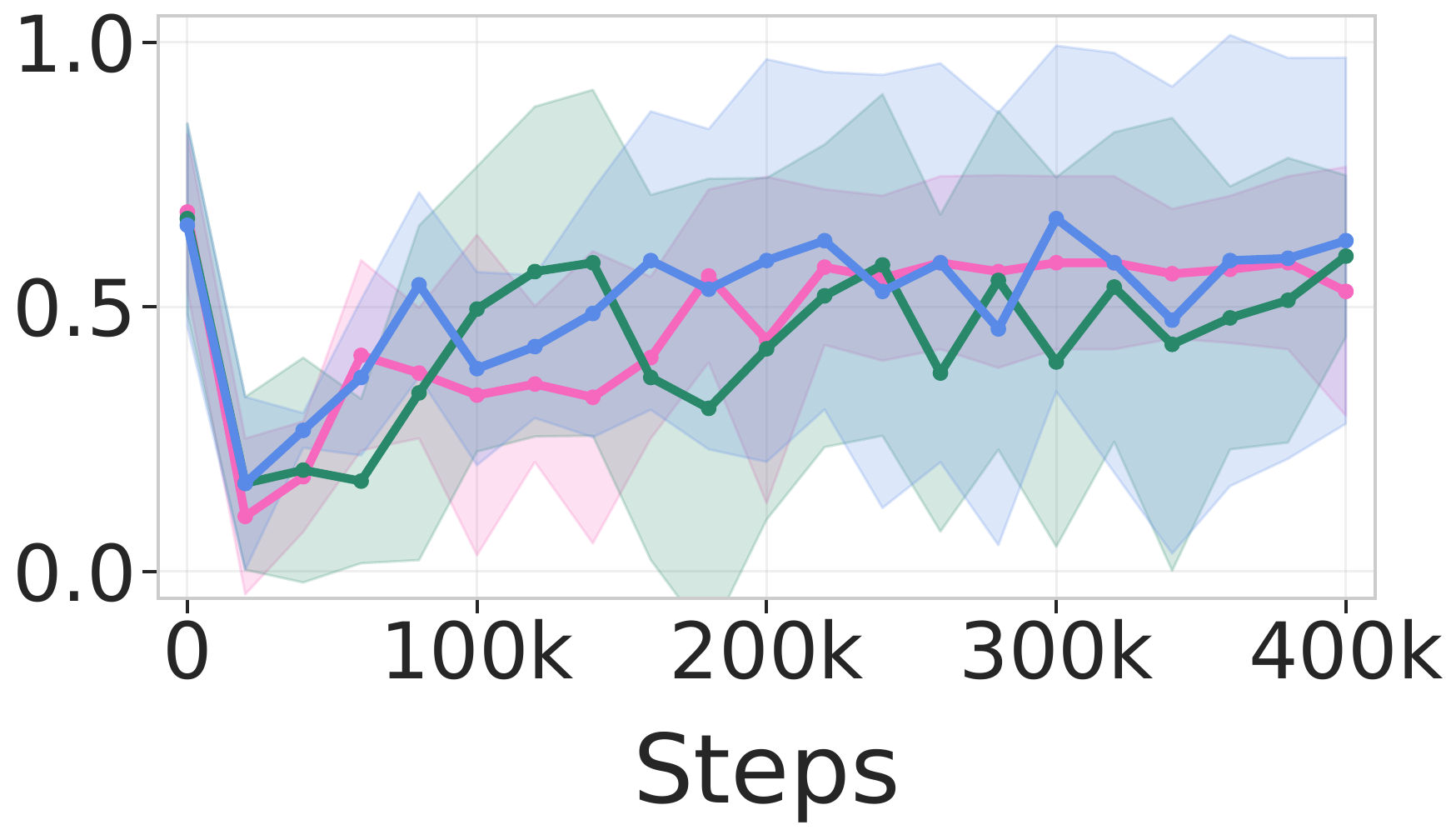}
    \end{subfigure}\hspace{0.1em}
    \begin{subfigure}[c]{0.21\textwidth}
        \includegraphics[width=1.0\linewidth]{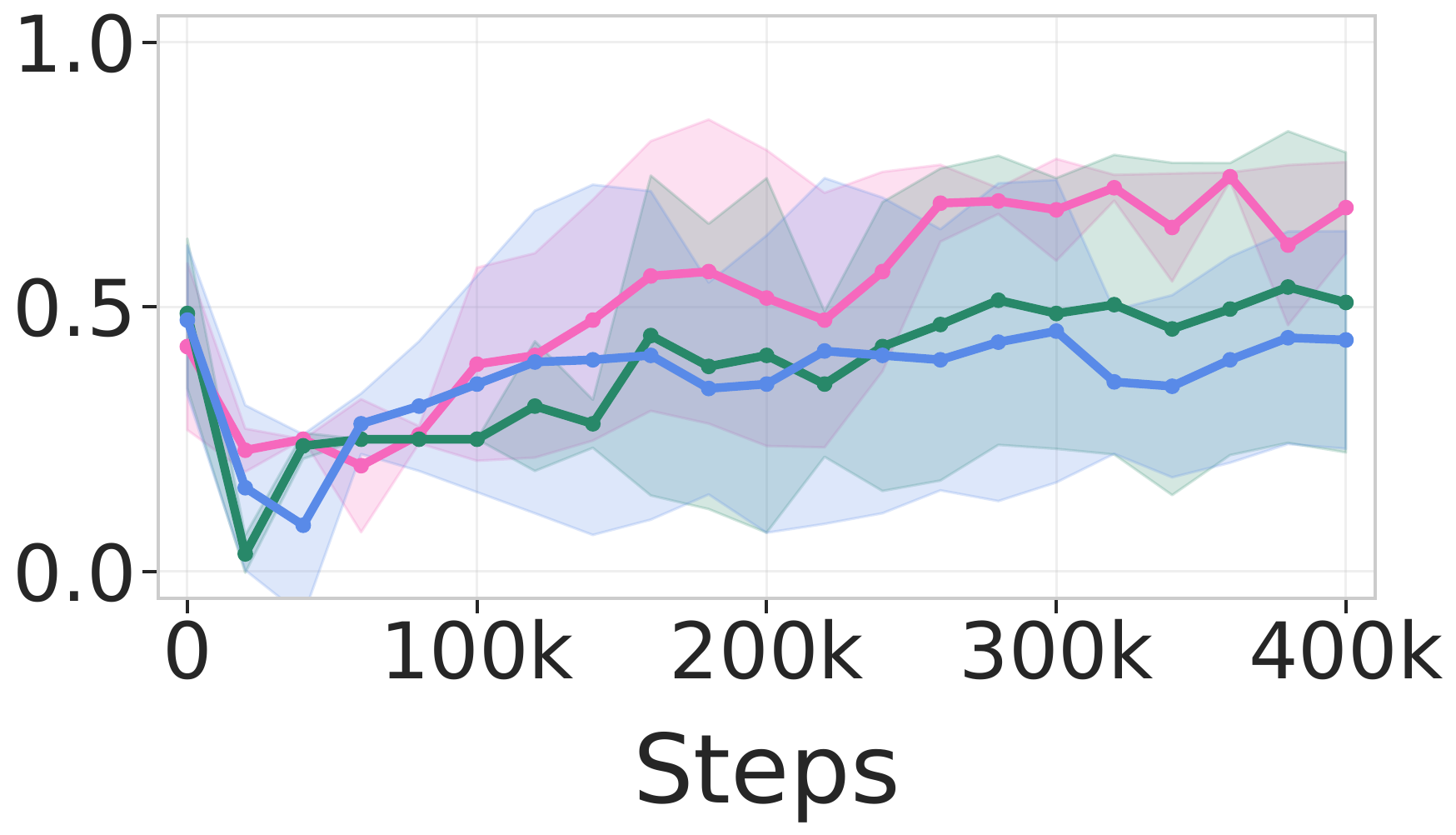}
    \end{subfigure}\hspace{0.1em}
    \begin{subfigure}[c]{0.21\textwidth}
        \includegraphics[width=1.0\linewidth]{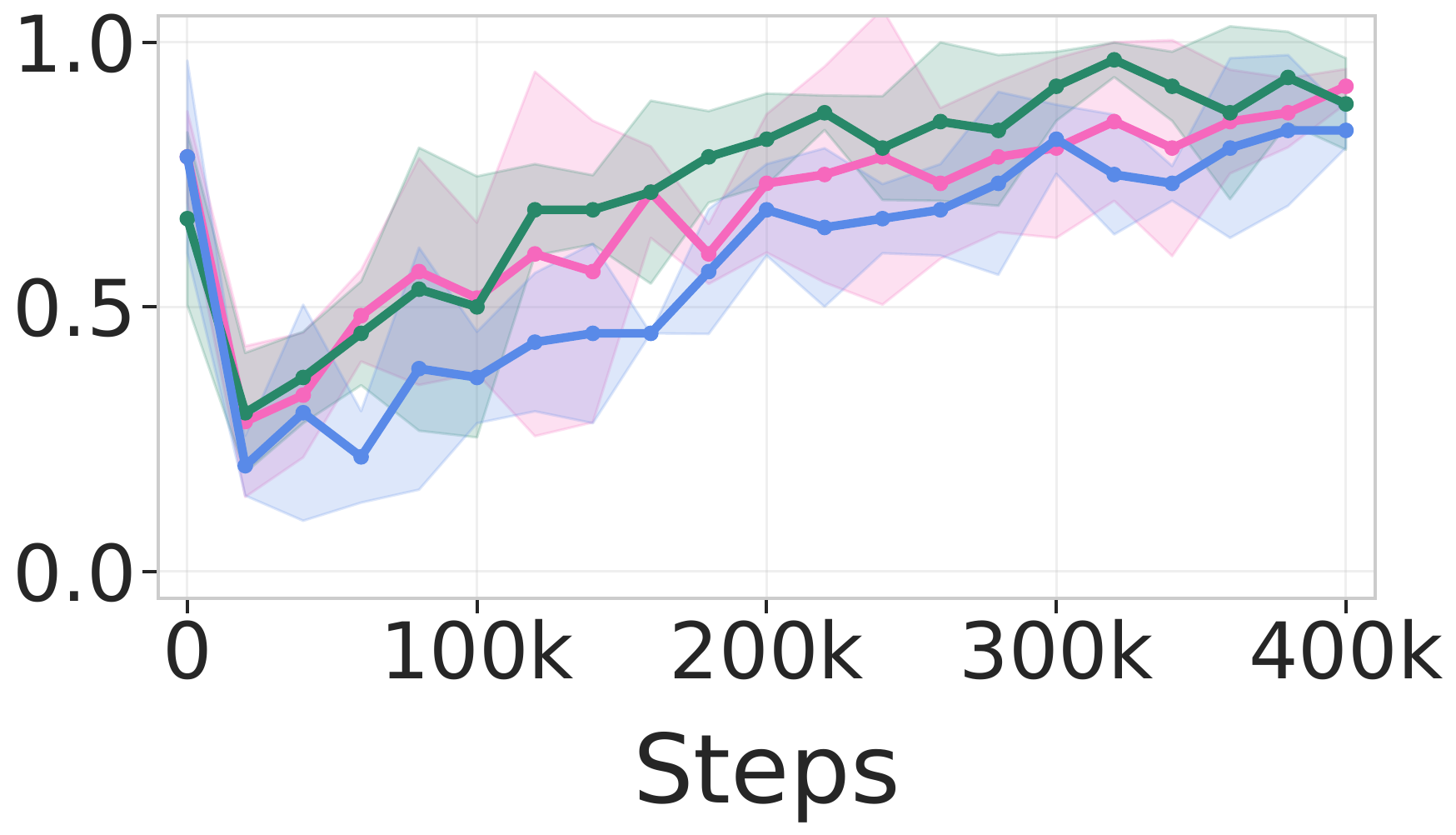}
    \end{subfigure}\\[1ex]
    \begin{subfigure}[c]{0.1\textwidth}
        \includegraphics[width=1.0\textwidth]{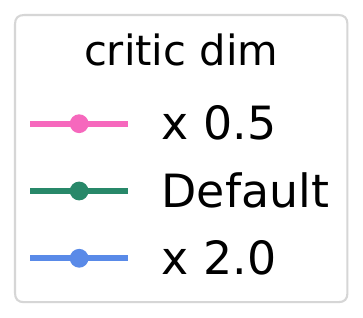}
    \end{subfigure}\hspace{0.1em}
    \begin{subfigure}[c]{0.21\textwidth}
        \includegraphics[width=1.0\linewidth]{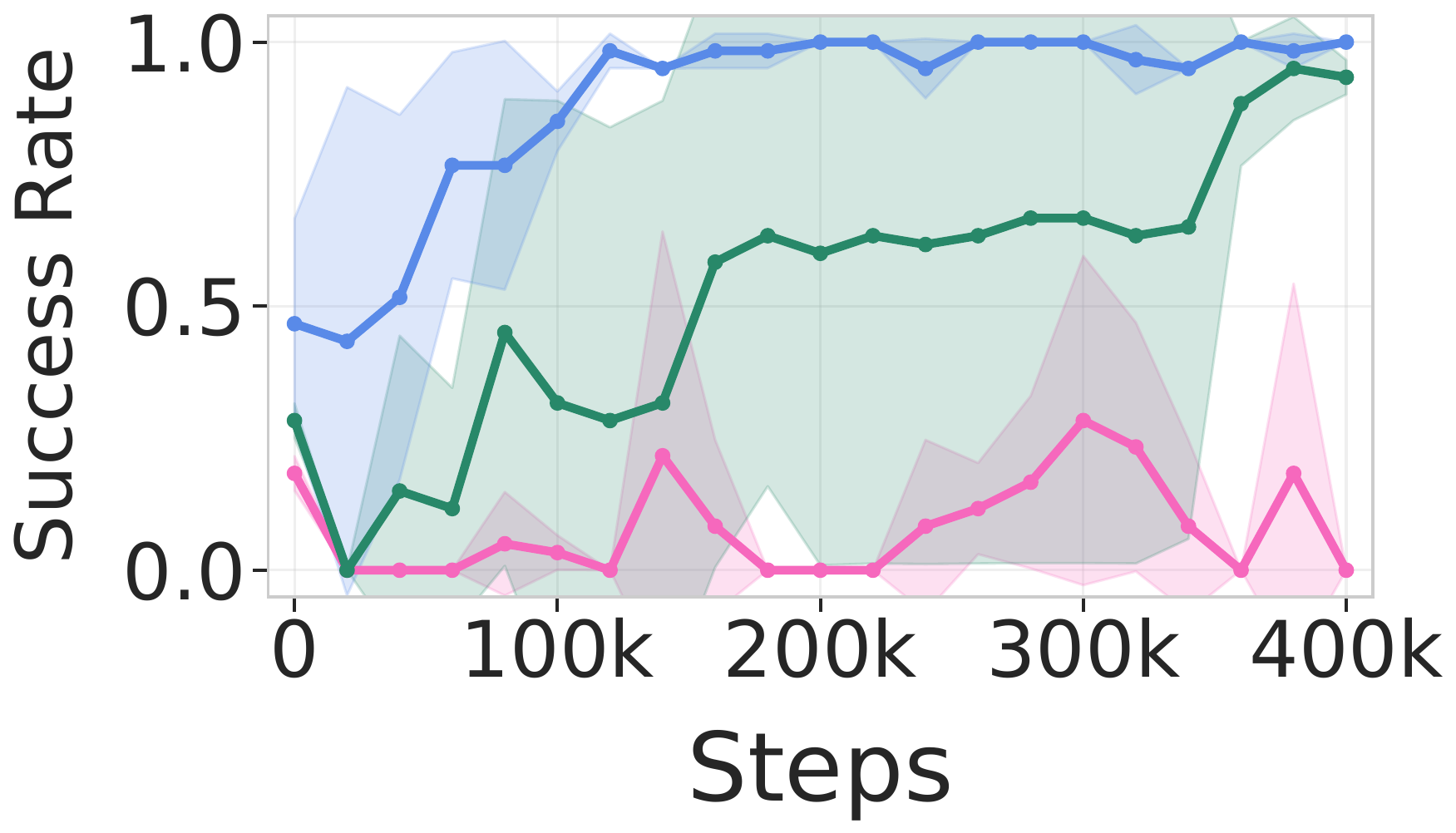}
    \end{subfigure}\hspace{0.1em}
    \begin{subfigure}[c]{0.21\textwidth}
        \includegraphics[width=1.0\linewidth]{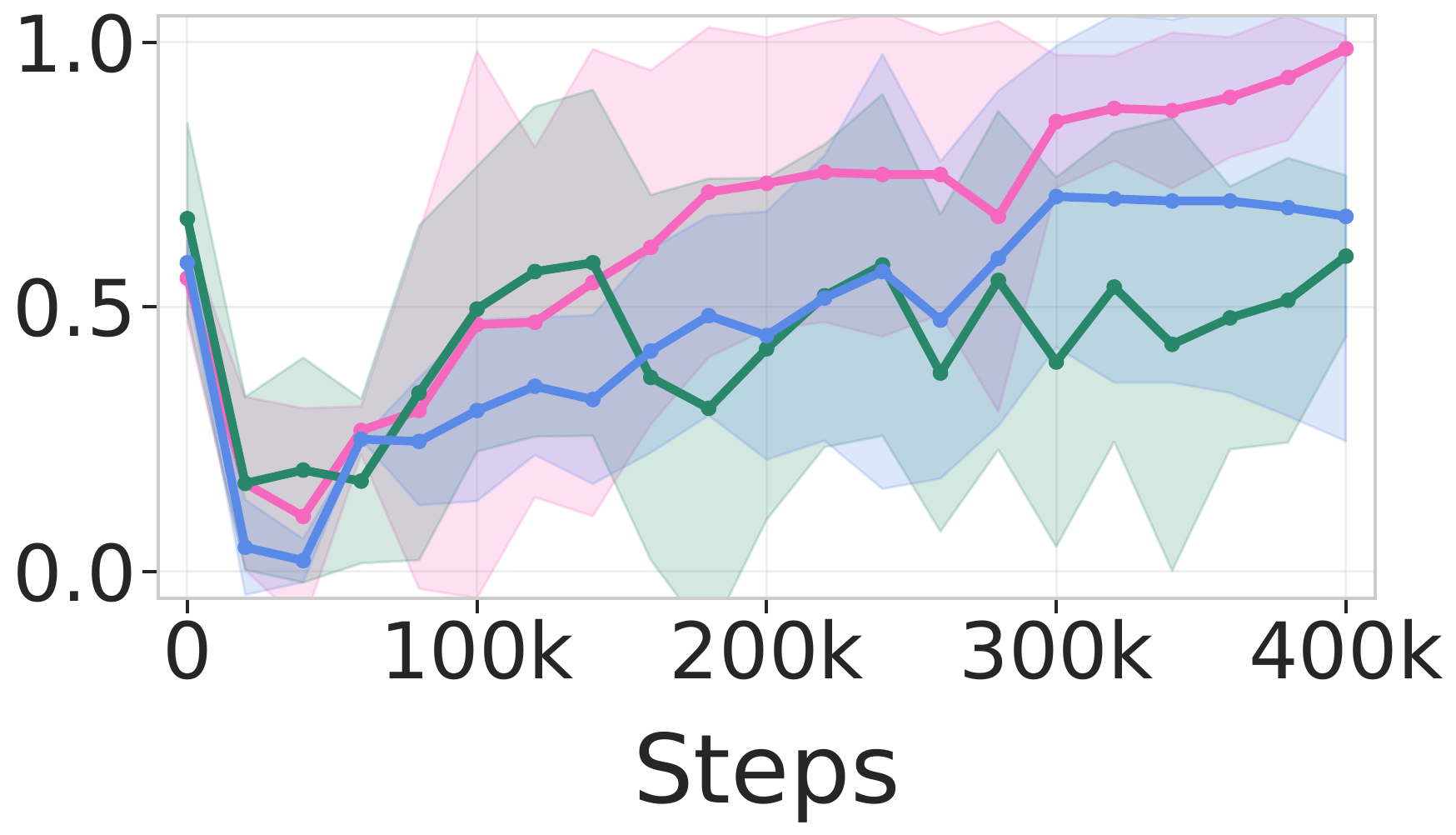}
    \end{subfigure}\hspace{0.1em}
    \begin{subfigure}[c]{0.21\textwidth}
        \includegraphics[width=1.0\linewidth]{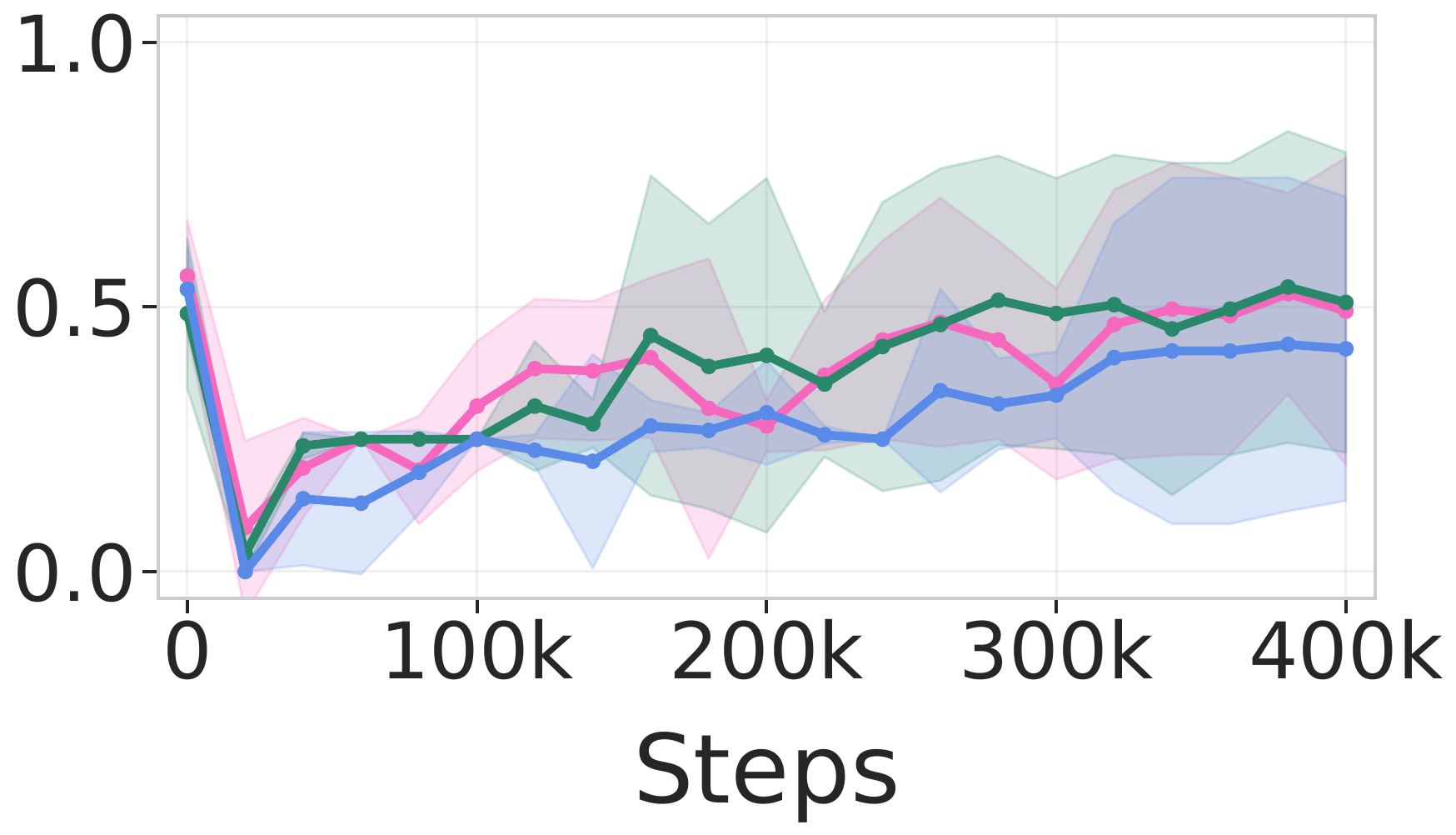}
    \end{subfigure}\hspace{0.1em}
    \begin{subfigure}[c]{0.21\textwidth}
        \includegraphics[width=1.0\linewidth]{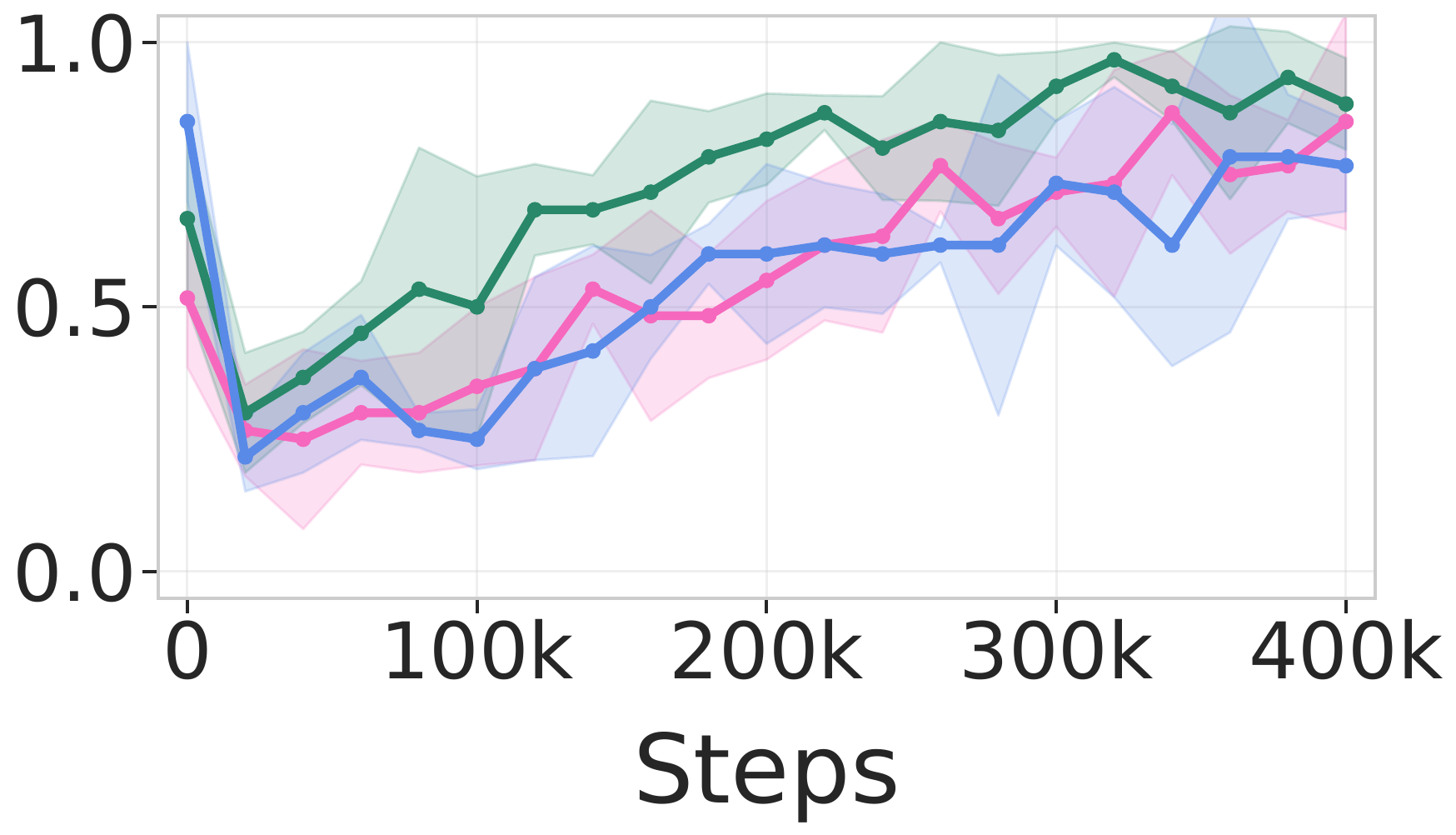}
    \end{subfigure}\\[1ex]
    \begin{subfigure}[c]{0.1\textwidth}
        \includegraphics[width=1.0\textwidth]{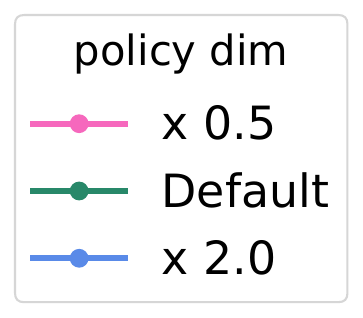}
    \end{subfigure}\hspace{0.1em}
    \begin{subfigure}[c]{0.21\textwidth}
        \includegraphics[width=1.0\linewidth]{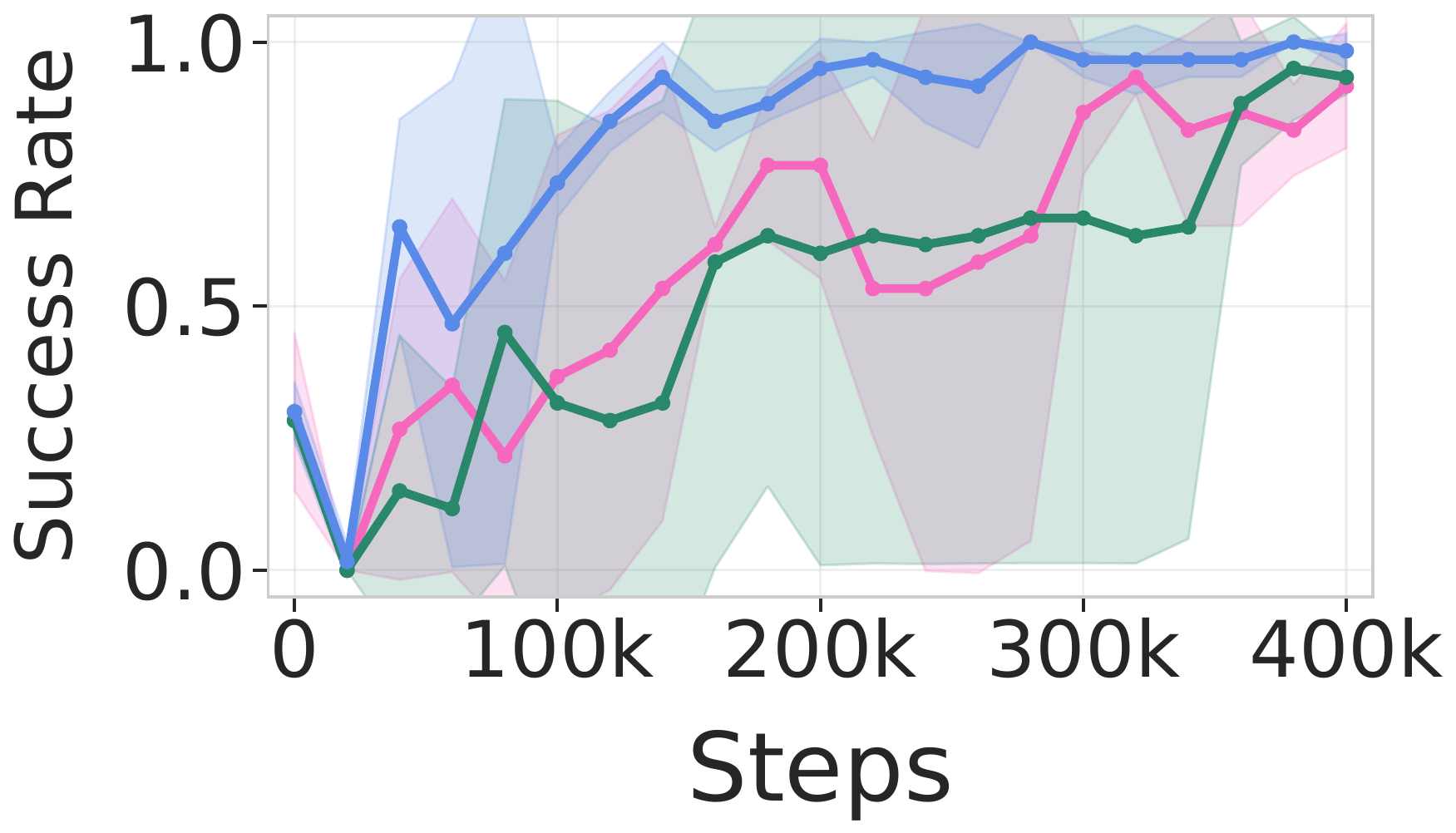}
    \end{subfigure}\hspace{0.1em}
    \begin{subfigure}[c]{0.21\textwidth}
        \includegraphics[width=1.0\linewidth]{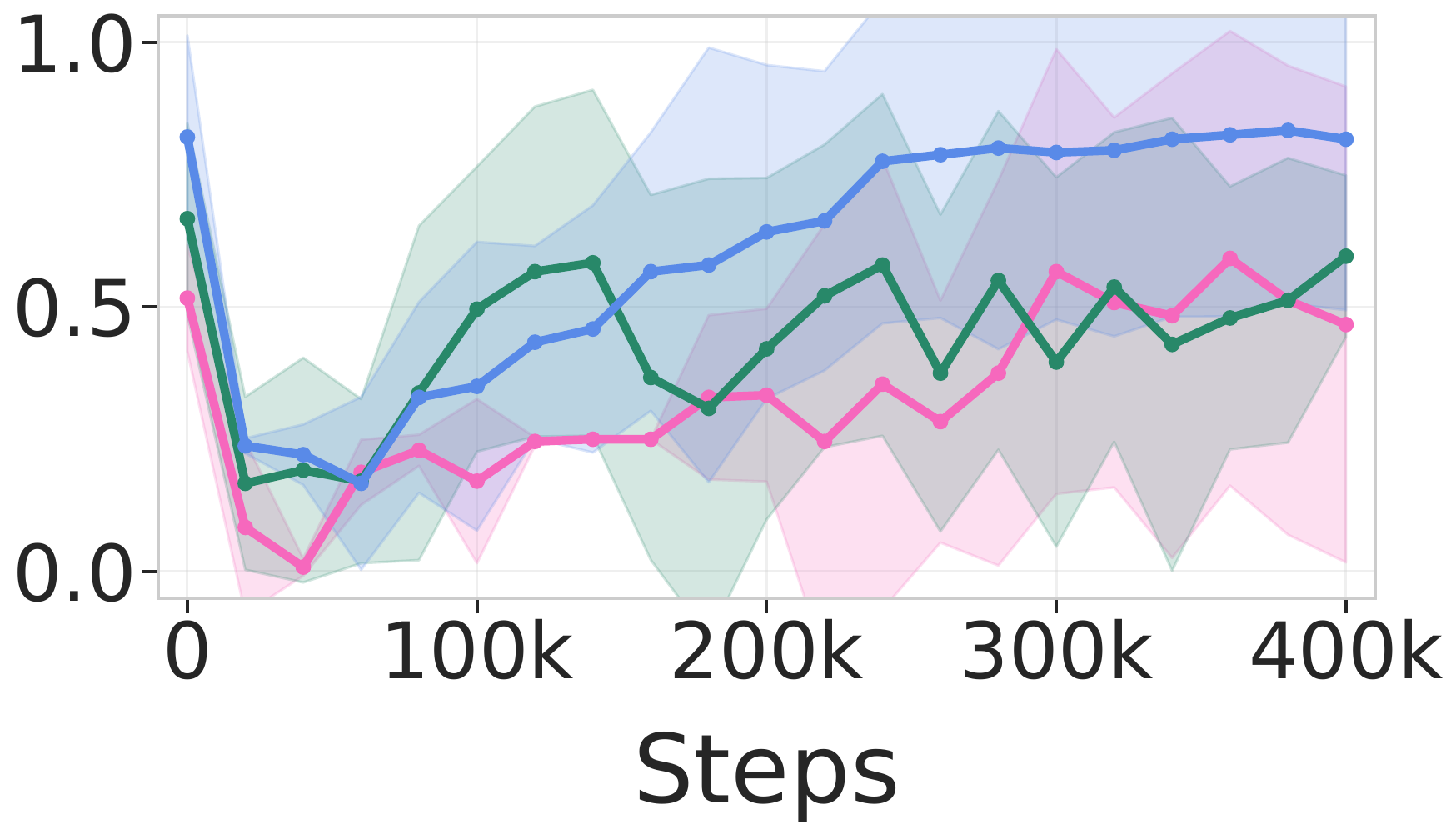}
    \end{subfigure}\hspace{0.1em}
    \begin{subfigure}[c]{0.21\textwidth}
        \includegraphics[width=1.0\linewidth]{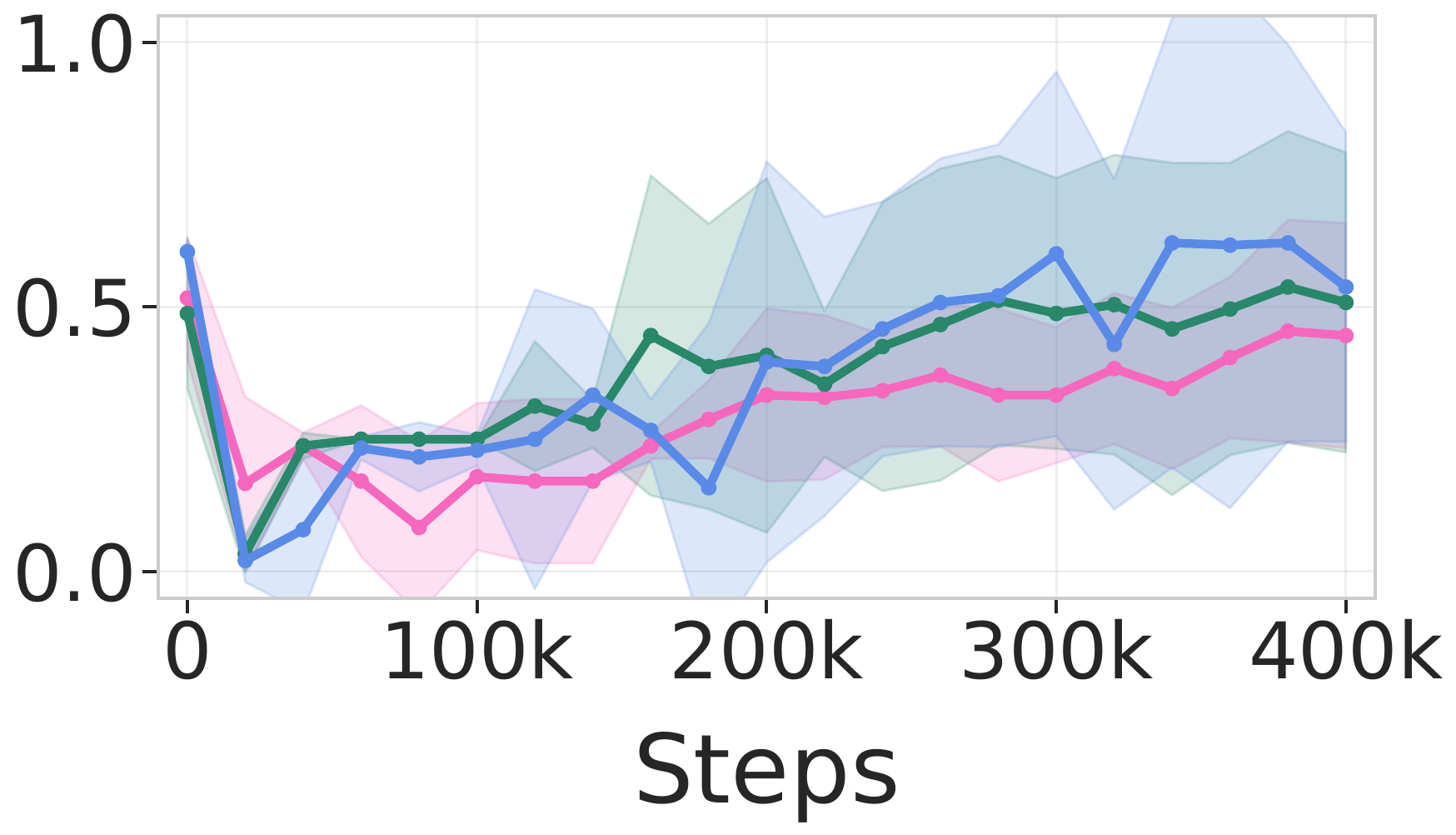}
    \end{subfigure}\hspace{0.1em}
    \begin{subfigure}[c]{0.21\textwidth}
        \includegraphics[width=1.0\linewidth]{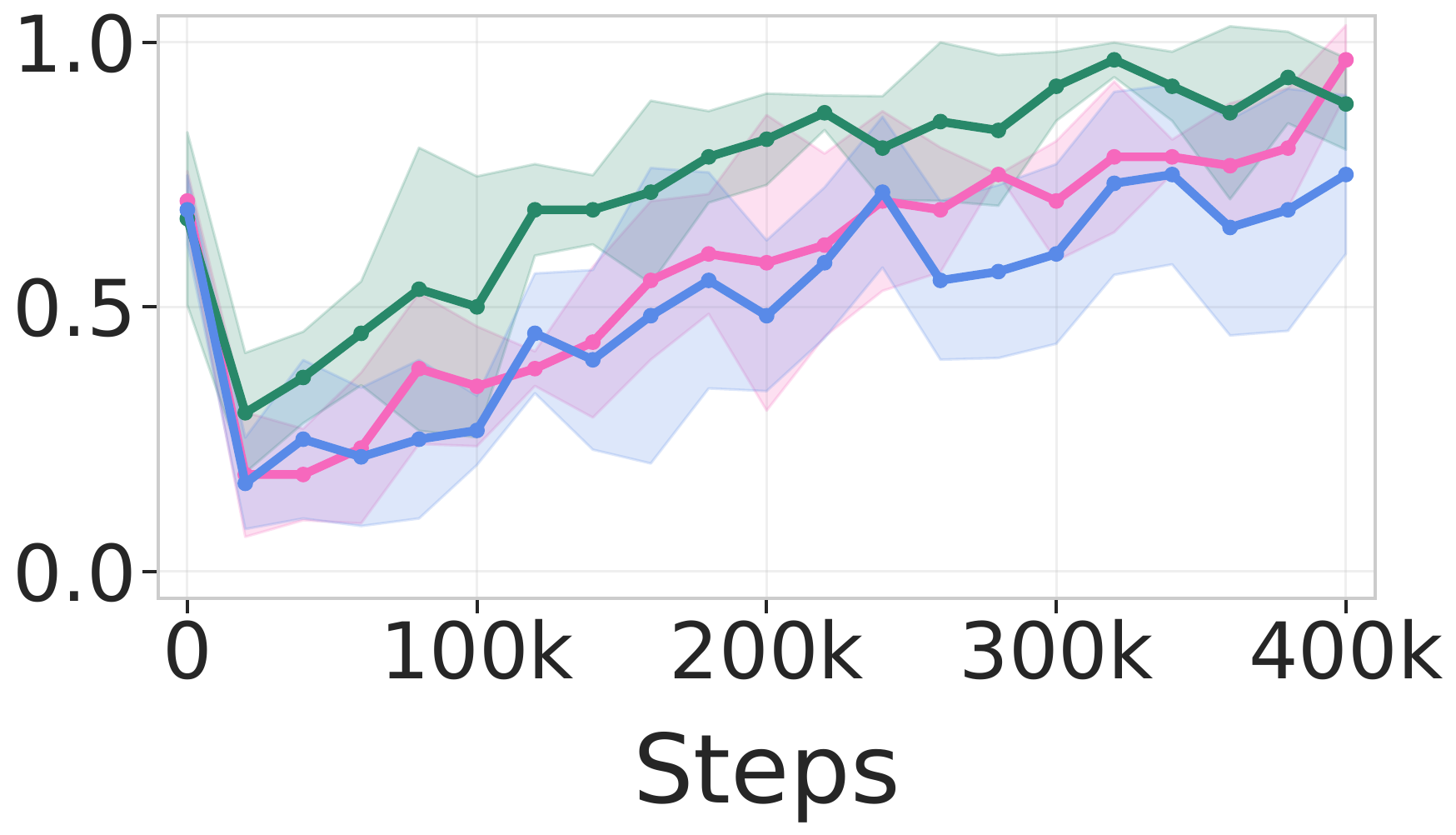}
    \end{subfigure}\\[1ex]
    \begin{subfigure}[c]{0.1\textwidth}
        \includegraphics[width=1.0\textwidth]{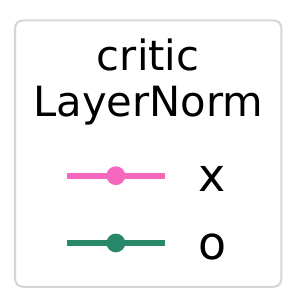}
    \end{subfigure}\hspace{0.1em}
    \begin{subfigure}[c]{0.21\textwidth}
        \includegraphics[width=1.0\linewidth]{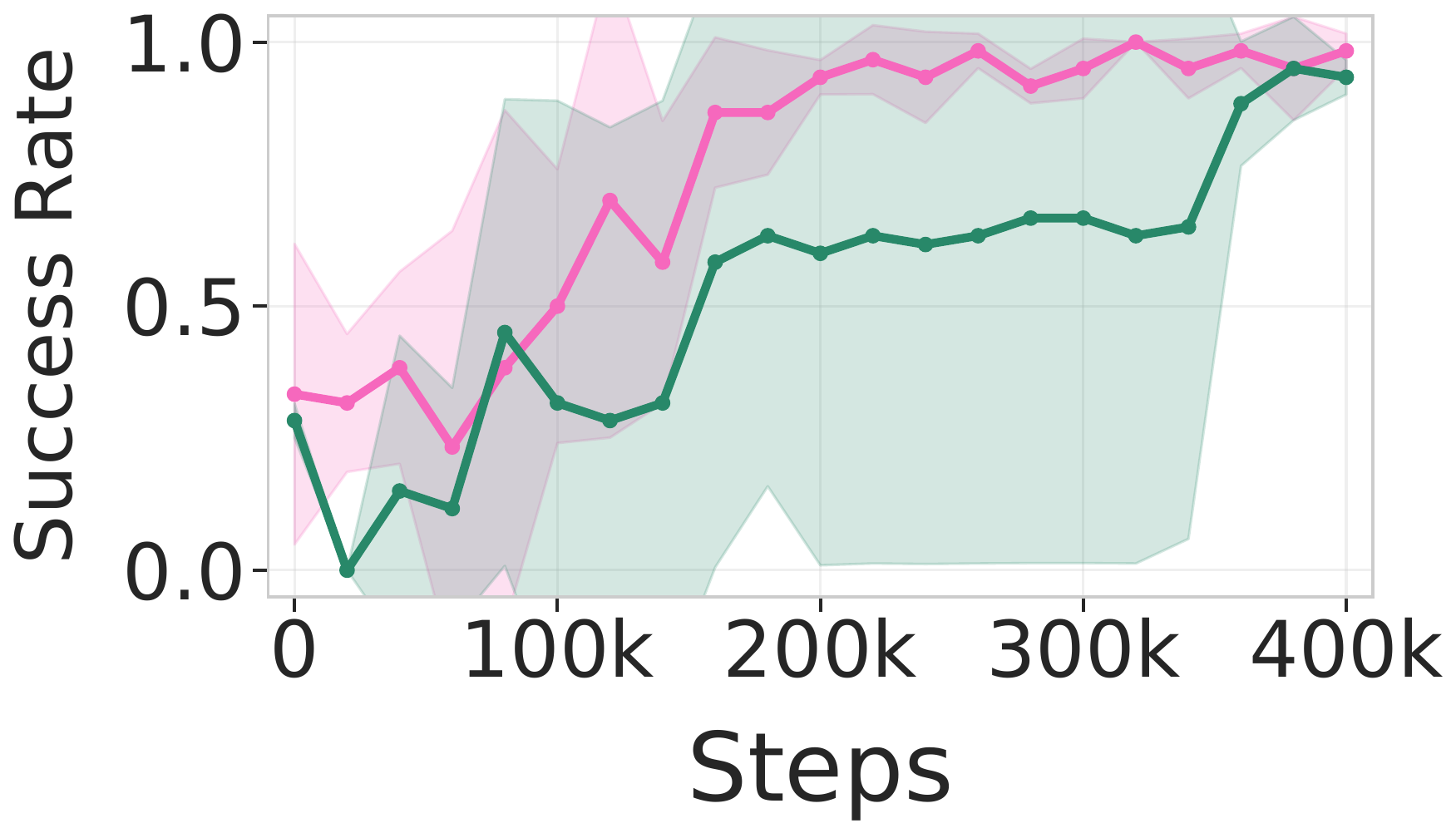}
    \end{subfigure}\hspace{0.1em}
    \begin{subfigure}[c]{0.21\textwidth}
        \includegraphics[width=1.0\linewidth]{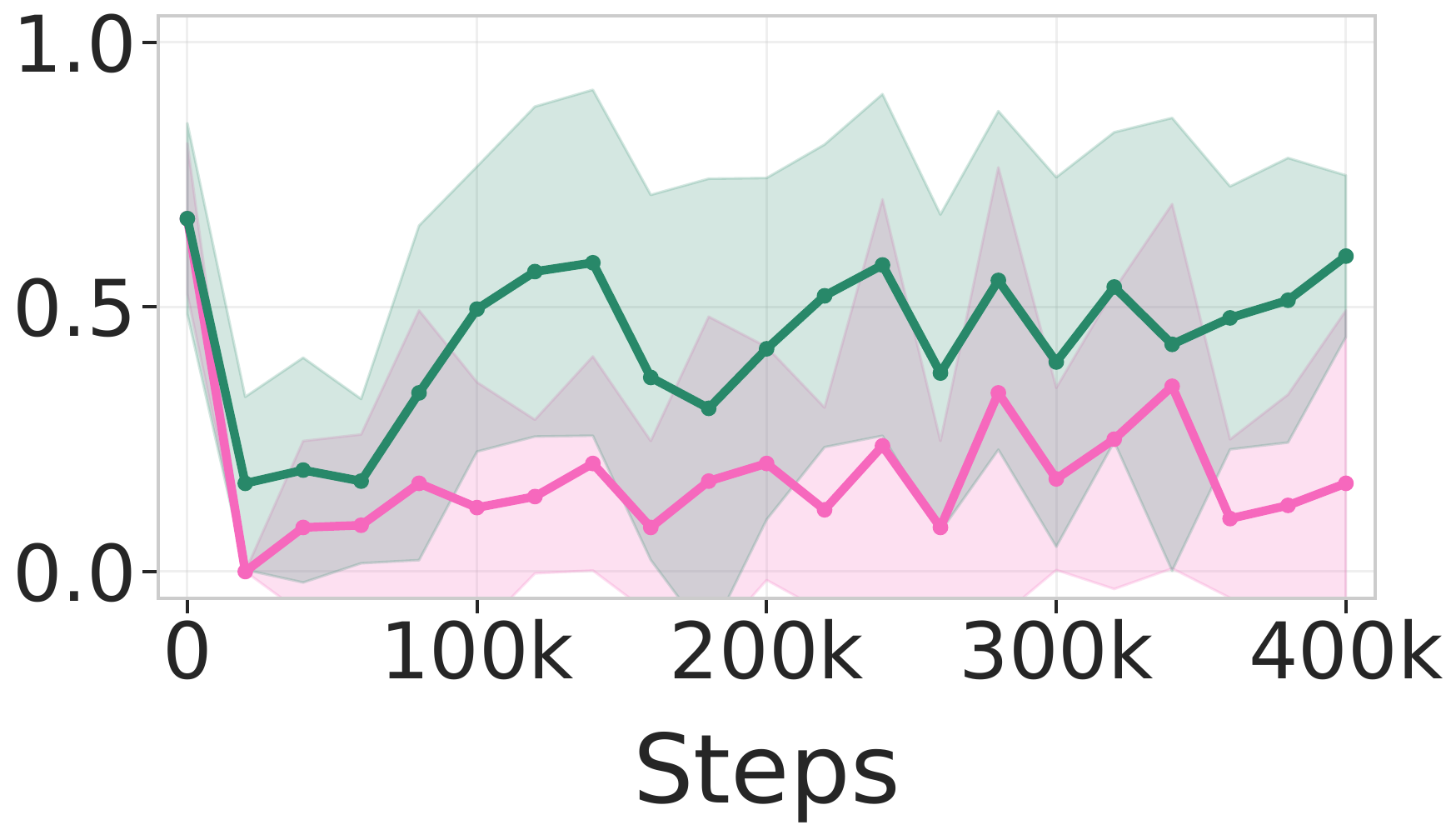}
    \end{subfigure}\hspace{0.1em}
    \begin{subfigure}[c]{0.21\textwidth}
        \includegraphics[width=1.0\linewidth]{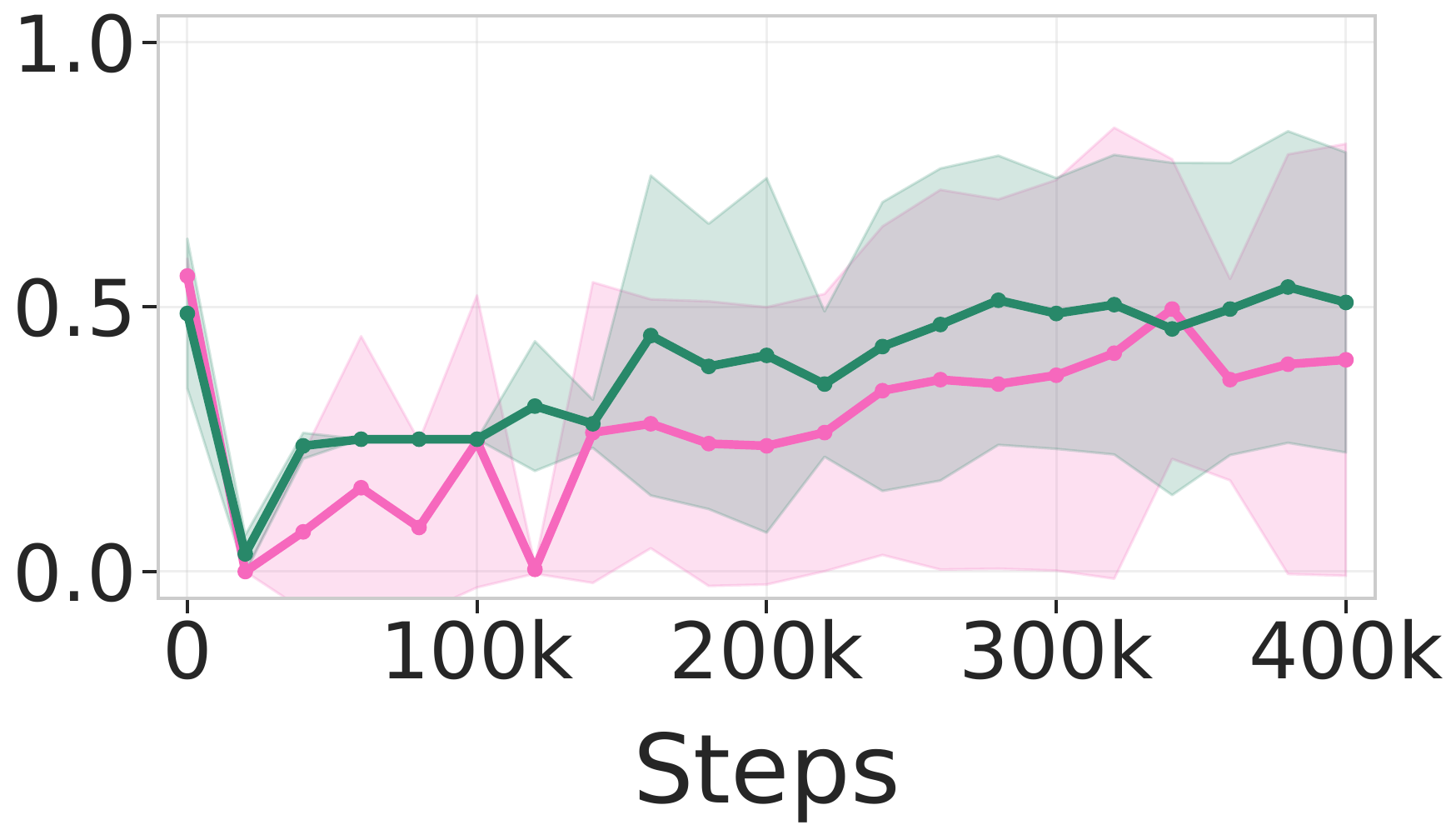}
    \end{subfigure}\hspace{0.1em}
    \begin{subfigure}[c]{0.21\textwidth}
        \includegraphics[width=1.0\linewidth]{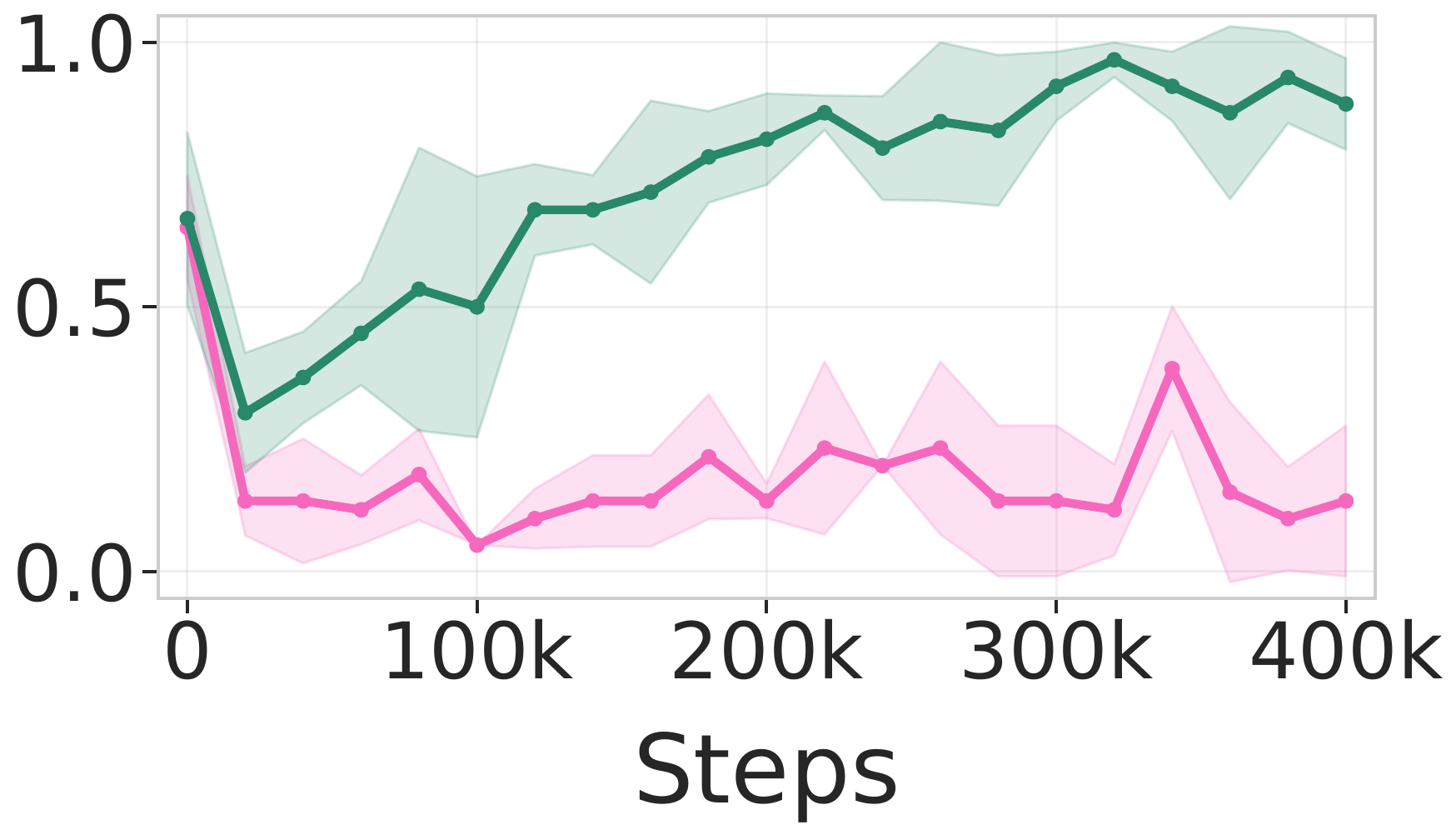}
    \end{subfigure}\\[1ex]
    \begin{subfigure}[c]{0.1\textwidth}
        \includegraphics[width=1.0\textwidth]{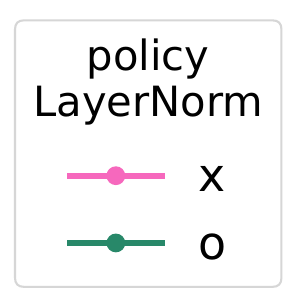}
    \end{subfigure}\hspace{0.1em}
    \begin{subfigure}[c]{0.21\textwidth}
        \includegraphics[width=1.0\linewidth]{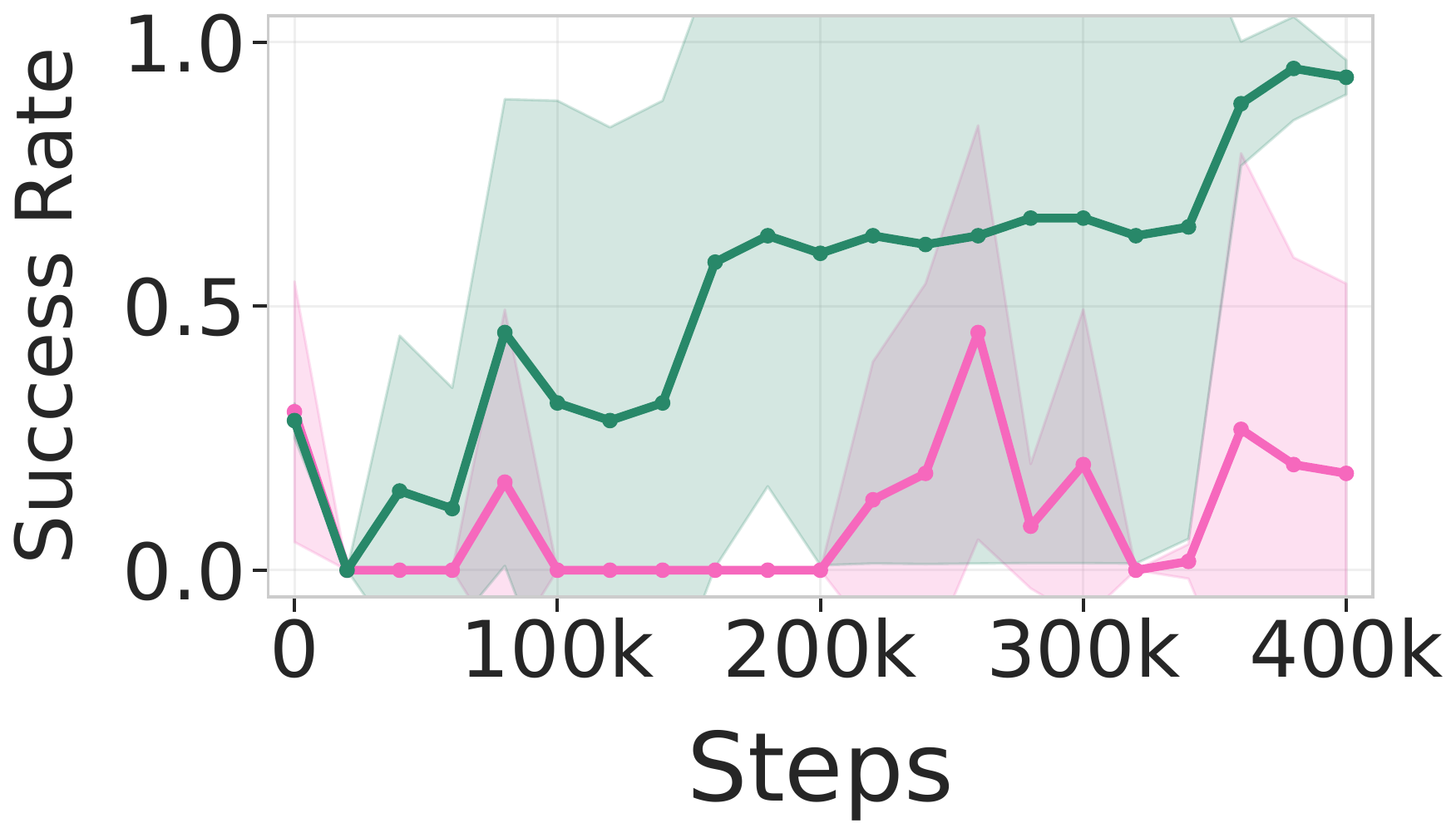}
    \end{subfigure}\hspace{0.1em}
    \begin{subfigure}[c]{0.21\textwidth}
        \includegraphics[width=1.0\linewidth]{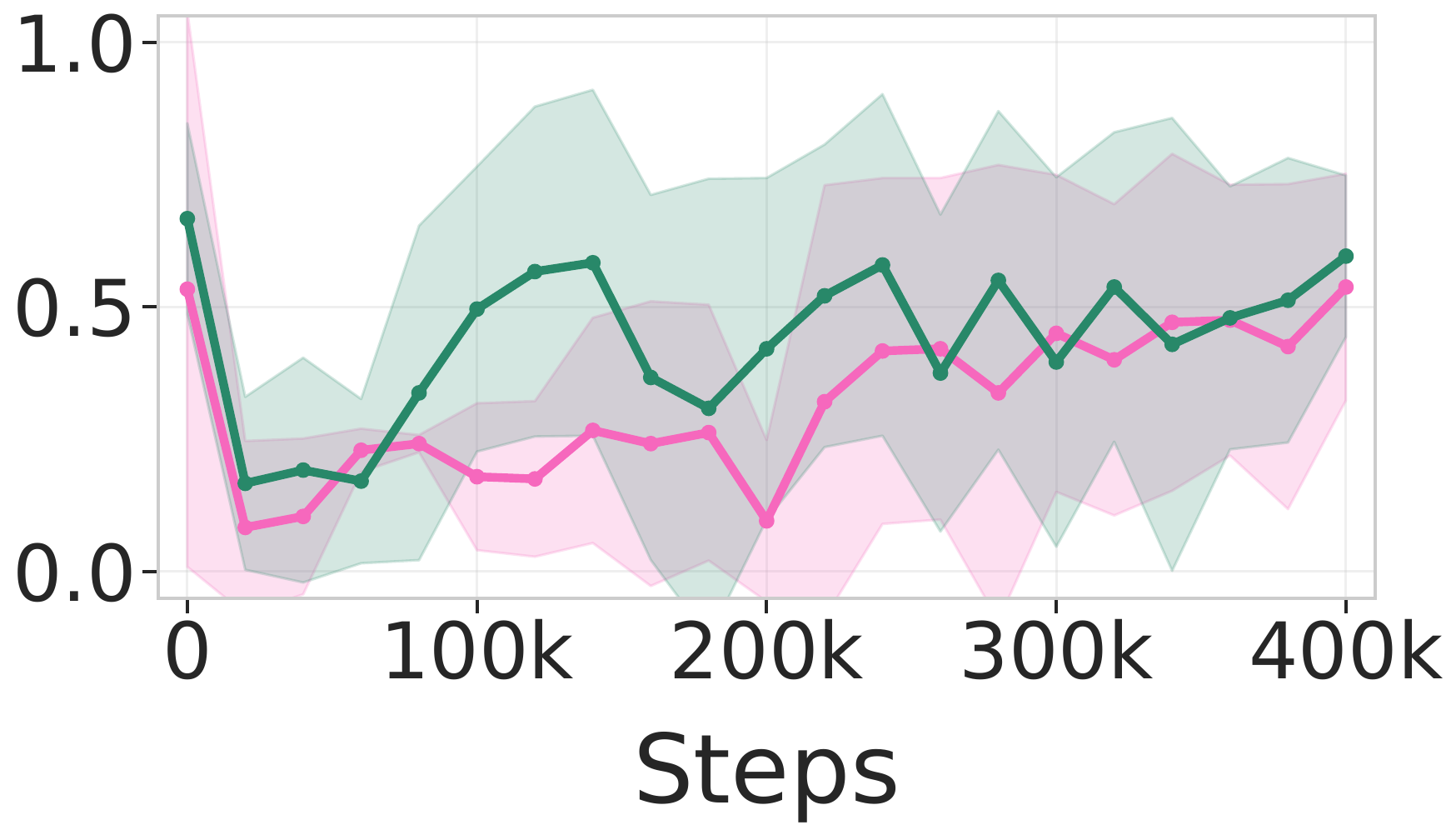}
    \end{subfigure}\hspace{0.1em}
    \begin{subfigure}[c]{0.21\textwidth}
        \includegraphics[width=1.0\linewidth]{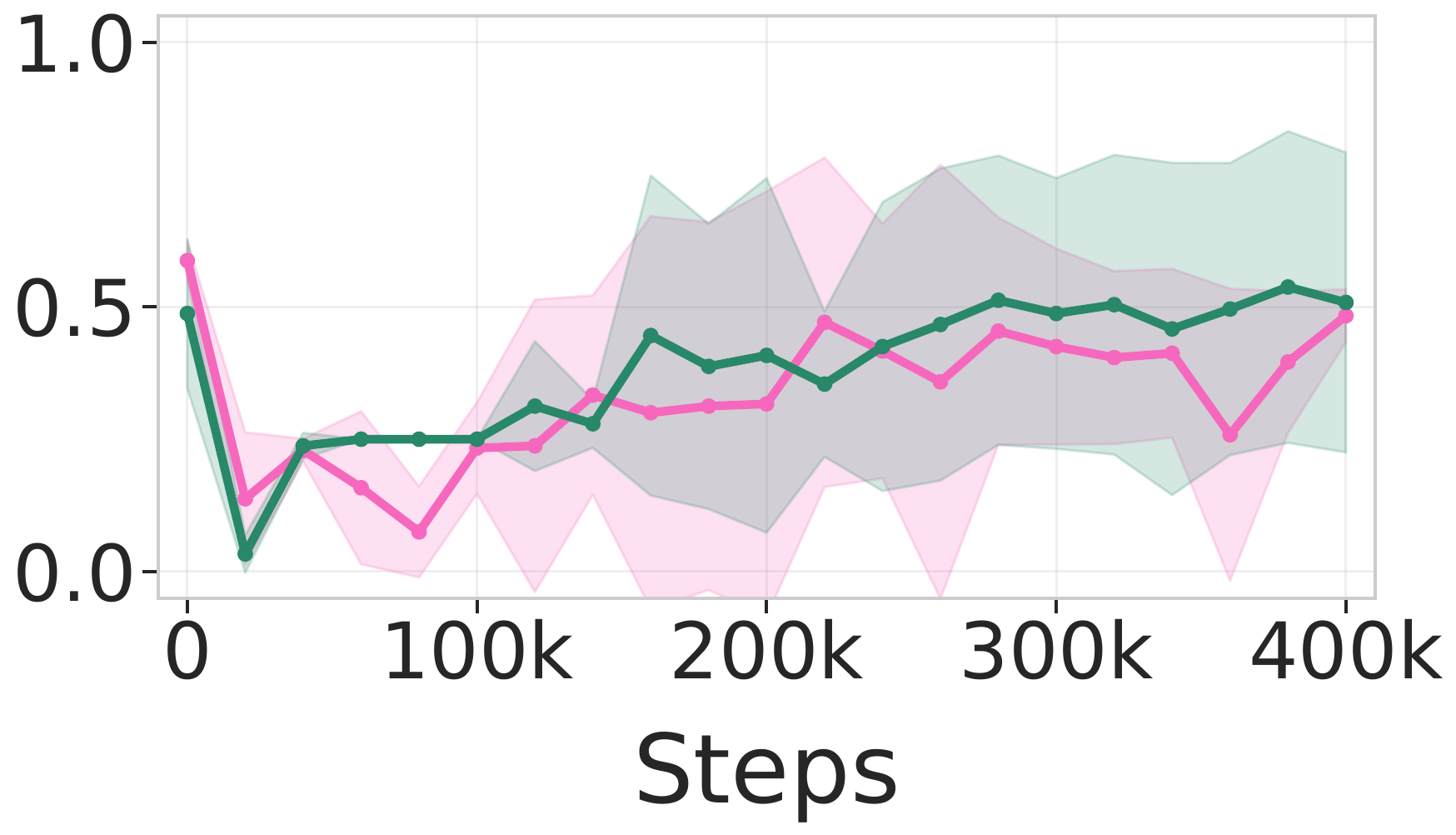}
    \end{subfigure}\hspace{0.1em}
    \begin{subfigure}[c]{0.21\textwidth}
        \includegraphics[width=1.0\linewidth]{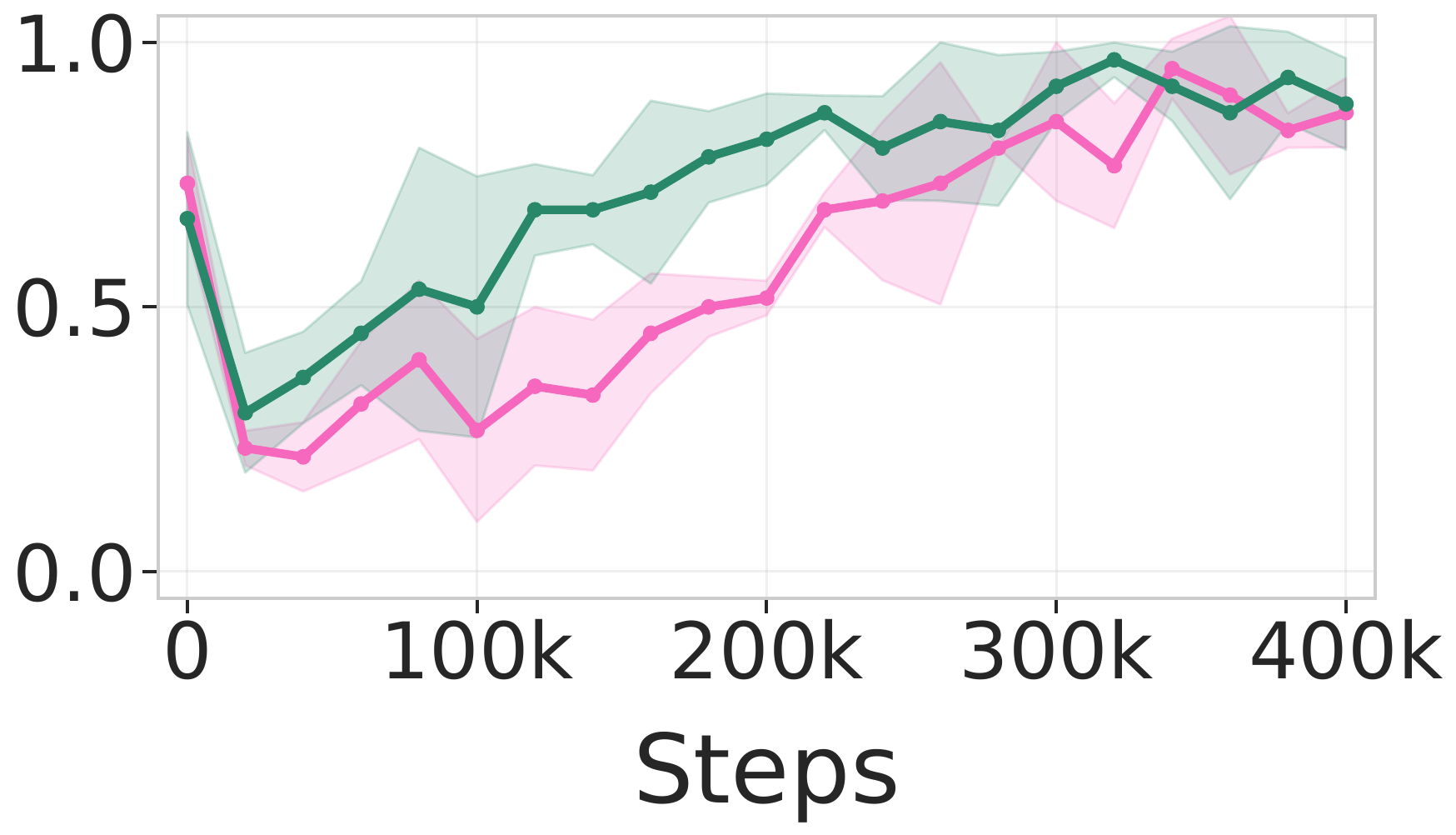}
    \end{subfigure}\\[1ex]
    \begin{subfigure}[c]{0.1\textwidth}
        \includegraphics[width=1.0\textwidth]{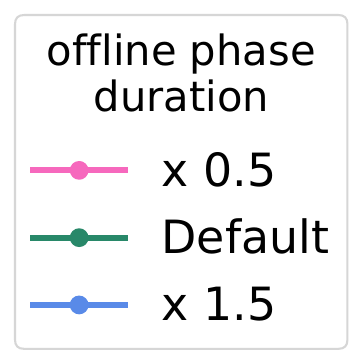}
    \end{subfigure}\hspace{0.1em}
    \begin{subfigure}[c]{0.21\textwidth}
        \includegraphics[width=1.0\linewidth]{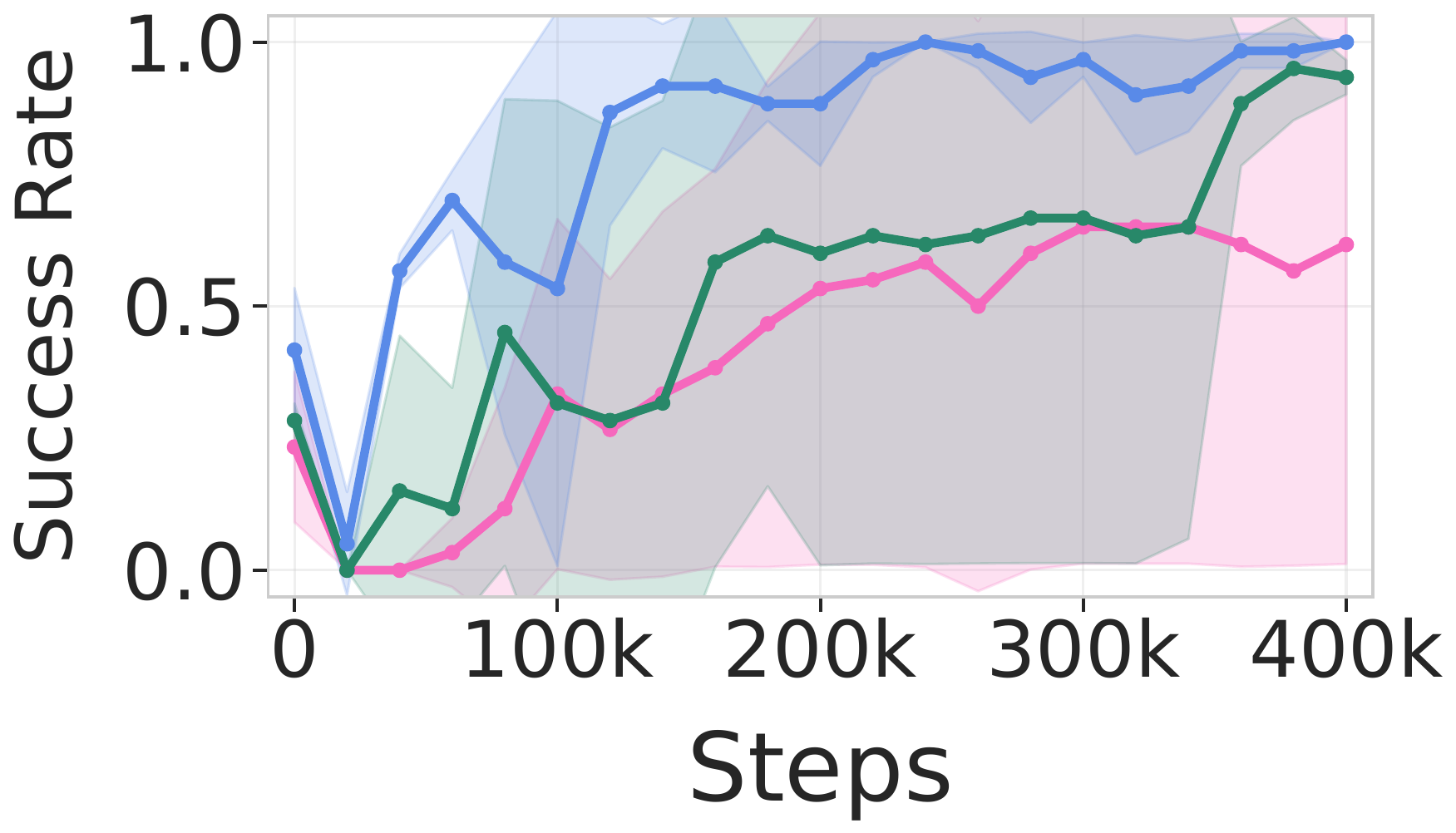}
    \end{subfigure}\hspace{0.1em}
    \begin{subfigure}[c]{0.21\textwidth}
        \includegraphics[width=1.0\linewidth]{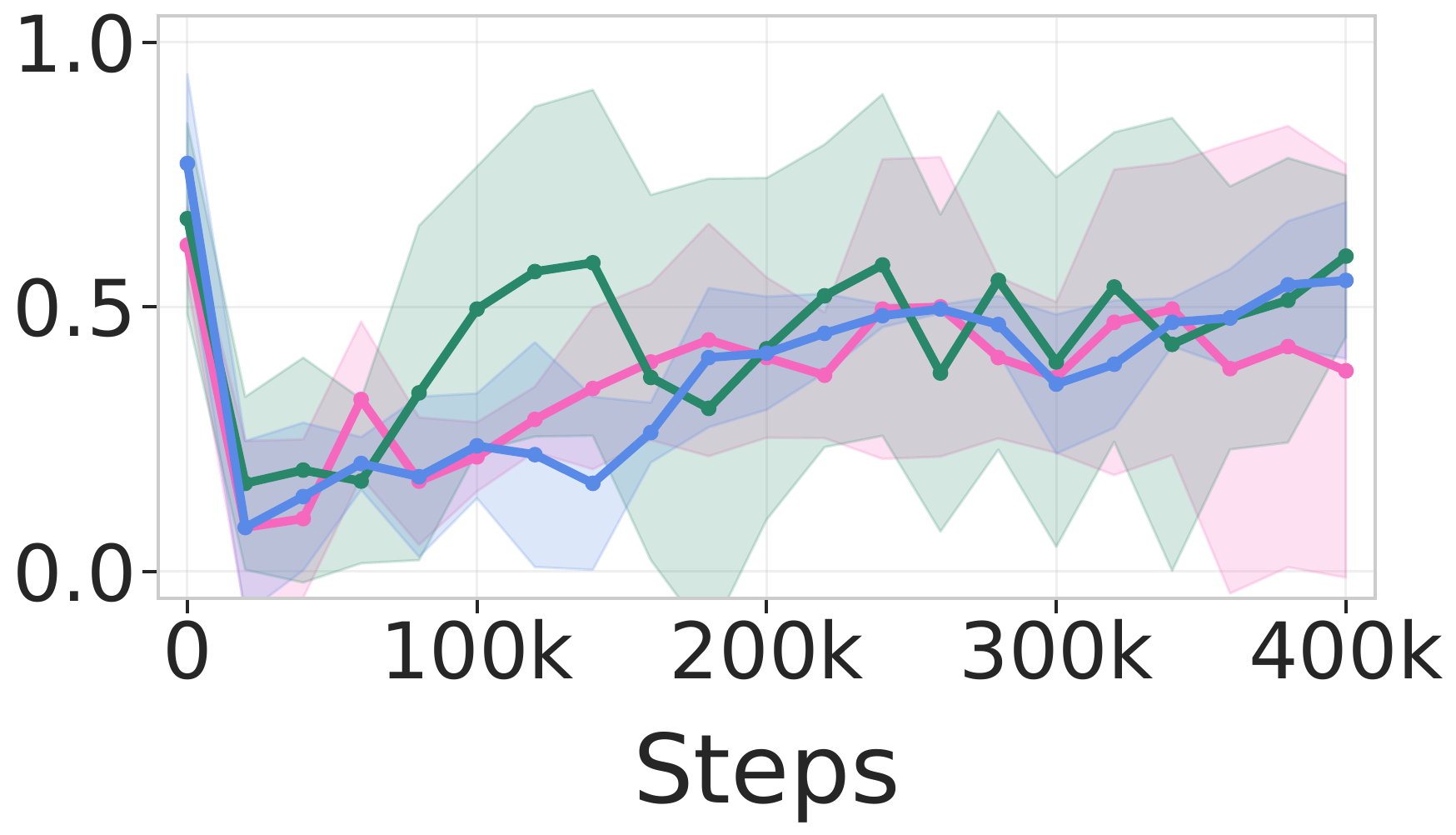}
    \end{subfigure}\hspace{0.1em}
    \begin{subfigure}[c]{0.21\textwidth}
        \includegraphics[width=1.0\linewidth]{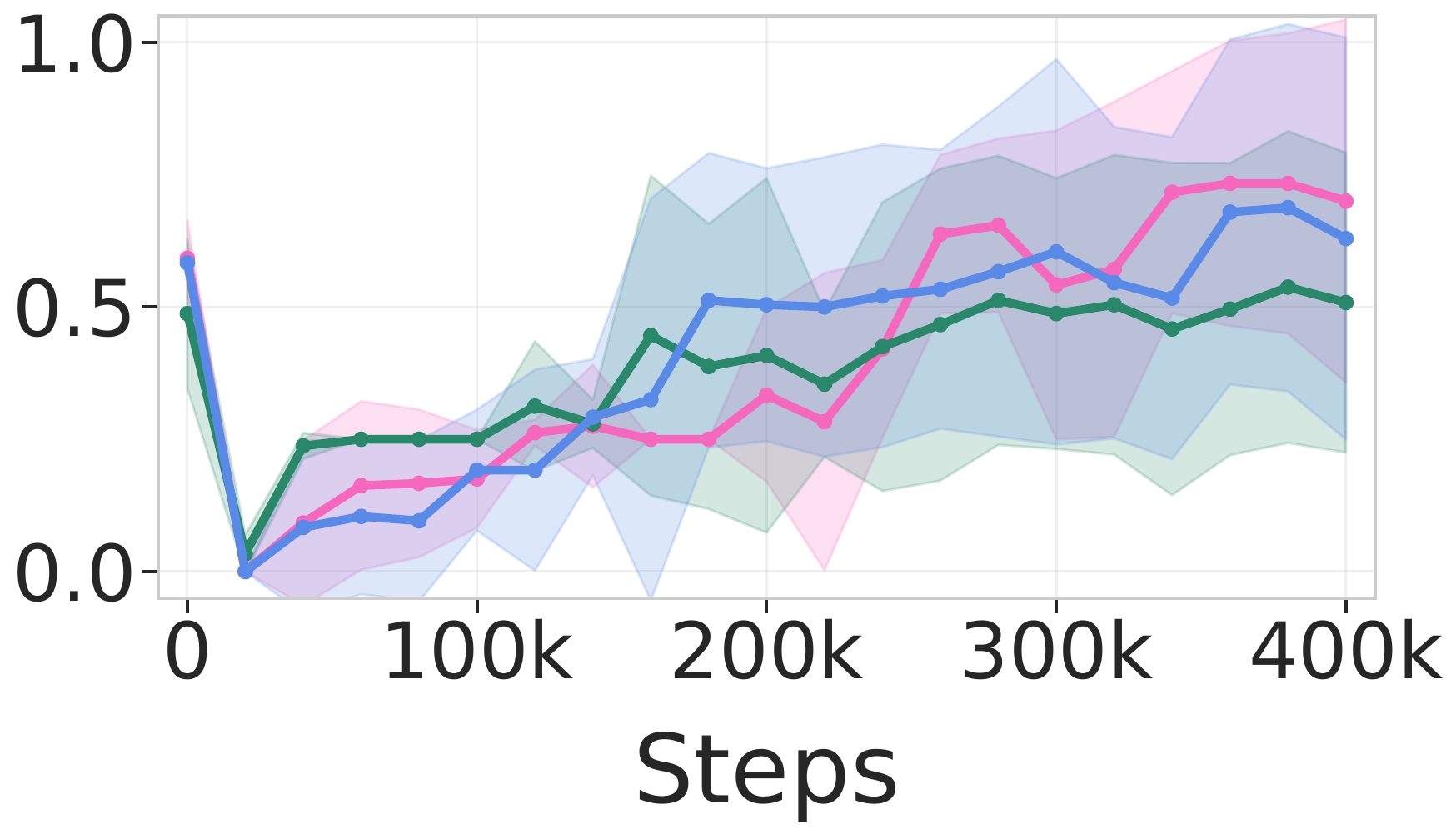}
    \end{subfigure}\hspace{0.1em}
    \begin{subfigure}[c]{0.21\textwidth}
        \includegraphics[width=1.0\linewidth]{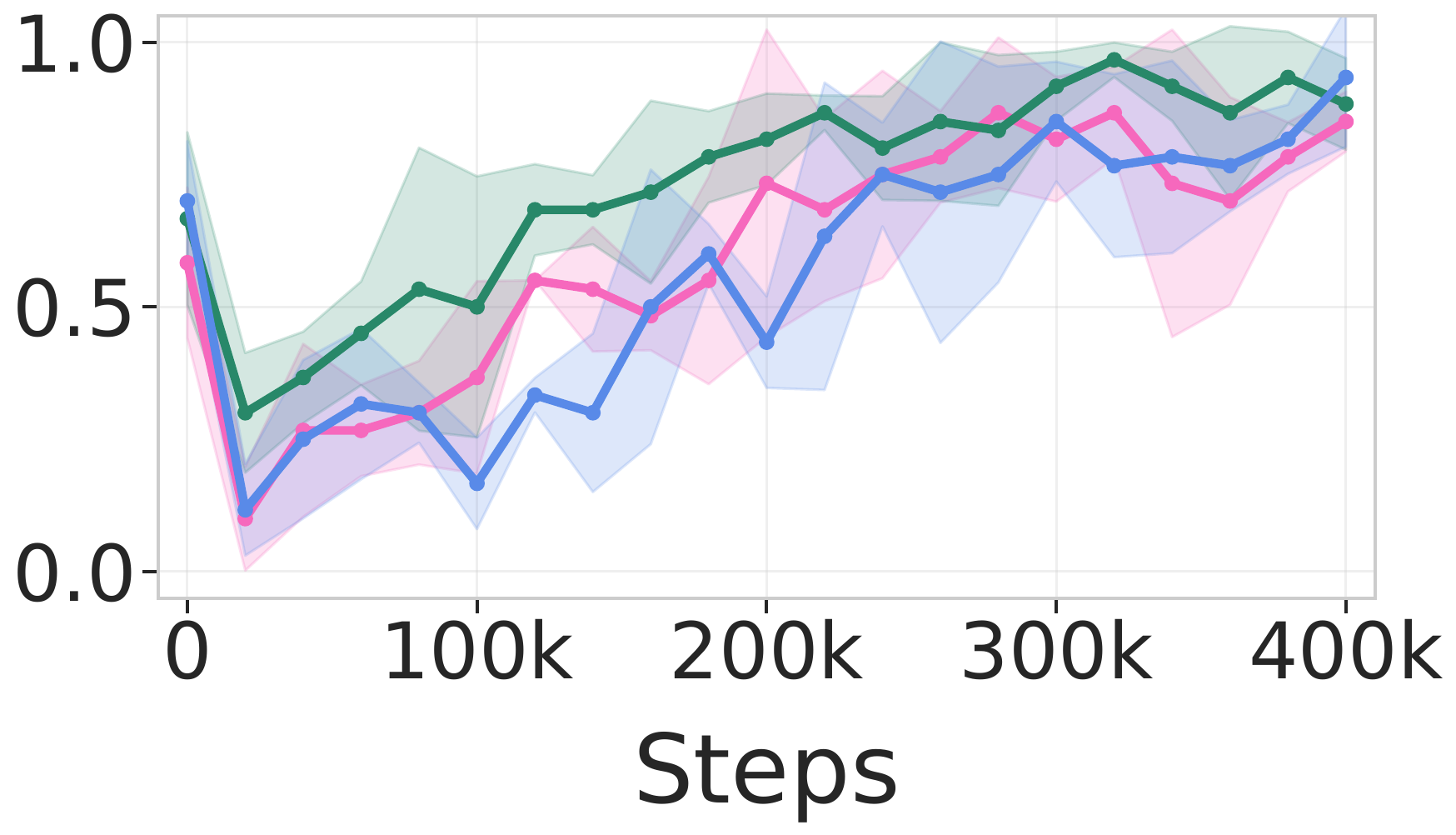}
    \end{subfigure}\\[1ex]
    \caption{\textbf{Hyperparameter and architectural tweaks do not consistently mitigate catastrophic failure.}
    From top to bottom, the plots correspond to: batch size, actor learning rate, critic learning rate, critic hidden dimension, actor hidden dimension, LayerNorm in the critic, LayerNorm in the actor, and the duration of the offline pretraining phase. Across all settings and tasks, none of these factors consistently mitigates catastrophic failure.}
    \label{fig:other_attempts_ablation}
\end{figure}

In Section~\ref{sec:analysis_catastrophic_failure}, we examined whether tuning the UTD ratio, varying warmup length, adjusting the Cal-QL conservative weight $\alpha_{\text{Cal-QL}}$, or regularizing the critic shift via $D_{\mathrm{KL}}\big(\mathrm{softmax}(Q^{\mathrm{offline}})\,\Vert\, \mathrm{softmax}(Q^{\mathrm{online}})\big)$ could mitigate catastrophic failure. None of these interventions proved effective. Here, we extend this investigation to additional settings: batch size, actor learning rate, critic learning rate, critic layer hidden dimension size, actor layer hidden dimension size, LayerNorm in the critic, LayerNorm in the actor, and the duration of the offline pretraining phase. As shown in Figure~\ref{fig:other_attempts_ablation}, none of these choices consistently reduces catastrophic failure across the four tasks, reinforcing our conclusion that such collapses are not reliably resolved by existing stabilization strategies.

\newpage
\section{Can Q-Driven Metrics Explain Catastrophic Failure?}
\label{sec:appendix:q_catastropic_failure_analysis}

\begin{figure}[h]
    \centering
    \begin{subfigure}[c]{0.22\textwidth}
        \includegraphics[width=1.0\linewidth]{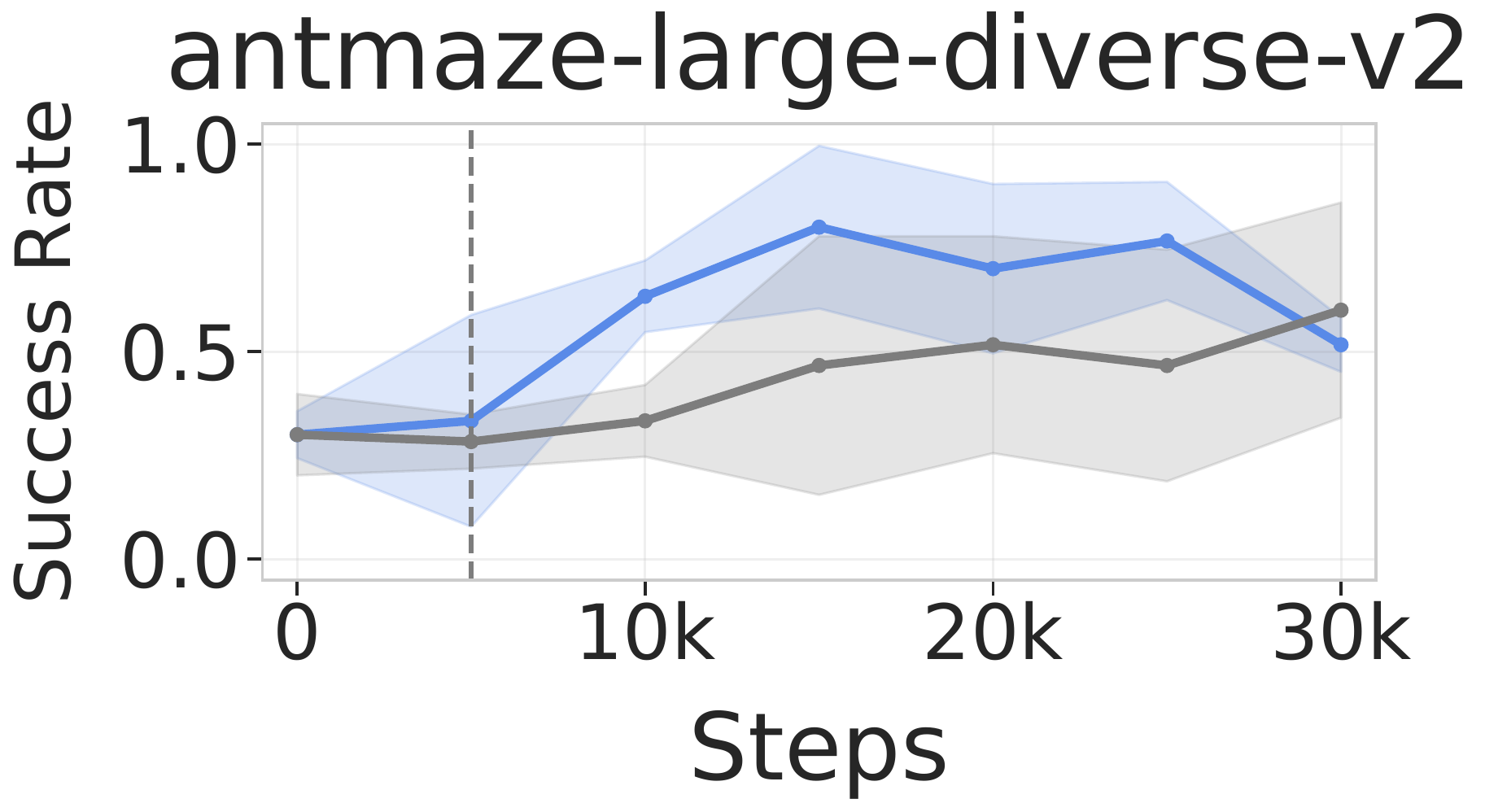}
    \end{subfigure}\hspace{0.1em}
    \begin{subfigure}[c]{0.22\textwidth}
        \includegraphics[width=1.0\linewidth]{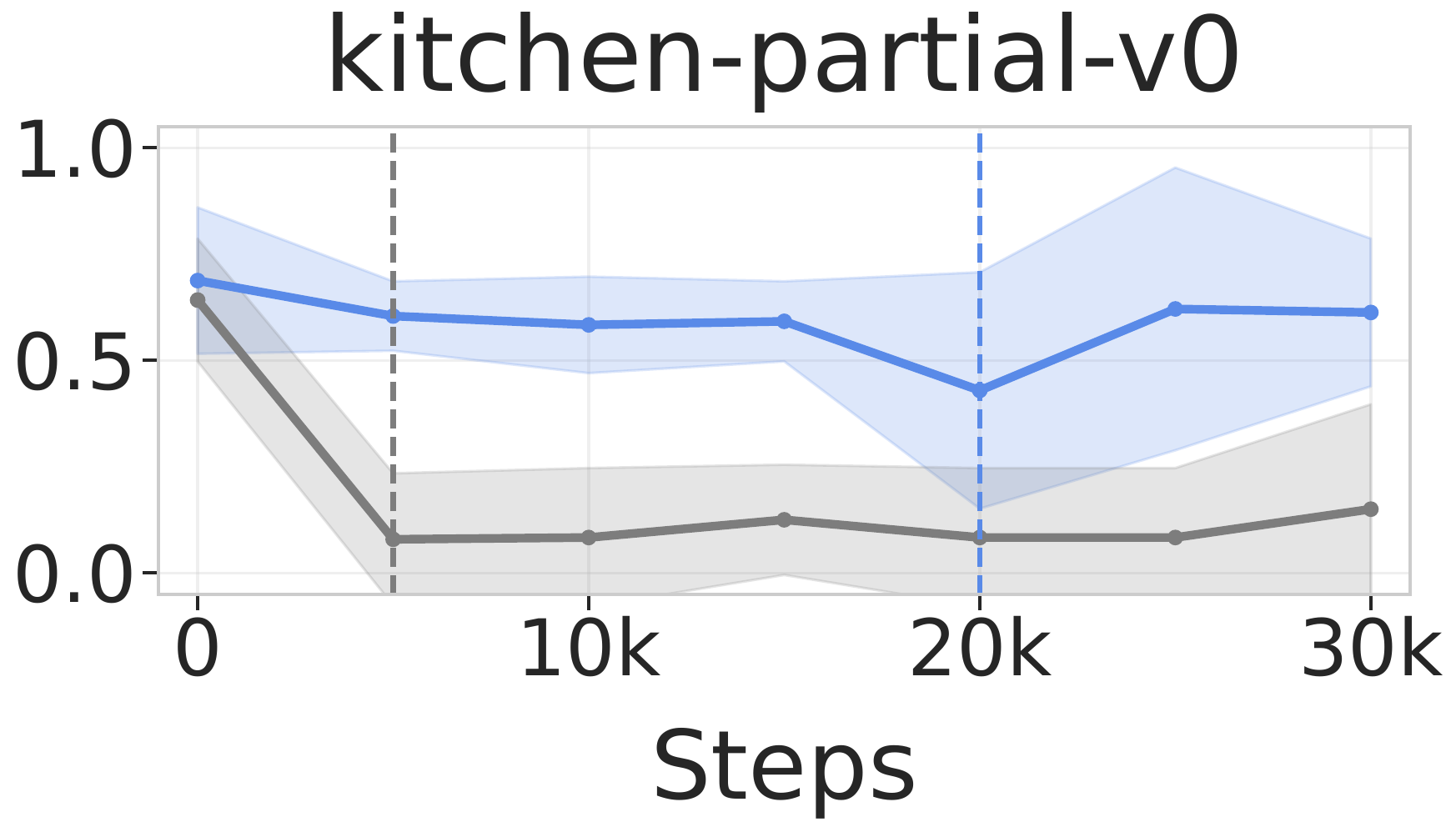}
    \end{subfigure}\hspace{0.1em}
    \begin{subfigure}[c]{0.22\textwidth}
        \includegraphics[width=1.0\linewidth]{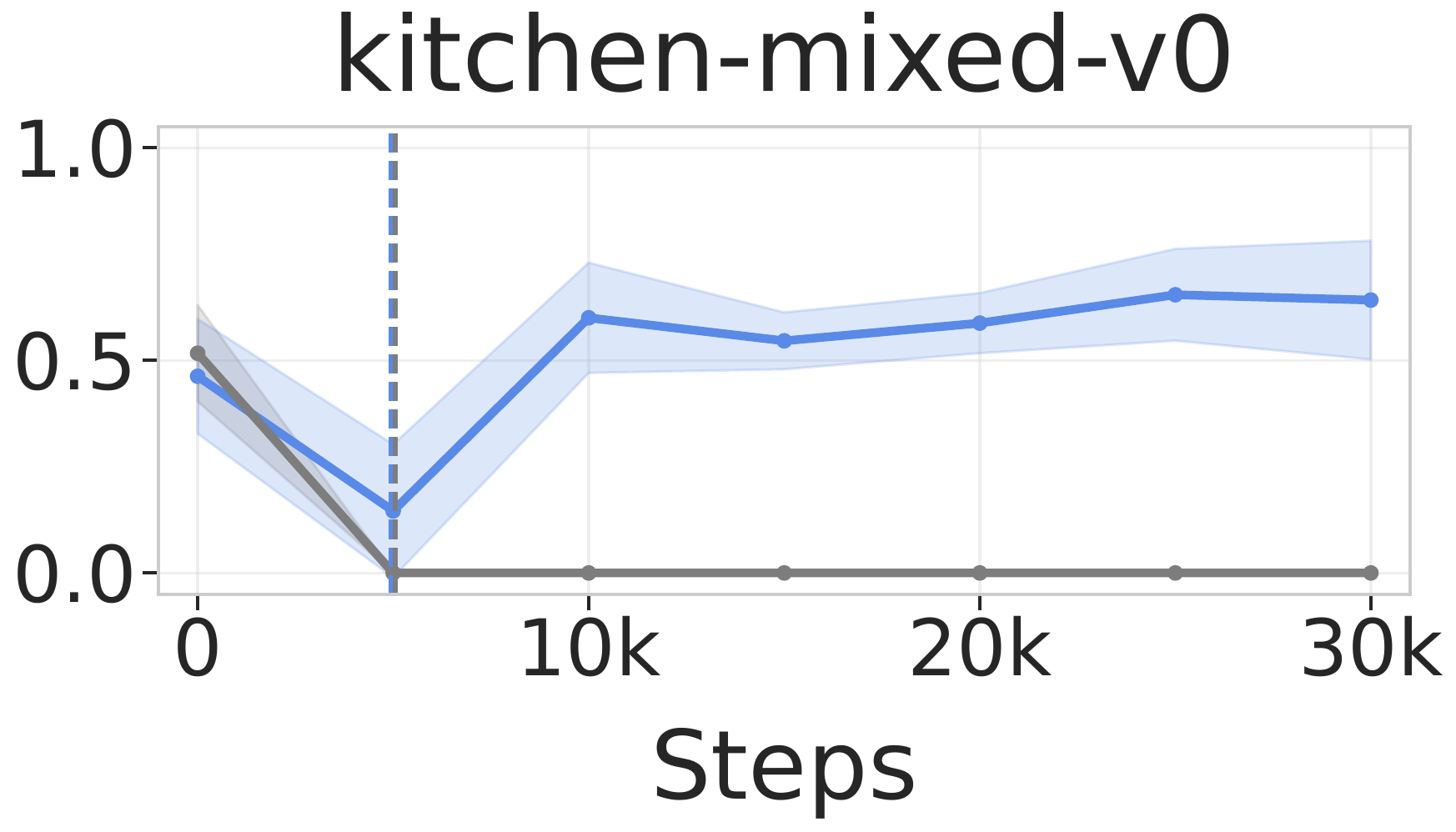}
    \end{subfigure}\hspace{0.1em}
    \begin{subfigure}[c]{0.22\textwidth}
        \includegraphics[width=1.0\linewidth]{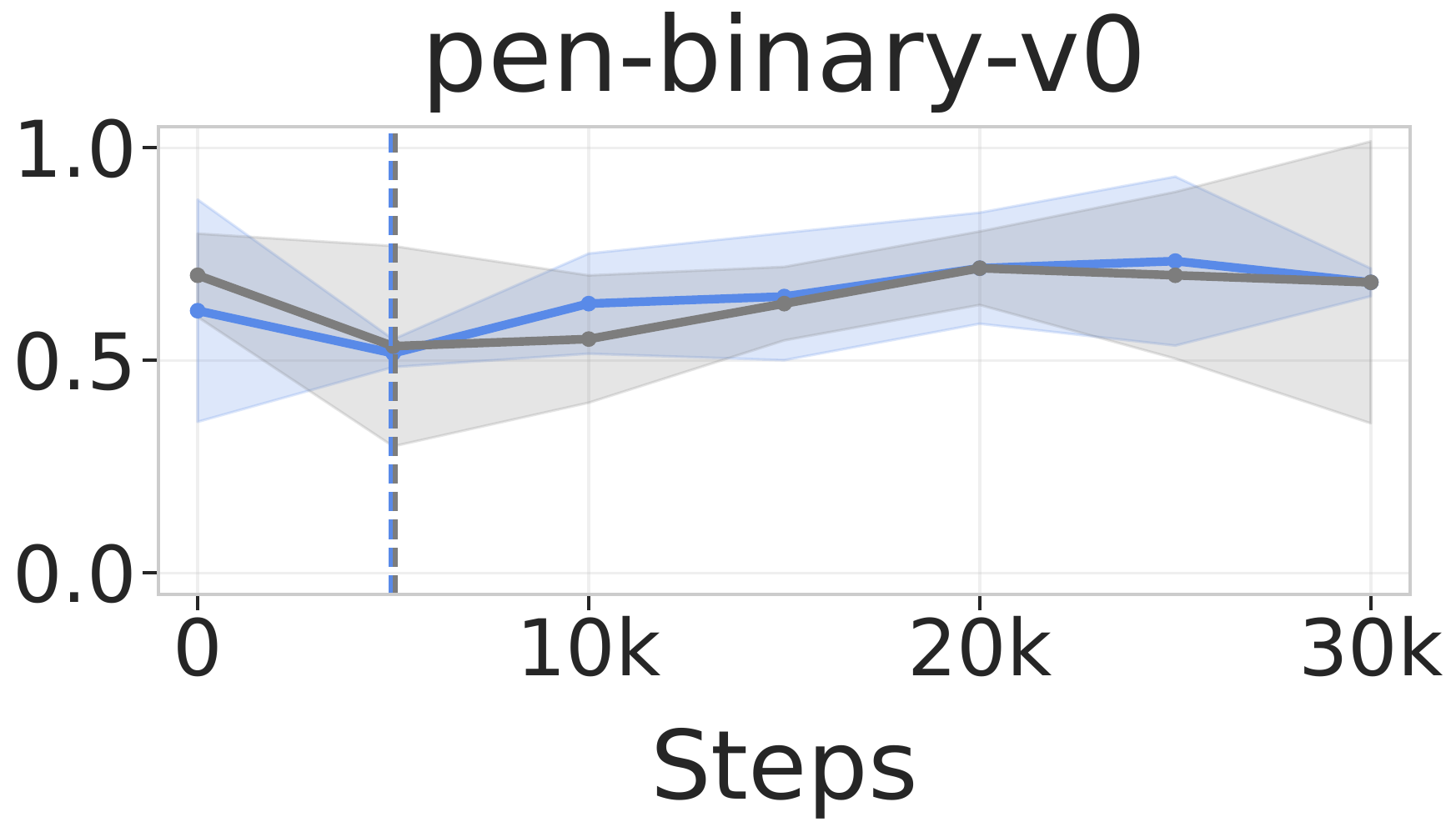}
    \end{subfigure}\\[1ex]
    \begin{subfigure}[c]{0.22\textwidth}
        \includegraphics[width=1.0\linewidth]{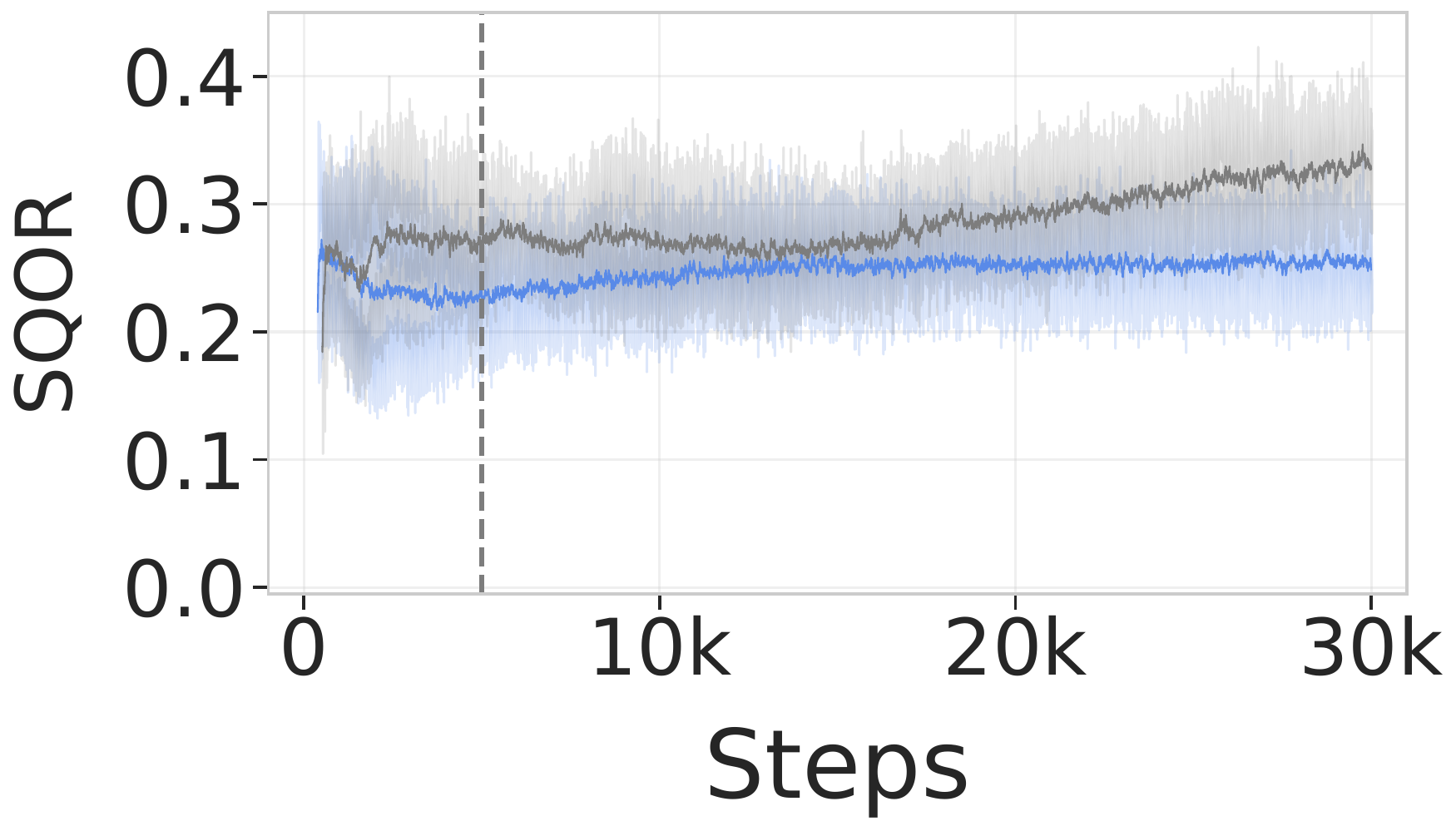}
    \end{subfigure}\hspace{0.1em}
    \begin{subfigure}[c]{0.22\textwidth}
        \includegraphics[width=1.0\linewidth]{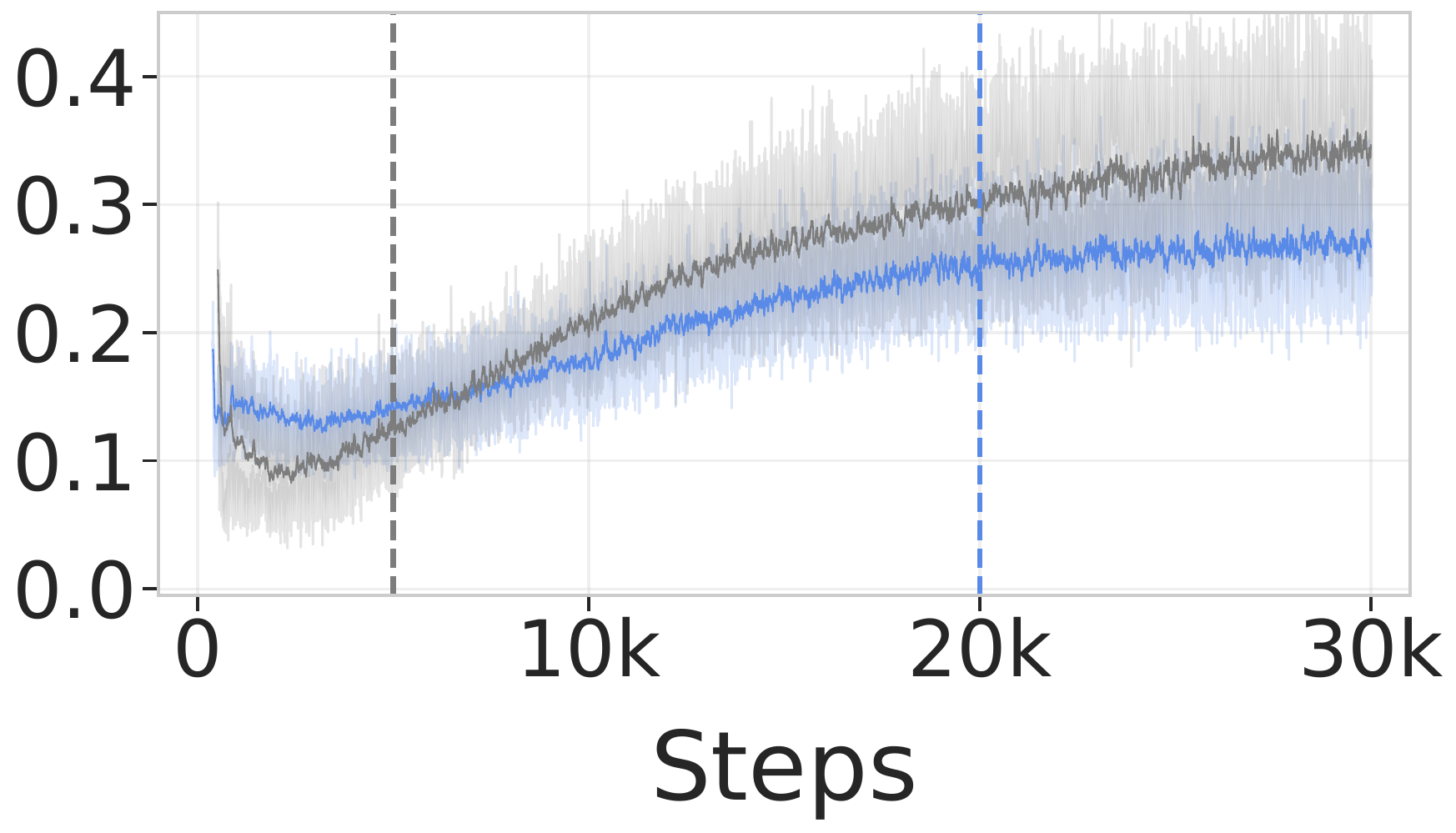}
    \end{subfigure}\hspace{0.1em}
    \begin{subfigure}[c]{0.22\textwidth}
        \includegraphics[width=1.0\linewidth]{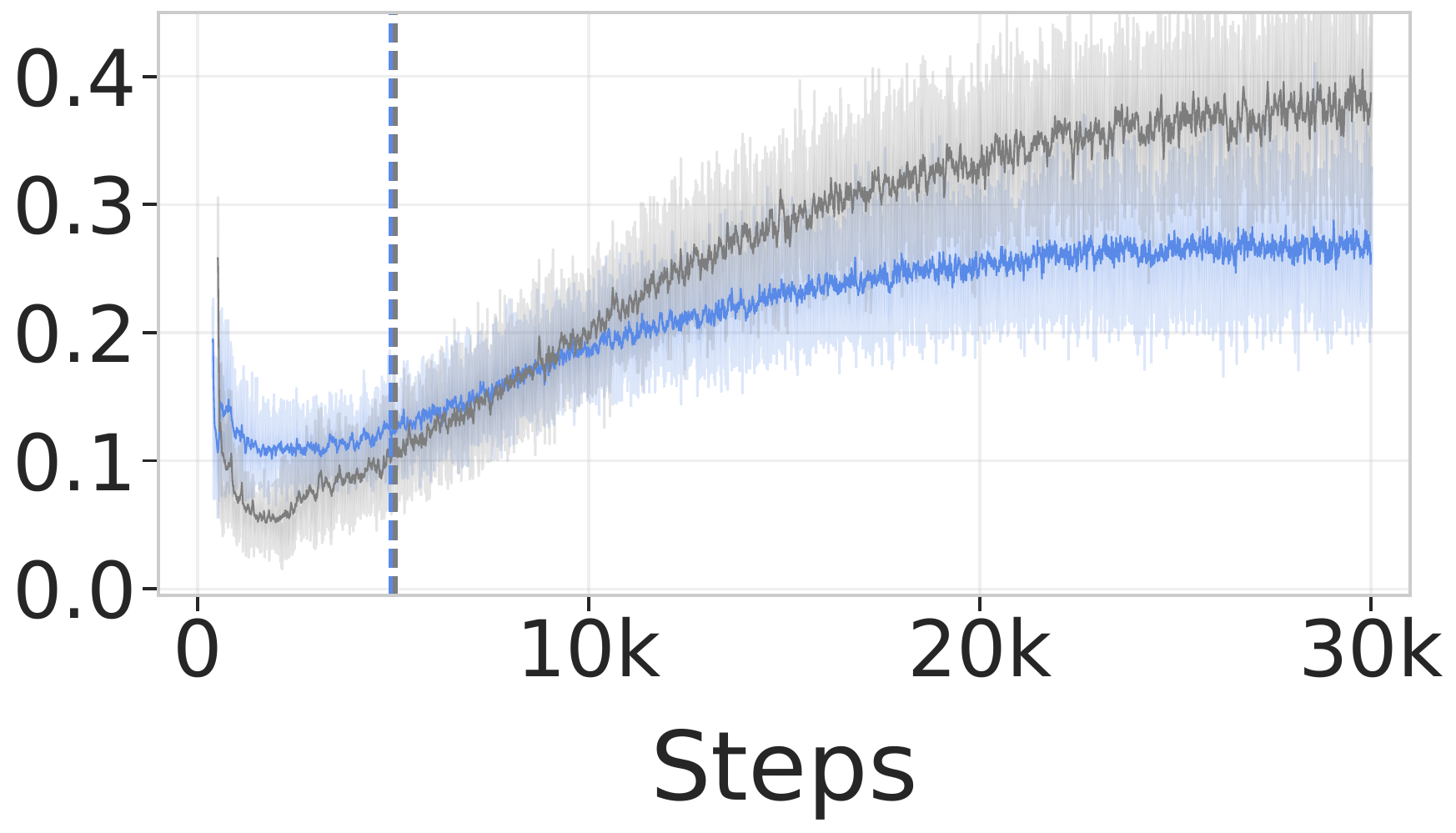}
    \end{subfigure}\hspace{0.1em}
    \begin{subfigure}[c]{0.22\textwidth}
        \includegraphics[width=1.0\linewidth]{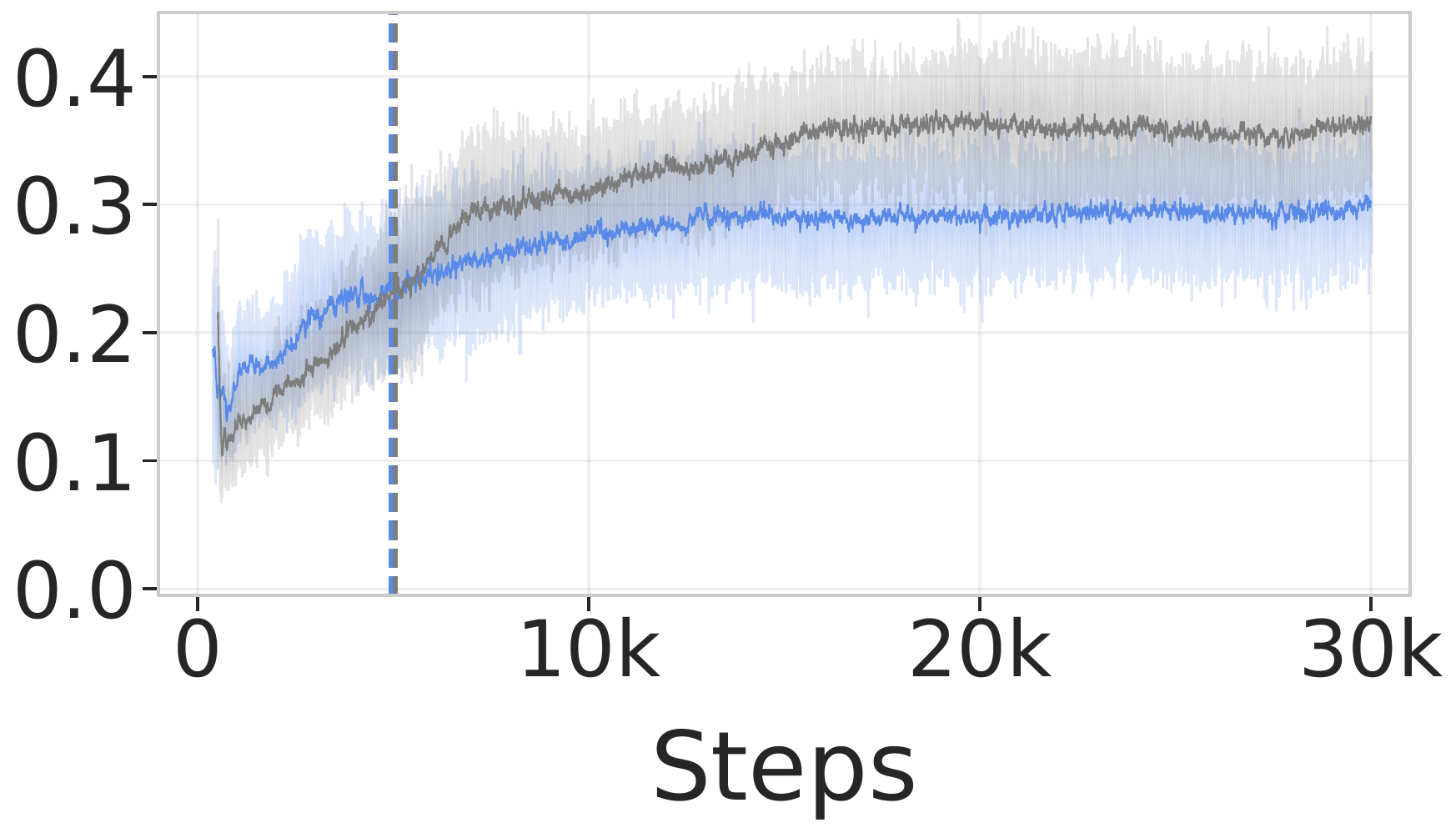}
    \end{subfigure}\\[1ex]
    \begin{subfigure}[c]{0.22\textwidth}
        \includegraphics[width=1.0\linewidth]{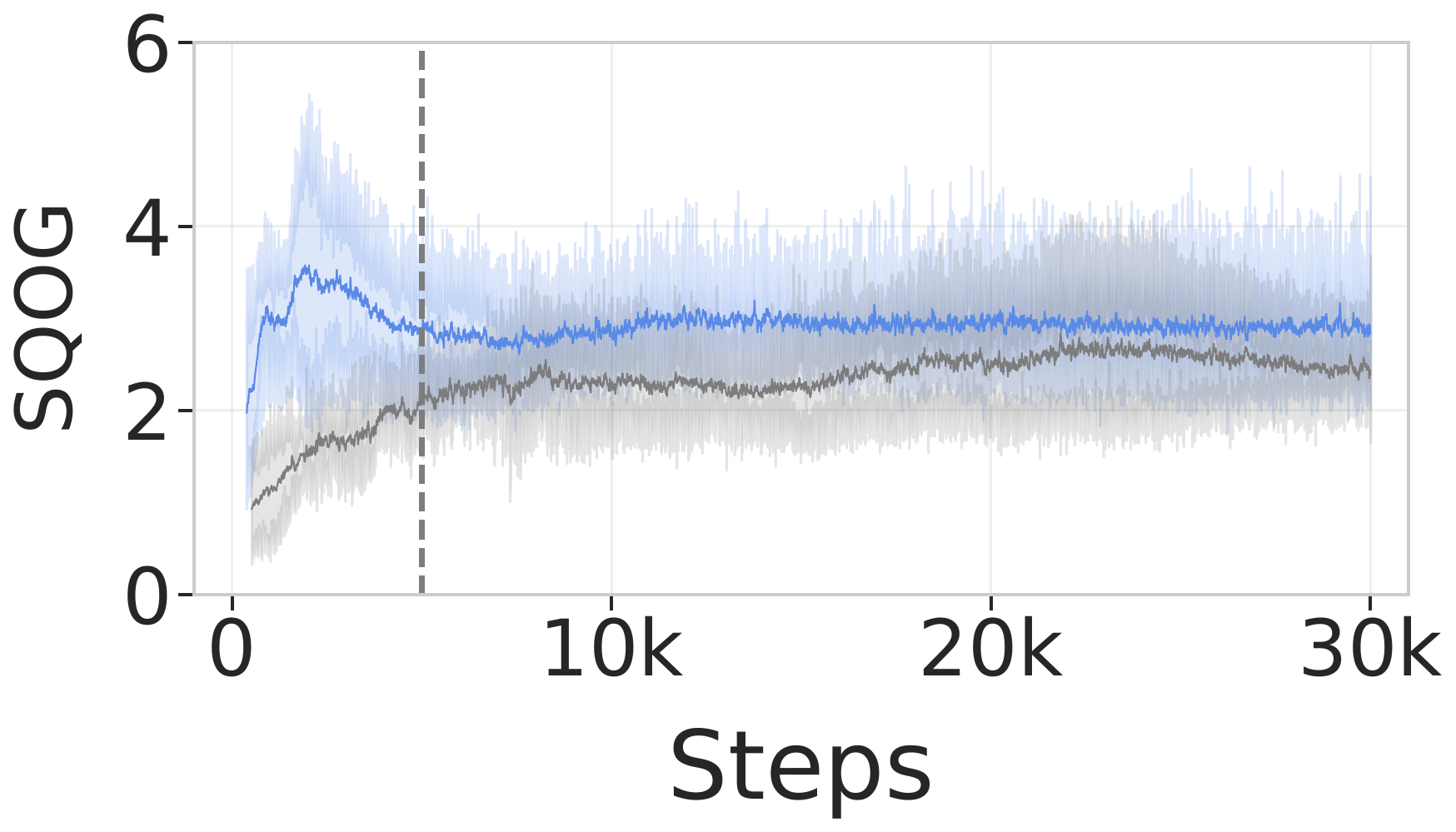}
    \end{subfigure}\hspace{0.1em}
    \begin{subfigure}[c]{0.22\textwidth}
        \includegraphics[width=1.0\linewidth]{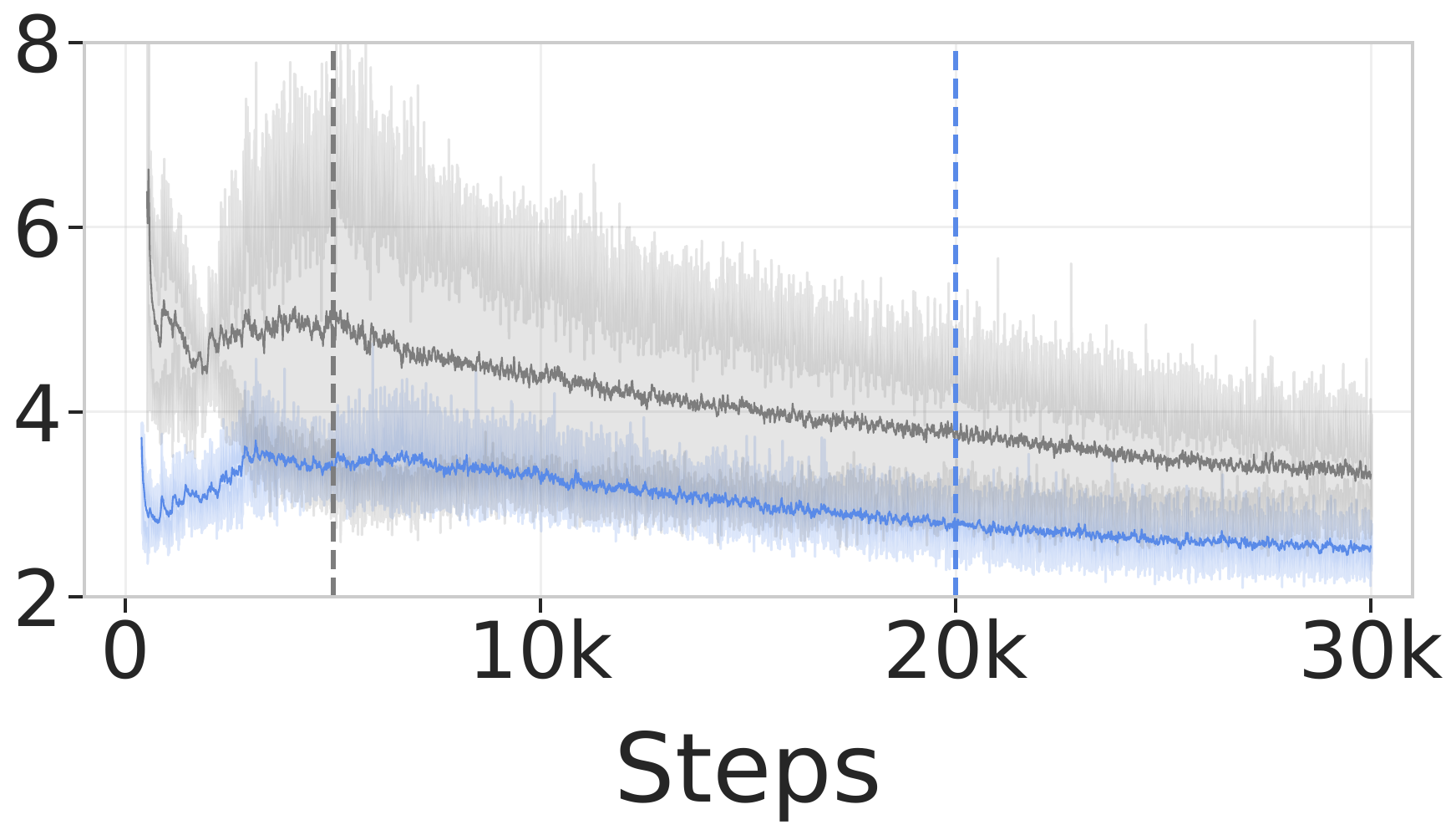}
    \end{subfigure}\hspace{0.1em}
    \begin{subfigure}[c]{0.22\textwidth}
        \includegraphics[width=1.0\linewidth]{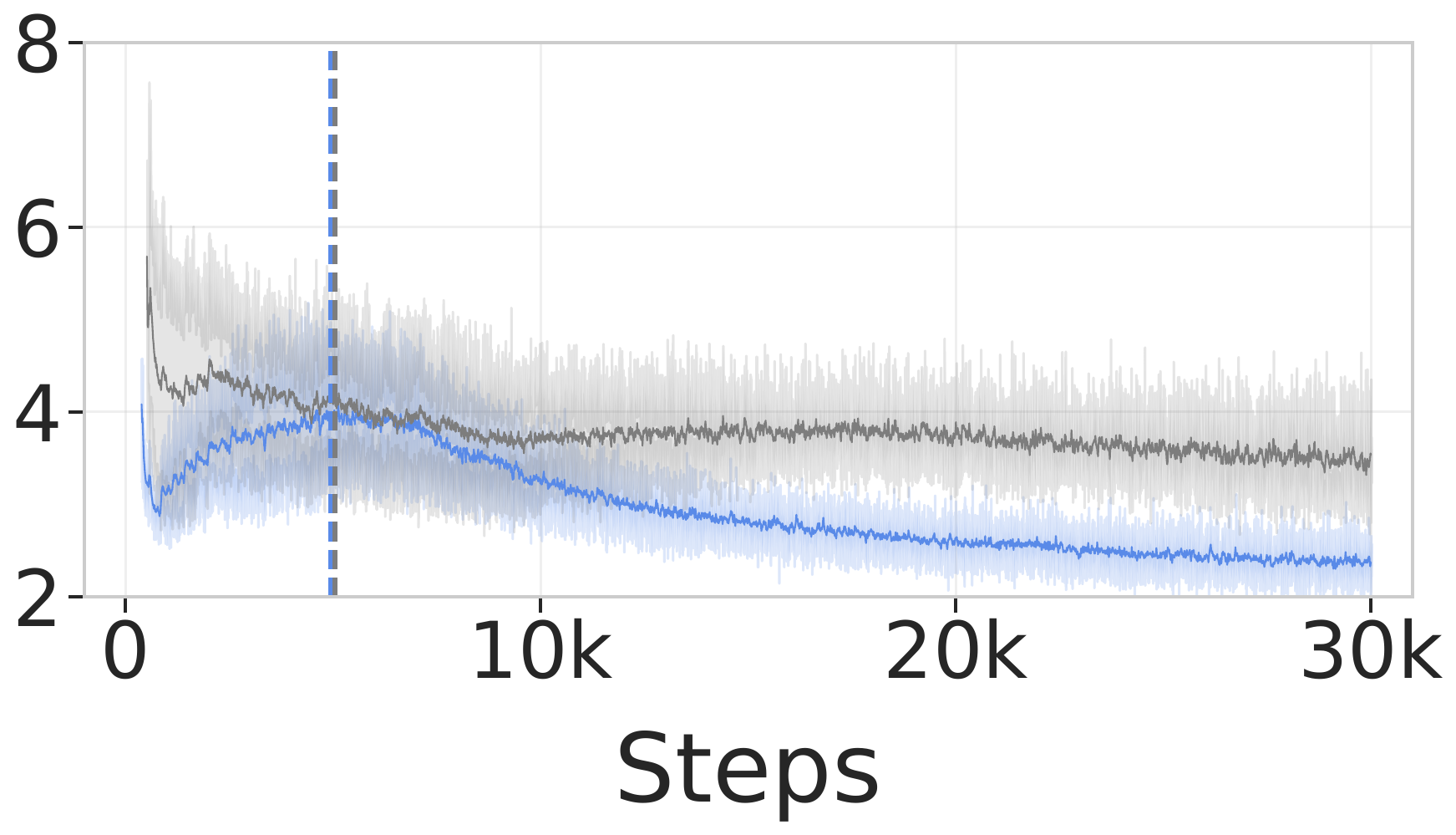}
    \end{subfigure}\hspace{0.1em}
    \begin{subfigure}[c]{0.22\textwidth}
        \includegraphics[width=1.0\linewidth]{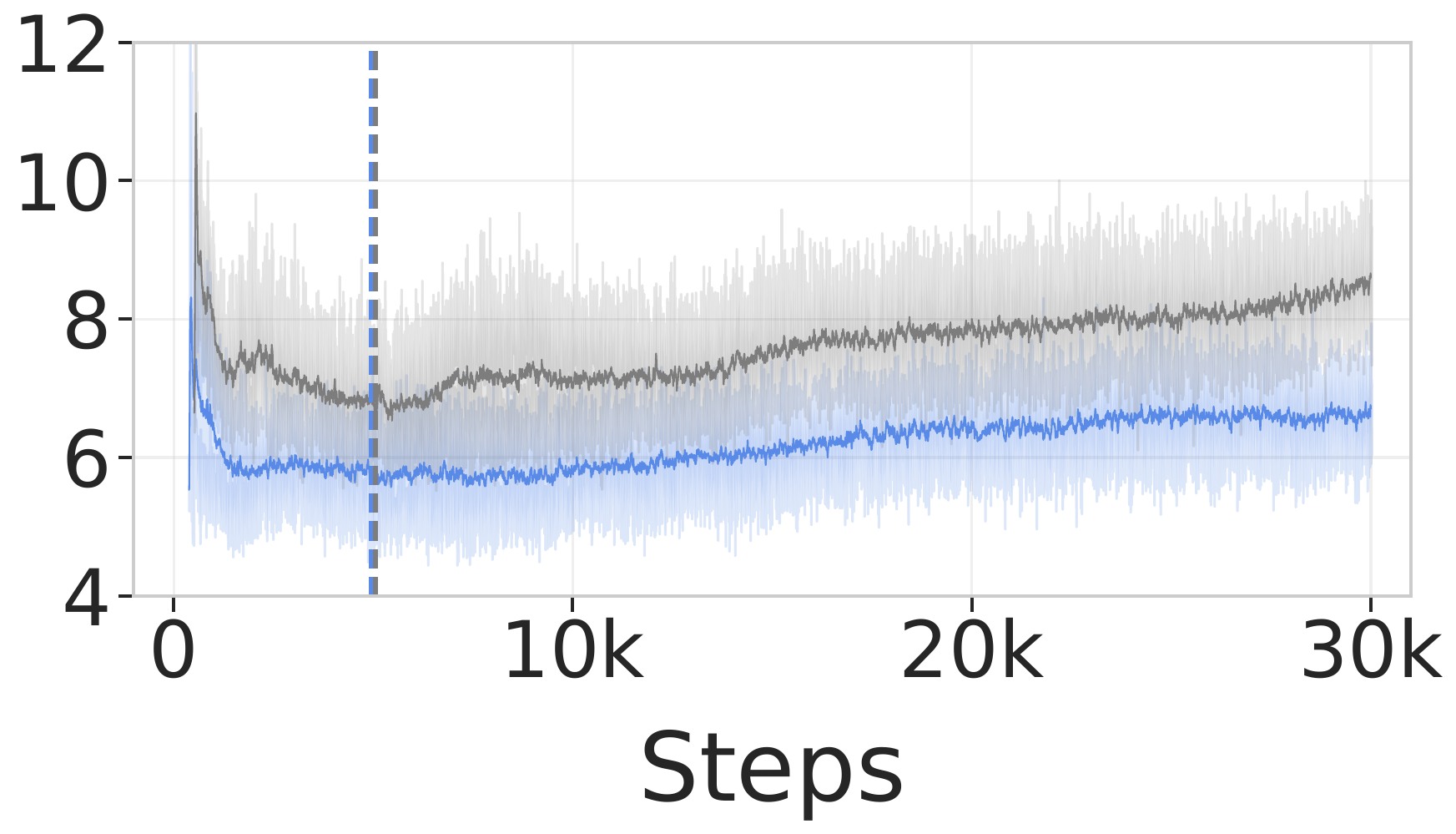}
    \end{subfigure}\\[1ex]

    \includegraphics[width=0.5\linewidth]
    {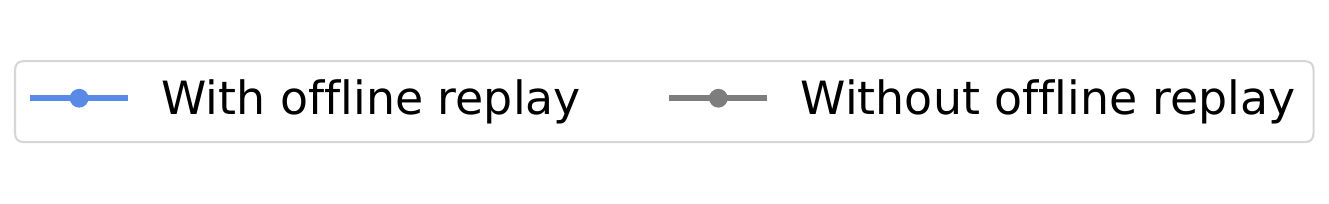}
    \caption{\textbf{SQOR and SQOG do not reliably predict catastrophic failure.}
    We compare success rate, the spurious Q-optimism ratio (SQOR), and the spurious
    Q-optimism gap (SQOG) across four tasks. Both metrics show task-dependent and
    inconsistent alignment with early performance collapse. The vertical dashed line marks
    the timestep at which the largest performance drop occurs.}
    \label{fig:appendix:additional_analysis_cat_failure}
\end{figure}

In Section~\ref{sec:analysis_catastrophic_failure}, we showed that explicitly slowing the
critic's distributional shift does not reliably prevent catastrophic failure. Here, we
further examine whether other Q-driven diagnostics can explain or predict early collapse.
We focus on the setting in Figure~\ref{fig:offline_data_alpha_analysis}(b), where online
fine-tuning proceeds without offline replay while retaining the conservative regularizer.

We evaluate four tasks: antmaze-large-diverse, kitchen-partial, kitchen-mixed, and
pen-binary. Since catastrophic failure typically occurs early in online fine-tuning, we
report all metrics over the first $30\mathrm{K}$ online steps. Due to computational
constraints, this analysis is run with three random seeds.

For a minibatch of states $s$, let $a_t\sim\pi_t(\cdot|s)$ denote an action sampled from
the current online policy and let $a_{\mathrm{off}}\sim\pi_{\mathrm{off}}(\cdot|s)$ denote
an action sampled from the offline-pretrained policy. Let $Q_t$ denote the critic at online
step $t$, and let $Q_{\mathrm{final}}$ denote the critic checkpoint after $400\mathrm{K}$
online steps for the same random seed. We define
\[
\Delta_t(s)
=
Q_t(s,a_t)-Q_t(s,a_{\mathrm{off}}),
\qquad
\Delta_{\mathrm{final}}(s)
=
Q_{\mathrm{final}}(s,a_t)-Q_{\mathrm{final}}(s,a_{\mathrm{off}}).
\]
Intuitively, $\Delta_t$ measures the current critic's relative preference between the
online-policy action and the offline-policy action, while $\Delta_{\mathrm{final}}$
provides a post-hoc reference after online fine-tuning has stabilized.

\textbf{Spurious Q-optimism ratio and mismatch magnitude.}
We first examine two preference-based Q diagnostics. Let $\mathcal{B}_t$ denote the
minibatch of states sampled from the online replay buffer at step $t$, and let
$\mathrm{sgn}(\cdot)$ denote the sign function. The first metric is the spurious
Q-optimism ratio:
\[
\mathrm{SQOR}(t)
=
\frac{1}{|\mathcal{B}_t|}
\sum_{s\in\mathcal{B}_t}
\mathbf{1}\left[
\Delta_t(s)>0
\;\wedge\;
\Delta_{\mathrm{final}}(s)<0
\right].
\]
SQOR measures the fraction of batch states where the current critic favors the
current-policy action, while the final critic favors the offline-policy action. The second
metric measures the magnitude of preference mismatch. Define the set of mismatched states
within the current online batch as
\[
\mathcal{M}_t
=
\left\{
s\in\mathcal{B}_t:
\mathrm{sgn}(\Delta_t(s))
\neq
\mathrm{sgn}(\Delta_{\mathrm{final}}(s))
\right\}.
\]
We define the spurious Q-optimism gap as
\[
\mathrm{SQOG}(t)
=
\frac{1}{|\mathcal{M}_t|}
\sum_{s\in\mathcal{M}_t}
\left|
\Delta_t(s)-\Delta_{\mathrm{final}}(s)
\right|,
\]
with the value set to zero when $\mathcal{M}_t$ is empty.

As shown in Figure~\ref{fig:appendix:additional_analysis_cat_failure}, neither SQOR nor
SQOG consistently tracks catastrophic failure across tasks and stress settings. In some
cases, the metrics increase without a corresponding collapse; in others, early performance
drops are not preceded by a clear rise in either metric. This suggests that these
preference-based Q diagnostics alone do not provide a reliable explanation for
catastrophic failure.

\begin{figure}[h]
    \centering
    \begin{subfigure}[c]{0.22\textwidth}
        \includegraphics[width=1.0\linewidth]{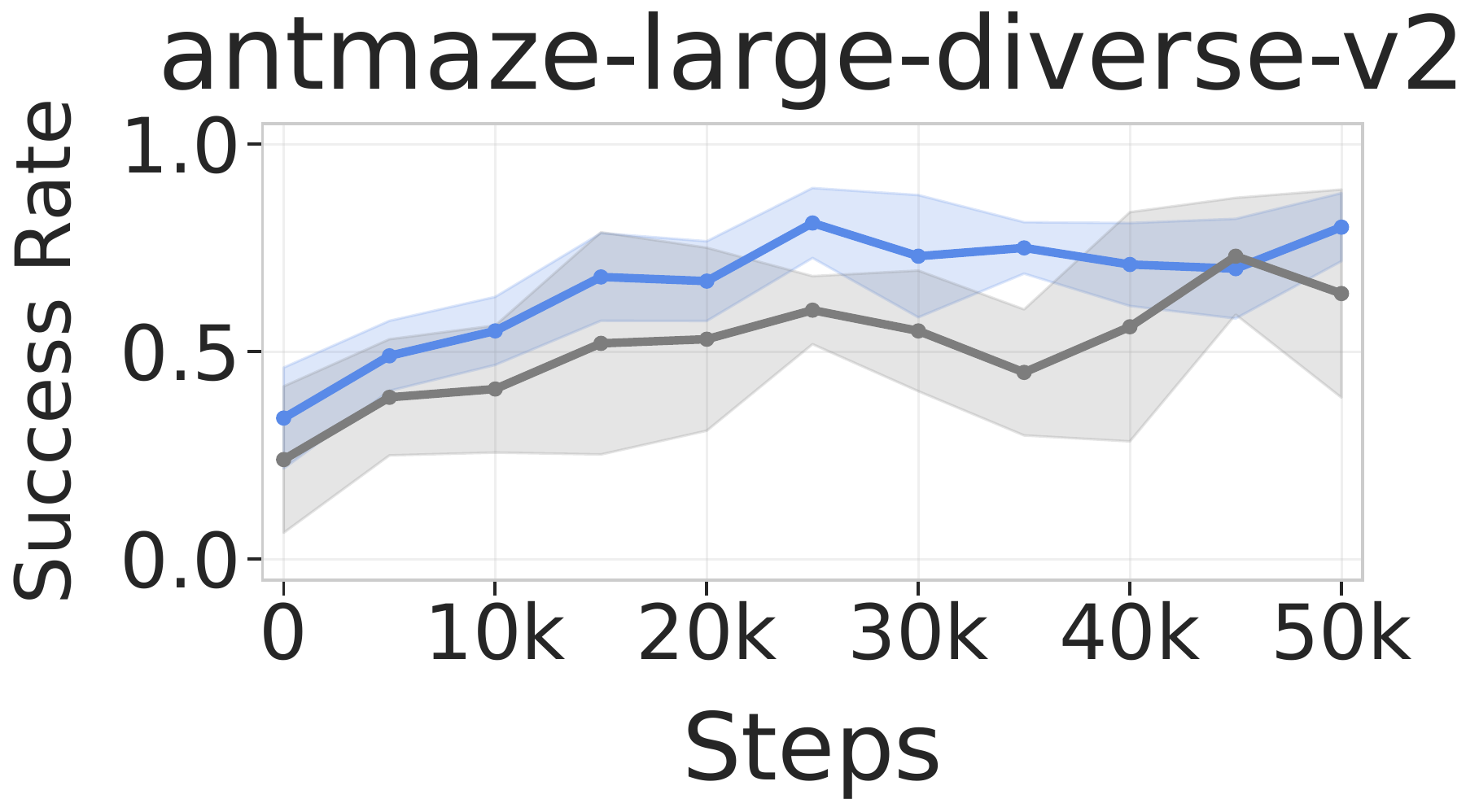}
    \end{subfigure}\hspace{0.1em}
    \begin{subfigure}[c]{0.22\textwidth}
        \includegraphics[width=1.0\linewidth]{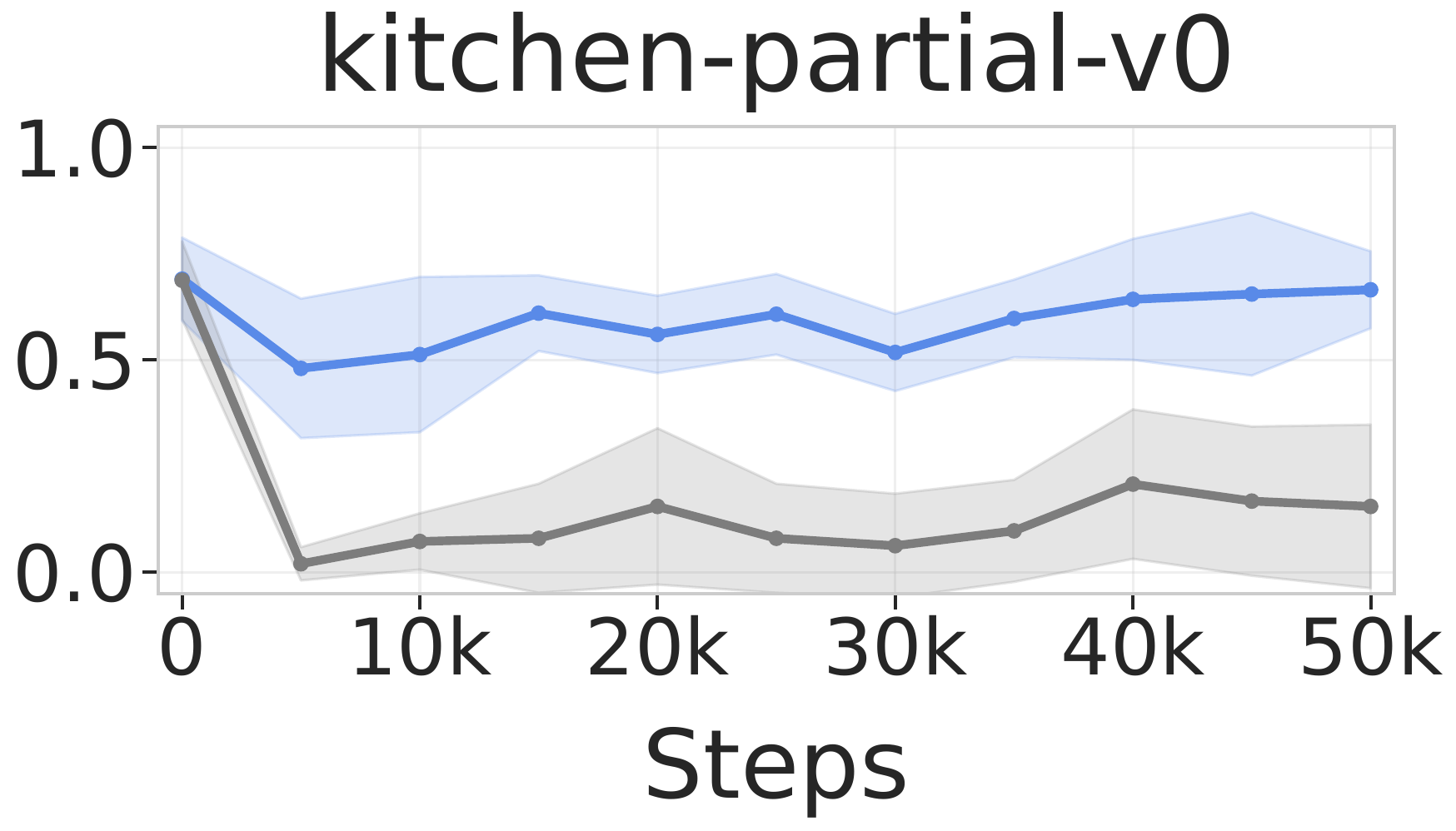}
    \end{subfigure}\hspace{0.1em}
    \begin{subfigure}[c]{0.22\textwidth}
        \includegraphics[width=1.0\linewidth]{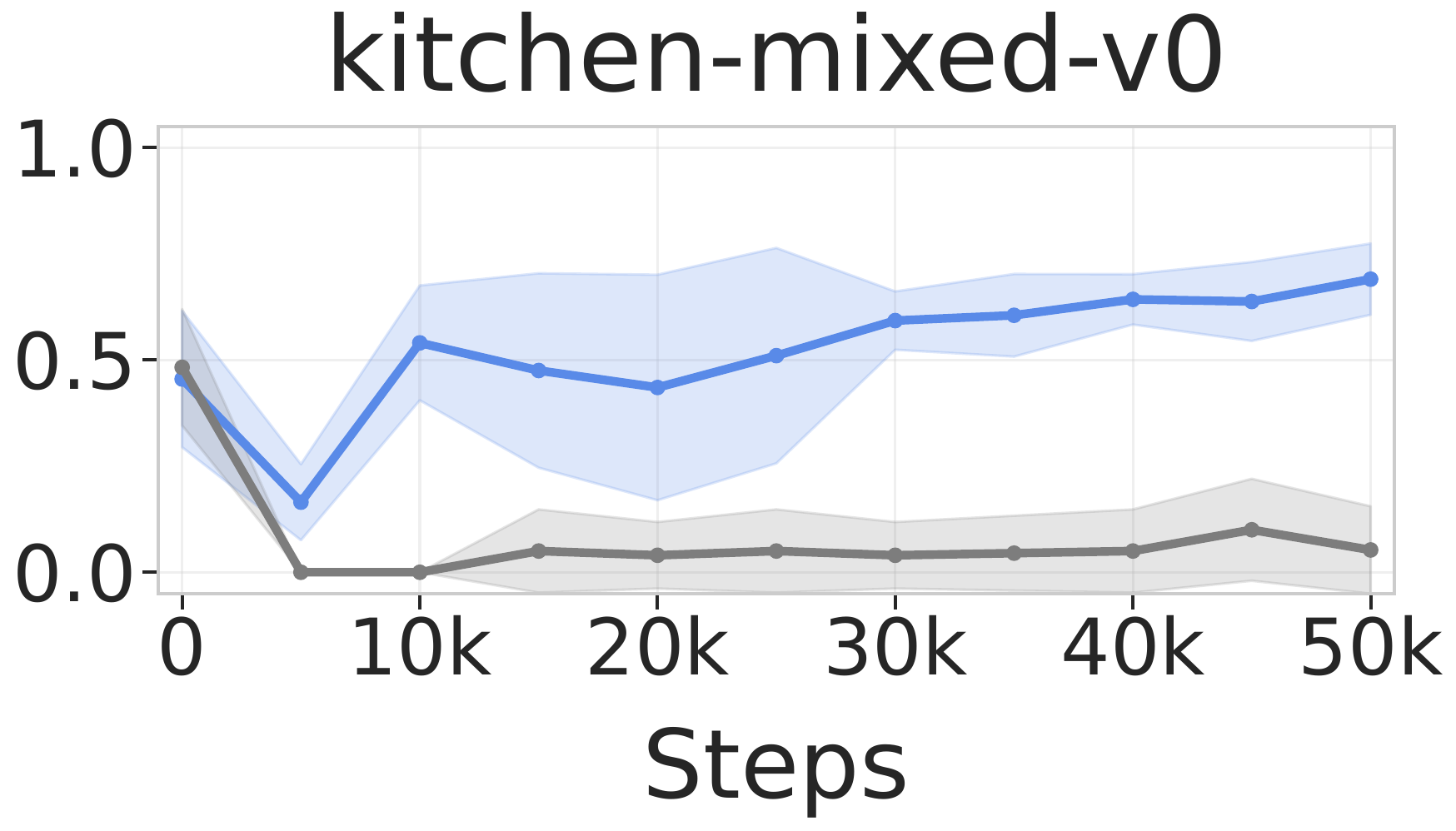}
    \end{subfigure}\hspace{0.1em}
    \begin{subfigure}[c]{0.22\textwidth}
        \includegraphics[width=1.0\linewidth]{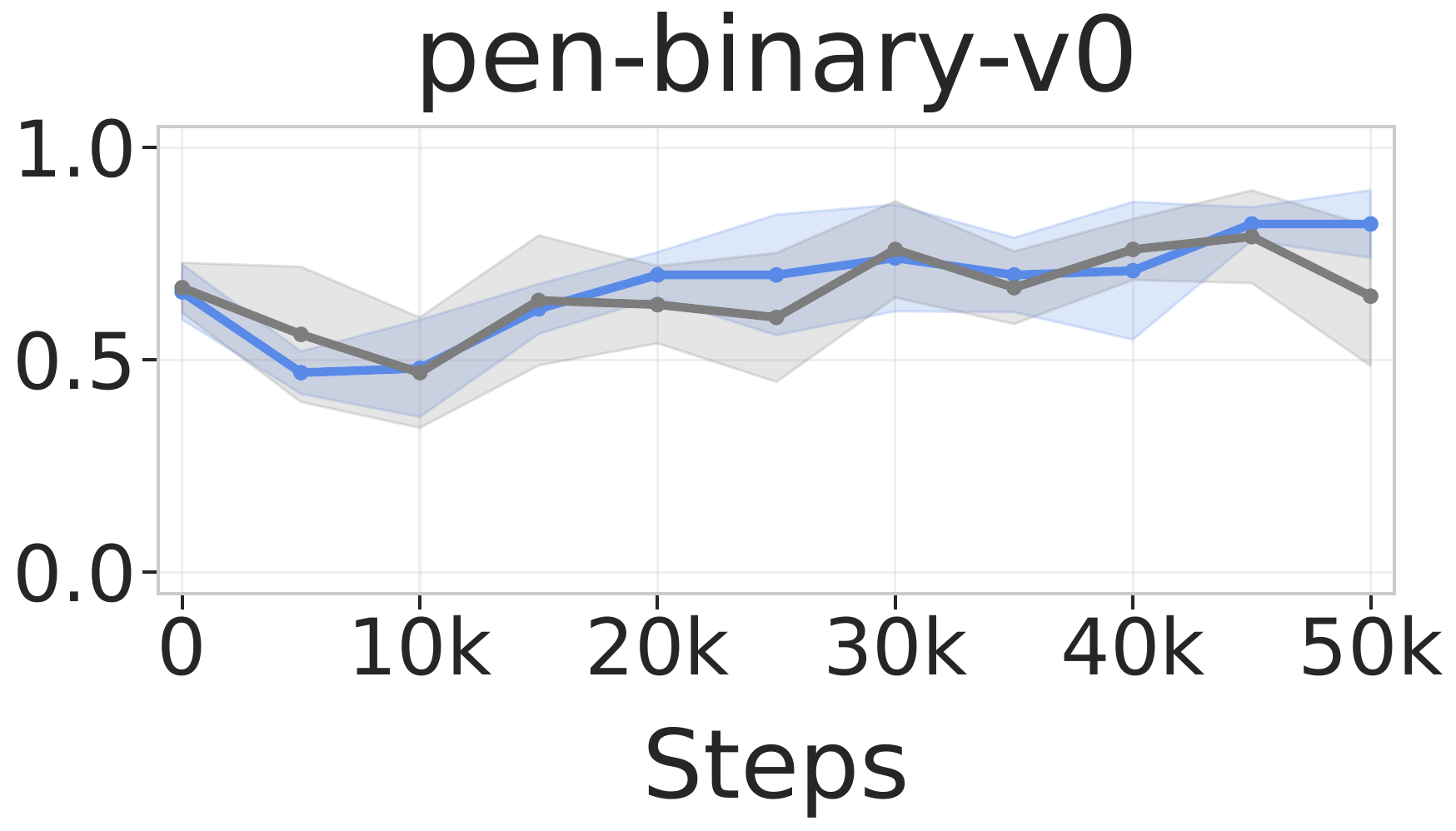}
    \end{subfigure}\\[1ex]
    \begin{subfigure}[c]{0.22\textwidth}
        \includegraphics[width=1.0\linewidth]{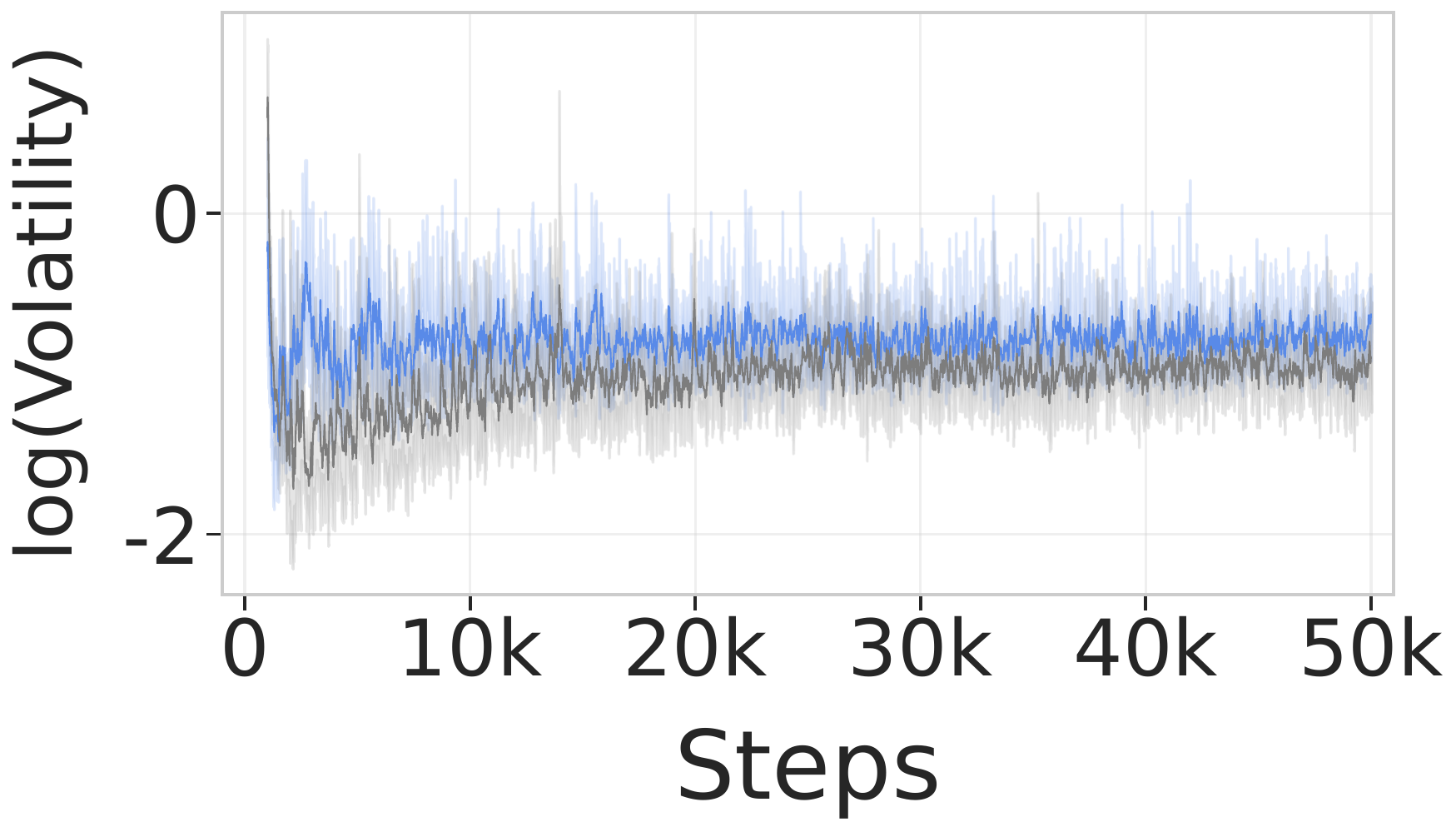}
    \end{subfigure}\hspace{0.1em}
    \begin{subfigure}[c]{0.22\textwidth}
        \includegraphics[width=1.0\linewidth]{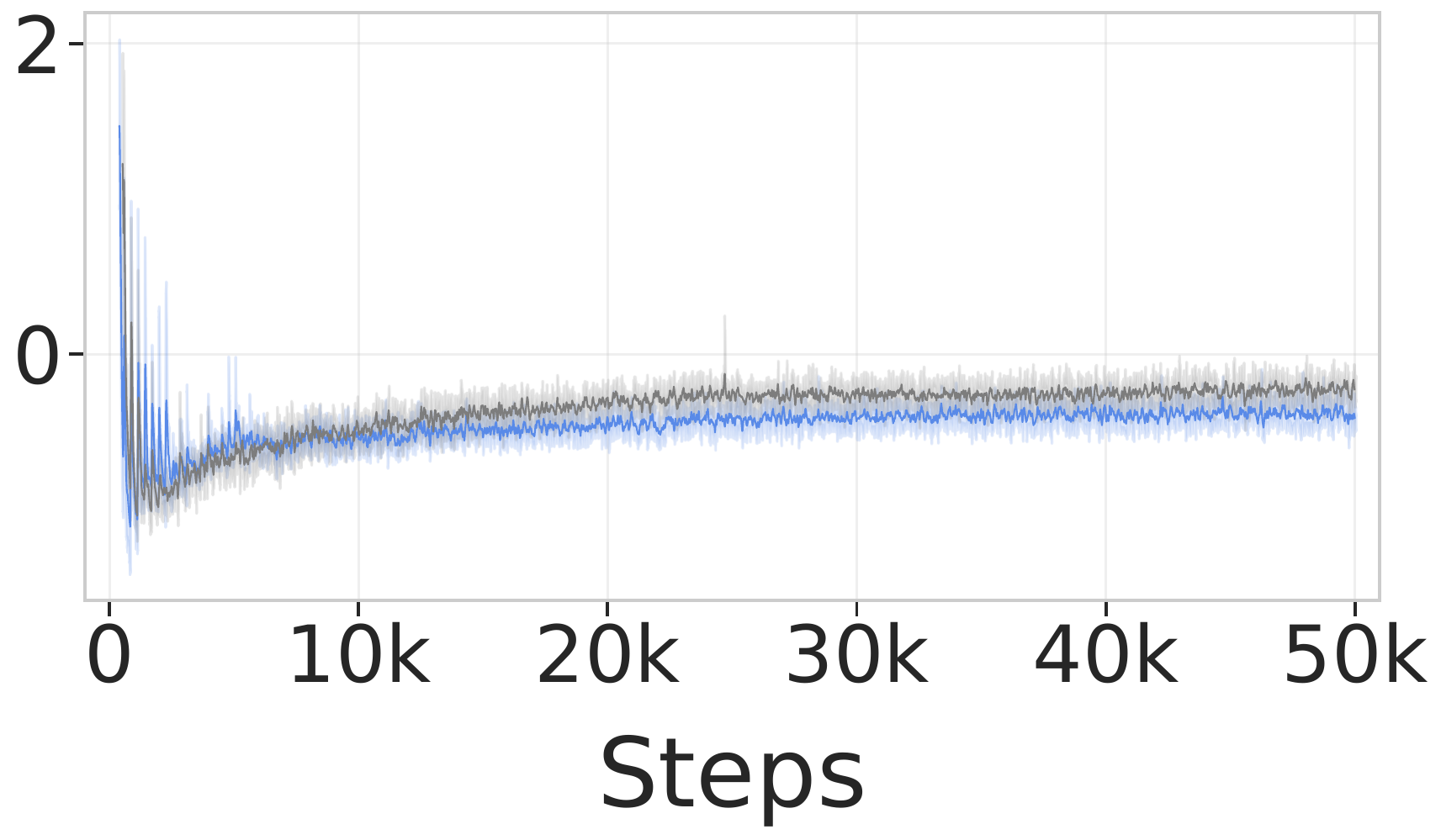}
    \end{subfigure}\hspace{0.1em}
    \begin{subfigure}[c]{0.22\textwidth}
        \includegraphics[width=1.0\linewidth]{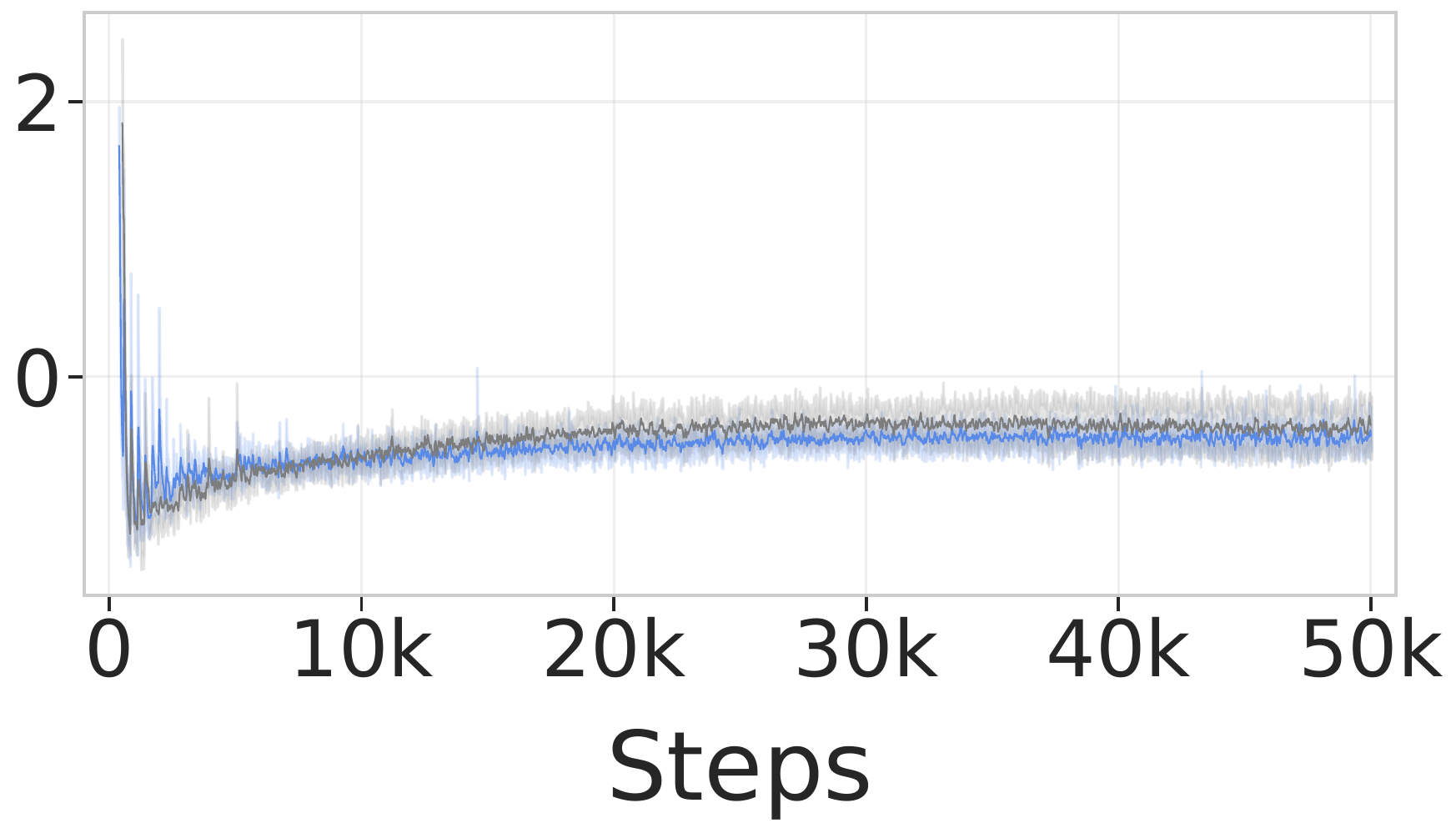}
    \end{subfigure}\hspace{0.1em}
    \begin{subfigure}[c]{0.22\textwidth}
        \includegraphics[width=1.0\linewidth]{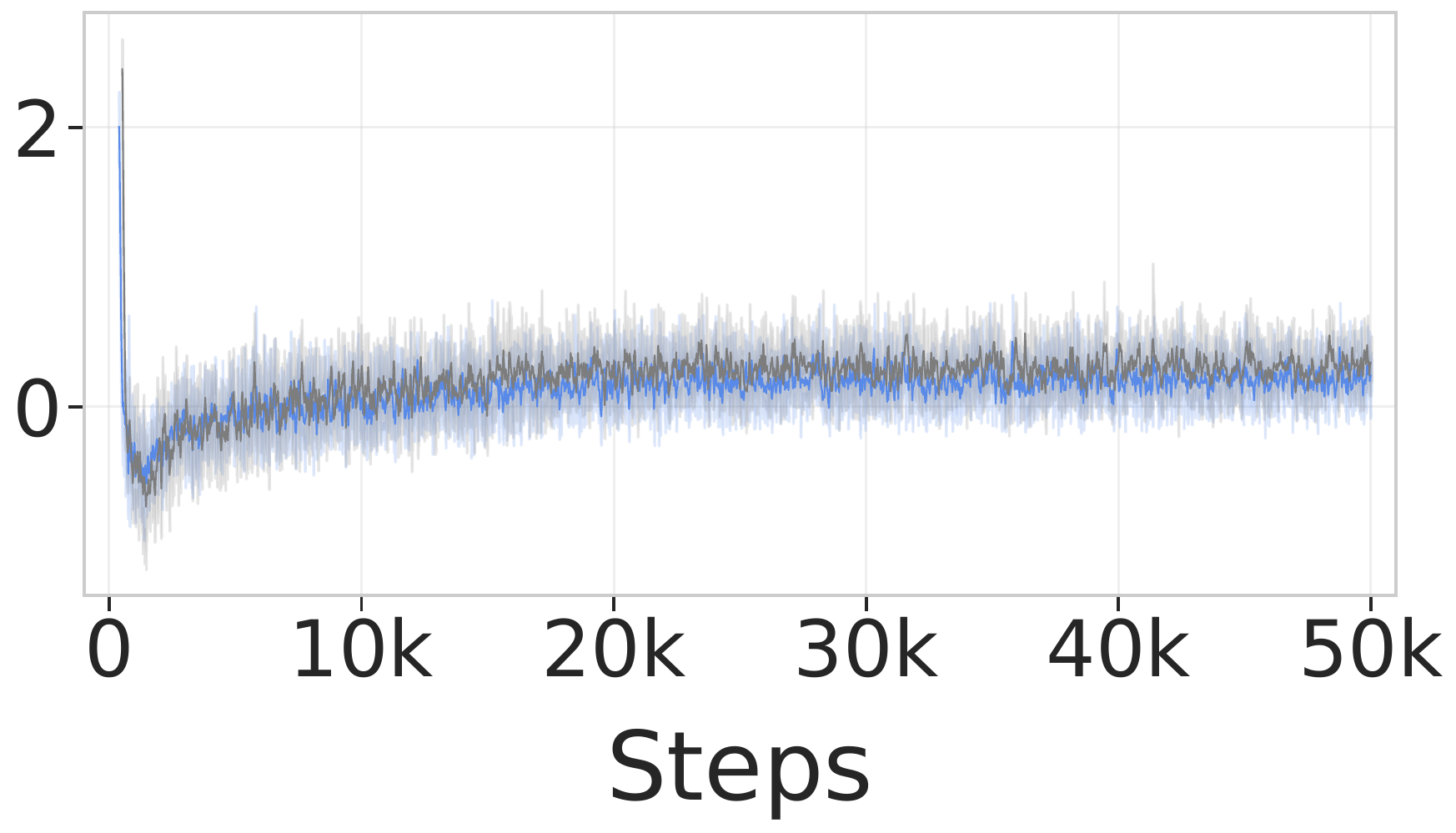}
    \end{subfigure}\\[1ex]

    \includegraphics[width=0.5\linewidth]
    {NeurIPS_figure/analysis/analysis_legend_appendix.pdf}
        \vspace{-1em}

    \caption{\textbf{Q-update volatility does not reliably predict catastrophic failure.}
    We compare success rate and critic-update volatility across four tasks. Volatility shows weak and inconsistent correspondence with early performance collapse.}
    \label{fig:appendix:additional_analysis_volatility}
\end{figure}

\textbf{Q-update volatility.}
We also examine whether abrupt critic updates explain early collapse. We define volatility
as the square root of the bias-corrected exponential moving average of squared one-step
Q-updates~\citep{kingma2014adam}:
\[
\mathrm{Volatility}_t
=
\sqrt{\frac{m_t}{1-\beta^t}},
\qquad
m_t
=
\beta m_{t-1}
+
(1-\beta)
\mathbb{E}_{(s,a)\sim\mathcal{B}_t}
\left[
\left(\Delta Q_t(s,a)\right)^2
\right],
\]
where $\Delta Q_t(s,a)=Q_{t+1}(s,a)-Q_t(s,a)$ and $\beta=0.9$. This metric measures the
aggregate magnitude of critic changes over minibatch state-action pairs.

Figure~\ref{fig:appendix:additional_analysis_volatility} shows that Q-update volatility
also fails to consistently align with catastrophic failure. High volatility does not always
lead to collapse, and severe collapse can occur without a distinctive volatility pattern.
Together with the critic-shift regularization results in Section~\ref{sec:analysis_catastrophic_failure}, these analyses do not rule out critic instability as a contributing factor. Rather, they suggest that critic-side stabilization alone may be insufficient to explain or prevent catastrophic failure, motivating our focus on policy-side stabilization through KL anchoring.

\newpage
\section{Additional Ablation Studies}
\label{sec:appendix:ablation_studies}

\subsection{Sensitivity to Batch Size and UTD Ratio}

\begin{figure}[H]
    \centering
    \begin{subfigure}[c]{0.1\textwidth}
        \includegraphics[width=1.0\textwidth]{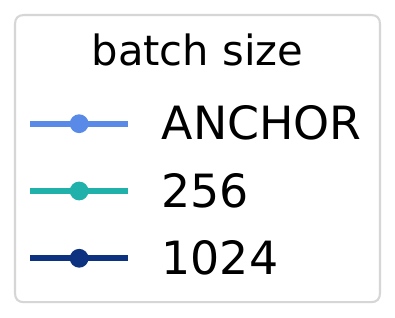}
    \end{subfigure}\hspace{0.1em}
    \begin{subfigure}[c]{0.26\textwidth}
        \includegraphics[width=1.0\linewidth]{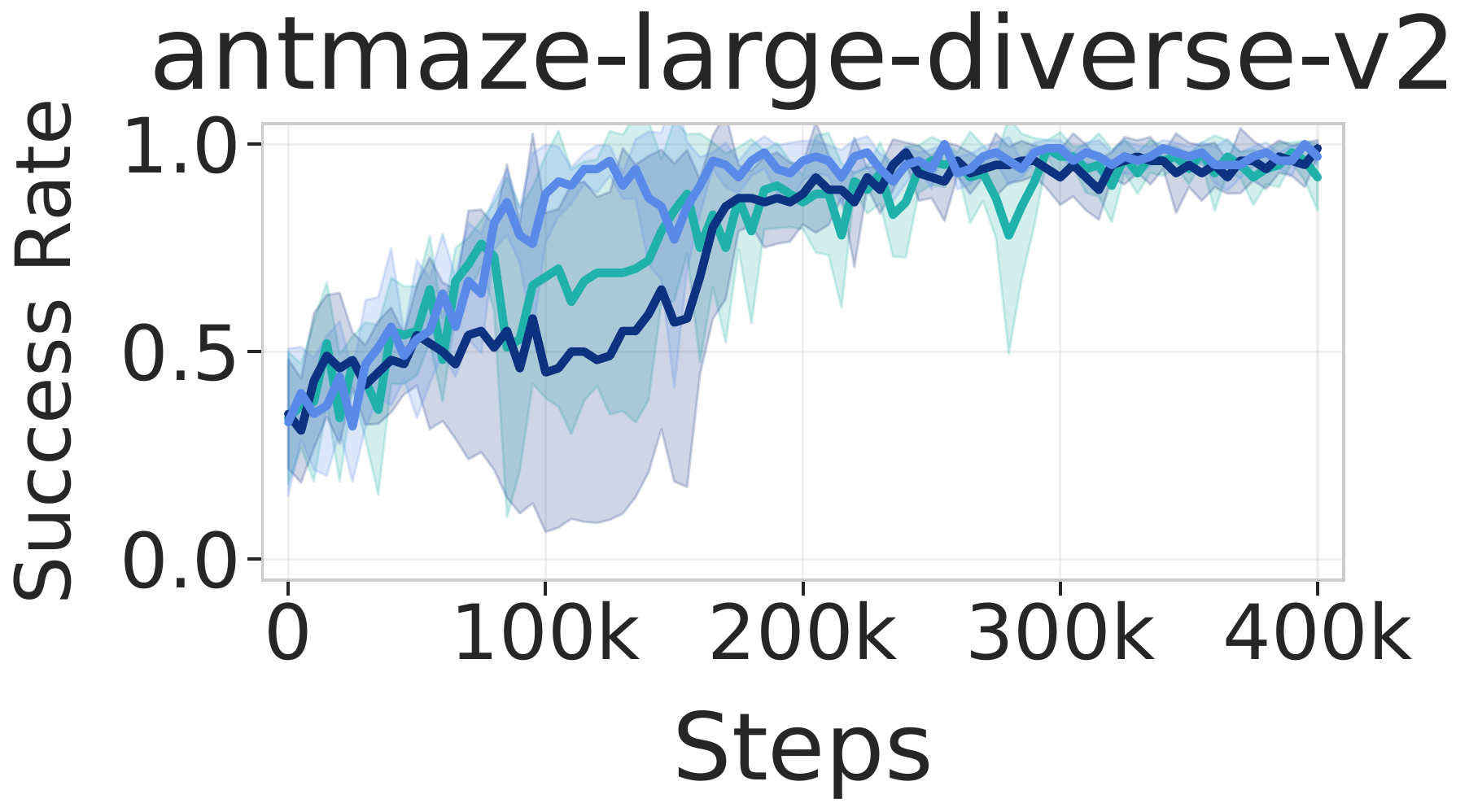}
    \end{subfigure}\hspace{0.1em}
    \begin{subfigure}[c]{0.26\textwidth}
        \includegraphics[width=1.0\linewidth]{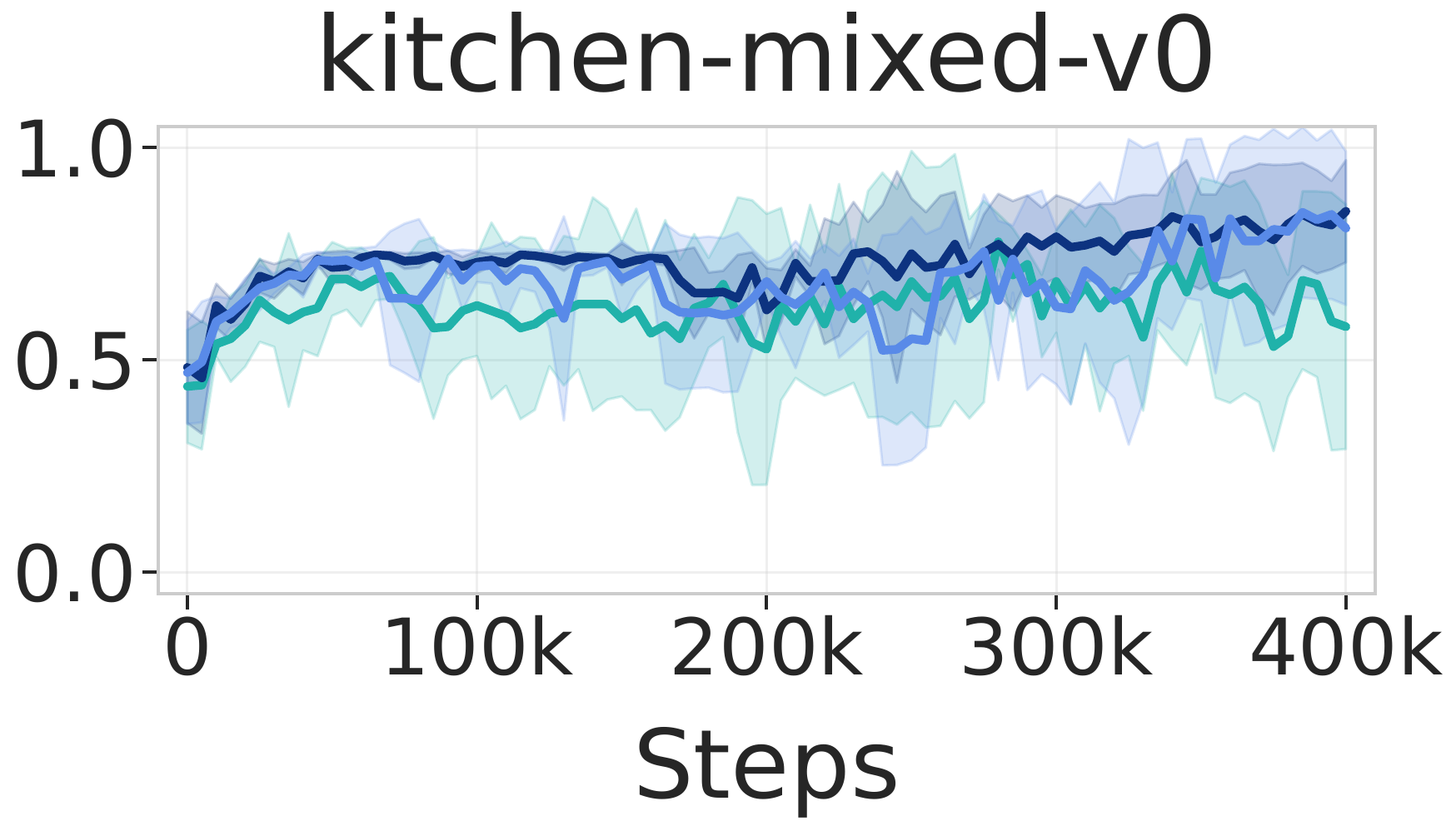}
    \end{subfigure}\hspace{0.1em}
    \begin{subfigure}[c]{0.26\textwidth}
        \includegraphics[width=1.0\linewidth]{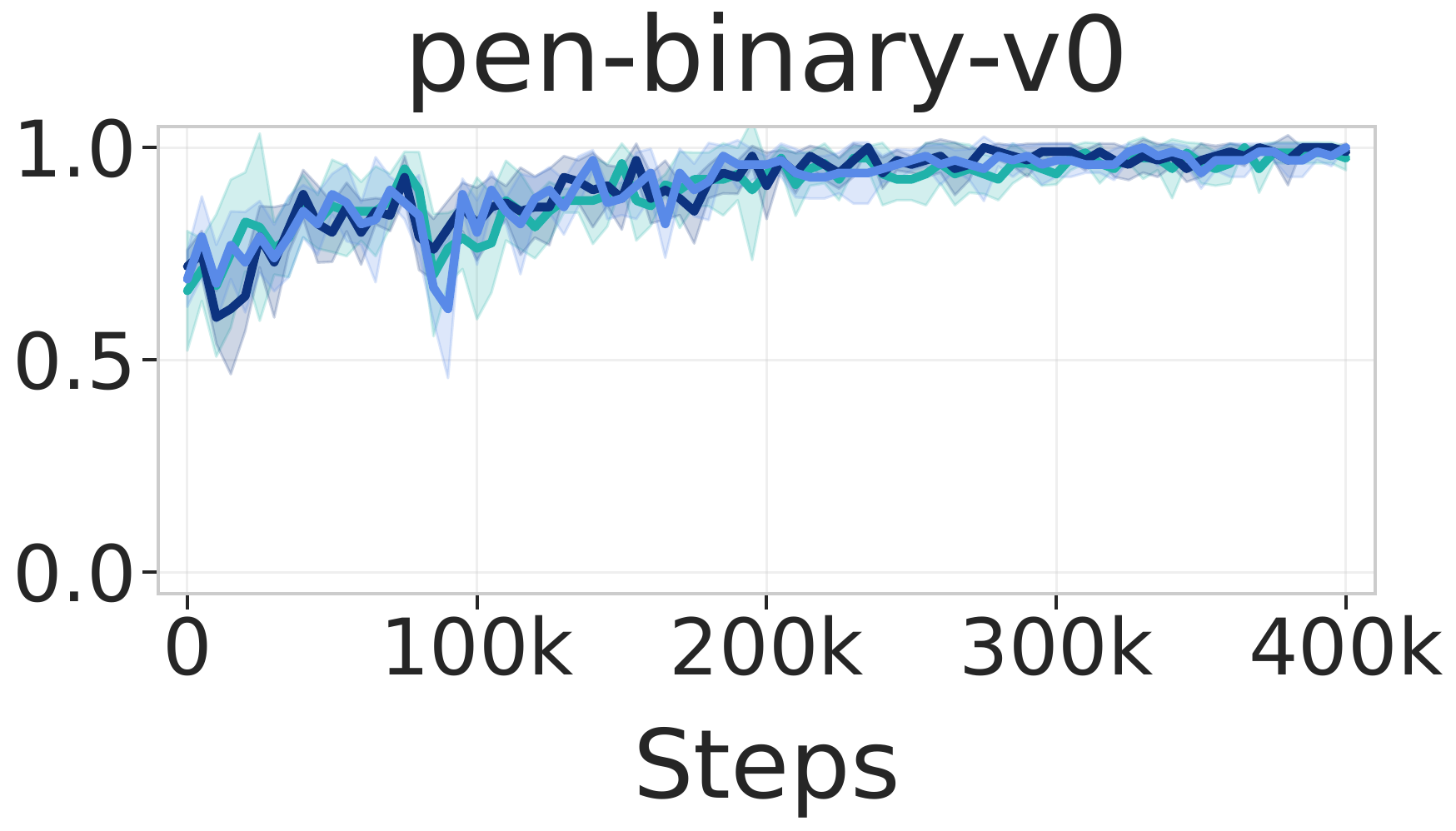}
    \end{subfigure}\\[1ex]
    \begin{subfigure}[c]{0.1\textwidth}
        \includegraphics[width=1.0\textwidth]{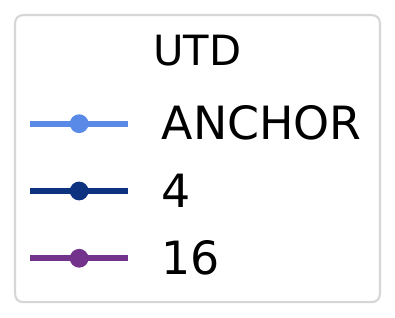}
    \end{subfigure}\hspace{0.1em}
    \begin{subfigure}[c]{0.26\textwidth}
        \includegraphics[width=1.0\linewidth]{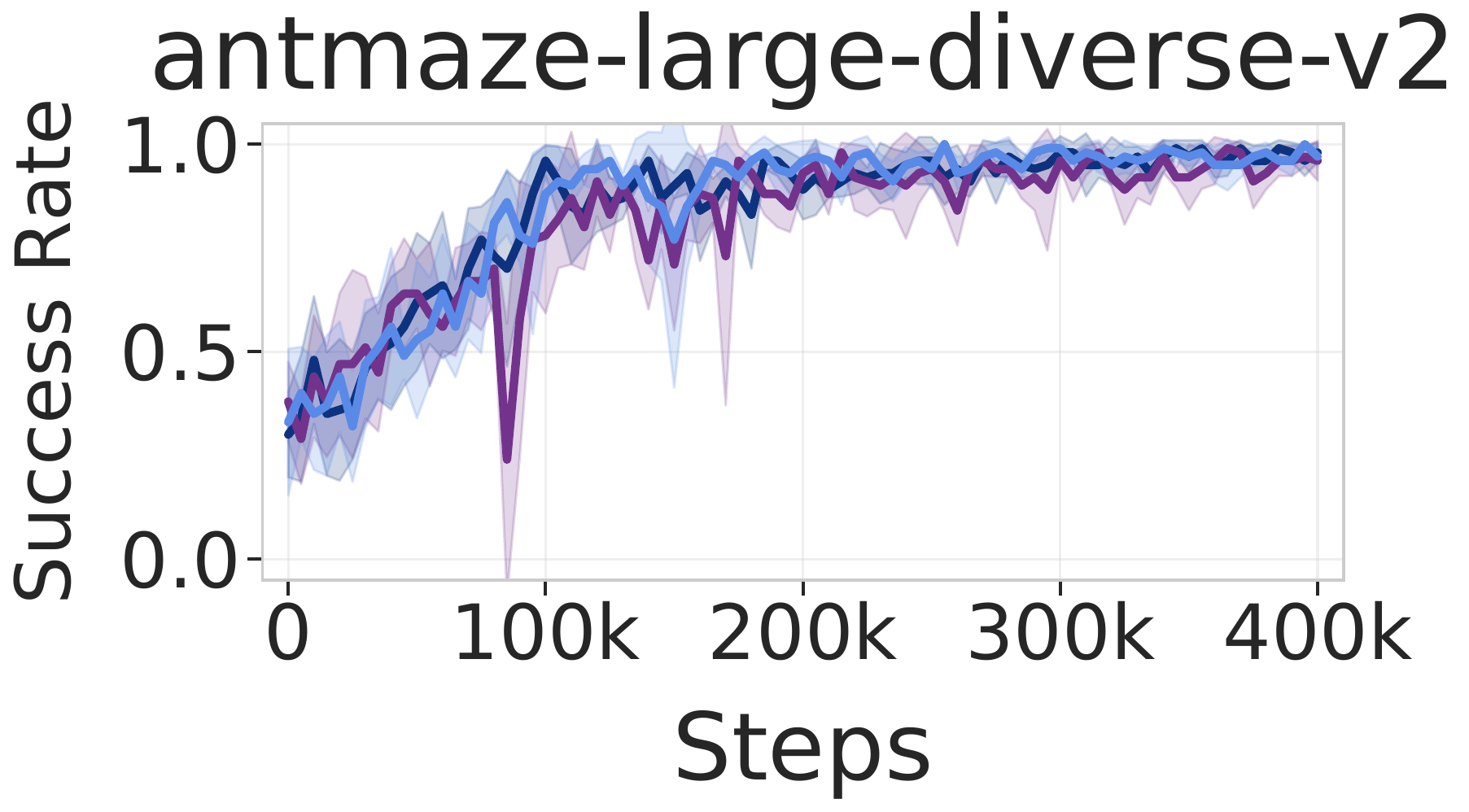}
    \end{subfigure}\hspace{0.1em}
    \begin{subfigure}[c]{0.26\textwidth}
        \includegraphics[width=1.0\linewidth]{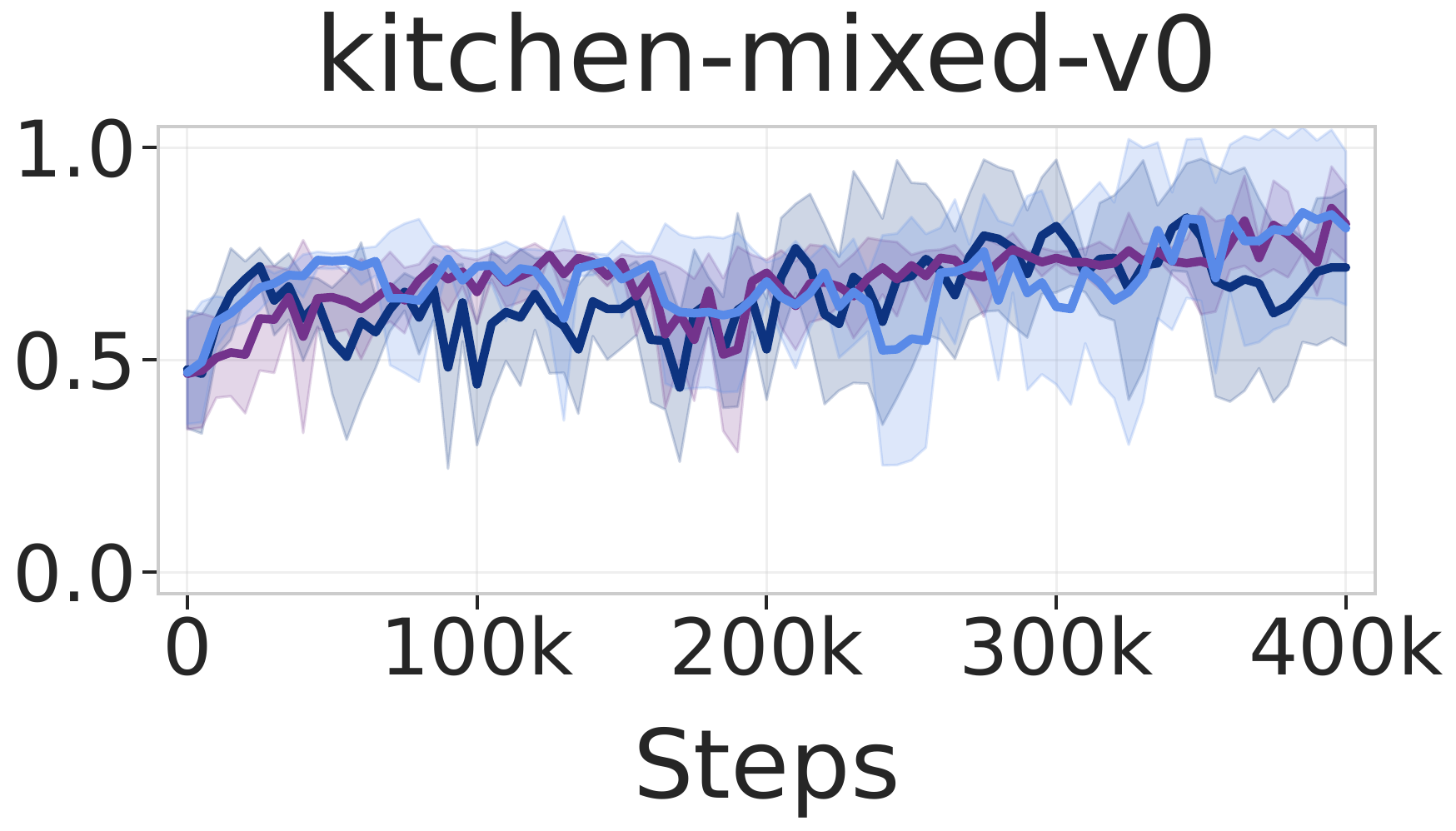}
    \end{subfigure}\hspace{0.1em}
    \begin{subfigure}[c]{0.26\textwidth}
        \includegraphics[width=1.0\linewidth]{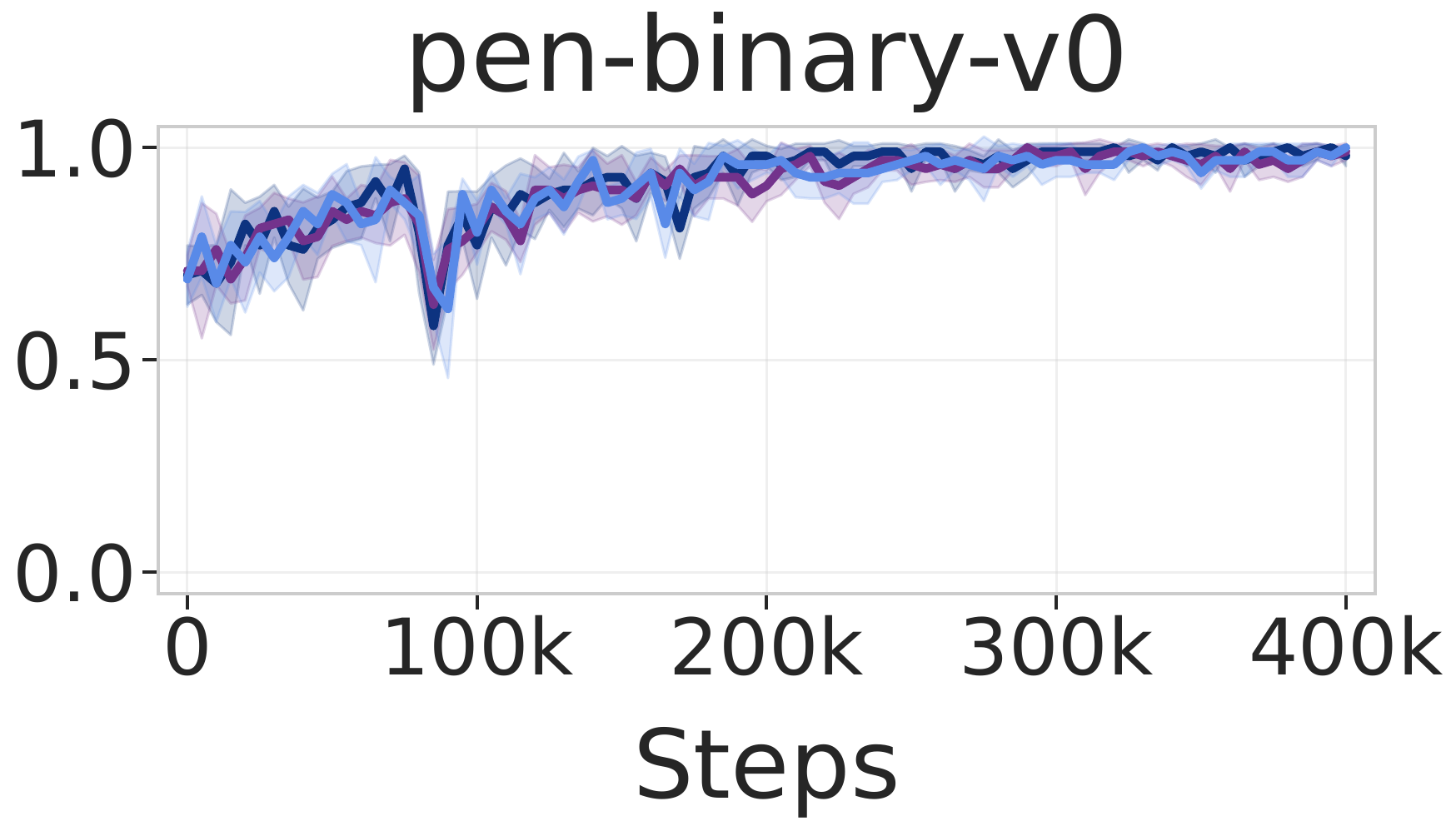}
    \end{subfigure}\hspace{0.1em}
    \caption{\textbf{Sensitivity to batch size and UTD ratio.}
    We vary the batch size (top) and UTD ratio (bottom) across three tasks. ANCHOR is
    generally robust to these choices, although larger batch size and larger UTD can
    increase learning instability.}
    \label{fig:appendix_anchor_ablation}
\end{figure}

Figure~\ref{fig:appendix_anchor_ablation} further evaluates the sensitivity of ANCHOR to
batch size and UTD ratio. Overall, ANCHOR remains robust across the tested variants in
kitchen-mixed and pen-binary, with similar early stability and asymptotic performance. In
antmaze-large-diverse, however, the larger batch size of $1024$ and the larger UTD ratio of
$16$ lead to greater learning instability and weaker prevention of catastrophic failure.
Nevertheless, compared with the baselines in Figure~\ref{fig:baseline_performance}, these
variants still achieve competitive catastrophic failure and asymptotic performance. These
results indicate that ANCHOR is not highly sensitive to batch size or UTD ratio, although
overly aggressive update configurations can reduce early stability.

\subsection{A Larger Initial KL Weight Yields Stronger Early Anchoring}
\begin{figure}[H]
    \centering
    \hspace{-1.0em}
    \begin{subfigure}[c]{0.08\textwidth}
        \hfill
        \includegraphics[width=1.0\linewidth]{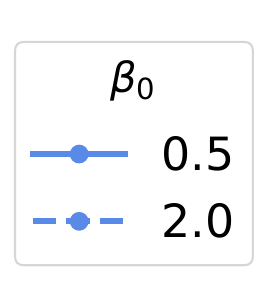}
    \end{subfigure}\hspace{1.0em}
    \begin{subfigure}[c]{0.24\textwidth}
        \includegraphics[width=1.0\linewidth]{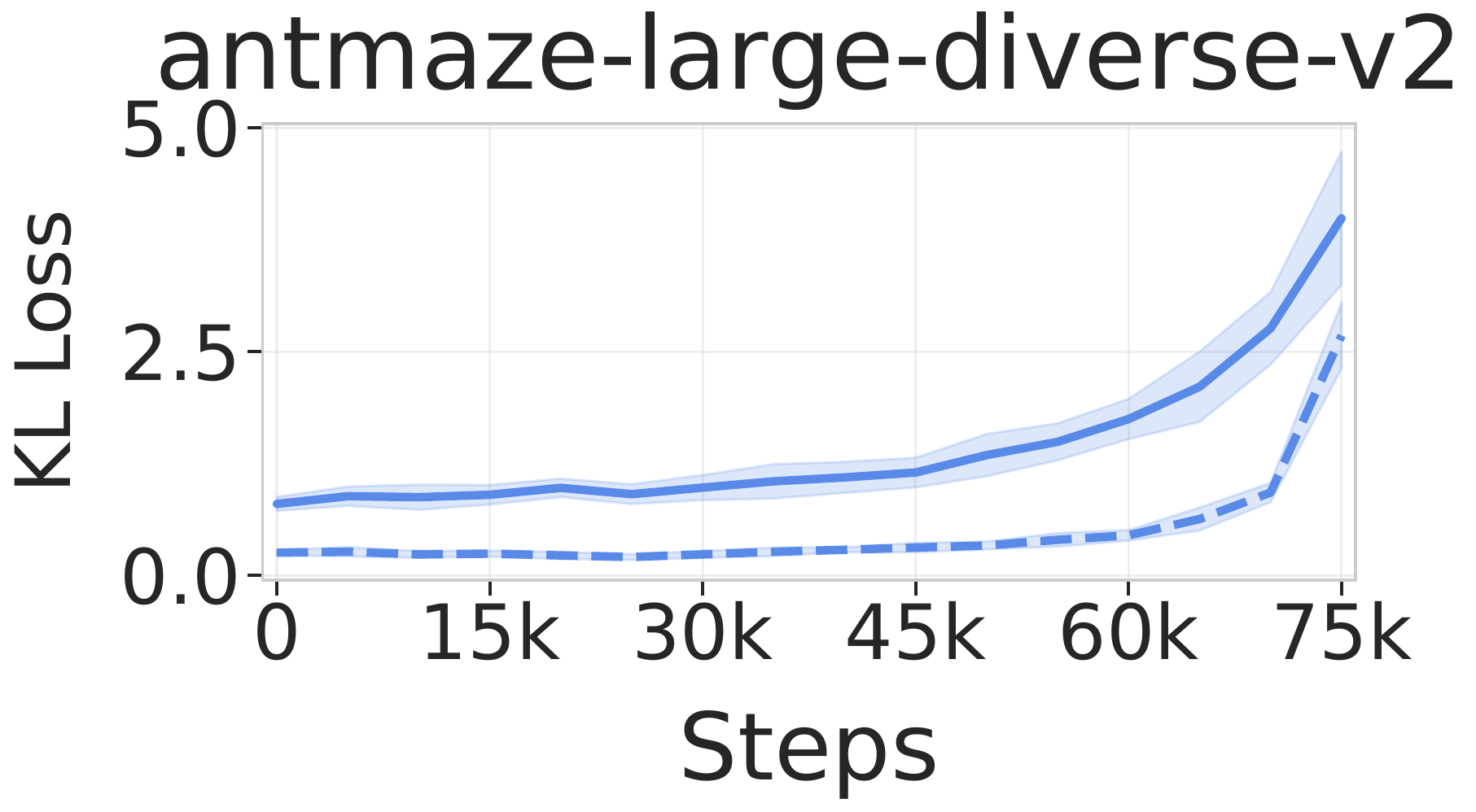}
    \end{subfigure}\hspace{0.1em}
    \begin{subfigure}[c]{0.24\textwidth}
        \includegraphics[width=1.0\linewidth]{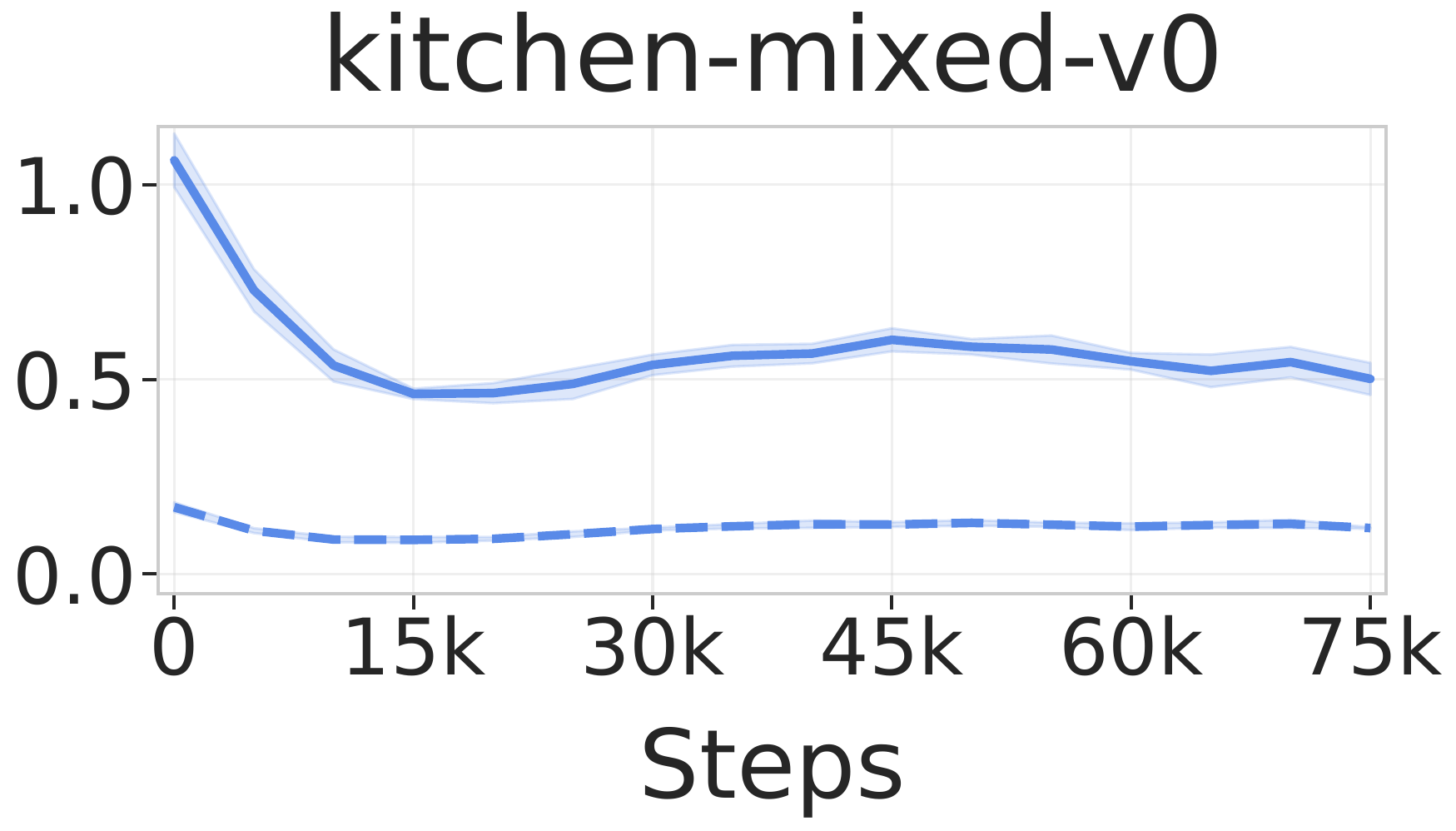}
    \end{subfigure}\hspace{0.1em}
    \begin{subfigure}[c]{0.24\textwidth}
        \includegraphics[width=1.0\linewidth]{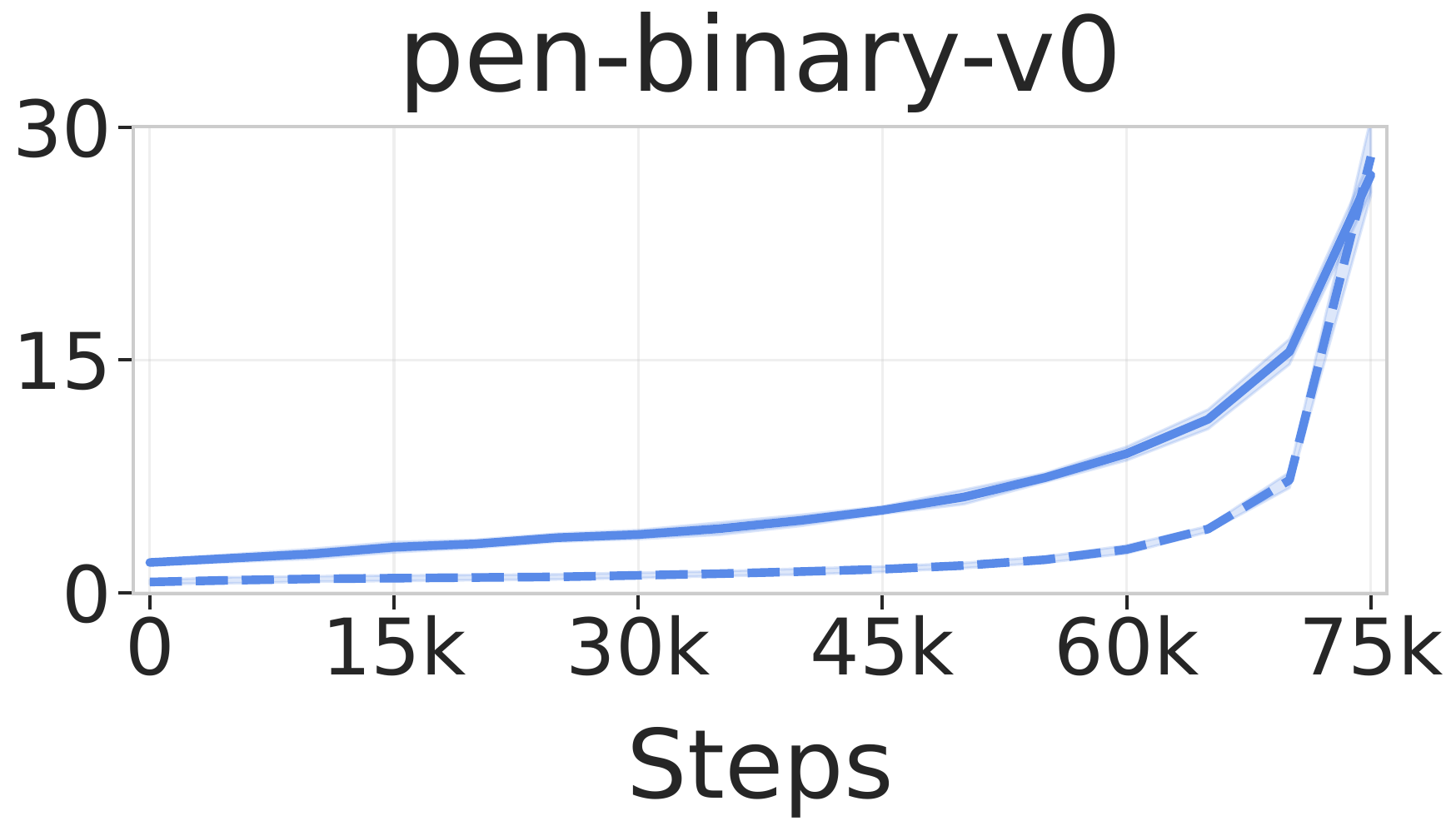}
    \end{subfigure}\hspace{0.1em}
    \caption{\textbf{Effect of the initial KL weight.}
    A larger initial KL weight ($\beta_0=2.0$) produces lower KL loss early in online fine-tuning, indicating that the KL penalty more strongly constrains deviation from the offline policy. As the KL coefficient is annealed, the gap between $\beta_0=0.5$ and $\beta_0=2.0$ gradually narrows.}
    \label{fig:kl_anchoring}
    \vspace{-1em}
\end{figure}

We check whether increasing the KL coefficient actually reduces the measured deviation between the online actor and the offline-pretrained policy. This is not automatic in practice, since the KL term is optimized jointly with the online RL objective and may not be minimized effectively under online updates. We compare two initial KL weights, $\beta_0=0.5$ and $\beta_0=2.0$. Figure~\ref{fig:kl_anchoring} reports the KL loss during online fine-tuning.

Across all three tasks, the larger initial KL weight consistently produces lower KL loss during the early phase of online adaptation. This verifies that $\beta_0$ controls the effective strength of actor anchoring in practice: increasing $\beta_0$ keeps the online actor closer to the offline-pretrained policy despite simultaneous optimization of the RL objective. As training proceeds and the KL coefficient is annealed, the KL losses under the two settings become increasingly similar. This behavior is consistent with the design of ANCHOR, where KL anchoring is strongest early in online fine-tuning and gradually weakens as standard online improvement takes over.
\section{Can ANCHOR Extend to Other Offline RL Backbones?}
\label{sec:appendix:ours_with_other_offline_rl}

\begin{figure}[H]
    \centering
    \begin{subfigure}[c]{0.45\textwidth}
        \includegraphics[width=1.0\linewidth]{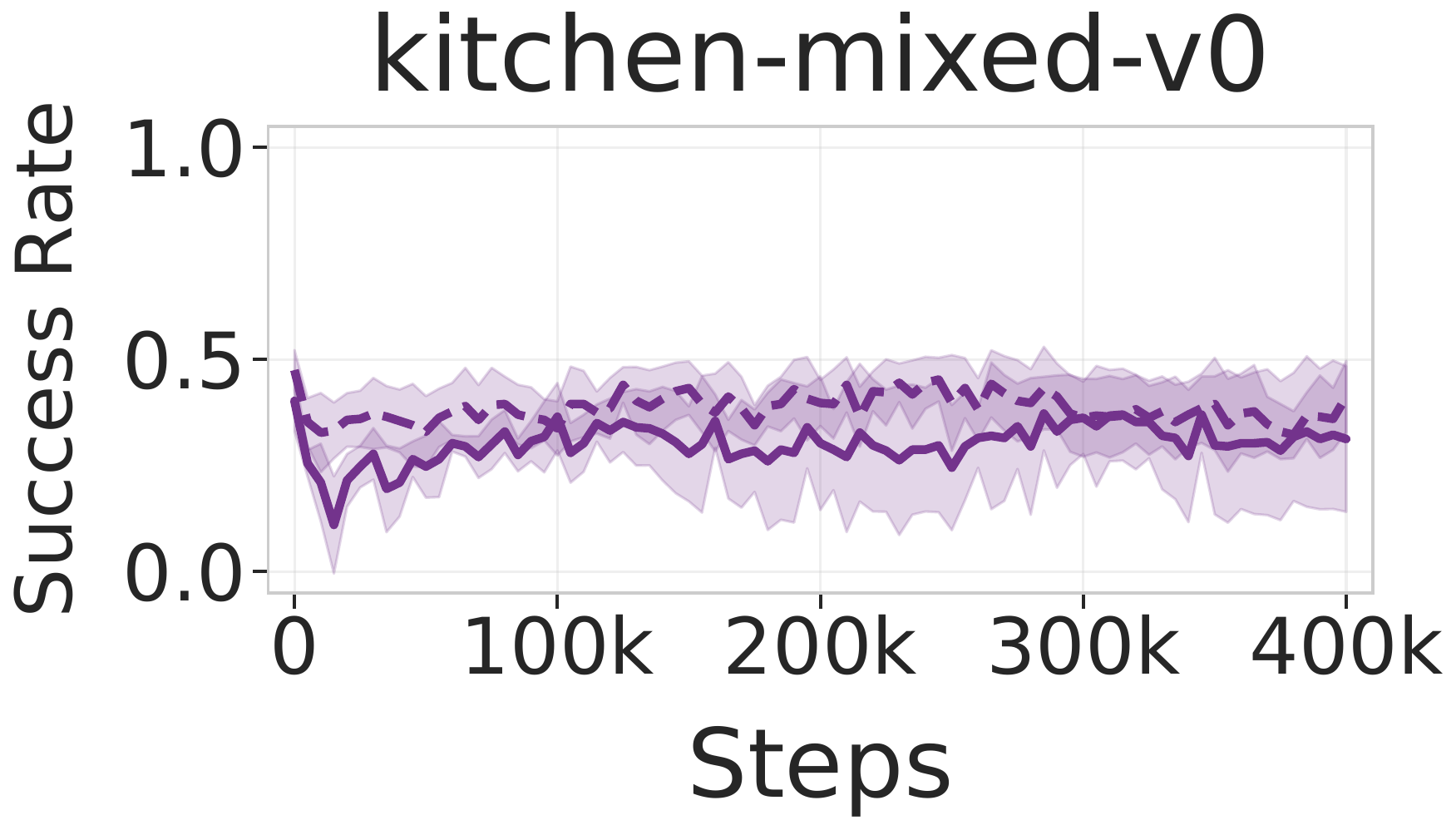}
    \end{subfigure}\hspace{0.1em}
    \begin{subfigure}[c]{0.45\textwidth}
        \includegraphics[width=1.0\linewidth]{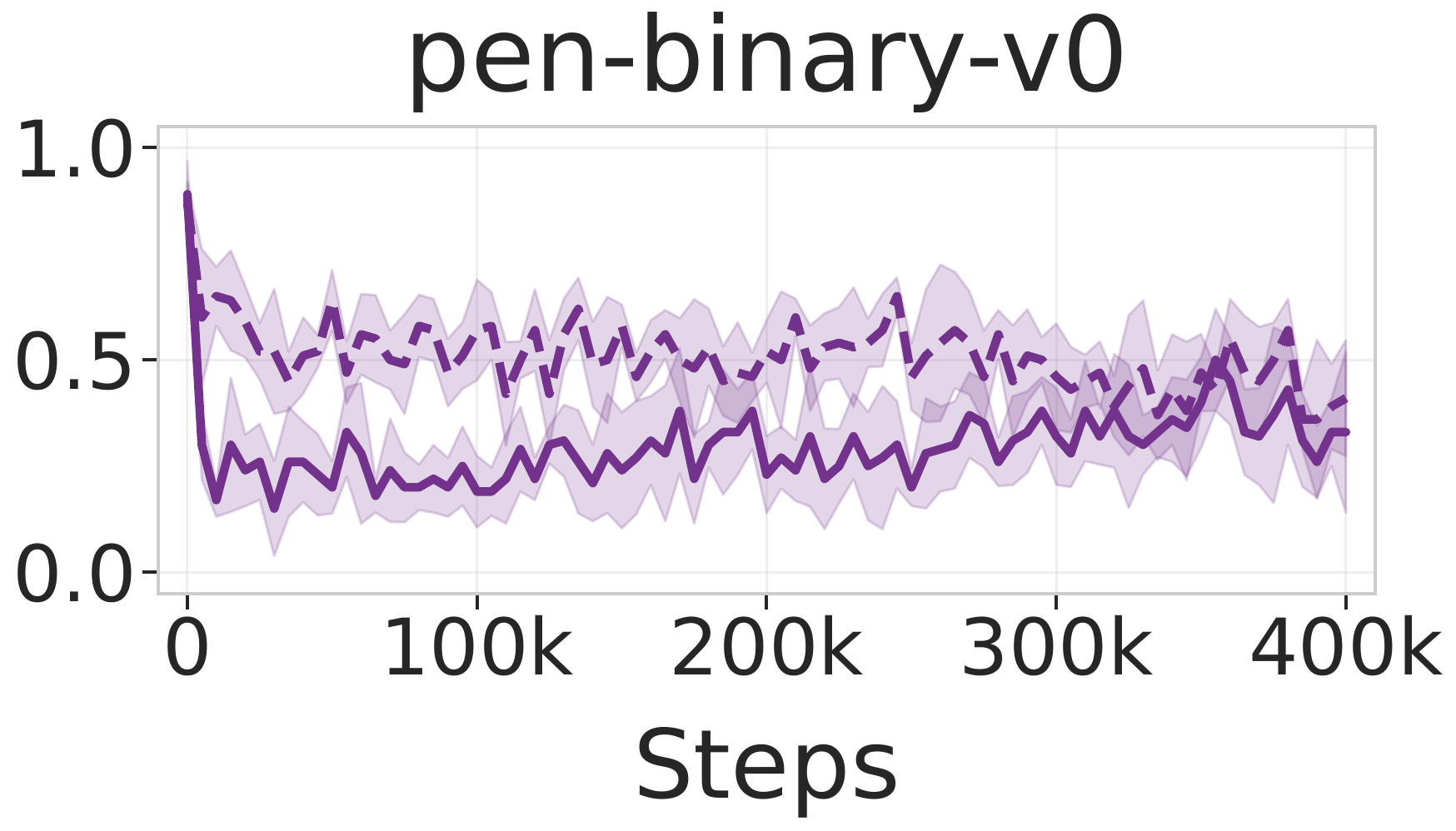}
    \end{subfigure}\\[1ex]
    \begin{minipage}{0.4\textwidth}
        \centering
        \includegraphics[width=\textwidth]{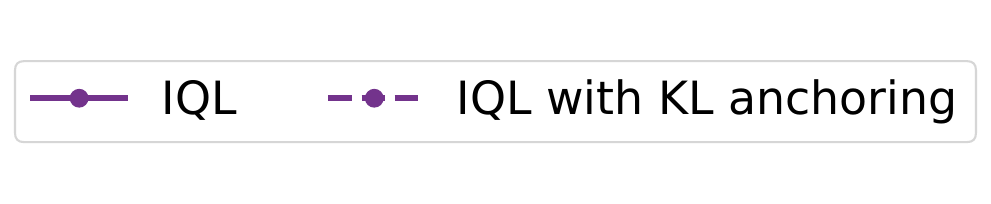}
    \end{minipage}
    \caption{\textbf{Applying KL anchoring to IQL.}
    Adding KL anchoring with annealing to IQL reduces the initial performance drop, but does not improve asymptotic performance over the IQL baseline.}
    \label{fig:iql_kl_annealing}
\end{figure}

We additionally test whether the actor-side KL anchoring used in ANCHOR can be applied to another offline RL backbone. Specifically, we apply KL anchoring with annealing to IQL. Since IQL does not use an explicit CQL-style conservative regularizer, we only add the actor-side KL term toward the offline-pretrained policy and anneal its coefficient during online fine-tuning. We conduct experiments on kitchen-mixed and pen-binary, where IQL exhibits nontrivial pretrained behavior with an initial success rate above $0.4$. For both
tasks, we use $\beta_0=5.0$ and $T_{\mathrm{KL}}=320\mathrm{K}$.

Figure~\ref{fig:iql_kl_annealing} shows that KL anchoring can also stabilize IQL during the early online phase. Compared to naive online fine-tuning of IQL, the KL-anchored variant reduces the initial performance drop, indicating that actor-side anchoring is not specific to the CQL backbone. However, the asymptotic performance does not improve substantially over the IQL baseline. We hypothesize that this is because IQL retains its offline inductive bias during online fine-tuning, which can limit its ability to adapt to the online data distribution even when the actor is stabilized by KL anchoring. Developing backbone-specific mechanisms that remove or relax such offline biases while preserving early stability is an interesting direction for future work.
\section{Additional Environments and Failure Cases}
\label{sec:appendix:additional_environments}

We further evaluate ANCHOR in three additional environments that highlight challenging
regimes for offline-to-online adaptation: kitchen-complete, relocate-binary, and
antmaze-ultra-diverse. The first two environments have near-zero initial success after
offline pretraining, leaving little useful pretrained behavior to preserve. In contrast,
antmaze-ultra-diverse is a larger and more complex maze than antmaze-large-diverse, where
sparse rewards and severe exploration mismatch make online improvement difficult. The offline pretraining duration follows the domain-specific setup used in the main experiments: 250K steps for kitchen-complete, 40K steps for relocate-binary, and 1M steps for antmaze-ultra-diverse.

\begin{figure}[h]
    \centering
    \begin{subfigure}[c]{0.30\textwidth}
        \includegraphics[width=1.0\linewidth]{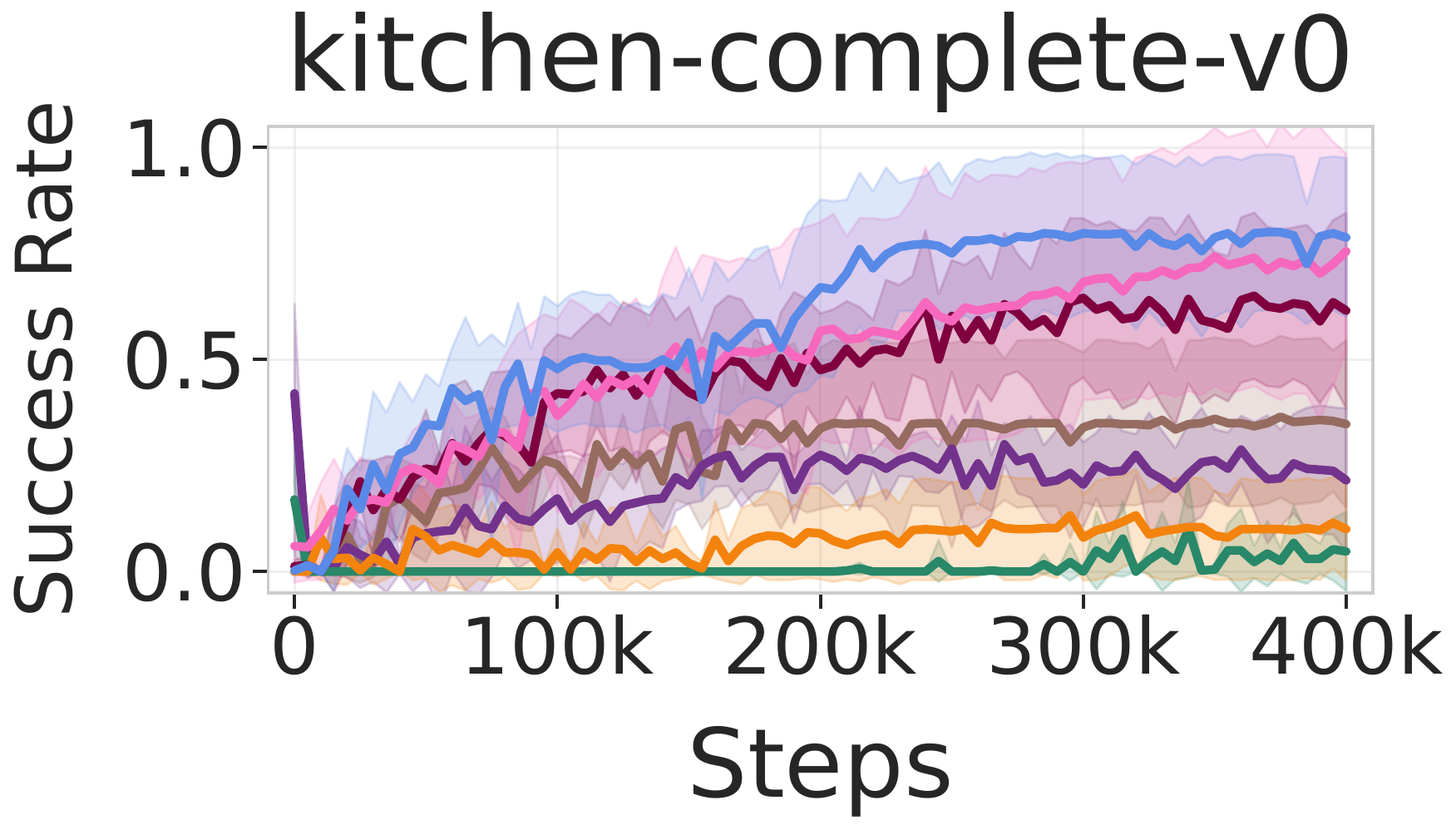}
    \end{subfigure}\hspace{0.1em}
    \begin{subfigure}[c]{0.30\textwidth}
        \includegraphics[width=1.0\linewidth]{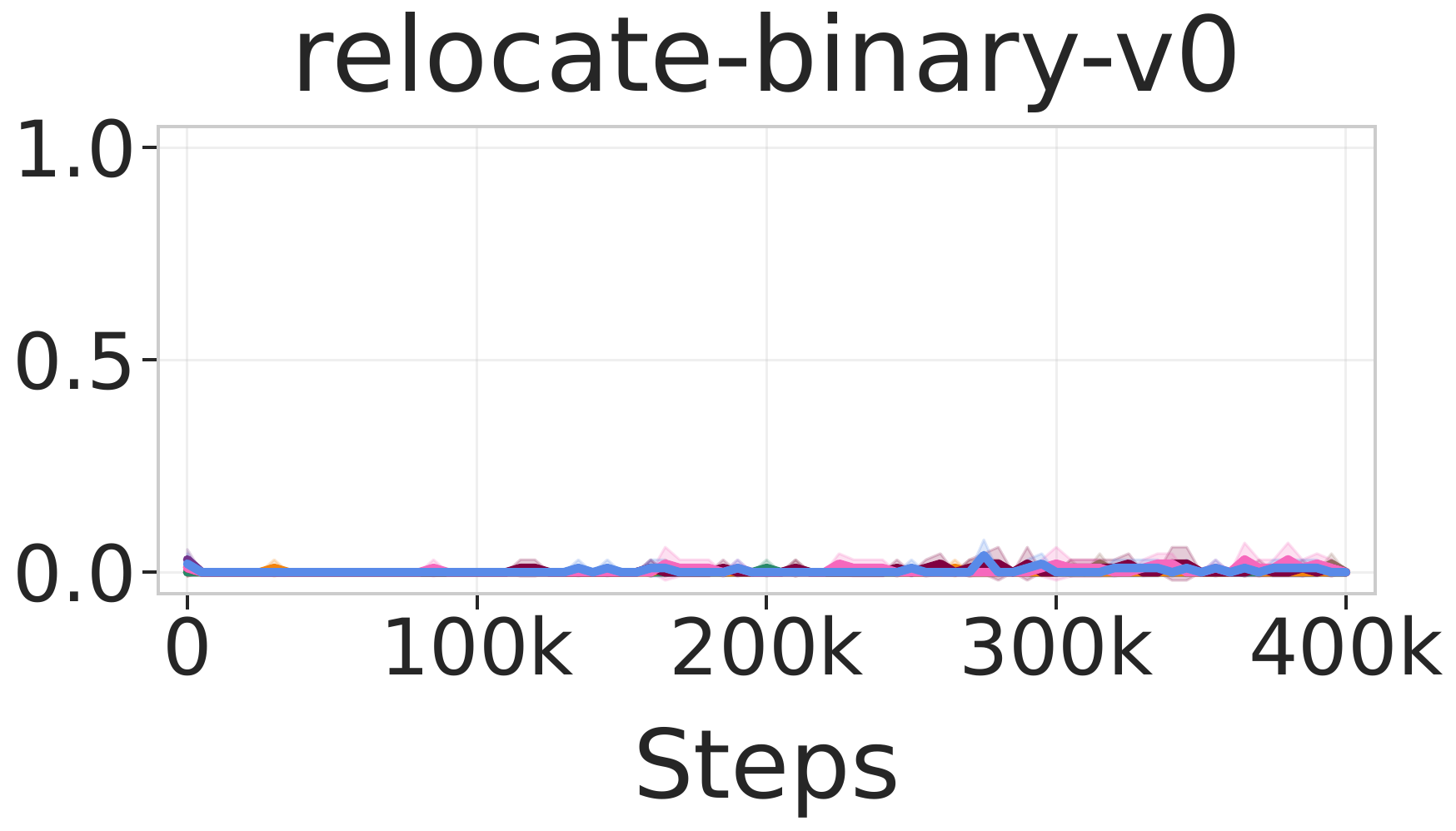}
    \end{subfigure}\hspace{0.1em}
    \begin{subfigure}[c]{0.30\textwidth}
        \includegraphics[width=1.0\linewidth]{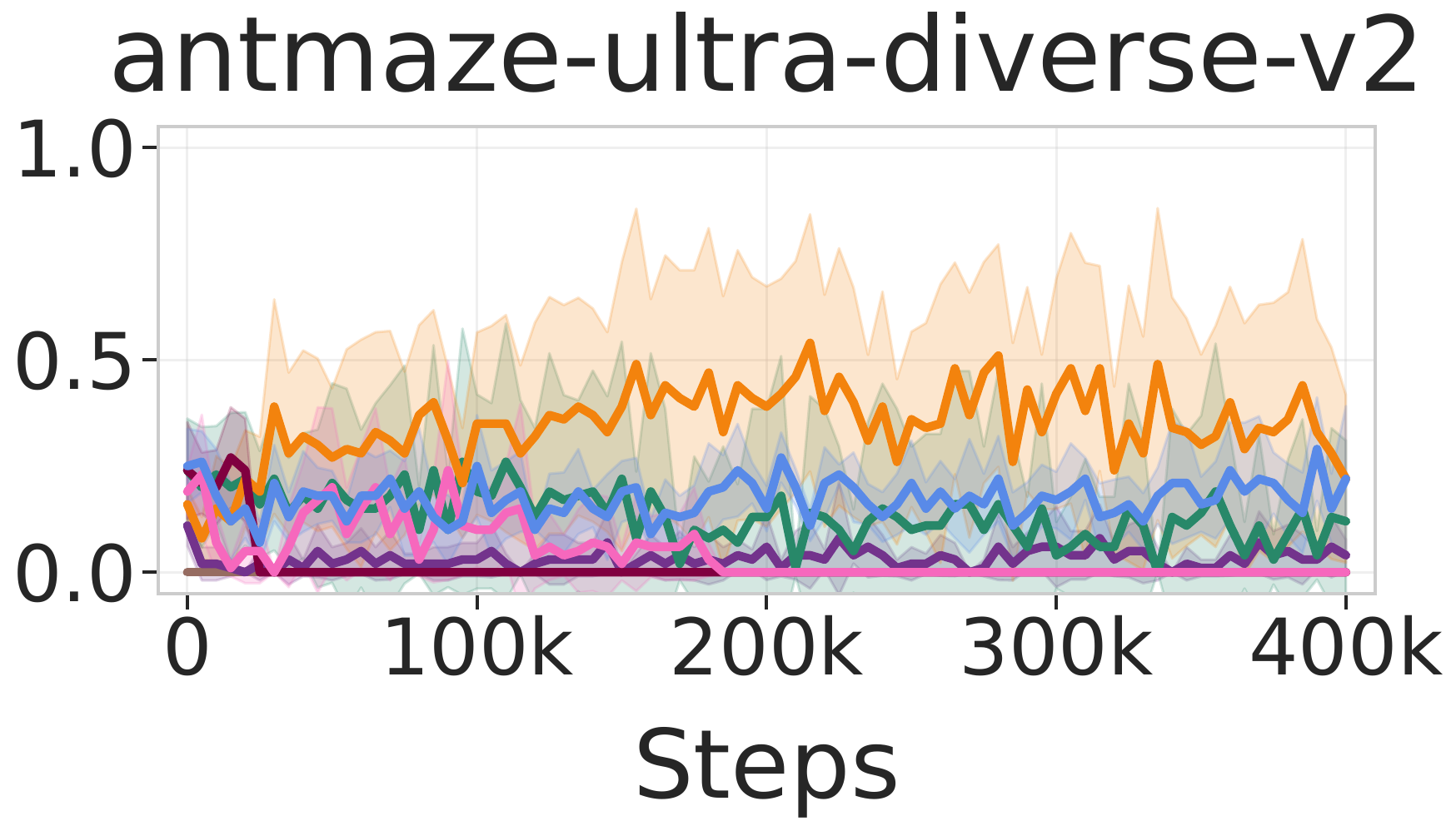}
    \end{subfigure}\\[1ex]
    \begin{minipage}{1.0\textwidth}
        \centering
        \includegraphics[width=0.9\linewidth]{NeurIPS_figure/baseline/baseline_legend.pdf}
    \end{minipage}
    \caption{\textbf{Additional environments reveal limitations of KL-anchored adaptation.}
    In kitchen-complete, ANCHOR remains competitive despite near-zero initial success. In
    relocate-binary, all methods fail to make meaningful progress. In antmaze-ultra-diverse,
    most methods do not improve over the offline-pretrained policy, suggesting that harder
    sparse-reward exploration requires mechanisms beyond KL anchoring alone.}
    \label{fig:additional_environments}
\end{figure}

For all three environments, we set the residual conservative regularization coefficient
$\alpha_{\mathrm{on}}$ to zero during online fine-tuning. Since kitchen-complete and
relocate-binary start with near-zero success rates, we set the initial KL weight to
$\beta_0=0$, reflecting the fact that there is little useful pretrained behavior to
preserve. In kitchen-complete, ANCHOR attains the highest mean final performance and
matches WSRL and PORL when confidence intervals are considered, while outperforming the
remaining baselines. This behavior is consistent with Lemma~\ref{thm:2}: when the
pretrained policy is weak, keeping the actor close to the offline policy can be costly, and
the KL constraint should be relaxed or removed to enable online improvement. In
relocate-binary, however, all methods remain near zero success throughout online
fine-tuning, suggesting that this environment requires additional exploration or adaptation
mechanisms beyond the scope of ANCHOR.

For antmaze-ultra-diverse, we use $\beta_0=1.0$ and set the KL annealing interval to
$T_{\mathrm{KL}}=400\mathrm{K}$, covering the full online fine-tuning phase. Despite this
longer anchoring schedule, ANCHOR's final asymptotic performance remains below its initial
offline-pretrained performance. Except for Cal-QL, none of the methods achieve meaningful
improvement over the offline-pretrained success rate. This suggests that in extremely
challenging sparse-reward environments, preserving the pretrained policy alone may be
insufficient; the agent may also require exploration or online adaptation mechanisms.

\begin{figure}[h]
    \centering
    \begin{subfigure}[c]{0.50\textwidth}
        \includegraphics[width=1.0\linewidth]{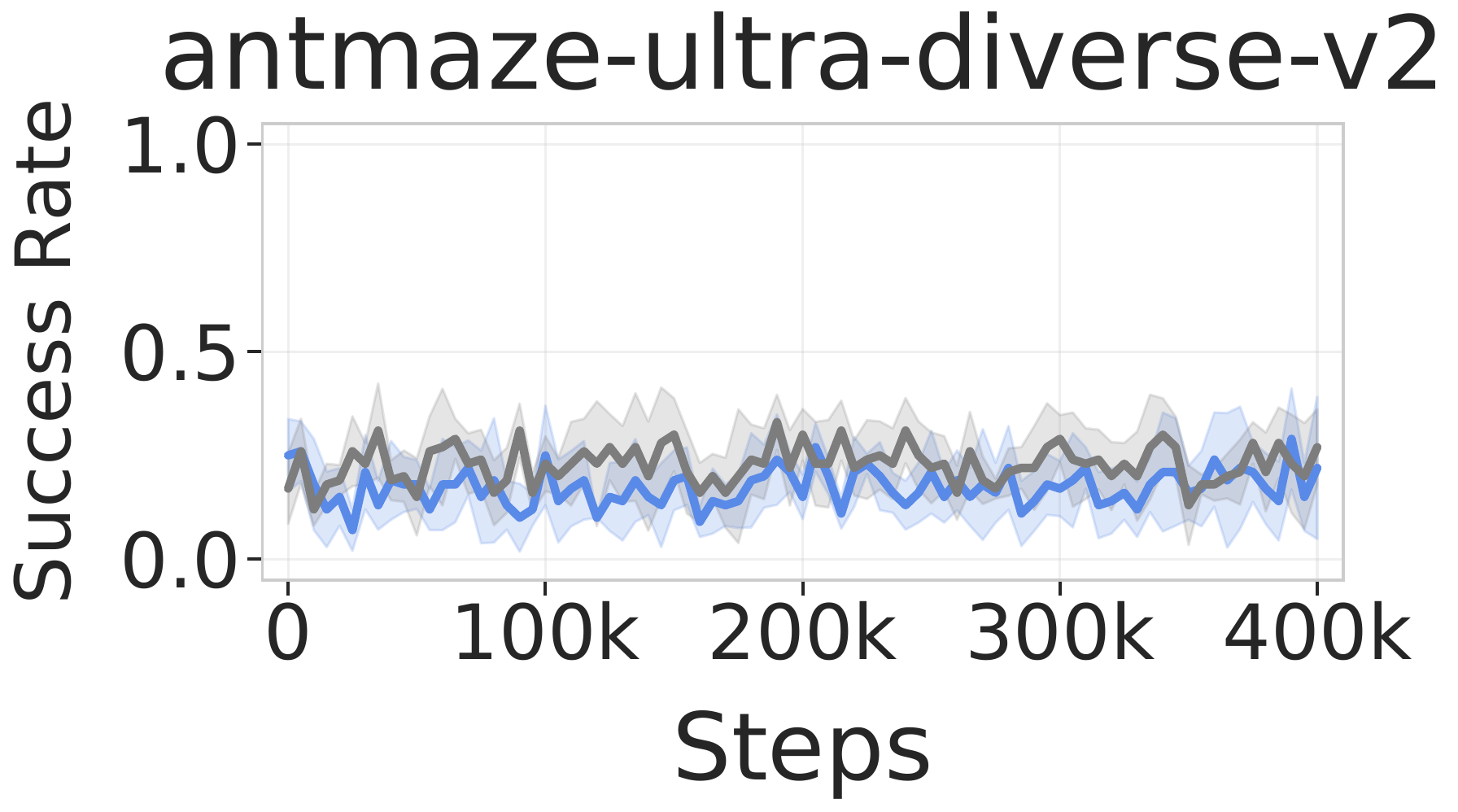}
    \end{subfigure}\hspace{0.1em}
    \begin{minipage}{1.0\textwidth}
        \centering
        \includegraphics[width=0.55\linewidth]{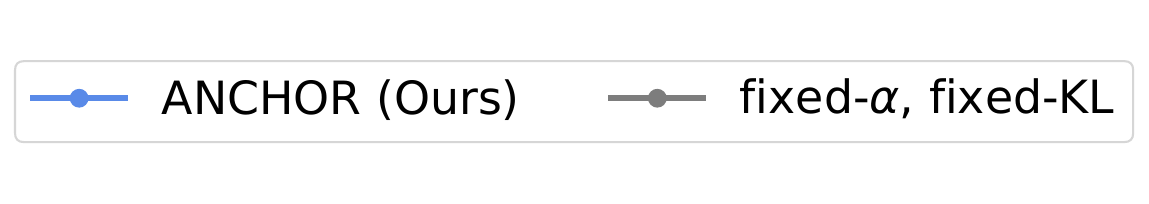}
    \end{minipage}
    \caption{\textbf{Stronger preservation is insufficient in antmaze-ultra-diverse.}
    Keeping both the conservative coefficient and KL coefficient fixed yields a higher mean
    success rate than ANCHOR during online fine-tuning, but does not significantly improve
    asymptotic performance.}
    \label{fig:antmaze_ultra_diverse_variant}
\end{figure}

To further test whether stronger preservation of offline-pretrained behavior helps in
antmaze-ultra-diverse, we evaluate an additional variant that keeps the conservative
coefficient at its offline-pretraining value, $\alpha=1.0$, and keeps the KL coefficient
fixed at $\beta_t=1.0$ without annealing. As shown in
Figure~\ref{fig:antmaze_ultra_diverse_variant}, this variant achieves a higher mean success
rate than ANCHOR across much of online fine-tuning, but its asymptotic performance does not
improve significantly. These results indicate that, in extremely challenging exploration
settings, simply retaining conservative regularization or maintaining a fixed KL anchor is
not sufficient. Developing mechanisms for online exploration and adaptation in
such environments remains an important direction for future work.
\section{Related Works}
\label{sec:appendix:related_works}

\textbf{Offline RL.}
Online reinforcement learning requires interaction with the environment, which can be expensive, unsafe, or impractical in domains such as robotics, healthcare, and recommender systems~\citep{singh2022reinforcement,liu2020reinforcement,chen2024opportunities}. Offline RL addresses this limitation by learning policies from a static dataset collected by a behavior policy~\citep{prudencio2023survey}. However, because the dataset provides limited coverage of the state-action space, offline-trained policies can suffer when they encounter unseen or poorly covered regions at deployment~\citep{prudencio2023survey}.

A large body of work has been developed to address this challenge, including regularization-based methods~\citep{fujimoto2021minimalist,park2025flow,tarasov2023revisiting,kumar2020cql}, uncertainty-aware methods~\citep{an2021uncertainty,nikulin2023anti}, model-based methods~\citep{kidambi2020morel,yu2020mopo,yu2021combo,chitnis2024iql_td_mpc,park2024model}, one-step methods~\citep{brandfonbrener2021offline,eysenbach2022contrastive,park2025flow}, weighted-regression methods~\citep{peng2019advantage,wang2020critic}, and in-sample maximization methods~\citep{garg2023extreme,kostrikov2022iql,xu2023offline}. Among these approaches, we build on Conservative Q-Learning (CQL)~\citep{kumar2020cql}, which directly mitigates value overestimation by penalizing high Q-values for actions outside the dataset support. We also include Implicit Q-Learning (IQL)~\citep{kostrikov2022iql} as a baseline, as it improves over the behavior policy through value-based policy extraction without explicitly evaluating out-of-distribution actions.

\textbf{Offline-to-Online RL.}
Offline-to-online RL studies how to further improve an offline-pretrained agent through online interaction. Although offline RL methods such as CQL~\citep{kumar2020cql}, IQL~\citep{kostrikov2022iql}, and related approaches~\citep{kostrikov2021offline,tarasov2023revisiting} can be fine-tuned online, simply continuing the offline objective often yields limited improvement~\citep{nakamoto2023calql}. This has motivated methods that explicitly adapt the offline-trained agent to the online learning regime.

Existing approaches include relaxing excessive conservatism in value estimates~\citep{nakamoto2023calql,luo2024optimistic,hu2024bayesian}, introducing an intermediate adaptation phase between offline pretraining and online fine-tuning~\citep{zhou2025wsrl,shin2025online,xiao2025porl}, using multiple Q-functions~\citep{lee2022offline,zhao2023enoto}, tuning the update-to-data (UTD) ratio~\citep{feng2024suf,xiao2025porl}, and incorporating uncertainty estimation~\citep{guo2023simple,Wen2024RO2O}. Another line of work focuses on leveraging pretrained policies directly, rather than updating both the actor and critic~\citep{uchendu2023jump,xiao2025porl,hu2023imitation}.

A key challenge in this transition is catastrophic failure: performance can drop sharply at the start of online fine-tuning due to distribution shift and unstable value learning~\citep{Wen2024RO2O,nakamoto2023calql}. Prior work has addressed this issue through uncertainty estimation~\citep{Wen2024RO2O}, calibration penalties~\citep{nakamoto2023calql}, and actor-critic alignment~\citep{ACA}. In contrast, ANCHOR stabilizes the early online phase by anchoring the actor to the offline-pretrained policy with a KL penalty. The KL coefficient is linearly annealed to zero, allowing the method to move from conservative checkpoint adaptation to unconstrained online improvement. For the critic, ANCHOR immediately reduces the CQL coefficient to a small residual value $\alpha_{\mathrm{on}}$ and keeps it fixed during online fine-tuning, avoiding excessive conservatism while retaining mild regularization against overestimation. Thus, ANCHOR addresses catastrophic failure with a simple adaptation schedule while preserving strong asymptotic performance, without requiring additional uncertainty models or complex auxiliary modules.

\textbf{Trust-region and KL-regularized policy optimization.}
Our actor-side KL regularization is related in form to trust-region methods such as
TRPO~\citep{schulman2015trust}, which constrain the KL divergence between consecutive
policies to stabilize online policy updates. However, ANCHOR differs in both purpose and
reference distribution. TRPO imposes a local trust region around the previous policy at
each update to support stable on-policy improvement. In contrast, ANCHOR regularizes the
online actor toward a fixed offline-pretrained policy during the early phase of
offline-to-online adaptation. This anchor is not intended to enforce a monotonic policy
improvement guarantee; rather, it preserves useful pretrained behavior when offline replay
is unavailable and prevents abrupt policy drift into regions where the critic may be
unreliable. Moreover, ANCHOR anneals the KL coefficient to zero, explicitly removing the
constraint once the online actor and critic have adapted. Thus, while both approaches use
KL-based control of policy updates, ANCHOR uses KL regularization as a temporary
checkpoint-adaptation mechanism for offline-to-online RL, rather than as a per-update trust
region for standard online RL.

\textbf{Policy-side approaches and scheduling precedents.}
PEX~\citep{zhang2023pex} freezes the offline policy and adds an online policy, using Q-based action selection to combine them. This retains the offline policy as a component of a composite policy. PROTO~\citep{li2023proto} instead uses an evolving reference in an iterative policy-regularized objective. Improving TD3-BC~\citep{beeson2022improving} first relaxes the offline BC constraint, then uses online-buffer action matching with a decaying coefficient during fine-tuning. These works already study stable adaptation and relaxation of behavioral constraints. ANCHOR's distinction is its temporary penalty against one fixed checkpoint, separated from CQL's critic penalty, in a setting without offline replay. This is a design choice among related stabilization mechanisms, not the invention of policy regularization or scheduling.

\textbf{Expressive policies and evaluation.}
Flow Q-learning~\citep{park2025flow} combines an expressive flow model with value-based policy improvement and evaluates offline-to-online adaptation. OGBench~\citep{park2025ogbench} supplies manipulation and navigation settings beyond the original D4RL suite. Our flow-policy extension uses an action-distance proxy and therefore supports a broader checkpoint-anchoring principle rather than an exact-KL implementation for all backbones. RND~\citep{burda2019rnd} is used only to characterize visitation; it does not provide an intrinsic reward in ANCHOR. We follow the aggregate reliability methodology of~\citet{agarwal2021deep} to distinguish cross-task evidence from individual task comparisons.

\section{Details on Baseline Algorithms}
\label{sec:appendix:details_on_baseline_algorithms}

\textbf{Soft Actor-Critic (SAC)}~\citep{haarnoja2018sac}. SAC extends standard actor-critic methods by incorporating entropy maximization into the RL objective. Instead of maximizing only the expected return, SAC also maximizes the entropy of the policy to encourage exploration and prevent premature convergence to deterministic policies. The resulting objective is to maximize the expected sum of rewards and entropies:
\begin{equation*}
J(\pi) = \mathbb{E}_{\pi} \left[ \sum_{t=0}^{\infty} \gamma^t \left( r(s_t, a_t) + \alpha \mathcal{H}(\pi(\cdot|s_t)) \right) \right],
\end{equation*}
where $\mathcal{H}(\pi(\cdot|s_t)) = -\mathbb{E}_{a_t \sim \pi}[ \log \pi(a_t|s_t) ]$ denotes the entropy of the policy at state $s_t$, and $\alpha > 0$ is the temperature parameter that balances reward maximization and entropy.

\textbf{Conservative Q-Learning (CQL)}~\citep{kumar2020cql}. CQL learns a Q-function by explicitly regularizing Q-values to mitigate overestimation issues common in offline RL~\citep{prudencio2023survey}:
\begin{align*}
\mathcal{L}(\theta) &= \alpha \underbrace{\Big(\mathbb{E}_{s\sim \mathcal{D},a \sim \pi}[Q_{\theta}(s,a)] - \mathbb{E}_{s,a \sim \mathcal{D}}[Q_{\theta}(s,a)]\Big)}_{\text{Conservative regularizer}} + \mathcal{L}_{\text{TD}}(\theta)
\end{align*}
Here, $\pi$ represents the current policy, $\mathcal{D}$ is the offline dataset, and $\alpha$ controls the intensity of the conservative regularization. The term $\mathcal{L}_{TD}$ is the temporal difference loss used in Q-learning methods, while the conservative regularization penalizes Q-values for state-action pairs not present in the offline dataset $\mathcal{D}$.

\textbf{Implicit Q-learning (IQL)}~\citep{kostrikov2022iql}. The IQL algorithm learns a state-value function $V_{\theta_V}: \mathcal{S} \rightarrow \mathbb{R}$ and an action-value function $Q_{\theta_Q}: \mathcal{S} \times \mathcal{A} \rightarrow \mathbb{R}$ by minimizing:
\begin{equation*}
\mathcal{L}_V(\theta_V) = \mathbb{E}_{(s,a)\sim \mathcal{D}} \left[\ell_{\kappa}^2 \left(V_{\theta_V}(s) - Q_{\bar{\theta}_Q}(s, a)\right)\right],
\end{equation*}
\begin{equation*}
\mathcal{L}_Q(\theta_Q) = \mathbb{E}_{(s,a,r,s')\sim \mathcal{D}} \left[\left(Q_{\theta_Q}(s, a) - r - \gamma V_{\theta_V}(s')\right)^2\right],
\end{equation*}
where the expectile loss is defined as $\ell_{\kappa}^2(x) = |\kappa - \mathbf{1}[x < 0]|x^2$~\citep{newey1987expectile}, and $\bar{\theta}_Q$ represents the target network parameters. Unlike standard MSE loss, expectile loss asymmetrically weights positive and negative errors, with $\kappa > 0.5$ emphasizing higher Q-values. By doing so, IQL evaluates only actions from the dataset, which enables policy improvement without querying out-of-distribution actions, thereby inducing implicit conservatism.

\textbf{Calibrated Q-Learning (Cal-QL)}~\citep{nakamoto2023calql}. Cal-QL adjusts CQL by calibrating the learned Q-function relative to a reference policy to avoid initial degradation caused by overly pessimistic Q-values during fine-tuning. The conservative regularization term in CQL is altered by substituting $\mathbb{E}_{s\sim \mathcal{D},a \sim \pi}[Q_{\theta}(s,a)]$ with $\mathbb{E}_{s\sim \mathcal{D},a \sim \pi}[\max(Q_\theta(s,a), V^\mu(s))]$, where $V^\mu(s)$ represents the value function of the reference policy $\mu$, which can be estimated via Monte-Carlo return. This calibration prevents the conservative penalty from further suppressing actions whose estimated values are already below the reference value, thereby mitigating the overly pessimistic Q-values that can hinder online fine-tuning.

\textbf{Policy-Only Reinforcement Learning Fine-Tuning (PORL)}~\citep{xiao2025porl}. PORL focuses on the challenge of fine-tuning online RL using only a pretrained policy, without relying on pretrained Q-functions or offline datasets. This approach is especially beneficial when pretrained Q-functions are either unreliable due to pessimism or unavailable, such as in imitation learning scenarios. PORL begins by training the randomly initialized Q-function at the beginning of the online fine-tuning phase. Training data is collected based on the online interaction of the pretrained policy using an epsilon-greedy exploration strategy. During this initial sampling phase, the Q-function is trained via temporal difference learning with a high UTD ratio. After this pre-sampling period, PORL transitions to standard SAC fine-tuning, updating both the policy and the Q-function.

\textbf{Warm-start RL (WSRL)}~\citep{zhou2025wsrl}. WSRL addresses the distribution shift issue from offline to online data in fine-tuning without offline data by introducing a warmup phase. This phase utilizes a frozen pretrained policy to collect online rollouts. The data collected during warmup bridges the distribution mismatch and helps recalibrate the offline Q-function to the online distribution, allowing the method to adapt in the online environment quickly. Once the warmup phase is complete, WSRL proceeds with conventional online RL using SAC, employing a high UTD ratio.

\section{Implementation Details}
\label{sec:appendix:implementation_details}

\subsection{Details on Environments}

\textbf{AntMaze}.
The AntMaze environment, part of the D4RL benchmark suite~\citep{fu2020d4rl}, tasks an 8-DoF ant quadruped robot with navigating through a large and complex maze to reach a designated goal position. The agent receives a binary reward of +1 only upon successfully reaching the goal. The observation space is 29-dimensional, including the robot's position, orientation, and velocity, while the action space is a continuous 8-dimensional vector, normalized to the range $[-1, 1]$. We use three variants: antmaze-large-diverse-v2, which contains trajectories collected by commanding the agent to random goals from random start positions; antmaze-large-play-v2, which contains trajectories directed to a specific location; and antmaze-ultra-diverse-v2, which uses a larger and more complex maze with the same diverse data collection strategy. The two large-maze variants share a maximum episode length of 1000 steps, while antmaze-ultra-diverse-v2 allows up to 700 steps.

\textbf{FrankaKitchen}.  
The FrankaKitchen environment contained in D4RL~\citep{fu2020d4rl} requires controlling a 9-DoF Franka Panda robotic arm to manipulate various kitchen appliances and configure the environment into a predefined target state. Each task consists of four subtasks, and the agent receives a reward between 0 and 4 based on the number of successfully completed subtasks. The observation space is 60-dimensional, encompassing joint positions and object states, and the action space is a 9-dimensional continuous vector normalized to $[-1, 1]$. We use three benchmark environments from D4RL: kitchen-partial-v0, kitchen-mixed-v0, and kitchen-complete-v0. kitchen-partial-v0 includes a mix of complete and incomplete demonstrations, where task elements involve operating the microwave, kettle, light switch, and slide cabinet. kitchen-mixed-v0 contains only incomplete demonstrations, with task components including the microwave, kettle, bottom burner, and light switch. kitchen-complete-v0 contains only complete demonstrations covering the same four subtasks as kitchen-partial-v0: microwave, kettle, light switch, and slide cabinet. All three environments use a maximum episode length of 280 steps.

\textbf{Adroit}.  
The Adroit suite~\citep{rajeswaran2017learning} evaluates dexterous manipulation using a high-DoF robotic hand. Specifically, we use three tasks from the D4RL benchmark: pen-binary-v0, door-binary-v0, and relocate-binary-v0. In pen-binary-v0, a 24-DoF shadow hand must reorient a pen to match a target pose, while in door-binary-v0, a 28-DoF hand must grasp and rotate a door handle to open it. In relocate-binary-v0, a 30-DoF hand must pick up an object and relocate it to a target position. All three environments have sparse binary rewards: a reward of +1 is given only upon successful task completion. The observation space is 45-dimensional for the pen task, 39-dimensional for the door and relocate tasks, consisting of hand joint angles and object poses. The action space is continuous, 24-dimensional for the pen task, 28-dimensional for the door task, and 30-dimensional for the relocate task, each normalized to $[-1, 1]$. The maximum episode lengths are 100 steps for the pen task and 200 steps for the door and relocate tasks.

\subsection{Network Architectures}

ANCHOR uses the same network architecture as the CQL baseline in the WSRL
codebase~\citep{zhou2025wsrl}. The agent follows an actor-critic framework. The actor takes an observation as input and outputs the mean and log standard deviation of a Gaussian action distribution. The critic takes the concatenated observation-action pair as input and uses an ensemble of ten Q-functions, each producing a scalar Q-value. For stability, two Q-functions are randomly subsampled during target computation, and the minimum of the two
values is used. Both the actor and critic are implemented as multilayer perceptrons (MLPs) with ReLU activations. A learnable temperature parameter controls the entropy regularization. The hidden dimensions for each domain are reported in
Table~\ref{tab:hyperparams}.

\subsection{Reproducing Baselines}
\label{sec:appendix:reproducing_baselines}

All baseline methods are reproduced under a unified experimental setup. For WSRL, Cal-QL,
CQL, IQL, and SAC, we use the implementations and hyperparameters provided in the WSRL
codebase~\citep{zhou2025wsrl}, with several stability-oriented modifications applied
consistently across methods. Specifically, (i) we increase the batch size from $256$ to
$512$; (ii) we use an ensemble of ten Q-functions and compute target values using the
minimum over two critics randomly subsampled at each update, following
REDQ~\citep{chen2021redq}; and (iii) we apply layer normalization~\citep{ba2016layernorm}
to both the policy and critic networks to mitigate uncontrolled extrapolation
effects~\citep{ball2023efficient}. Since no public implementation of PORL was available,
we reimplement it following~\citet{xiao2025porl} and use the authors' reported settings.

\subsection{Hyperparameters}
\label{sec:appendix:implementation_details:hyperparameters}

Table~\ref{tab:hyperparams} reports the hyperparameters used to train ANCHOR. For
hyperparameters not specific to ANCHOR, we follow the default CQL settings used
by~\citet{zhou2025wsrl}, except for the UTD ratio in door-binary, where we use the WSRL
setting. All main baseline runs and ANCHOR use 5K collection-only warmup steps. For ANCHOR-specific hyperparameters, we select the best-performing configuration per environment from
$\alpha_{\mathrm{on}}\in\{0.0,1.0\}$,
$T_{\mathrm{KL}}\in\{40\mathrm{K},80\mathrm{K},160\mathrm{K}\}$, and
$\beta_0\in\{0.2,0.4,1.0\}$.

\begin{table}[h]
\caption{\textbf{Hyperparameters of ANCHOR.}}
\vspace{0.5em}
\centering
\scriptsize
\resizebox{\textwidth}{!}{%
\begin{tabular}{lcccccc}
\toprule
& \multicolumn{2}{c}{\textbf{AntMaze}} 
& \multicolumn{2}{c}{\textbf{Adroit}} 
& \multicolumn{2}{c}{\textbf{Kitchen}} \\
\cmidrule{2-3} \cmidrule{4-5} \cmidrule{6-7}
& large diverse & large play & door & pen & partial & mixed \\
\midrule
\multicolumn{7}{l}{\textbf{ANCHOR}} \\
$\alpha_{\mathrm{on}}$
& 0.0 & 0.0
& 0.0 & 0.0
& 1.0 & 1.0 \\
$T_{\mathrm{KL}}$
& 80{,}000 & 80{,}000
& 40{,}000 & 80{,}000
& 160{,}000 & 160{,}000 \\
$\beta_{0}$
& 0.4 & 0.2
& 0.2 & 1.0
& 1.0 & 1.0 \\
\midrule
\multicolumn{7}{l}{\textbf{Optimization}} \\
Actor Learning Rate 
& \multicolumn{2}{c}{$1\times 10^{-4}$} 
& \multicolumn{2}{c}{$1\times 10^{-4}$} 
& \multicolumn{2}{c}{$1\times 10^{-4}$} \\
Critic Learning Rate 
& \multicolumn{2}{c}{$3\times 10^{-4}$} 
& \multicolumn{2}{c}{$3\times 10^{-4}$} 
& \multicolumn{2}{c}{$3\times 10^{-4}$} \\
Batch Size 
& \multicolumn{2}{c}{512} 
& \multicolumn{2}{c}{512} 
& \multicolumn{2}{c}{512} \\
UTD 
& 1 & 1
& 16 & 1
& 1 & 1 \\
Offline Steps 
& \multicolumn{2}{c}{1{,}000{,}000} 
& \multicolumn{2}{c}{40{,}000} 
& \multicolumn{2}{c}{250{,}000} \\
$\alpha_{\mathrm{off}}$
& \multicolumn{2}{c}{5.0} 
& \multicolumn{2}{c}{1.0} 
& \multicolumn{2}{c}{5.0} \\
Warmup Steps 
& \multicolumn{2}{c}{5{,}000} 
& \multicolumn{2}{c}{5{,}000} 
& \multicolumn{2}{c}{5{,}000} \\
\midrule
\multicolumn{7}{l}{\textbf{Architecture}} \\
Critic Network 
& \multicolumn{2}{c}{{[}256, 256, 256, 256{]}} 
& \multicolumn{2}{c}{{[}512, 512, 512{]}} 
& \multicolumn{2}{c}{{[}512, 512, 512{]}} \\
Actor Network 
& \multicolumn{2}{c}{{[}256, 256{]}} 
& \multicolumn{2}{c}{{[}512, 512{]}}
& \multicolumn{2}{c}{{[}512, 512, 512{]}} \\
Activations 
& \multicolumn{2}{c}{ReLU} 
& \multicolumn{2}{c}{ReLU} 
& \multicolumn{2}{c}{ReLU} \\
Q-ensemble 
& \multicolumn{2}{c}{10} 
& \multicolumn{2}{c}{10}
& \multicolumn{2}{c}{10} \\
\bottomrule
\end{tabular}%
}
\label{tab:hyperparams}
\end{table}

\subsection{Online Training Time Comparison}
\label{sec:appendix:online_training_time_comparison}

Table~\ref{tab:online_training_time_comparison} compares the online training time of
ANCHOR and the baselines. ANCHOR has a similar average training time to WSRL and PORL,
despite the additional actor-side KL regularization. While it is slower than lightweight
baselines such as IQL and SAC, these methods suffer from substantially weaker asymptotic
performance and/or larger catastrophic failure in our main results. Considering the gains
of ANCHOR in both early stability and final performance, its computational overhead is
moderate and comparable to other competitive offline-to-online methods.

\begin{table}[h]
\caption{\textbf{Online training time (in decimal hours) comparison across baselines.}}
\vspace{0.5em}

\centering
\scriptsize
\resizebox{\textwidth}{!}{%
\begin{tabular}{lccccccc}
\toprule
Task & \textbf{ANCHOR (Ours)} & WSRL & PORL & Cal-QL & CQL & IQL & SAC\\
\midrule
antmaze-large-diverse-v2 
& 2.75 $\pm$ 0.02
& 2.73 $\pm$ 0.01
& 3.35 $\pm$ 0.07
& 3.44 $\pm$ 0.01
& 2.82 $\pm$ 0.02
& 2.08 $\pm$ 0.02
& 1.53 $\pm$ 0.0 \\
antmaze-large-play-v2    
& 2.75 $\pm$ 0.01
& 3.48 $\pm$ 0.08
& 3.54 $\pm$ 0.06
& 3.26 $\pm$ 0.03 
& 2.34 $\pm$ 0.03 
& 2.09 $\pm$ 0.03
& 1.52 $\pm$ 0.0 \\
\midrule
kitchen-partial-v0       
& 4.31 $\pm$ 0.02 
& 3.37 $\pm$ 0.0
& 3.41 $\pm$ 0.03
& 4.35 $\pm$ 0.01
& 3.28 $\pm$ 0.05
& 1.41 $\pm$ 0.01
& 1.38 $\pm$ 0.0 \\
kitchen-mixed-v0         
& 3.88 $\pm$ 0.03
& 3.61 $\pm$ 0.01
& 3.51 $\pm$ 0.13 
& 4.36 $\pm$ 0.0
& 3.33 $\pm$ 0.02
& 1.46 $\pm$ 0.02
& 1.36 $\pm$ 0.0 \\
\midrule
door-binary-v0          
& 3.07 $\pm$ 0.01
& 2.9 $\pm$ 0.01
& 2.93 $\pm$ 0.01
& 3.55 $\pm$ 0.01
& 2.43 $\pm$ 0.01
& 1.28 $\pm$ 0.02
& 0.78 $\pm$ 0.01 \\
pen-binary-v0
& 2.33 $\pm$ 0.02
& 2.77 $\pm$ 0.01
& 2.79 $\pm$ 0.01
& 3.51 $\pm$ 0.01
& 2.29 $\pm$ 0.01
& 1.13 $\pm$ 0.1
& 0.67 $\pm$ 0.0 \\
\midrule
Average 
& 3.18
& 3.14
& 3.25
& 3.75
& 2.75
& 1.58 
& 1.21 \\
\bottomrule
\end{tabular}
}
\label{tab:online_training_time_comparison}
\end{table}

\subsection{Experimental Setup and Reproducibility}

All experiments were conducted on a compute node equipped with 8 NVIDIA A100 GPUs (80GB each), an AMD EPYC 7543 32-Core CPU, and 885 GB of RAM. The software environment was based on Python 3.10.18, Pytorch 2.7.0, JAX 0.4.25 with CUDA 12.2.

\section{Limitations}
\label{sec:appendix:limitations}

While ANCHOR provides a simple and effective recipe for offline-to-online adaptation without offline replay, several limitations remain.

\textbf{Scope of evaluation.}
Our experiments focus on single-agent continuous-control benchmarks, including AntMaze,
FrankaKitchen, Adroit, and selected OGBench tasks. We do not evaluate ANCHOR under hard safety constraints,
non-stationary dynamics, partial observability, or multi-agent settings. Therefore, its
effectiveness in these regimes is not guaranteed.

\textbf{Dependence on pretrained policy quality.}
ANCHOR is most effective when the offline-pretrained policy contains useful behavior to
preserve. When the pretrained policy has near-zero success, as in relocate-binary, KL anchoring provides limited benefit.
This suggests that ANCHOR is better suited for stabilizing useful pretrained behavior than
for solving tasks where the offline policy provides little meaningful guidance.

\textbf{Limited exploration in hard sparse-reward environments.}
ANCHOR controls policy drift during online fine-tuning, but it does not introduce a new
exploration mechanism. In challenging sparse-reward environments such as
antmaze-ultra-diverse, most methods fail to improve substantially over the offline-pretrained
policy. Addressing such settings may require combining KL anchoring with stronger
exploration or task-specific adaptation mechanisms.

\textbf{Hand-designed KL schedules.}
ANCHOR uses a manually specified initial KL weight and annealing interval. Although our
ablations show robustness to moderate changes in these hyperparameters, selecting the
schedule still requires some environment-level tuning. Future work could develop adaptive
KL schedules based on online uncertainty, critic reliability, or measured policy drift.

\textbf{Backbone dependence.}
The main benchmark instantiates ANCHOR on top of CQL. The FQL/IFQL extension uses a coupled action-distance proxy rather than an exact policy KL, and its stability benefits are mixed. The exact-KL occupancy interpretation does not automatically extend to this proxy. We also test actor-side KL
anchoring with IQL and find that it reduces early catastrophic failure, but does not improve
asymptotic performance. This suggests that different offline RL backbones may require
backbone-specific mechanisms for relaxing offline inductive biases during online adaptation.

\textbf{Statistical coverage.}
Unless otherwise noted, experiments are run with five random seeds due to computational
constraints, and some auxiliary analyses use fewer seeds. Although confidence intervals are
reported, rare failure modes may still be under-sampled.

\textbf{Protocol and evidence limitations.}
The additional policy-side comparisons adapt prior methods to CQL and online-only replay, but their supplied defaults do not match every critic setting of the main ANCHOR configuration. BC controls use offline data for actor regularization. The two-seed shared-duration check lacks a retained initial-weight value, and the OGBench summaries do not specify the final-score evaluation window. We report these limitations rather than treating these experiments as exact single-factor ablations. Aggregate bootstrap summaries are retained from the experiment record; raw seed-level traces are needed for independent recomputation.

\newpage

\section{Visualizations}
\label{sec:appendix:visualizations}
\begin{figure}[H]
    \centering
    \begin{subfigure}[b]{\textwidth}
        \centering
        \includegraphics[width=0.9\textwidth]{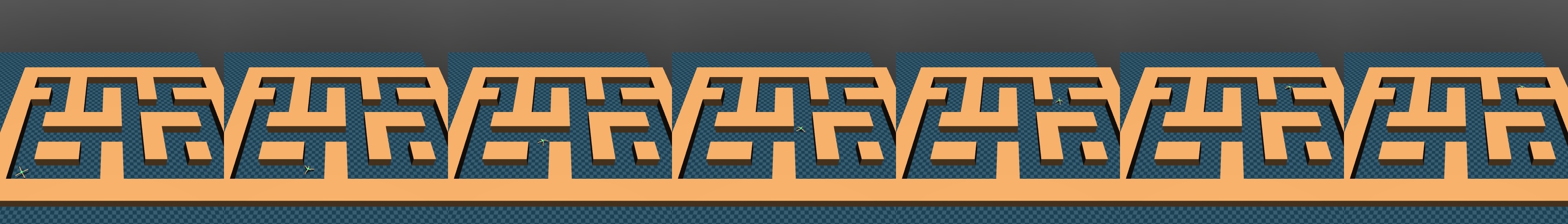}
        \caption{\textbf{antmaze-large-diverse-v2:} 8-DoF ant navigating through a maze to reach the goal.}
        \label{fig:viz_antmaze_diverse}
    \end{subfigure}
    \vspace{0.15cm}
    \begin{subfigure}[b]{\textwidth}
        \centering
        \includegraphics[width=0.9\textwidth]{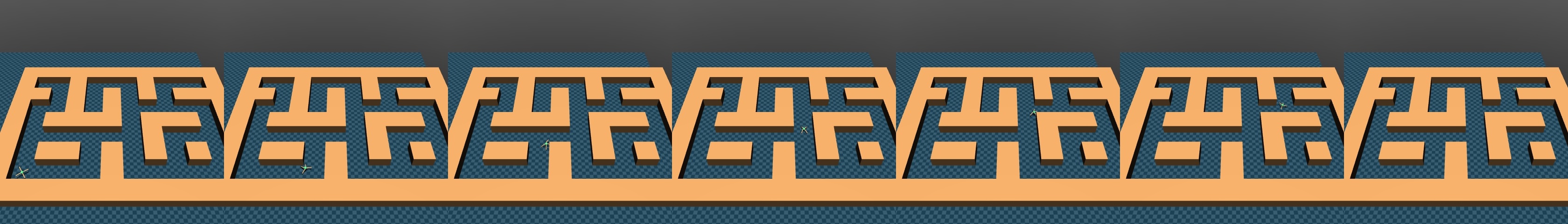}
        \caption{\textbf{antmaze-large-play-v2:} Same evaluation as diverse-v2, different training data.}
        \label{fig:viz_antmaze_play}
    \end{subfigure}
    \begin{subfigure}[b]{\textwidth}
        \centering
        \includegraphics[width=0.9\textwidth]{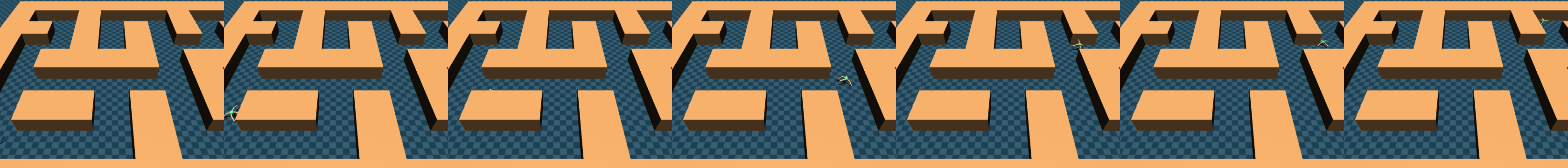}
        \caption{\textbf{antmaze-ultra-diverse-v2:} 8-DoF ant navigating through a larger maze than large-diverse-v2.}
        \label{fig:viz_antmaze_ultra}
    \end{subfigure}
    \vspace{0.15cm}
    \begin{subfigure}[b]{\textwidth}
        \centering
        \includegraphics[width=0.9\textwidth]{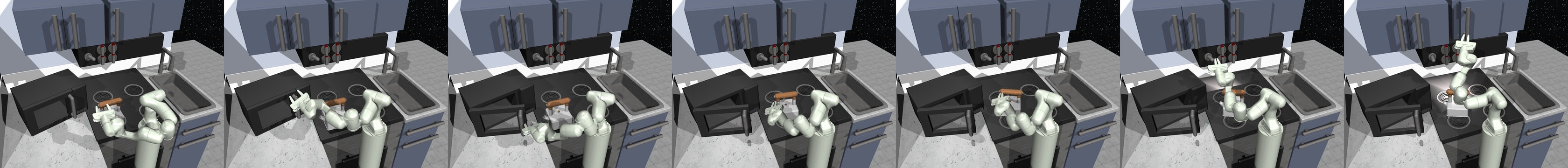}
        \caption{\textbf{kitchen-partial-v0:} Completing subtasks: microwave, kettle, light switch, and slide cabinet.}
        \label{fig:viz_kitchen_partial}
    \end{subfigure}
    \begin{subfigure}[b]{\textwidth}
        \centering
        \includegraphics[width=0.9\textwidth]{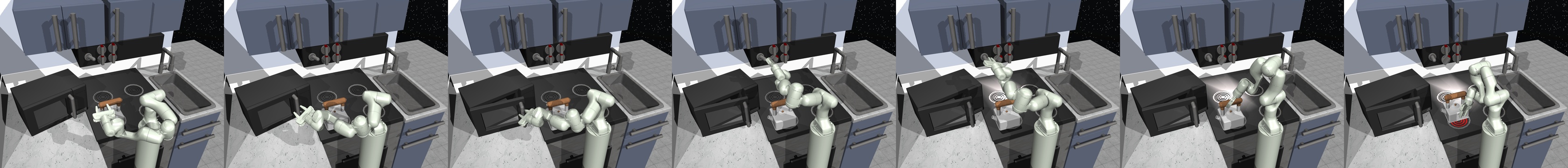}
        \caption{\textbf{kitchen-mixed-v0:} Completing subtasks: microwave, kettle, bottom burner, and light switch.}
        \label{fig:viz_kitchen_mixed}
    \end{subfigure}
        \begin{subfigure}[b]{\textwidth}
        \centering
        \includegraphics[width=0.9\textwidth]{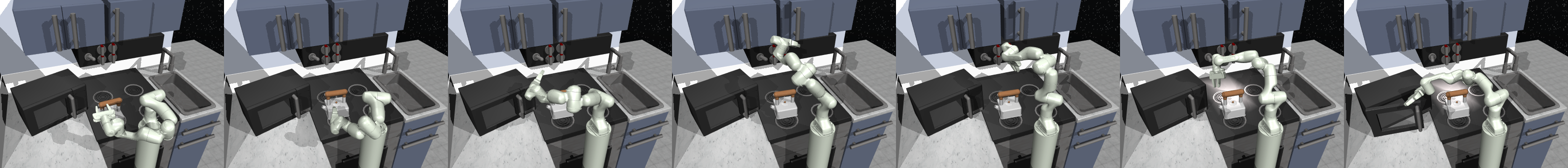}
        \caption{\textbf{kitchen-complete-v0:} Completing subtasks: microwave, kettle, light switch, and slide cabinet.}
        \label{fig:viz_kitchen_complete}
    \end{subfigure}

    \begin{subfigure}[b]{\textwidth}
        \centering
        \includegraphics[width=0.9\textwidth]{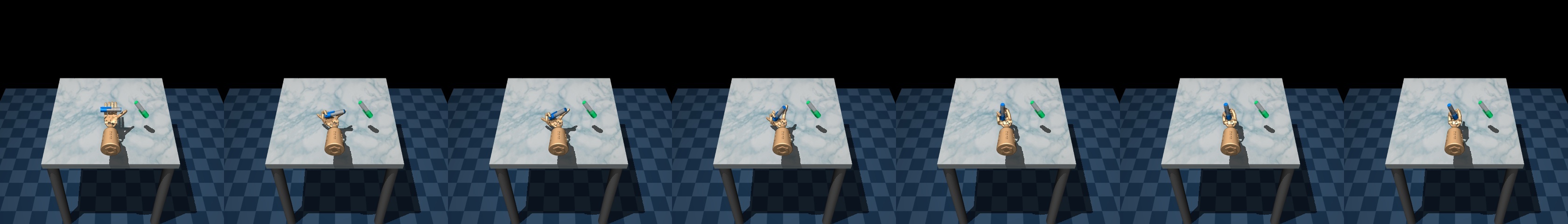}
        \caption{\textbf{pen-binary-v0:} 24-DoF shadow hand reorienting a pen to target pose.}
        \label{fig:viz_pen_binary}
    \end{subfigure}
    \begin{subfigure}[b]{\textwidth}
        \centering
        \includegraphics[width=0.9\textwidth]{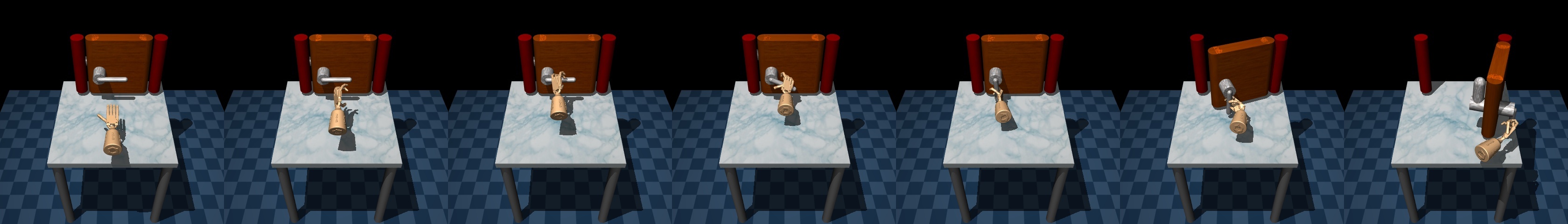}
        \caption{\textbf{door-binary-v0:} 28-DoF hand opening a door by rotating the handle.}
        \label{fig:viz_door_binary}
    \end{subfigure}

    \begin{subfigure}[b]{\textwidth}
        \centering
        \includegraphics[width=0.9\textwidth]{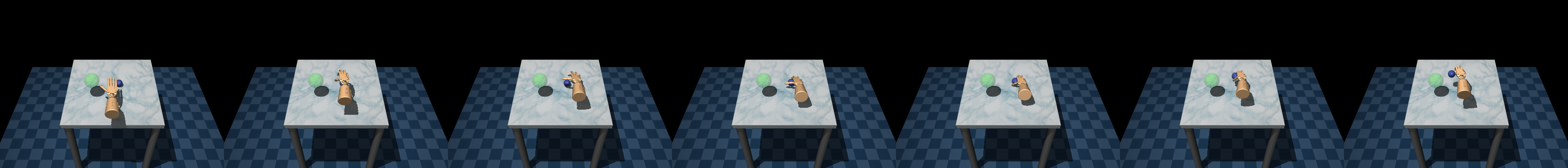}
        \caption{\textbf{relocate-binary-v0:} 30-DoF shadow hand relocating an object to a target location.}
        \label{fig:viz_relocate_binary}
    \end{subfigure}

        \caption{
    \textbf{Representative episodes across nine tasks.} Each row shows temporally ordered frames from a trajectory, demonstrating navigation (antmaze-ultra-diverse-v2, antmaze-large-diverse-v2, antmaze-large-play-v2), kitchen manipulation (kitchen-partial-v0, kitchen-mixed-v0, kitchen-complete-v0), and dexterous control (pen-binary-v0, door-binary-v0, relocate-binary-v0).}

    \label{fig:visualizations_all}
\end{figure}

\newpage
\section{Numerical Results}
\label{sec:appendix:numerical_results}

\subsection{Raw Performance and Catastrophic Failure Values}

\begin{table}[h]
\caption{\textbf{Performance at 400k online steps.} \textbf{Bold} indicates the best performance per task, and \underline{underline} indicates the second best.}
\vspace{0.5em}
\centering
\scriptsize
\resizebox{\textwidth}{!}{%
\begin{tabular}{lccccccc}
\toprule
Task & \textbf{ANCHOR (Ours)} & WSRL & PORL & Cal-QL & CQL & IQL & SAC\\
\midrule
antmaze-large-diverse-v2 
& \textbf{0.97} $\pm$ 0.01 
& 0.68 $\pm$ 0.17 
& 0.9 $\pm$ 0.07
& \textbf{0.97} $\pm$ 0.01 
& 0.92 $\pm$ 0.05 
& 0.06 $\pm$ 0.05 
& 0.0 $\pm$ 0.0 \\
antmaze-large-play-v2    
& \textbf{0.96} $\pm$ 0.02 
& \underline{0.95} $\pm$ 0.02 
& 0.47 $\pm$ 0.2
& 0.9 $\pm$ 0.07 
& \underline{0.95} $\pm$ 0.03 
& 0.01 $\pm$ 0.01
& 0.0 $\pm$ 0.0 \\
\midrule
kitchen-partial-v0       
& \textbf{1.0} $\pm$ 0.0 
& \underline{0.93} $\pm$ 0.06 
& 0.91 $\pm$ 0.06
& 0.7 $\pm$ 0.09 
& 0.44 $\pm$ 0.12 
& 0.4 $\pm$ 0.05 
& 0.49 $\pm$ 0.07 \\
kitchen-mixed-v0         
& \underline{0.81} $\pm$ 0.09 
& 0.72 $\pm$ 0.11
& \textbf{0.88} $\pm$ 0.06 
& 0.45 $\pm$ 0.12 
& 0.4 $\pm$ 0.06 
& 0.31 $\pm$ 0.09 
& 0.3 $\pm$ 0.05 \\
\midrule
door-binary-v0          
& \textbf{1.0} $\pm$ 0.0 
& \underline{0.8} $\pm$ 0.2 
& 0.6 $\pm$ 0.24
& 0.19 $\pm$ 0.19
& 0.18 $\pm$ 0.18
& 0.0 $\pm$ 0.0
& 0.2 $\pm$ 0.2 \\
pen-binary-v0
& \textbf{1.0} $\pm$ 0.0
& 0.97 $\pm$ 0.01
& \underline{0.99} $\pm$ 0.01
& 0.94 $\pm$ 0.02
& 0.81 $\pm$ 0.09
& 0.33 $\pm$ 0.1
& 0.78 $\pm$ 0.04 \\
\midrule
Mean of displayed tasks
& 0.957
& 0.842
& 0.792
& 0.692
& 0.617
& 0.185
& 0.295 \\
\bottomrule
\end{tabular}
}
\label{tab:final_perform}
\end{table}

\begin{table}[h]
\caption{\textbf{Performance at 350k online steps.} \textbf{Bold} indicates the best performance per task, and \underline{underline} indicates the second best.}
\vspace{0.5em}
\centering
\scriptsize
\resizebox{\textwidth}{!}{%
\begin{tabular}{lccccccc}
\toprule
Task & \textbf{ANCHOR (Ours)} & WSRL & PORL & Cal-QL & CQL & IQL & SAC\\
\midrule
antmaze-large-diverse-v2 
& \textbf{0.97} $\pm$ 0.01 
& 0.73 $\pm$ 0.18 
& 0.84 $\pm$ 0.07
& 0.95 $\pm$ 0.03 
& \underline{0.96} $\pm$ 0.02 
& 0.03 $\pm$ 0.03 
& 0.0 $\pm$ 0.0 \\
antmaze-large-play-v2    
& \underline{0.96} $\pm$ 0.02 
& \textbf{0.98} $\pm$ 0.01 
& 0.37 $\pm$ 0.23
& 0.89 $\pm$ 0.07 
& 0.95 $\pm$ 0.03 
& 0.01 $\pm$ 0.01
& 0.0 $\pm$ 0.0 \\
\midrule
kitchen-partial-v0       
& \textbf{1.0} $\pm$ 0.0 
& \underline{0.95} $\pm$ 0.05 
& 0.9 $\pm$ 0.06
& 0.7 $\pm$ 0.09 
& 0.44 $\pm$ 0.12 
& 0.39 $\pm$ 0.05 
& 0.47 $\pm$ 0.06 \\
kitchen-mixed-v0         
& \underline{0.83} $\pm$ 0.1 
& 0.7 $\pm$ 0.12
& \textbf{0.88} $\pm$ 0.06 
& 0.45 $\pm$ 0.12 
& 0.4 $\pm$ 0.06 
& 0.3 $\pm$ 0.08 
& 0.25 $\pm$ 0.03 \\
\midrule
door-binary-v0          
& \textbf{0.97} $\pm$ 0.01
& \underline{0.8} $\pm$ 0.2 
& 0.6 $\pm$ 0.24
& 0.17 $\pm$ 0.17
& 0.19 $\pm$ 0.19
& 0.0 $\pm$ 0.0
& 0.2 $\pm$ 0.2 \\
pen-binary-v0
& 0.94 $\pm$ 0.02
& \textbf{0.96} $\pm$ 0.02
& 0.95 $\pm$ 0.02
& \textbf{0.96} $\pm$ 0.01
& 0.72 $\pm$ 0.05
& 0.4 $\pm$ 0.05
& 0.84 $\pm$ 0.02 \\
\midrule
Mean of displayed tasks
& 0.945
& 0.853
& 0.757
& 0.687
& 0.610
& 0.188
& 0.293 \\
\bottomrule
\end{tabular}
}
\label{tab:perform_at_350k}
\end{table}

Tables~\ref{tab:final_perform} and \ref{tab:perform_at_350k} report success rates (mean~$\pm$~SE) after 400k and at 350k online fine-tuning, respectively, averaged over five seeds. The endpoint values provide context for the 350K--400K late-stage window, but do not by themselves establish convergence. Mean rows are unweighted arithmetic means of the six displayed task means (rounded to three decimals); aggregate SEs cannot be recovered without seed-level data and are therefore not assigned. These means are distinct from the IQM in Appendix~\ref{sec:appendix:reliability}.

\begin{table}[h]
\caption{\textbf{Performance after the offline phase.}}
\vspace{0.5em}
\centering
\scriptsize
\resizebox{\textwidth}{!}{%
\begin{tabular}{l ccccccc}
\toprule
Task & \textbf{ANCHOR (Ours)} & WSRL & PORL & Cal-QL & CQL & IQL \\
\midrule
antmaze-large-diverse-v2 
& 0.33 $\pm$ 0.09 
& 0.24 $\pm$ 0.05 
& 0.39 $\pm$ 0.04
& 0.26 $\pm$ 0.04
& 0.26 $\pm$ 0.05 
& 0.03 $\pm$ 0.02 \\
antmaze-large-play-v2    
& 0.29 $\pm$ 0.05
& 0.29 $\pm$ 0.02 
& 0.37 $\pm$ 0.02
& 0.24 $\pm$ 0.04 
& 0.28 $\pm$ 0.05 
& 0.0 $\pm$ 0.0 \\
\midrule
kitchen-partial-v0       
& 0.7 $\pm$ 0.03
& 0.73 $\pm$ 0.06 
& 0.7 $\pm$ 0.04
& 0.68 $\pm$ 0.05
& 0.7 $\pm$ 0.06 
& 0.3 $\pm$ 0.06 \\
kitchen-mixed-v0         
& 0.47 $\pm$ 0.06 
& 0.47 $\pm$ 0.06
& 0.49 $\pm$ 0.07
& 0.52 $\pm$ 0.06
& 0.58 $\pm$ 0.03 
& 0.4 $\pm$ 0.04 \\
\midrule
door-binary-v0          
& 0.26 $\pm$ 0.04
& 0.31 $\pm$ 0.19 
& 0.24 $\pm$ 0.03
& 0.21 $\pm$ 0.07
& 0.08 $\pm$ 0.08
& 0.05 $\pm$ 0.04 \\
pen-binary-v0
& 0.69 $\pm$ 0.03
& 0.66 $\pm$ 0.05
& 0.74 $\pm$ 0.03
& 0.75 $\pm$ 0.05
& 0.58 $\pm$ 0.03
& 0.89 $\pm$ 0.04 \\
\midrule
Mean of displayed tasks
& 0.457
& 0.450
& 0.488
& 0.443
& 0.413
& 0.278 \\
\bottomrule
\end{tabular}
}
\label{tab:perform_end_of_offline_phase}
\end{table}

Table~\ref{tab:perform_end_of_offline_phase} reports success rates (mean~$\pm$~SE) at the end of the offline pretraining phase, i.e., the start of online fine-tuning, averaged over five seeds. Initial performance differs across algorithms and runs; these values should not be interpreted as evidence of identical pretrained checkpoints. Low CF must be assessed alongside the initial success rate.

\newpage

\begin{table}[H]
\caption{\textbf{Catastrophic failure (0-400K window).}}
\vspace{0.5em}
\centering
\scriptsize
\begin{minipage}{\textwidth}
    \centering
    \resizebox{\textwidth}{!}{%
    \begin{tabular}{l cccccc}
    \toprule
    Task & \textbf{ANCHOR (Ours)} & WSRL & PORL & Cal-QL & CQL & IQL \\
    \midrule
    antmaze-large-diverse-v2 
    & 0.12 $\pm$ 0.07
    & 0.07 $\pm$ 0.04
    & 0.39 $\pm$ 0.04
    & 0.17 $\pm$ 0.07 
    & 0.21 $\pm$ 0.04
    & 0.03 $\pm$ 0.02 \\
    antmaze-large-play-v2    
    & 0.11 $\pm$ 0.09
    & 0.2 $\pm$ 0.06 
    & 0.37 $\pm$ 0.02
    & 0.16 $\pm$ 0.05
    & 0.16 $\pm$ 0.07
    & 0.0 $\pm$ 0.0 \\
    \midrule
    kitchen-partial-v0      
    & 0.15 $\pm$ 0.08
    & 0.68 $\pm$ 0.09
    & 0.7 $\pm$ 0.04
    & 0.68 $\pm$ 0.05
    & 0.7 $\pm$ 0.06
    & 0.19 $\pm$ 0.04 \\
    kitchen-mixed-v0         
    & 0.22 $\pm$ 0.11
    & 0.42 $\pm$ 0.09
    & 0.48 $\pm$ 0.07
    & 0.52 $\pm$ 0.06
    & 0.58 $\pm$ 0.03
    & 0.35 $\pm$ 0.04 \\
    \midrule
    door-binary-v0           
    & 0.26 $\pm$ 0.04 
    & 0.31 $\pm$ 0.19
    & 0.24 $\pm$ 0.03
    & 0.21 $\pm$ 0.07
    & 0.08 $\pm$ 0.08
    & 0.05 $\pm$ 0.04 \\
    pen-binary-v0            
    & 0.14 $\pm$ 0.04
    & 0.52 $\pm$ 0.04
    & 0.66 $\pm$ 0.04
    & 0.33 $\pm$ 0.08
    & 0.58 $\pm$ 0.03
    & 0.87 $\pm$ 0.03 \\
    \midrule
    Mean of displayed tasks
    & 0.167
    & 0.367
    & 0.473
    & 0.345
    & 0.385
    & 0.248 \\
    \bottomrule
    \end{tabular}
    }
    \label{tab:cat_failure}
\end{minipage}

\vspace{2em}

\begin{minipage}{\textwidth}
    \caption{\textbf{Catastrophic failure (0-100K window).}  }
    \vspace{0.5em}

    \centering
    \resizebox{\textwidth}{!}{%
    \begin{tabular}{l cccccc}
    \toprule
    Task & \textbf{ANCHOR (Ours)} & WSRL & PORL & Cal-QL & CQL & IQL \\
    \midrule
    antmaze-large-diverse-v2 
    & 0.1 $\pm$ 0.08
    & 0.07 $\pm$ 0.04
    & 0.39 $\pm$ 0.04
    & 0.17 $\pm$ 0.07 
    & 0.21 $\pm$ 0.04
    & 0.03 $\pm$ 0.02 \\
    antmaze-large-play-v2    
    & 0.11 $\pm$ 0.09
    & 0.2 $\pm$ 0.06 
    & 0.37 $\pm$ 0.02 
    & 0.16 $\pm$ 0.05
    & 0.16 $\pm$ 0.07
    & 0.0 $\pm$ 0.0 \\
    \midrule
    kitchen-partial-v0      
    & 0.06 $\pm$ 0.02
    & 0.68 $\pm$ 0.09
    & 0.7 $\pm$ 0.04
    & 0.68 $\pm$ 0.05
    & 0.7 $\pm$ 0.06
    & 0.18 $\pm$ 0.04 \\
    kitchen-mixed-v0         
    & 0.03 $\pm$ 0.01
    & 0.42 $\pm$ 0.09
    & 0.48 $\pm$ 0.07
    & 0.52 $\pm$ 0.06
    & 0.58 $\pm$ 0.03
    & 0.33 $\pm$ 0.03 \\
    \midrule
    door-binary-v0           
    & 0.26 $\pm$ 0.04 
    & 0.31 $\pm$ 0.19
    & 0.24 $\pm$ 0.03
    & 0.21 $\pm$ 0.07
    & 0.08 $\pm$ 0.08
    & 0.05 $\pm$ 0.04 \\
    pen-binary-v0            
    & 0.14 $\pm$ 0.04
    & 0.52 $\pm$ 0.04
    & 0.66 $\pm$ 0.04
    & 0.33 $\pm$ 0.08
    & 0.58 $\pm$ 0.03
    & 0.83 $\pm$ 0.04 \\
    \midrule
    Mean of displayed tasks
    & 0.117
    & 0.367
    & 0.473
    & 0.345
    & 0.385
    & 0.237 \\
    \bottomrule
    \end{tabular}
    }
     \label{tab:cat_failure_100k}
\end{minipage}
\end{table}

Tables~\ref{tab:cat_failure} and~\ref{tab:cat_failure_100k} report catastrophic failure
over the 0-400K and 0-100K windows, respectively. Per-task values are averaged over five seeds and reported as mean~$\pm$~SE; the final rows are arithmetic means of the displayed tasks without an inferred aggregate SE. The estimates are similar across the two windows,
indicating that most failures occur early in the offline-to-online transition. ANCHOR
achieves the lowest average catastrophic failure. Low CF for IQL on some tasks must be interpreted together with its low initial and final success. On door-binary, all methods reach zero success at some point during the main online run.

\subsection{Pretrained Policy Quality Analysis with Confidence Intervals}
\label{sec:appendix:pretrained_policy_quality_ci}

\begin{figure}[h]
    \centering
    \begin{subfigure}[c]{0.45\linewidth}
        \includegraphics[width=1.0\linewidth]{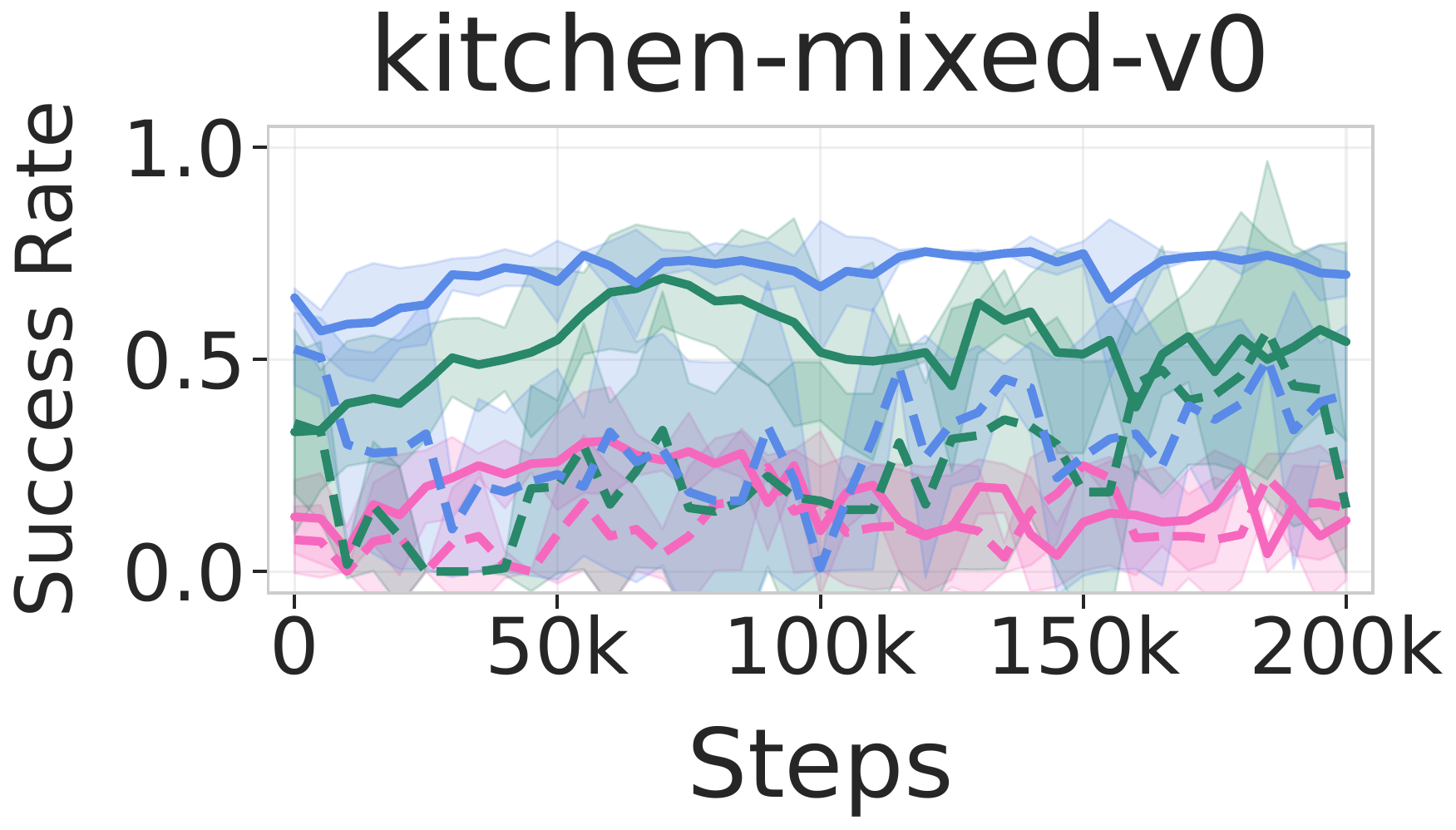}
    \end{subfigure}\hspace{0.1em}
    \begin{subfigure}[c]{0.45\linewidth}
        \includegraphics[width=1.0\linewidth]{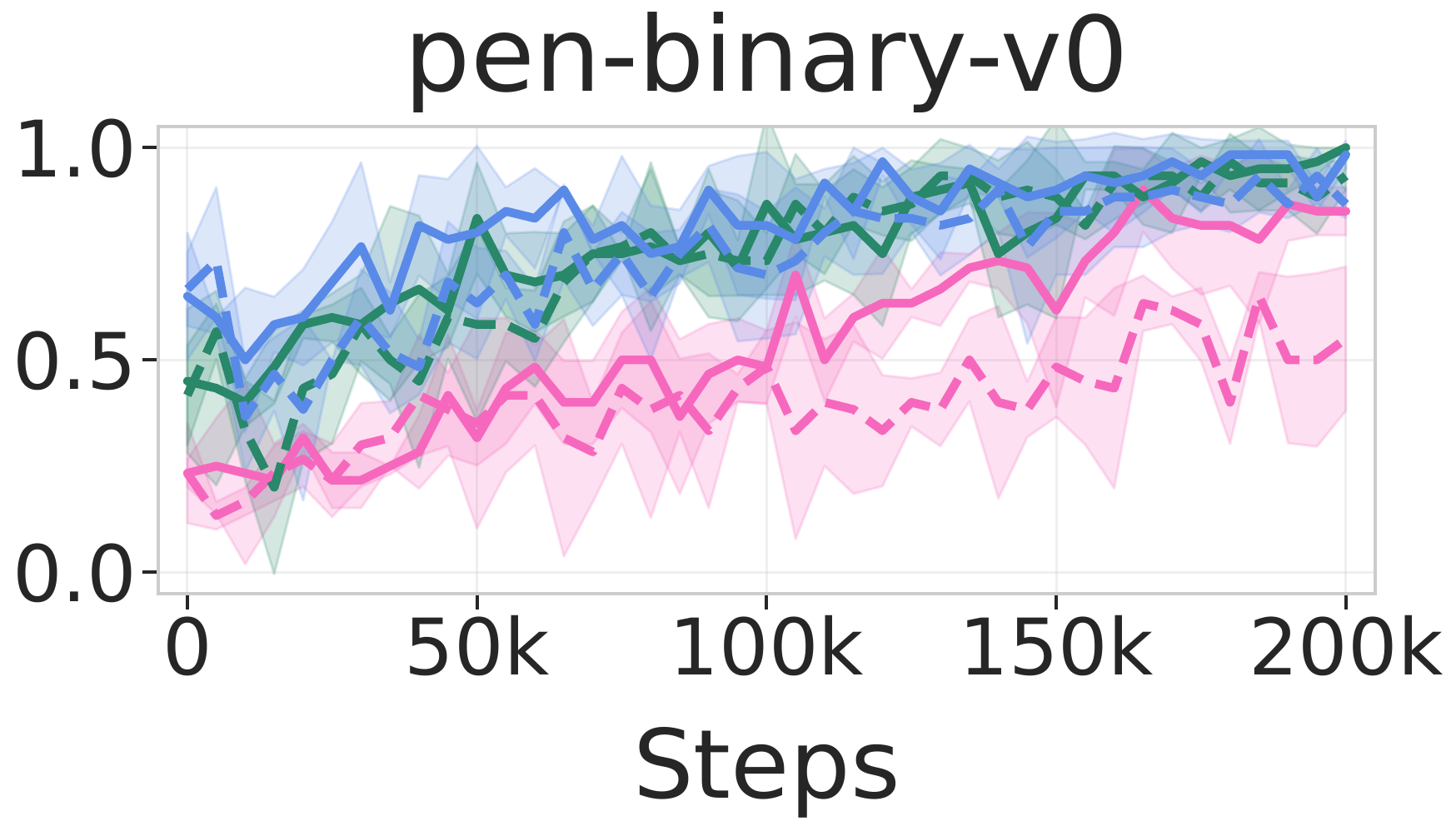}
    \end{subfigure}\\[1ex]
    \begin{minipage}{0.6\textwidth}
        \centering
        \includegraphics[width=\textwidth]{NeurIPS_figure/base_model_importance/legend.pdf}
    \end{minipage}
    \caption{\textbf{Benefit of ANCHOR depends on pretrained policy quality.} ANCHOR reduces catastrophic failure more clearly when the offline-pretrained policy starts from a higher initial success rate. Curves with the same color correspond to the same pretrained policy. Results are averaged over three seeds.}
    \label{fig:full_confidence_kl_anchoring_importance}
\end{figure}

Figure~\ref{fig:full_confidence_kl_anchoring_importance} reports the
pretrained-policy-quality analysis from Section~\ref{sec:experiments:kl_anchoring} with
full confidence intervals. The results are consistent with the interpretation of
Figure~\ref{fig:kl_anchoring_importance}: when the pretrained policy starts from a higher
initial success rate, the gap between ANCHOR and the no-KL variant becomes clearer. This
shows that ANCHOR reduces catastrophic failure more effectively when there is useful
pretrained behavior to preserve.

\clearpage
\section{Supplementary Diagnostic and Sensitivity Figures}
\label{sec:appendix:diagnostic_figures}
These figures accompany the original stabilization analysis in Section~\ref{sec:analysis_catastrophic_failure} and the hyperparameter analysis in Section~\ref{sec:experiments:kl_anchoring}.

\begin{figure}[ht]
    \centering
    \begin{subfigure}[c]{0.07\textwidth}
        \includegraphics[width=1.0\textwidth]{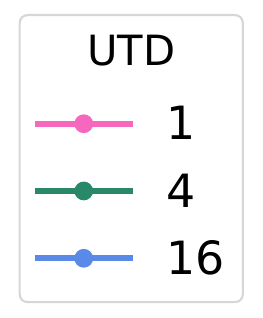}
    \end{subfigure}\hspace{0.1em}
    \begin{subfigure}[c]{0.22\textwidth}
        \includegraphics[width=1.0\linewidth]{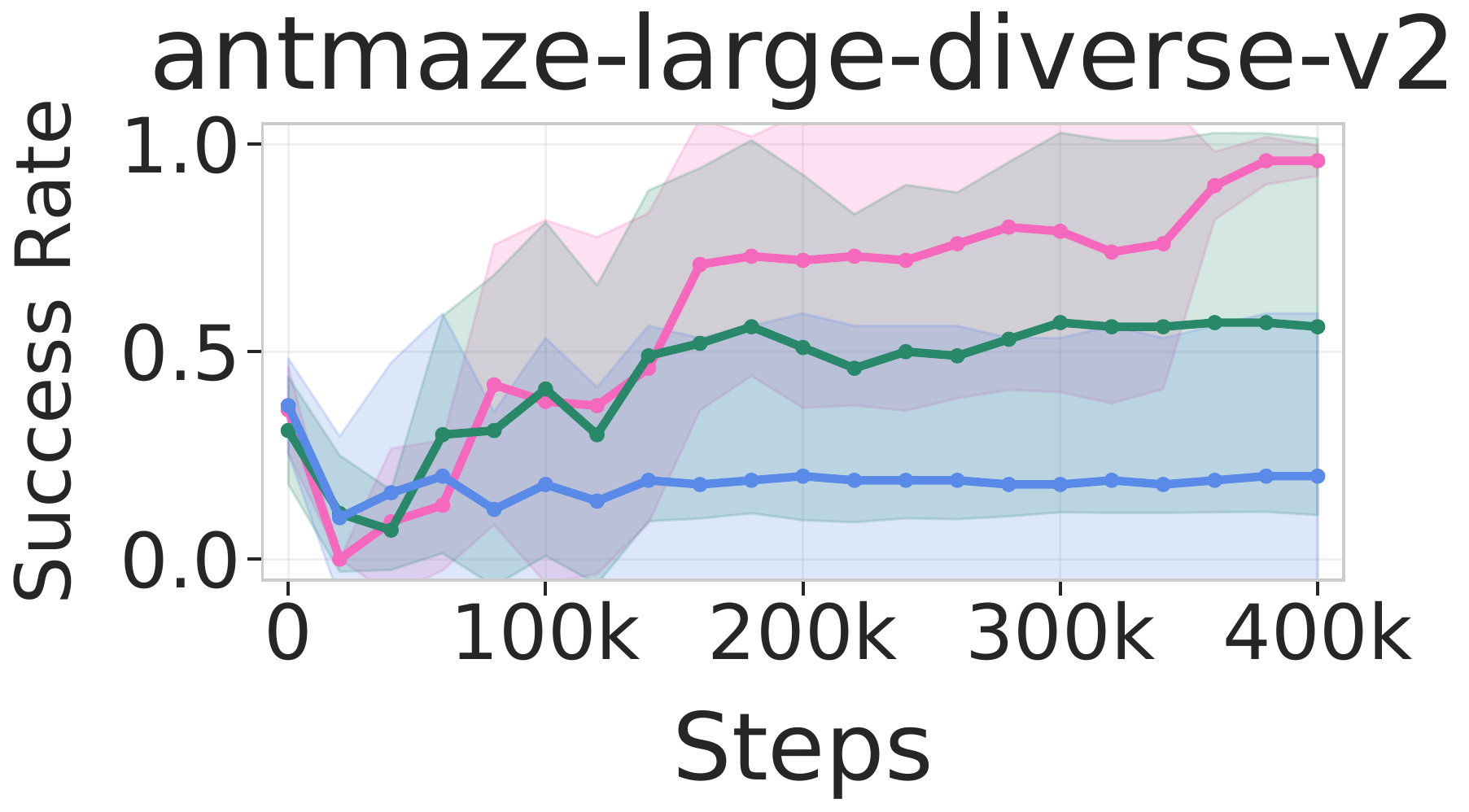}
    \end{subfigure}\hspace{0.1em}
    \begin{subfigure}[c]{0.22\textwidth}
        \includegraphics[width=1.0\linewidth]{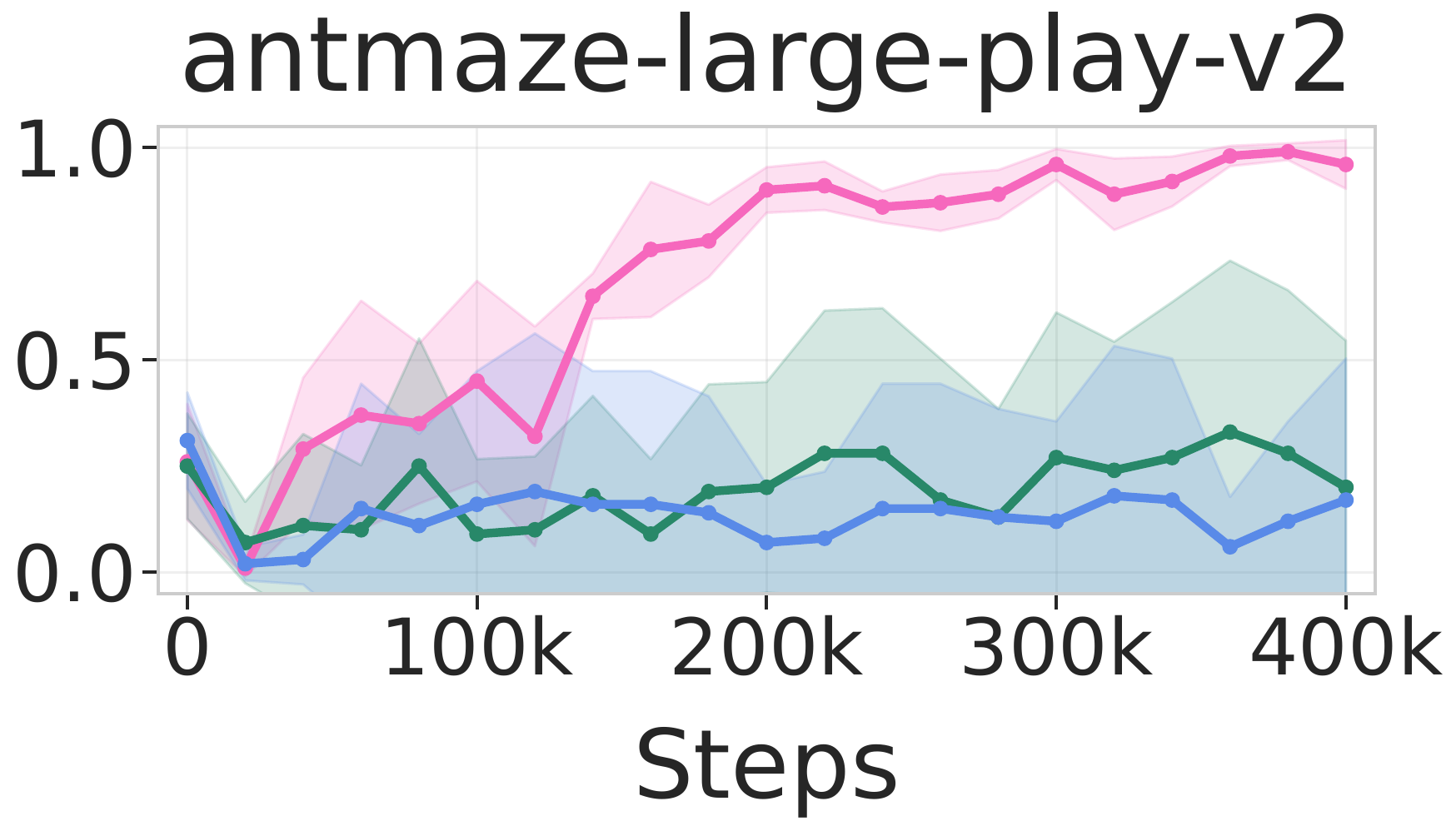}
    \end{subfigure}\hspace{0.1em}
    \begin{subfigure}[c]{0.22\textwidth}
        \includegraphics[width=1.0\linewidth]{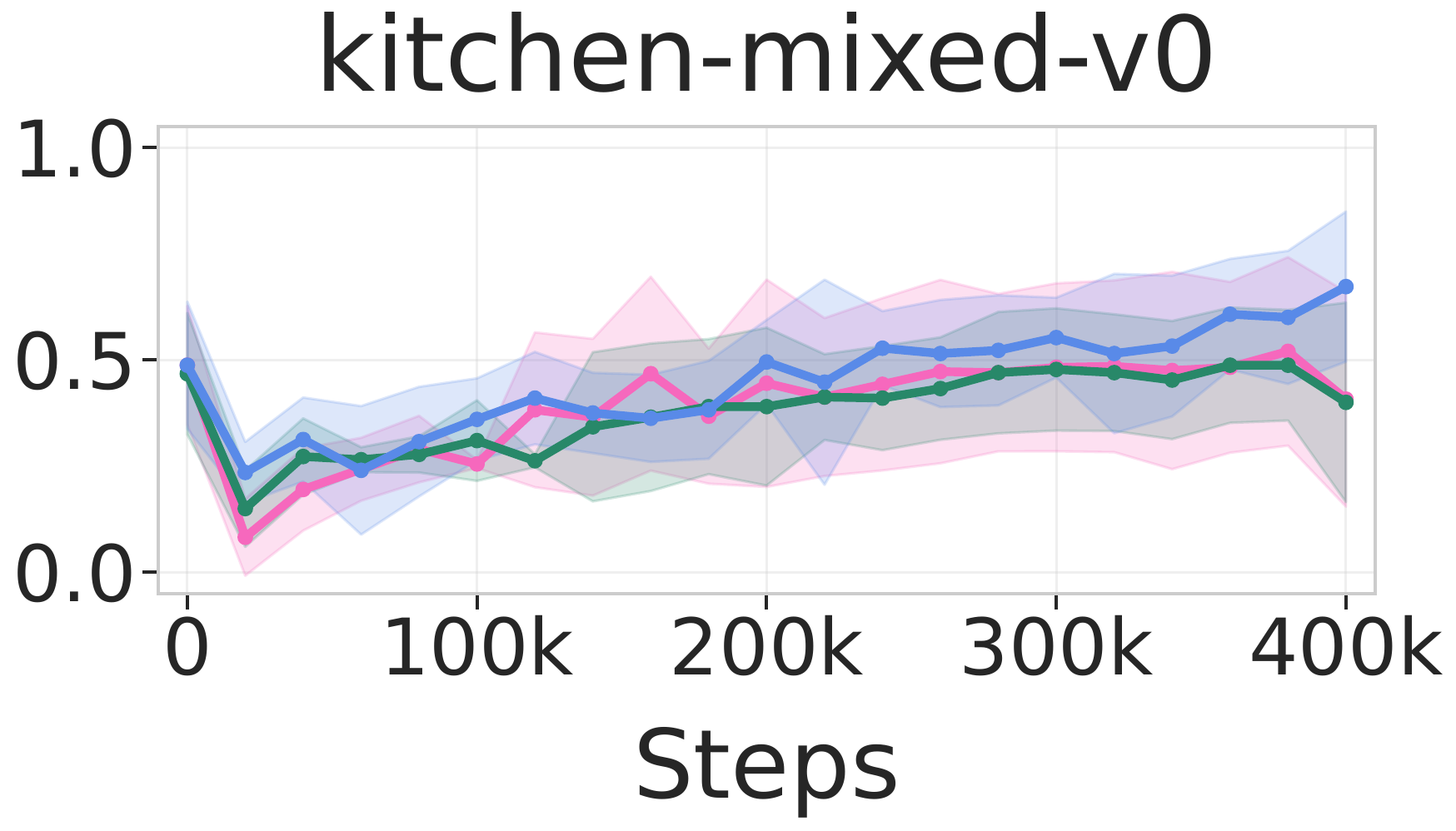}
    \end{subfigure}\hspace{0.1em}
    \begin{subfigure}[c]{0.22\textwidth}
        \includegraphics[width=1.0\linewidth]{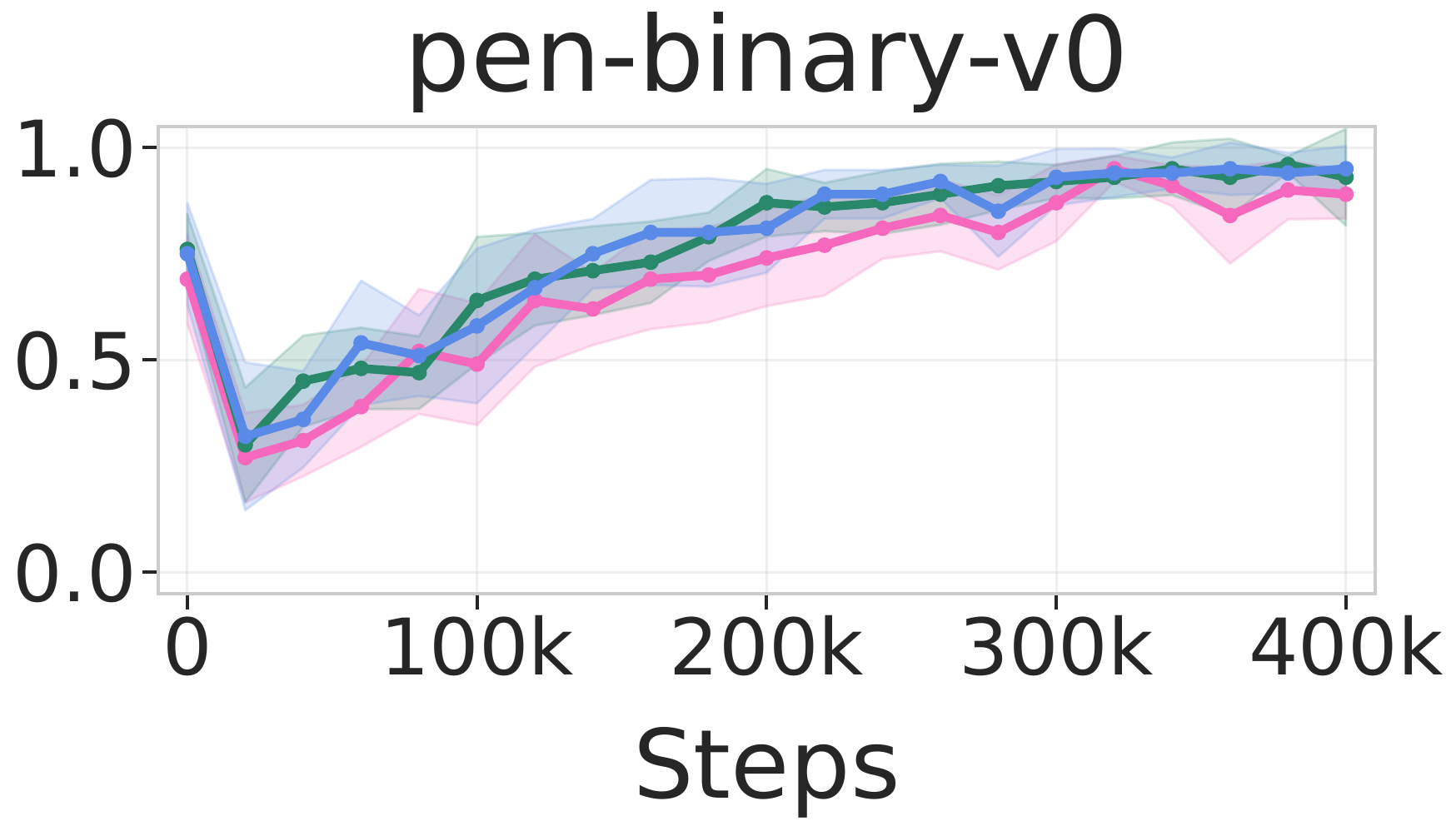}
    \end{subfigure}\\[1ex]
    \begin{subfigure}[c]{0.07\textwidth}
        \includegraphics[width=1.0\textwidth]{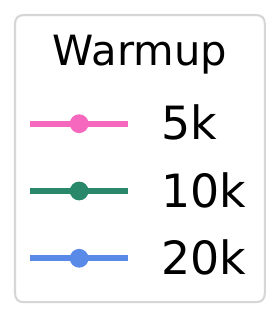}
    \end{subfigure}\hspace{0.1em}
    \begin{subfigure}[c]{0.22\textwidth}
        \includegraphics[width=1.0\linewidth]{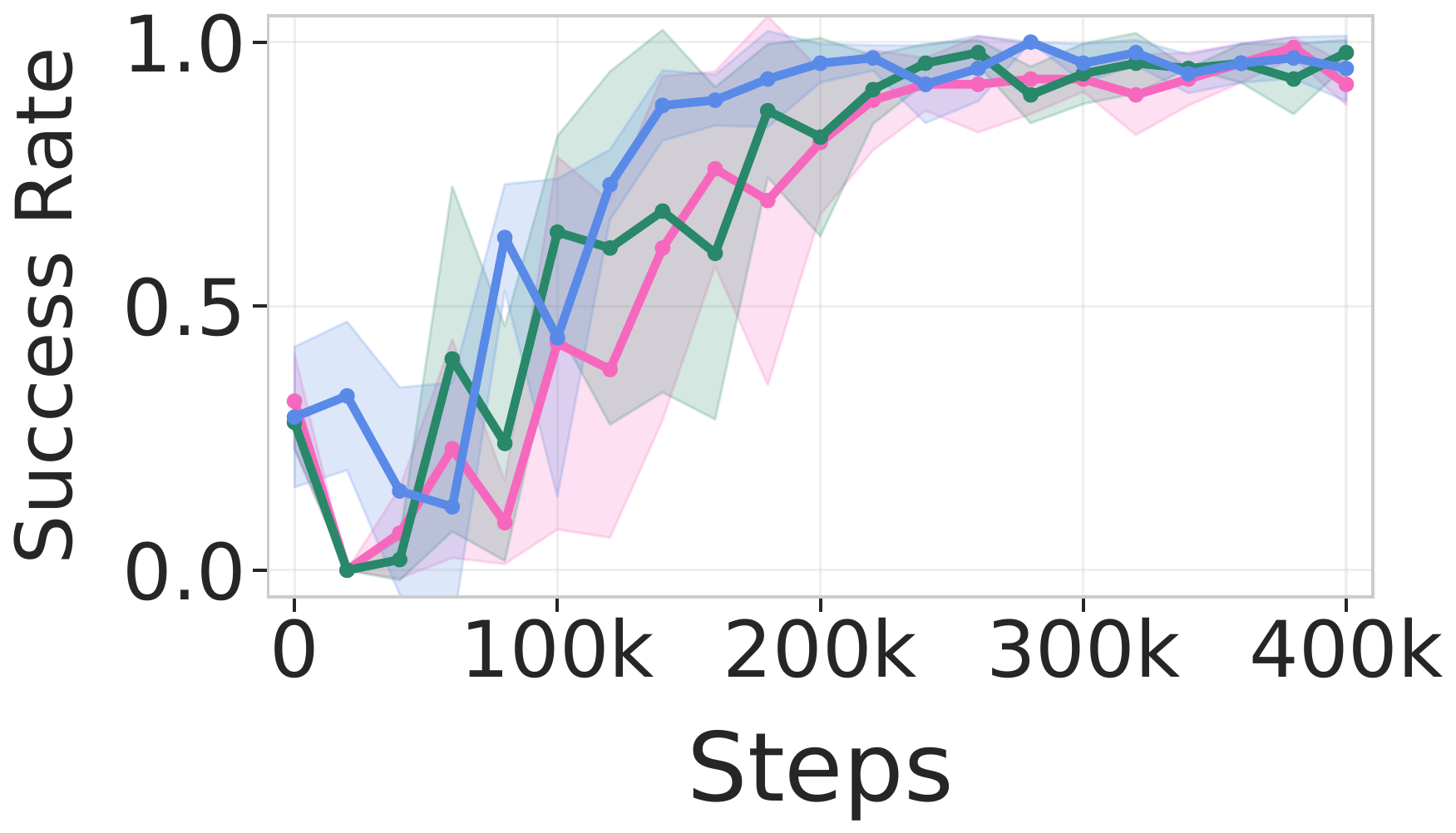}
    \end{subfigure}\hspace{0.1em}
    \begin{subfigure}[c]{0.22\textwidth}
        \includegraphics[width=1.0\linewidth]{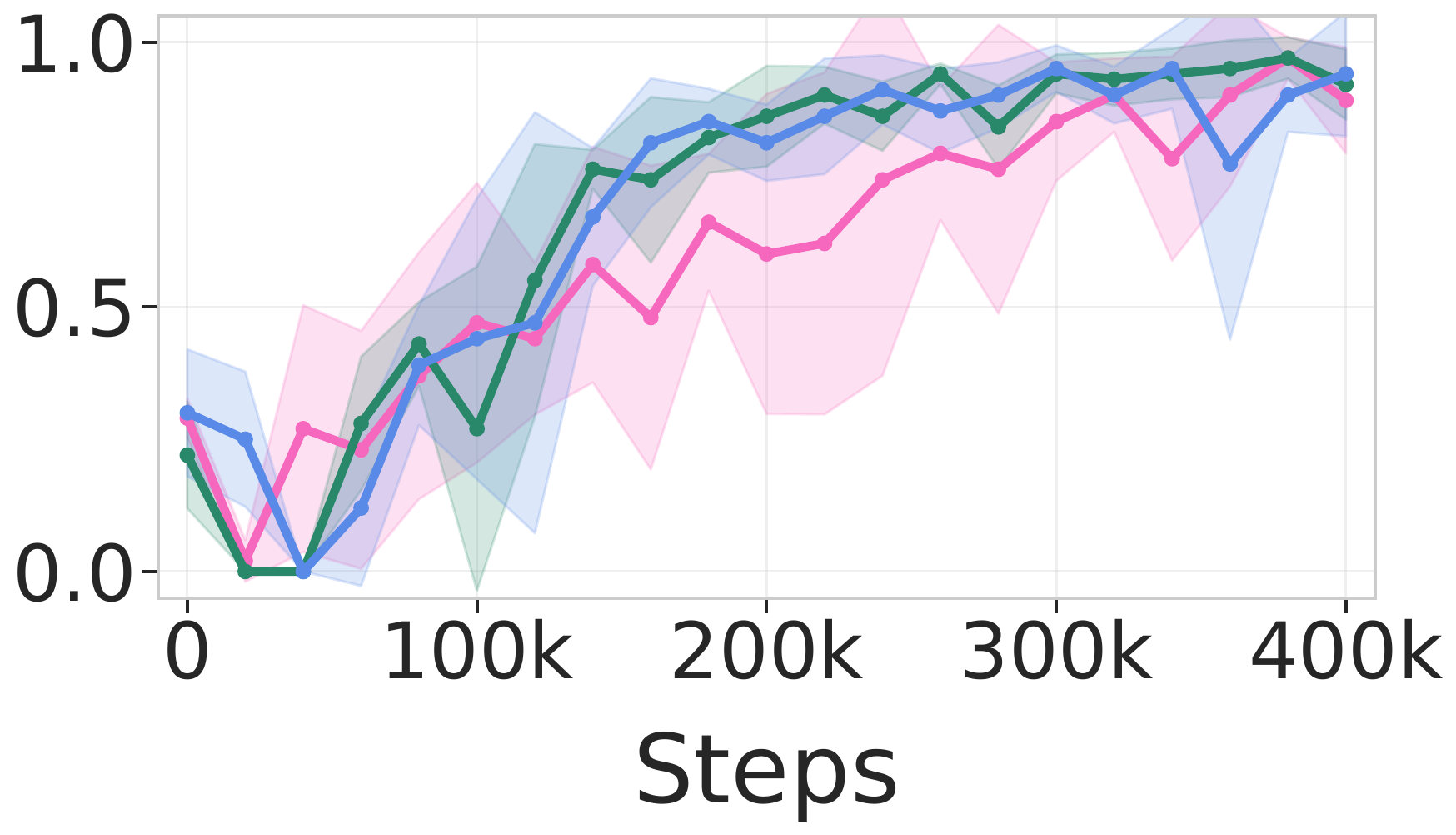}
    \end{subfigure}\hspace{0.1em}
    \begin{subfigure}[c]{0.22\textwidth}
        \includegraphics[width=1.0\linewidth]{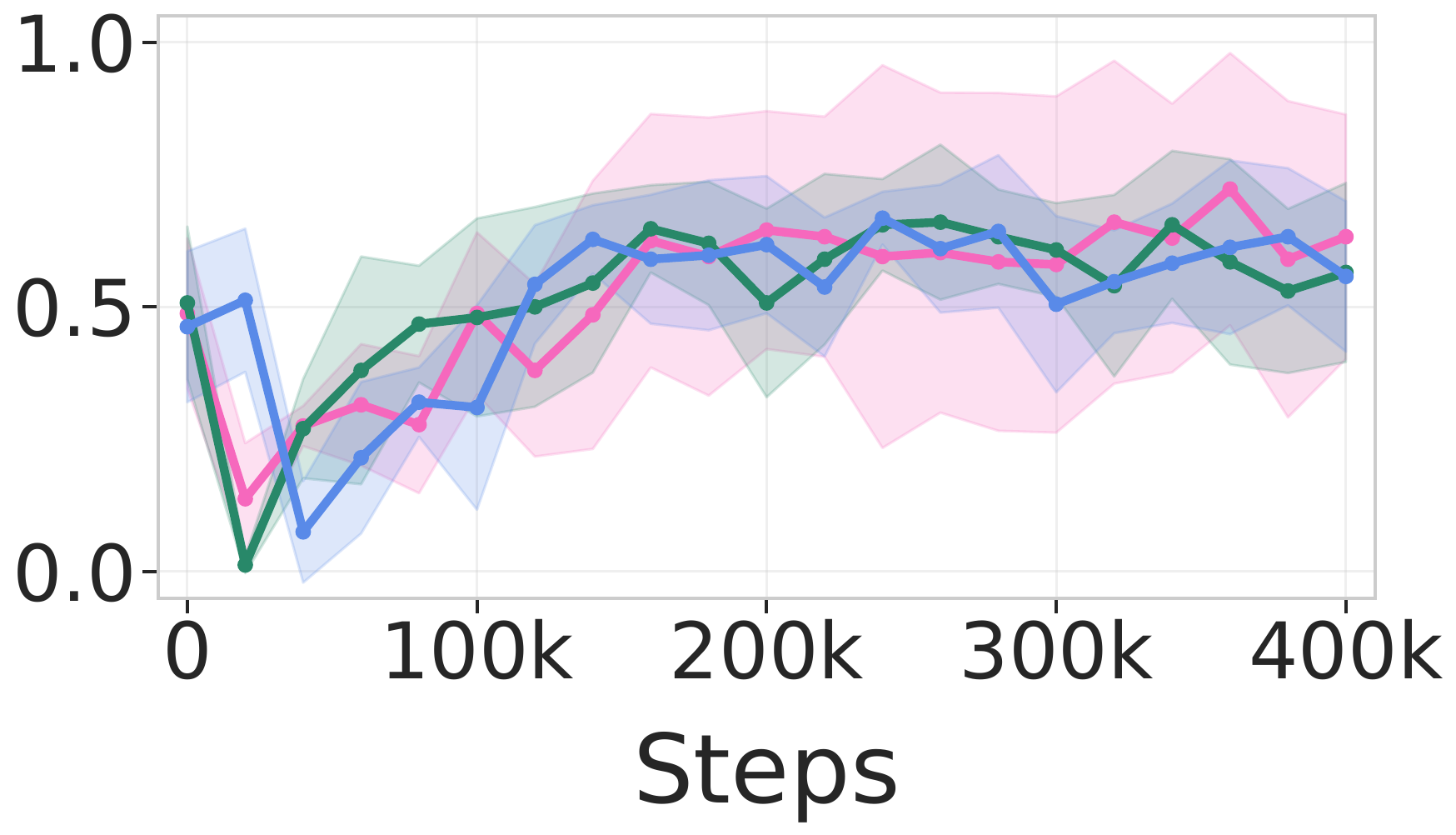}
    \end{subfigure}\hspace{0.1em}
    \begin{subfigure}[c]{0.22\textwidth}
        \includegraphics[width=1.0\linewidth]{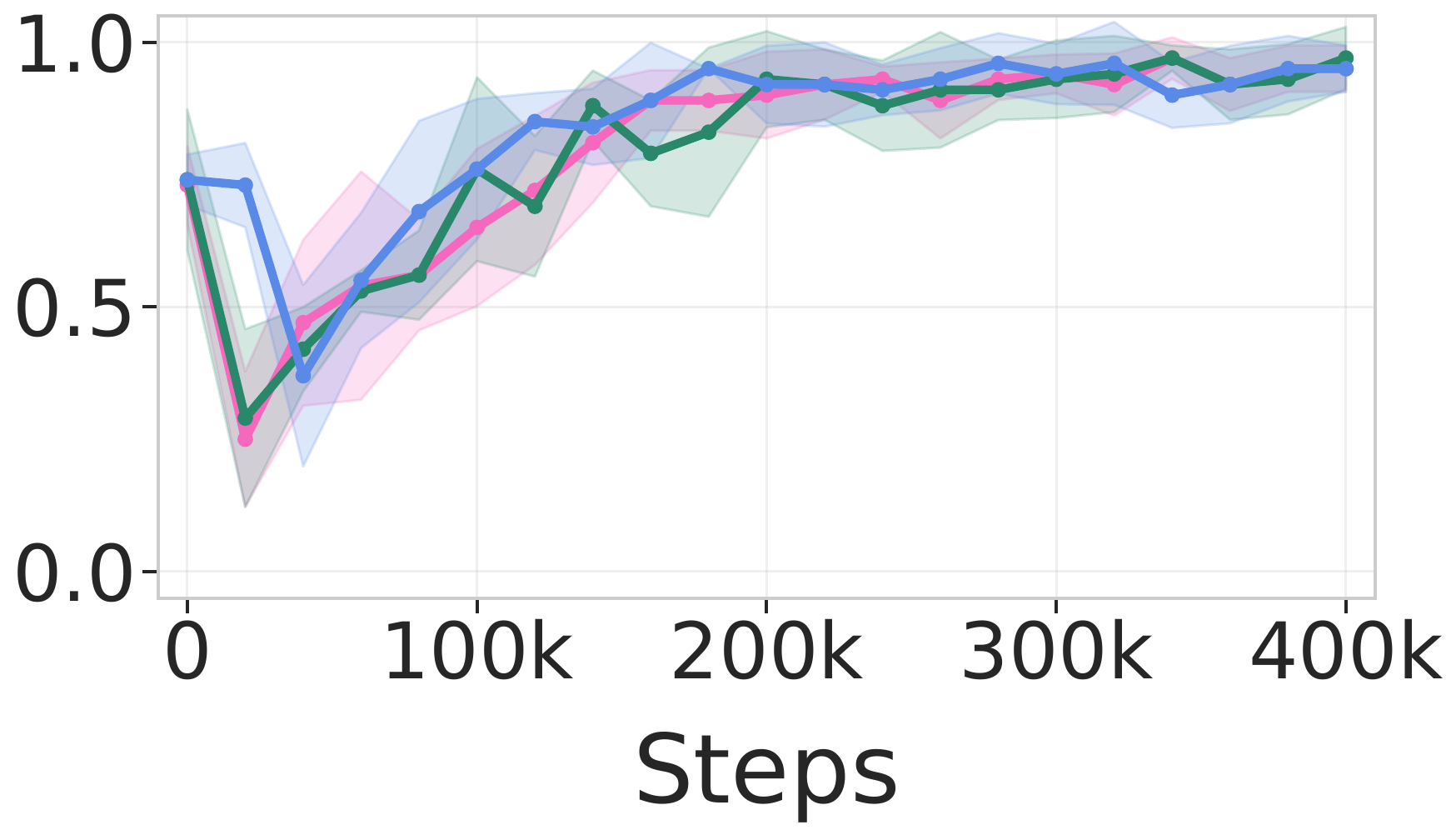}
    \end{subfigure}\\[1ex]
    \begin{subfigure}[c]{0.07\textwidth}
        \includegraphics[width=1.0\textwidth]{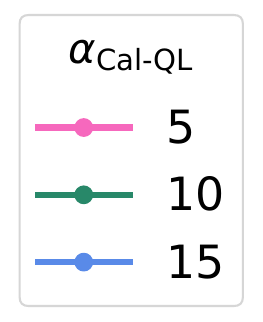}
    \end{subfigure}\hspace{0.1em}
    \begin{subfigure}[c]{0.22\textwidth}
        \includegraphics[width=1.0\linewidth]{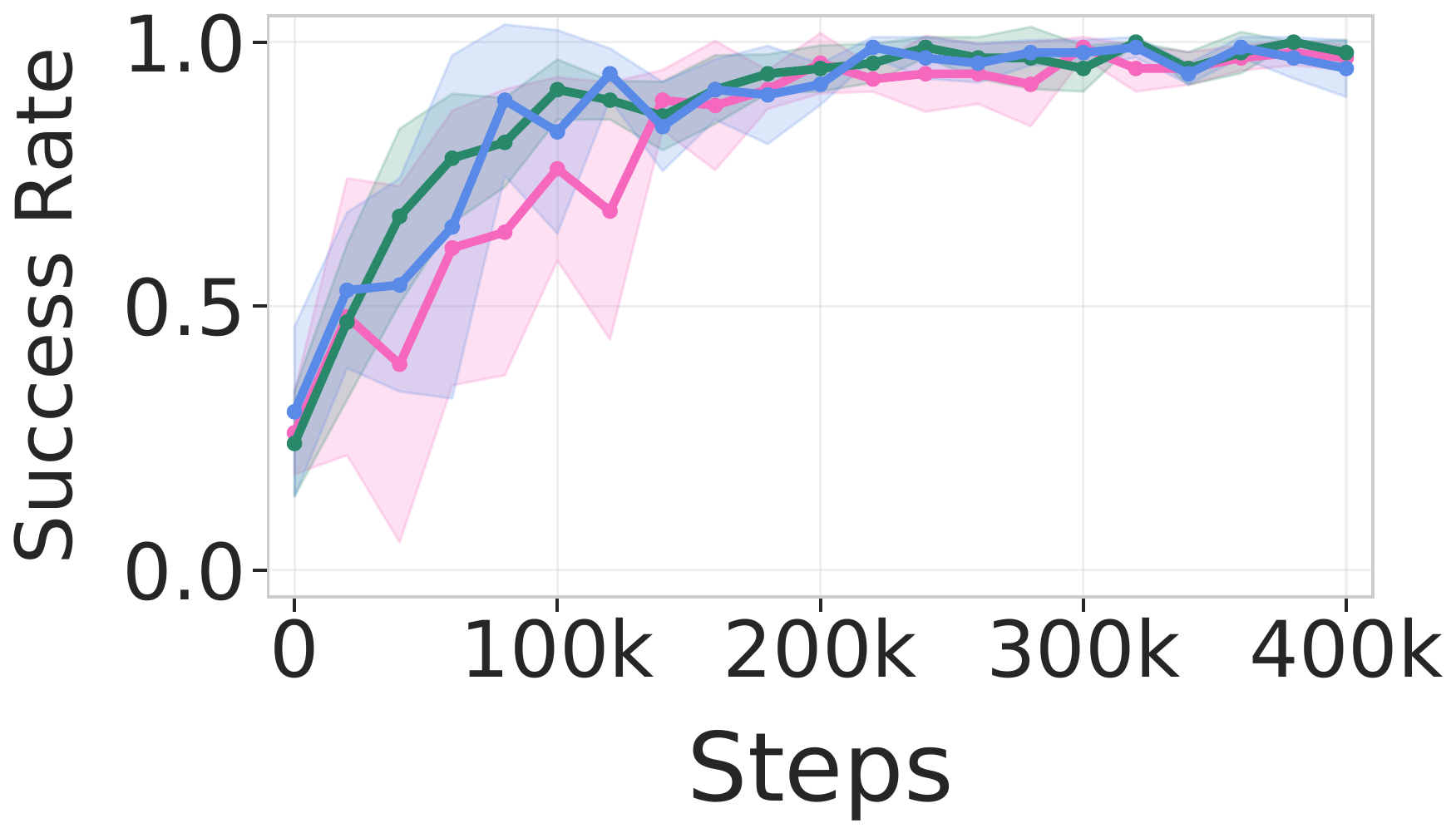}
    \end{subfigure}\hspace{0.1em}
    \begin{subfigure}[c]{0.22\textwidth}
        \includegraphics[width=1.0\linewidth]{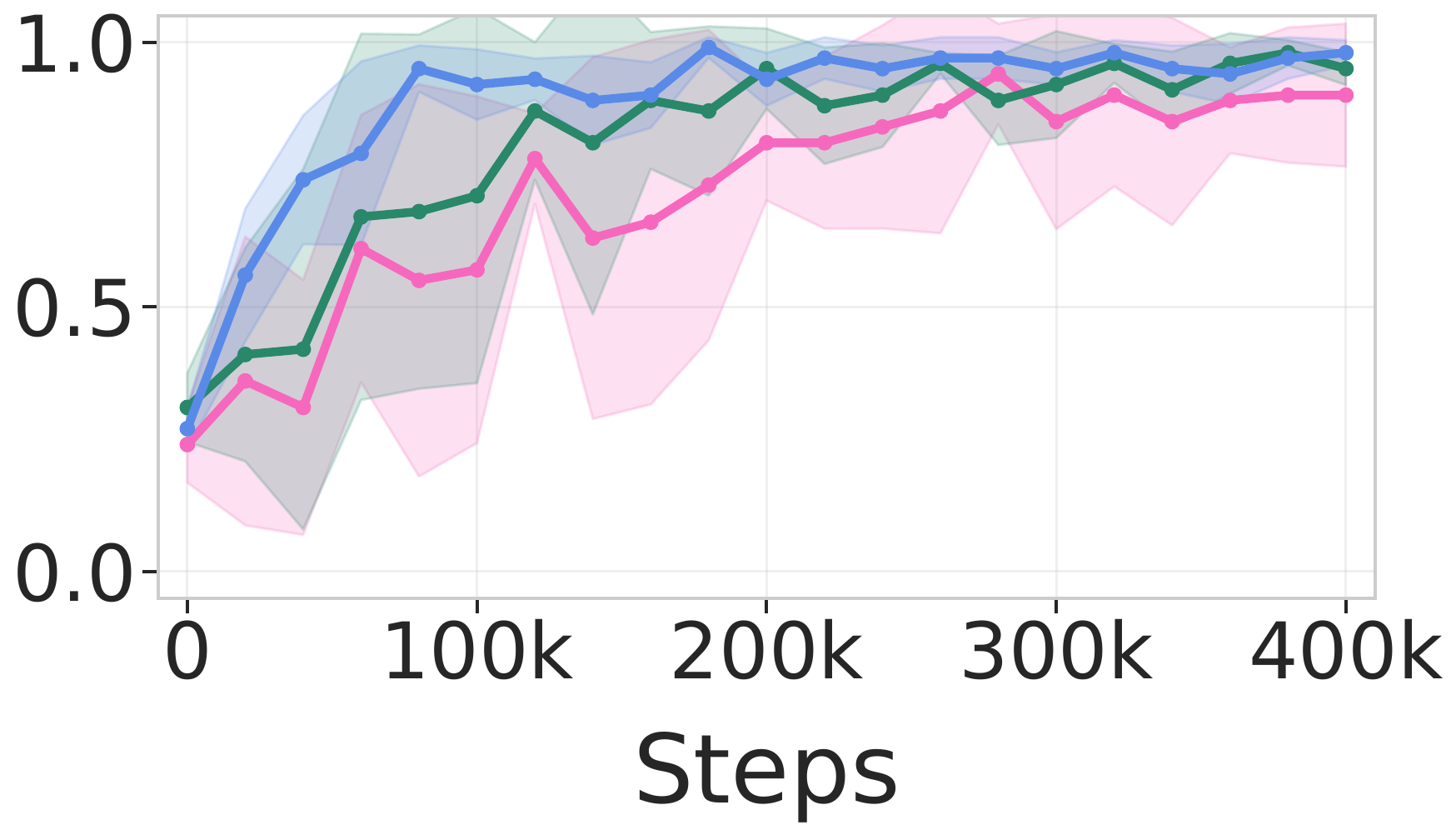}
    \end{subfigure}\hspace{0.1em}
    \begin{subfigure}[c]{0.22\textwidth}
        \includegraphics[width=1.0\linewidth]{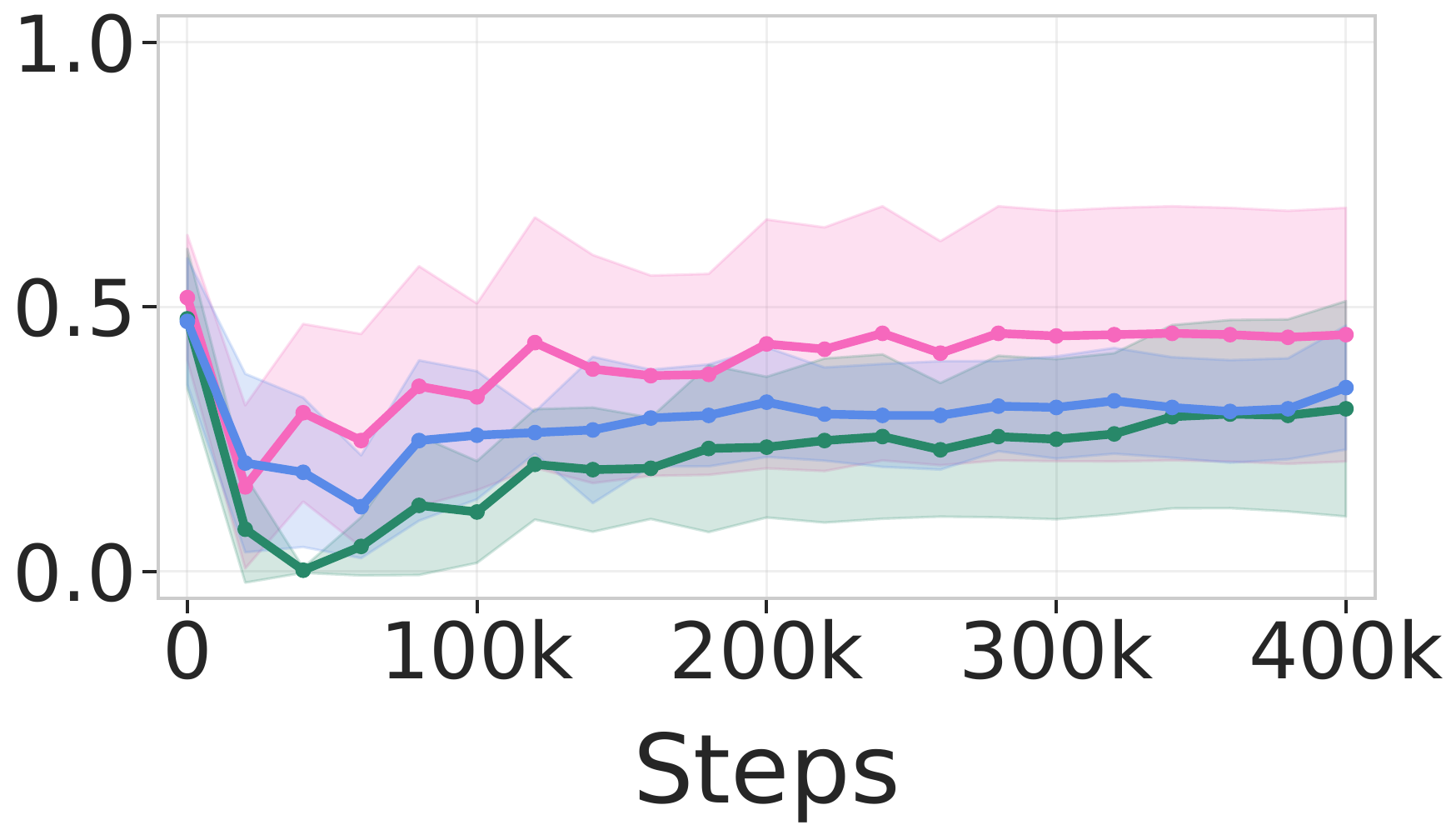}
    \end{subfigure}\hspace{0.1em}
    \begin{subfigure}[c]{0.22\textwidth}
        \includegraphics[width=1.0\linewidth]{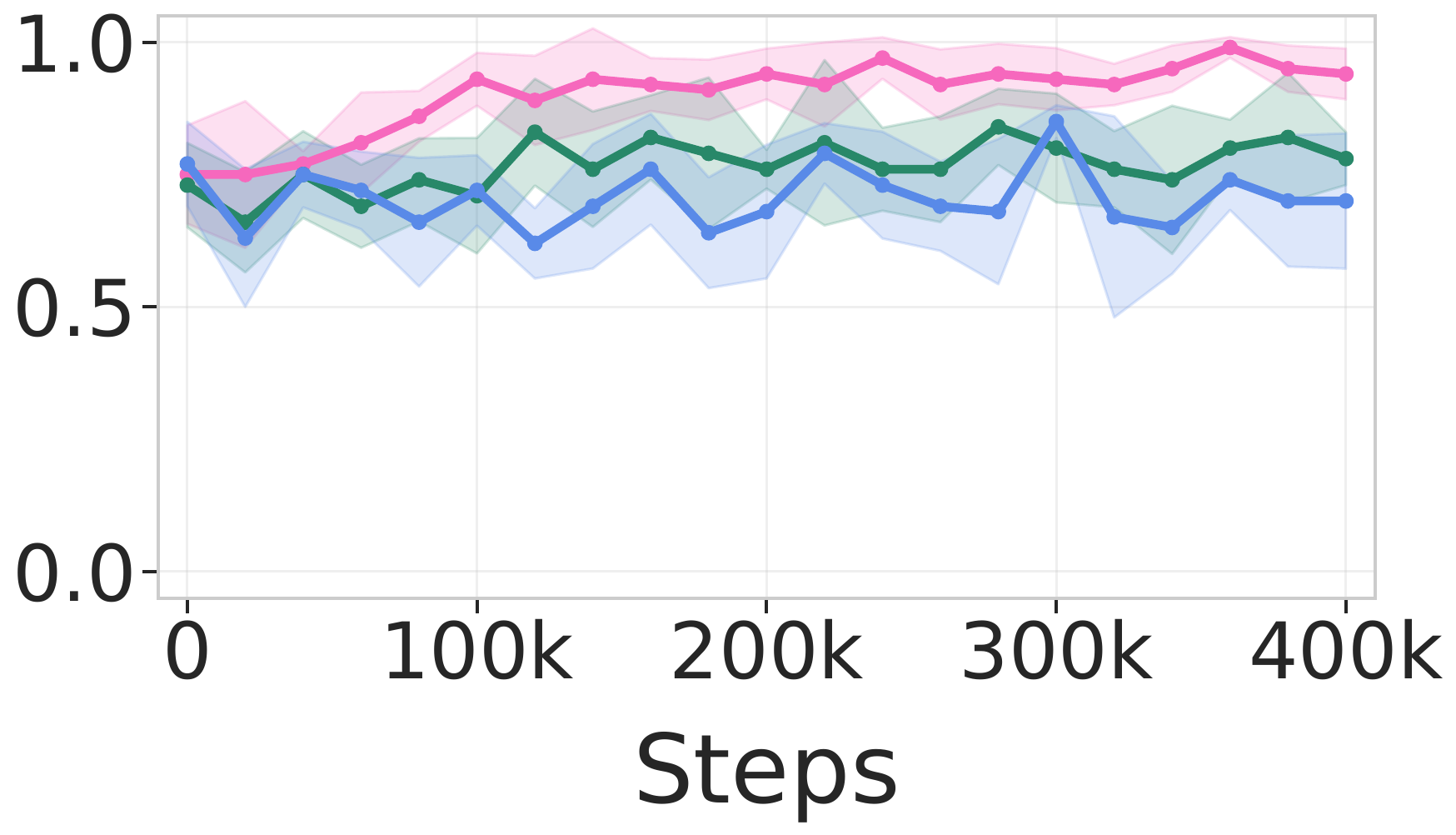}
    \end{subfigure}\\[1ex]
    \vspace{-0.5em}
    \caption{\textbf{Common stabilization strategies do not prevent catastrophic failure.}
    Effects of varying standard online fine-tuning choices across four tasks.
    \emph{Top}: UTD ratio; \emph{Middle}: warmup length; \emph{Bottom}: Cal-QL
    conservative-regularizer weight, $\alpha_{\mathrm{Cal\text{-}QL}}$.
    Increasing UTD, extending warmup, or tuning $\alpha_{\mathrm{Cal\text{-}QL}}$ does not
    reliably prevent early performance collapse.}
    \label{fig:hyperparams}
\end{figure}

\begin{figure}[ht]
    \centering
    \begin{subfigure}[c]{0.23\textwidth}
        \includegraphics[width=1.0\linewidth]{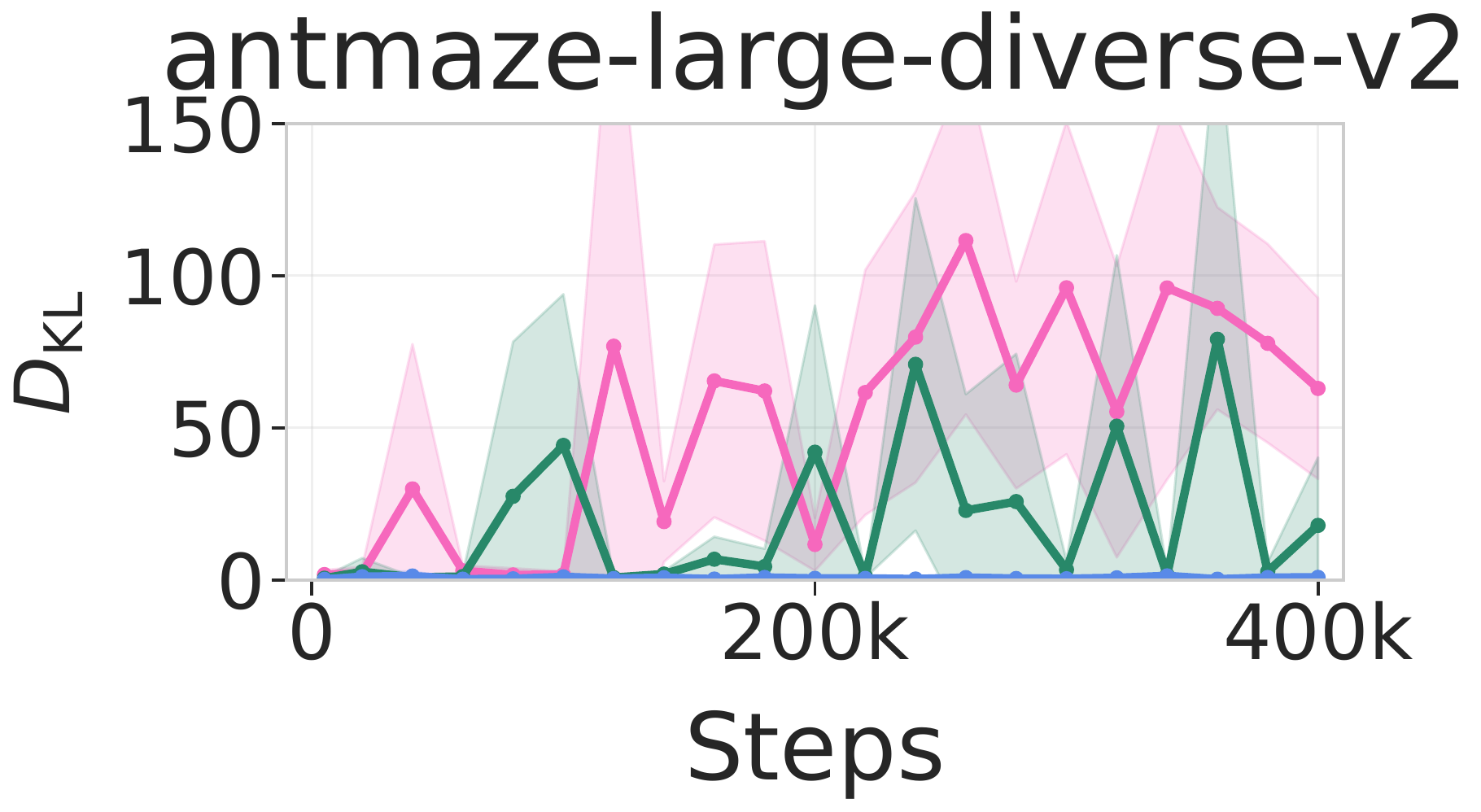}
    \end{subfigure}\hspace{0.1em}
    \begin{subfigure}[c]{0.23\textwidth}
        \includegraphics[width=1.0\linewidth]{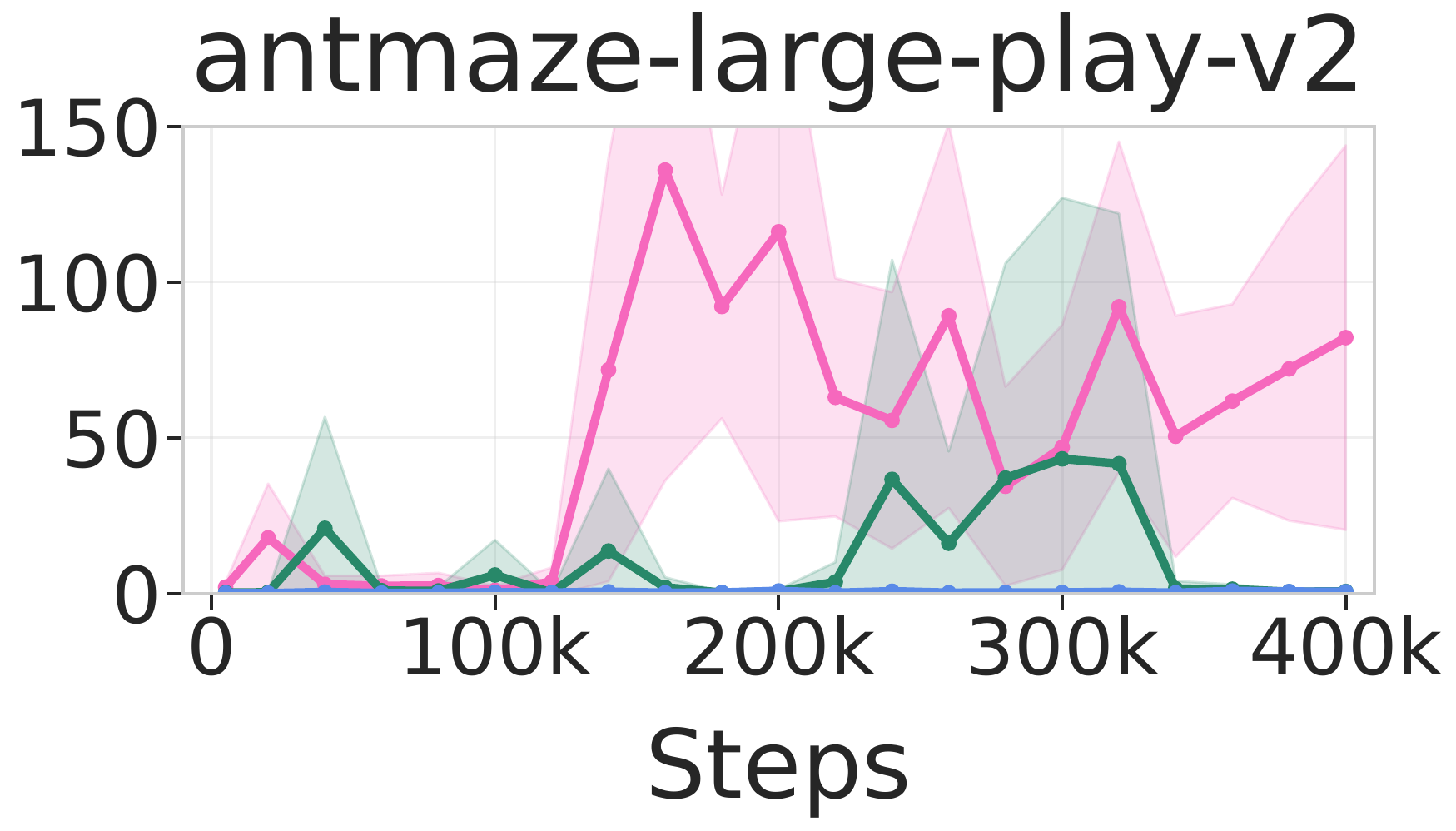}
    \end{subfigure}\hspace{0.1em}
    \begin{subfigure}[c]{0.23\textwidth}
        \includegraphics[width=1.0\linewidth]{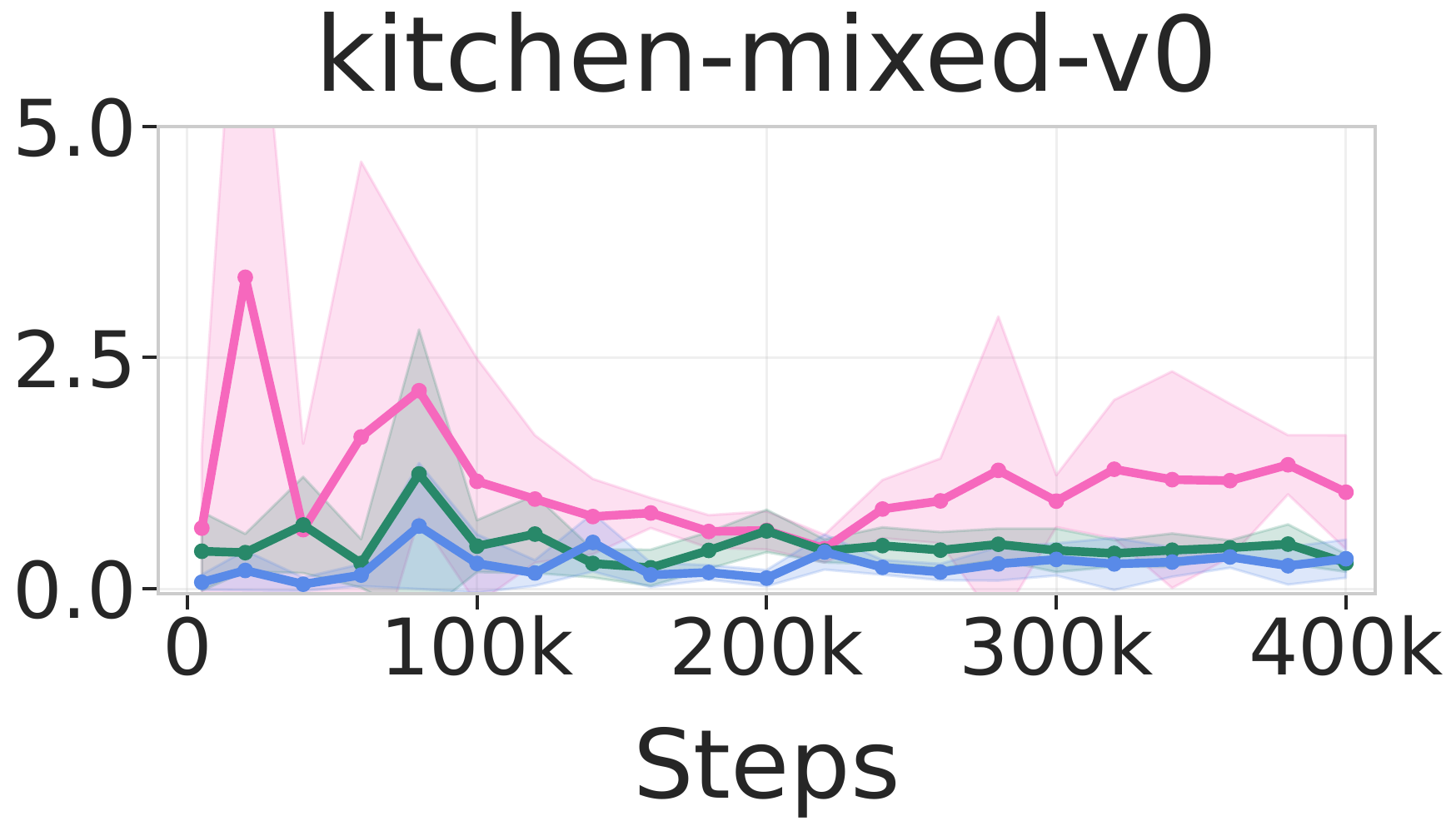}
    \end{subfigure}\hspace{0.1em}
    \begin{subfigure}[c]{0.23\textwidth}
        \includegraphics[width=1.0\linewidth]{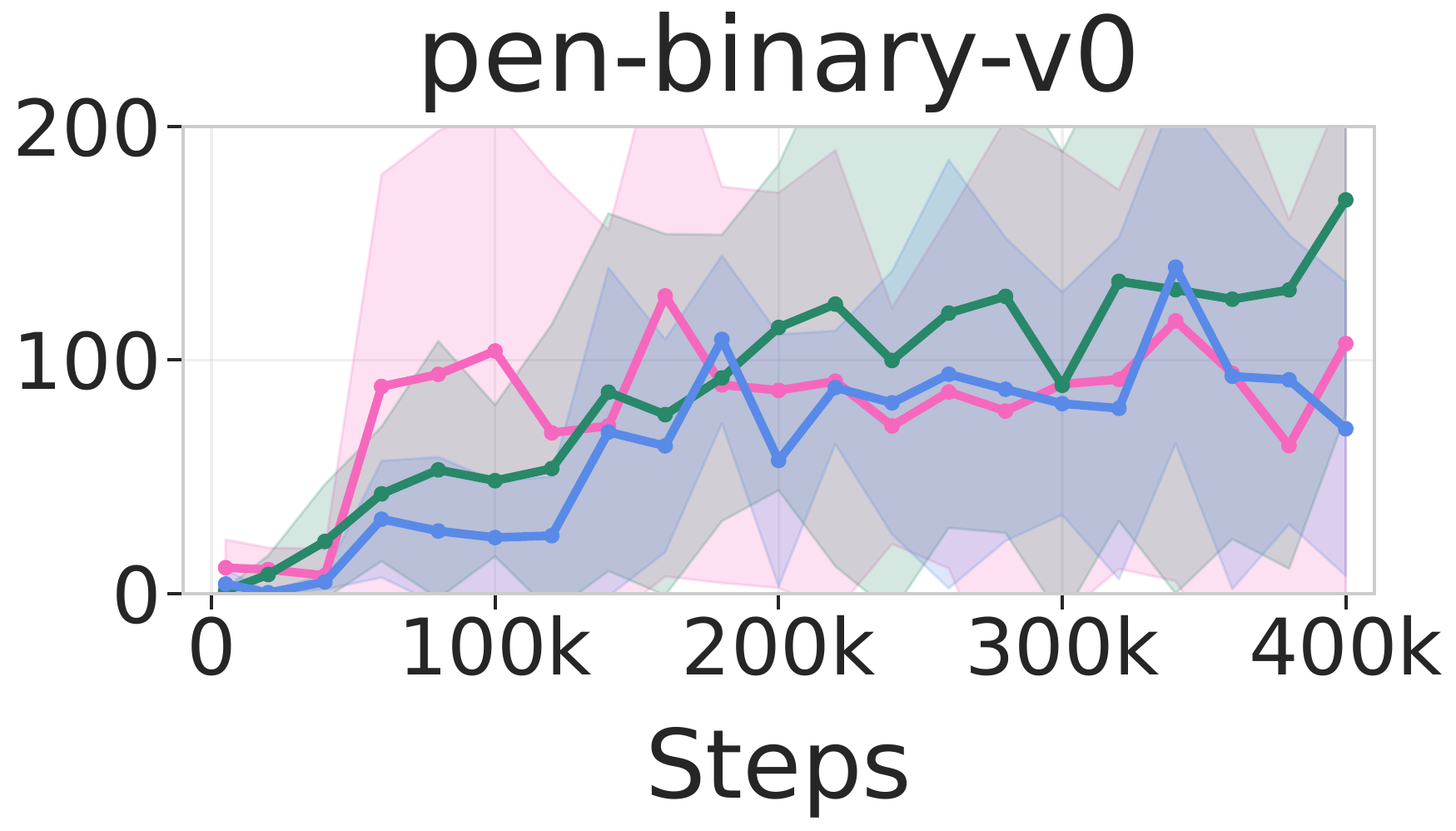}
    \end{subfigure}\\[1ex]
    \begin{subfigure}[c]{0.23\textwidth}
        \includegraphics[width=1.0\linewidth]{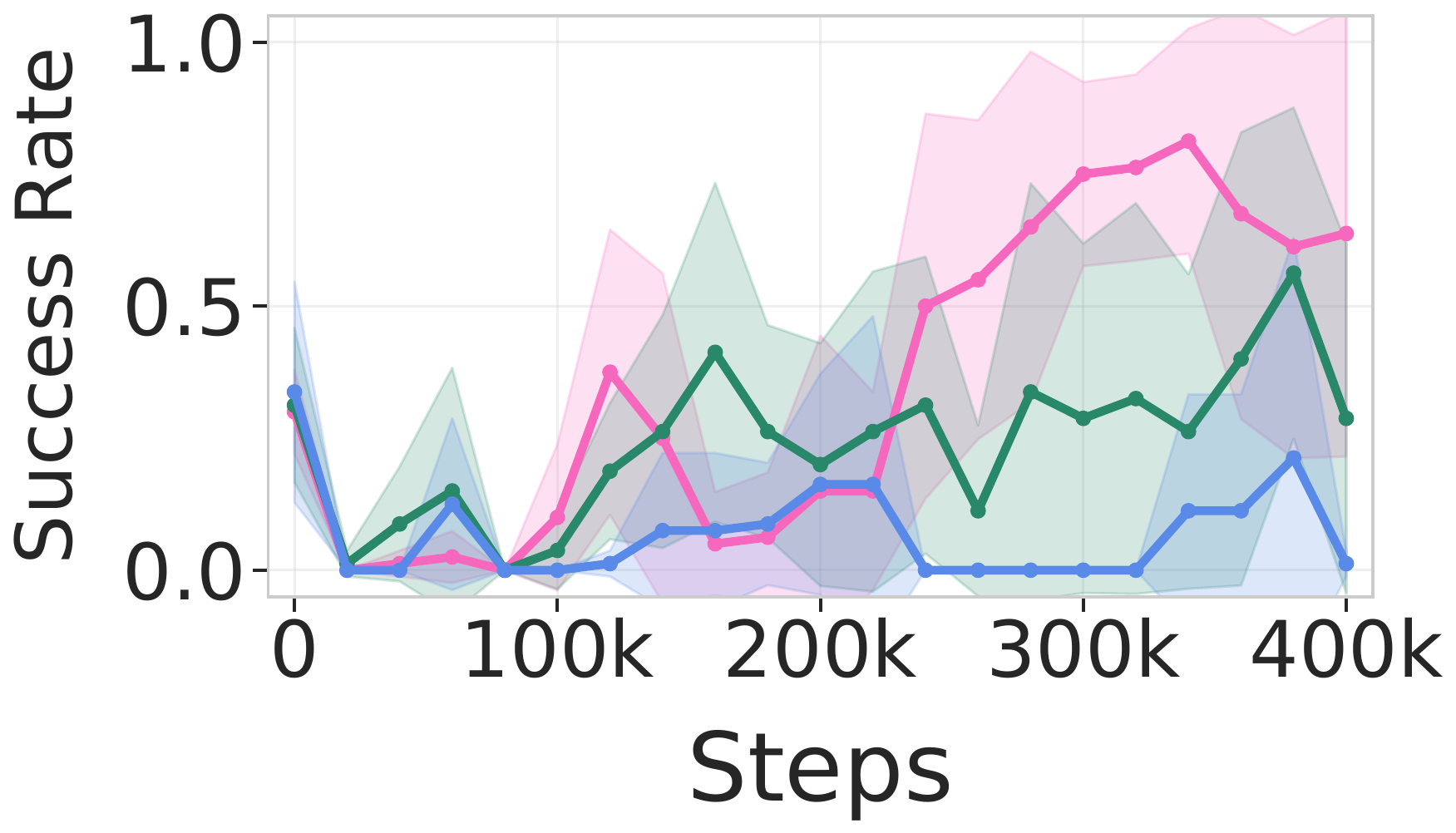}
    \end{subfigure}\hspace{0.1em}
    \begin{subfigure}[c]{0.23\textwidth}
        \includegraphics[width=1.0\linewidth]{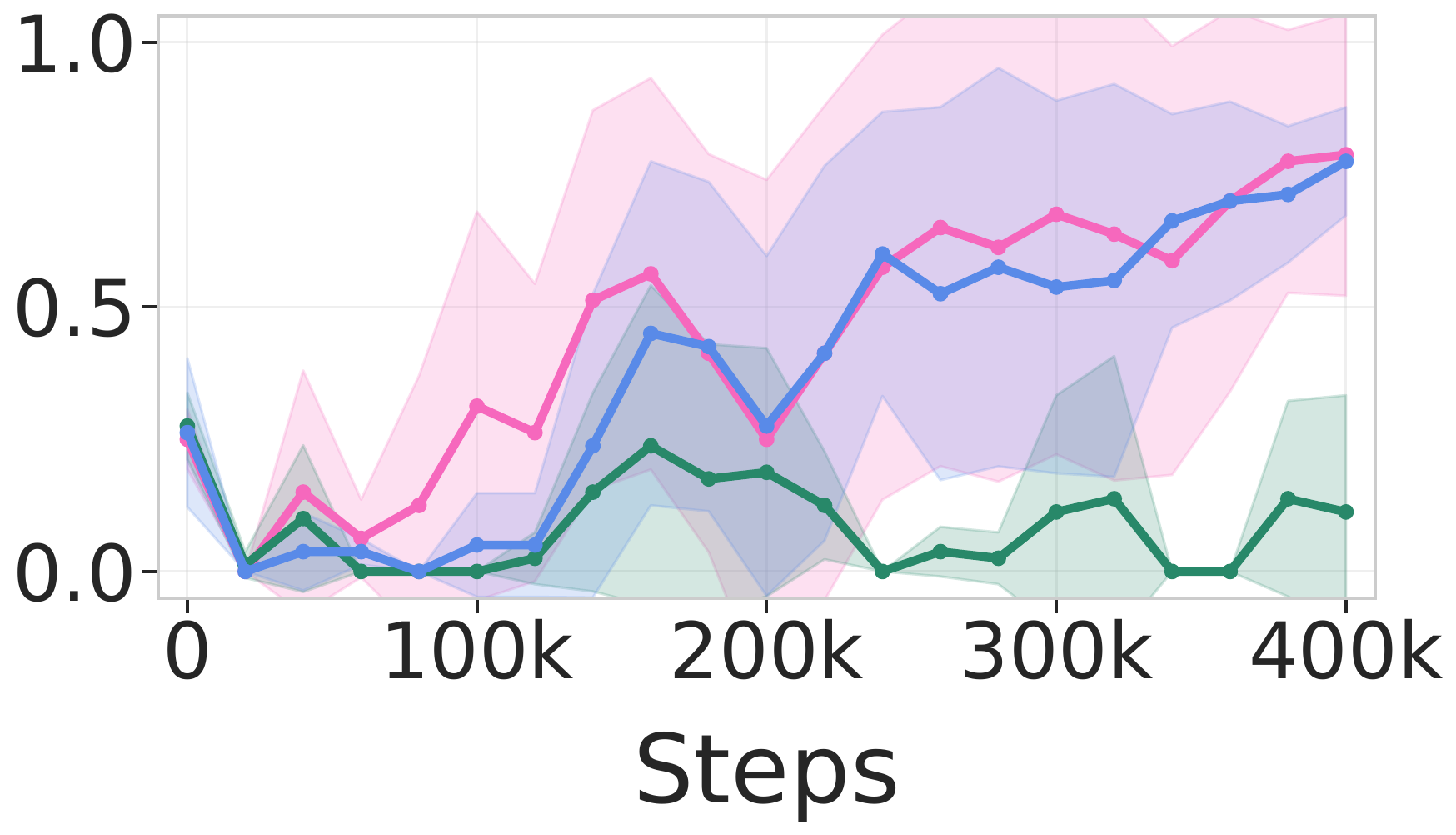}
    \end{subfigure}\hspace{0.1em}
    \begin{subfigure}[c]{0.23\textwidth}
        \includegraphics[width=1.0\linewidth]{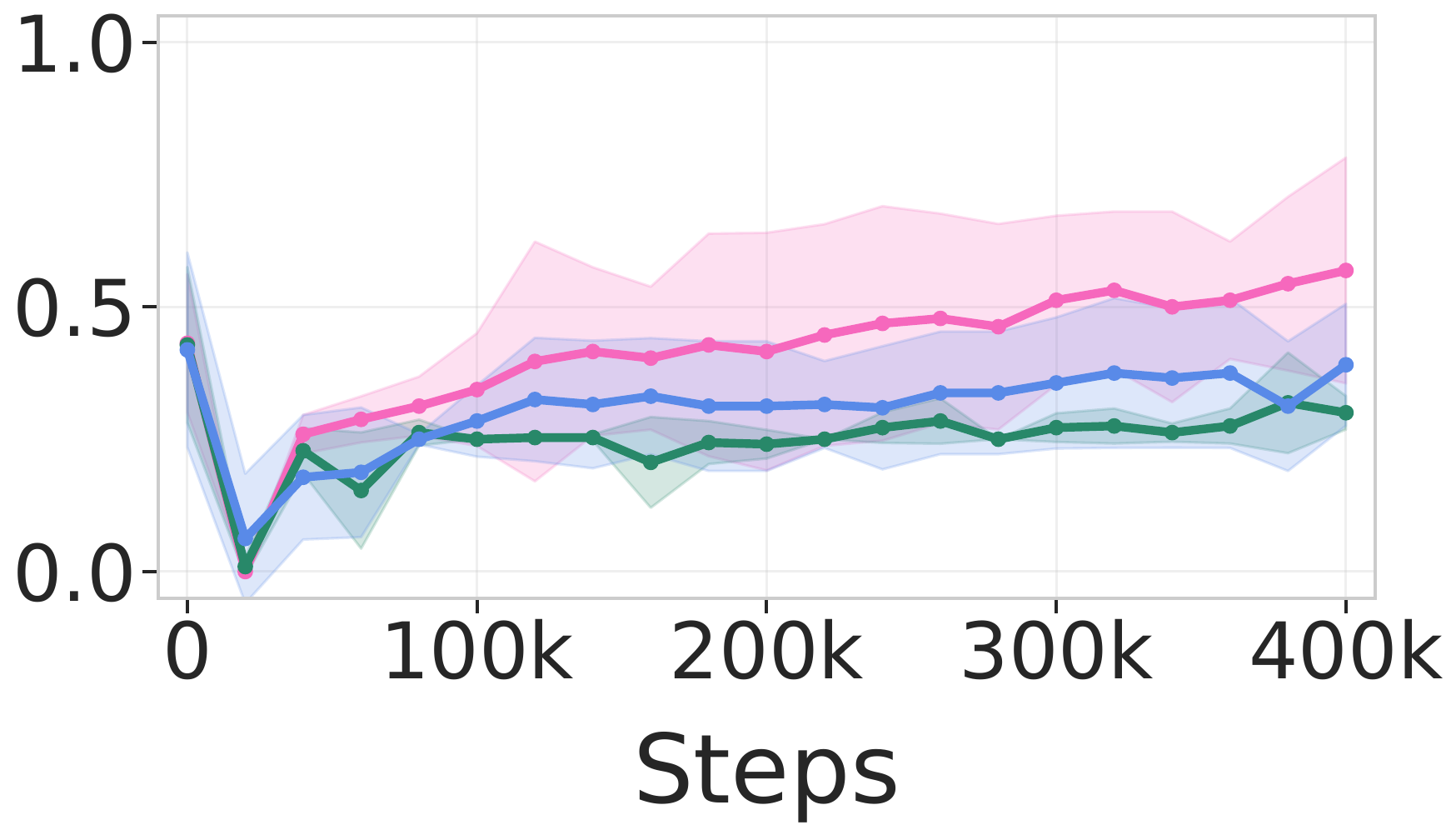}
    \end{subfigure}\hspace{0.1em}
    \begin{subfigure}[c]{0.23\textwidth}
        \includegraphics[width=1.0\linewidth]{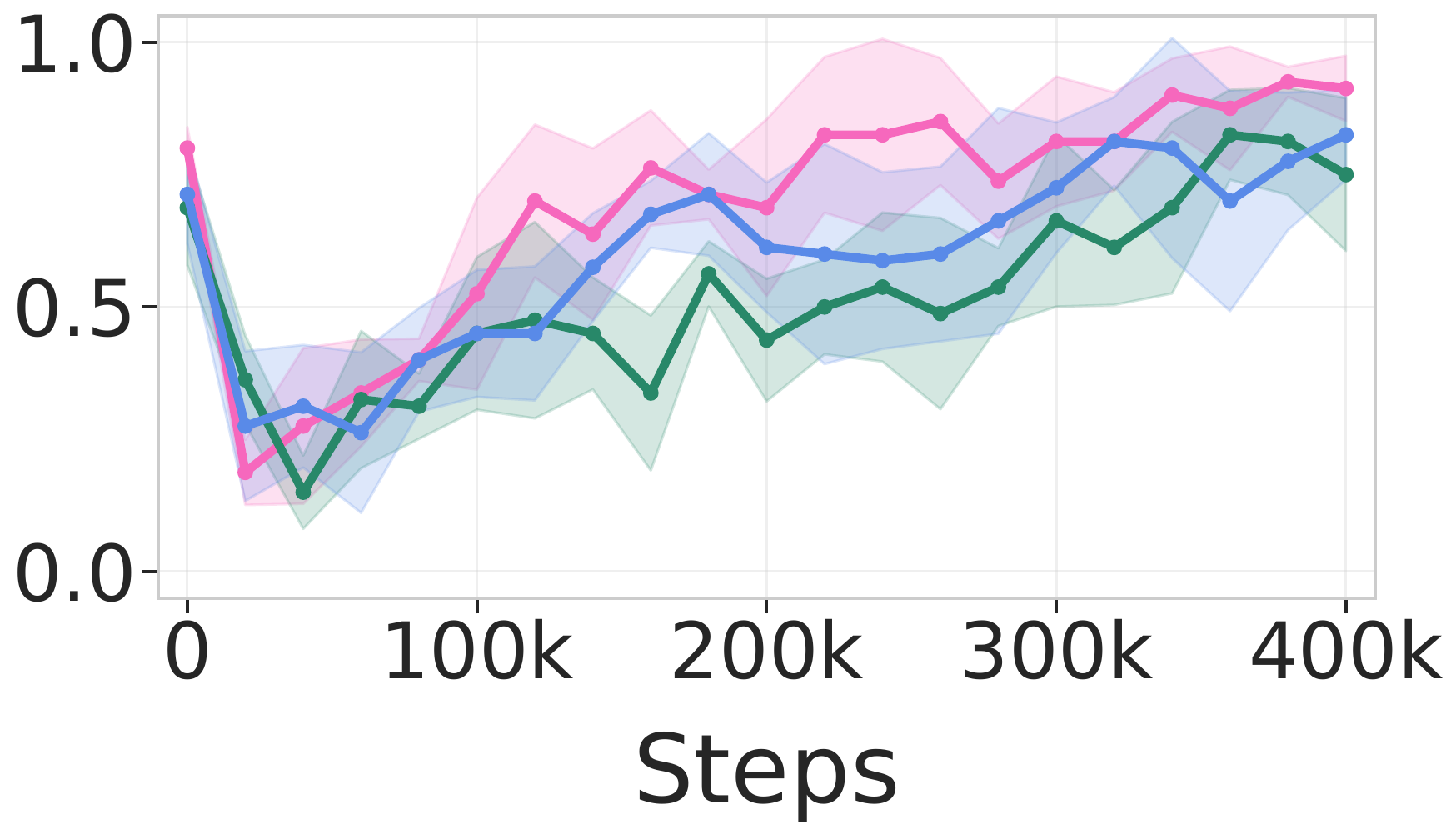}
    \end{subfigure}\\[1ex]
    \includegraphics[width=0.35\linewidth]{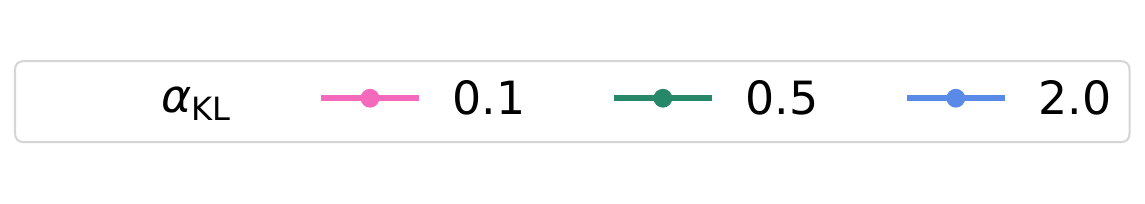}
    \vspace{-0.5em}
    \caption{\textbf{Critic-shift regularization does not prevent catastrophic failure.}
    Effects of varying the critic-shift penalty strength $\alpha_{\mathrm{KL}}$ across four
    tasks. \emph{Top}: critic distribution shift during online fine-tuning, measured by
    $D_{\mathrm{KL}}\big(\mathrm{softmax}(Q_{\mathrm{off}})\,\Vert\,
    \mathrm{softmax}(Q_{\theta})\big)$. \emph{Bottom}: corresponding success rates.
    Increasing $\alpha_{\mathrm{KL}}$ slows the critic shift, but does not reliably prevent
    early performance collapse.}
    \label{fig:critic_shift}
        \vspace{-0.5em}
\end{figure}

\clearpage

\textbf{How sensitive is ANCHOR to its hyperparameters?} We finally study the sensitivity of ANCHOR to three hyperparameters: the initial KL weight
$\beta_0$, the KL annealing interval $T_{\mathrm{KL}}$, and the residual online
conservative regularization strength $\alpha_{\mathrm{on}}$.

\begin{figure}[ht]
    \centering
    \begin{subfigure}[c]{0.1\textwidth}
        \includegraphics[width=1.0\textwidth]{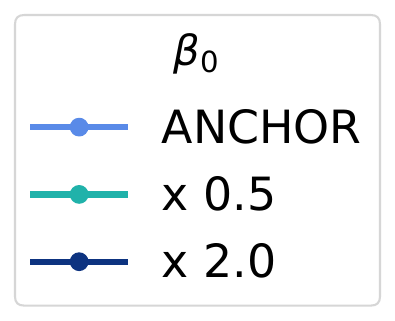}
    \end{subfigure}\hspace{0.1em}
    \begin{subfigure}[c]{0.26\textwidth}
        \includegraphics[width=1.0\linewidth]{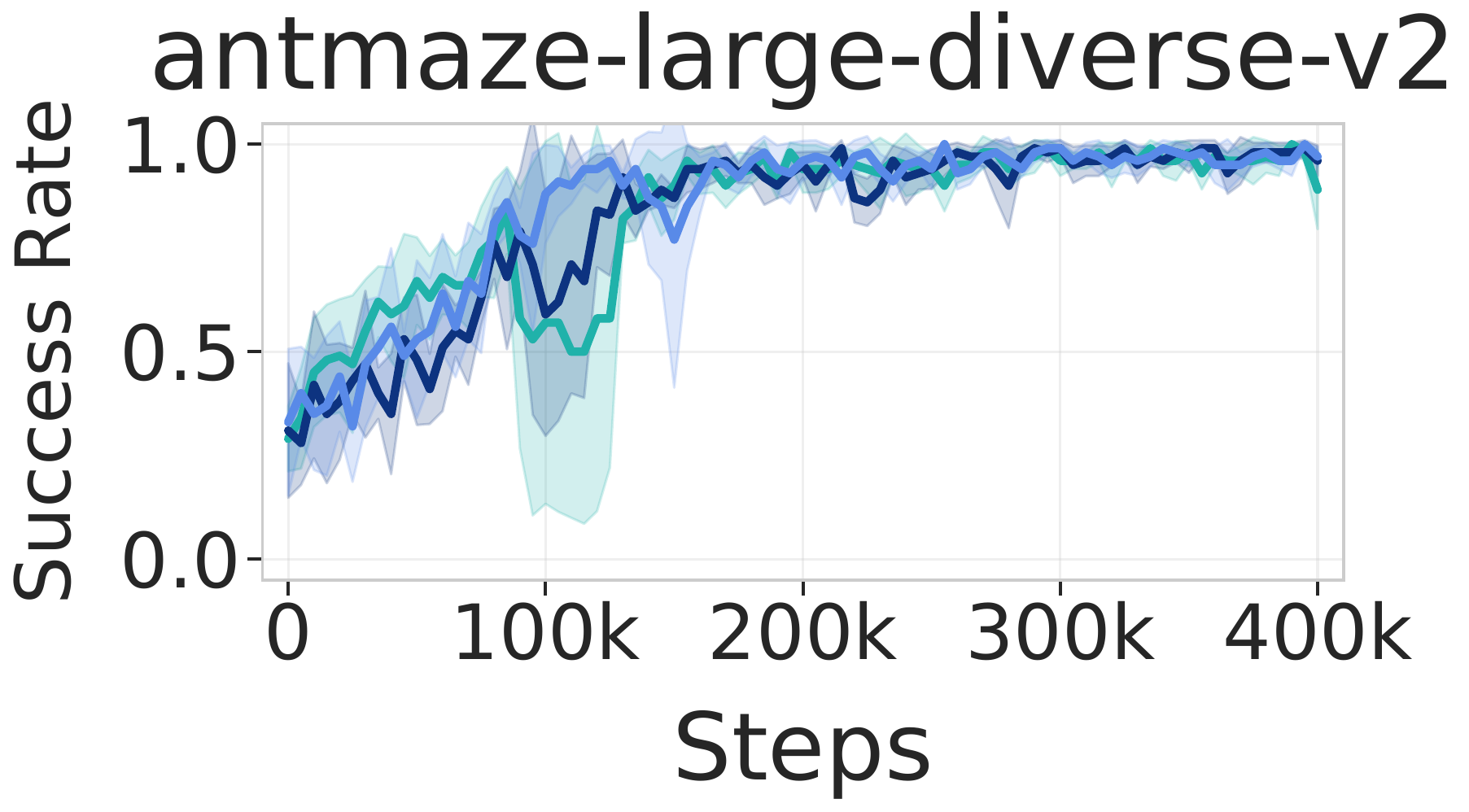}
    \end{subfigure}\hspace{0.1em}
    \begin{subfigure}[c]{0.26\textwidth}
        \includegraphics[width=1.0\linewidth]{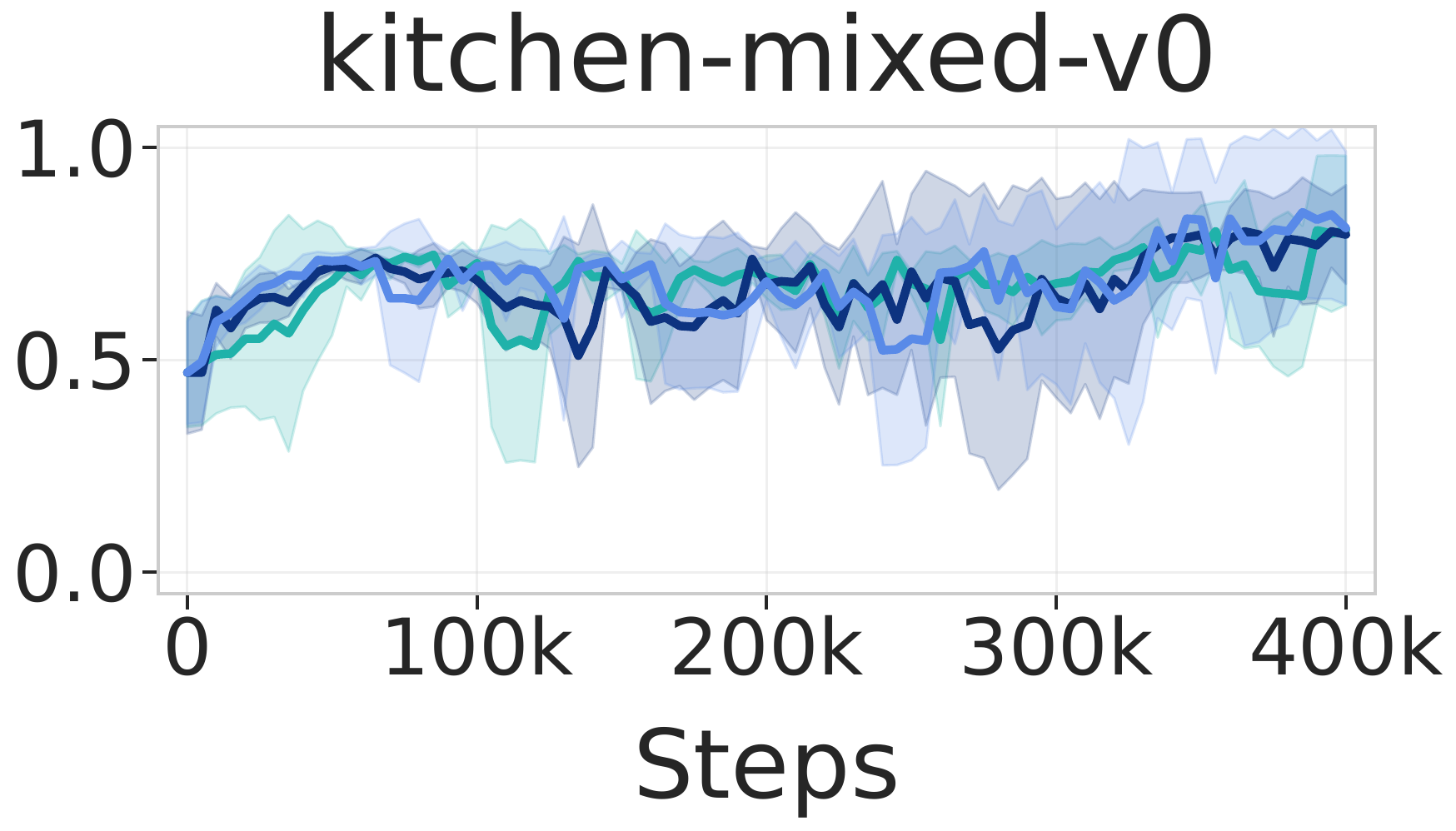}
    \end{subfigure}\hspace{0.1em}
    \begin{subfigure}[c]{0.26\textwidth}
        \includegraphics[width=1.0\linewidth]{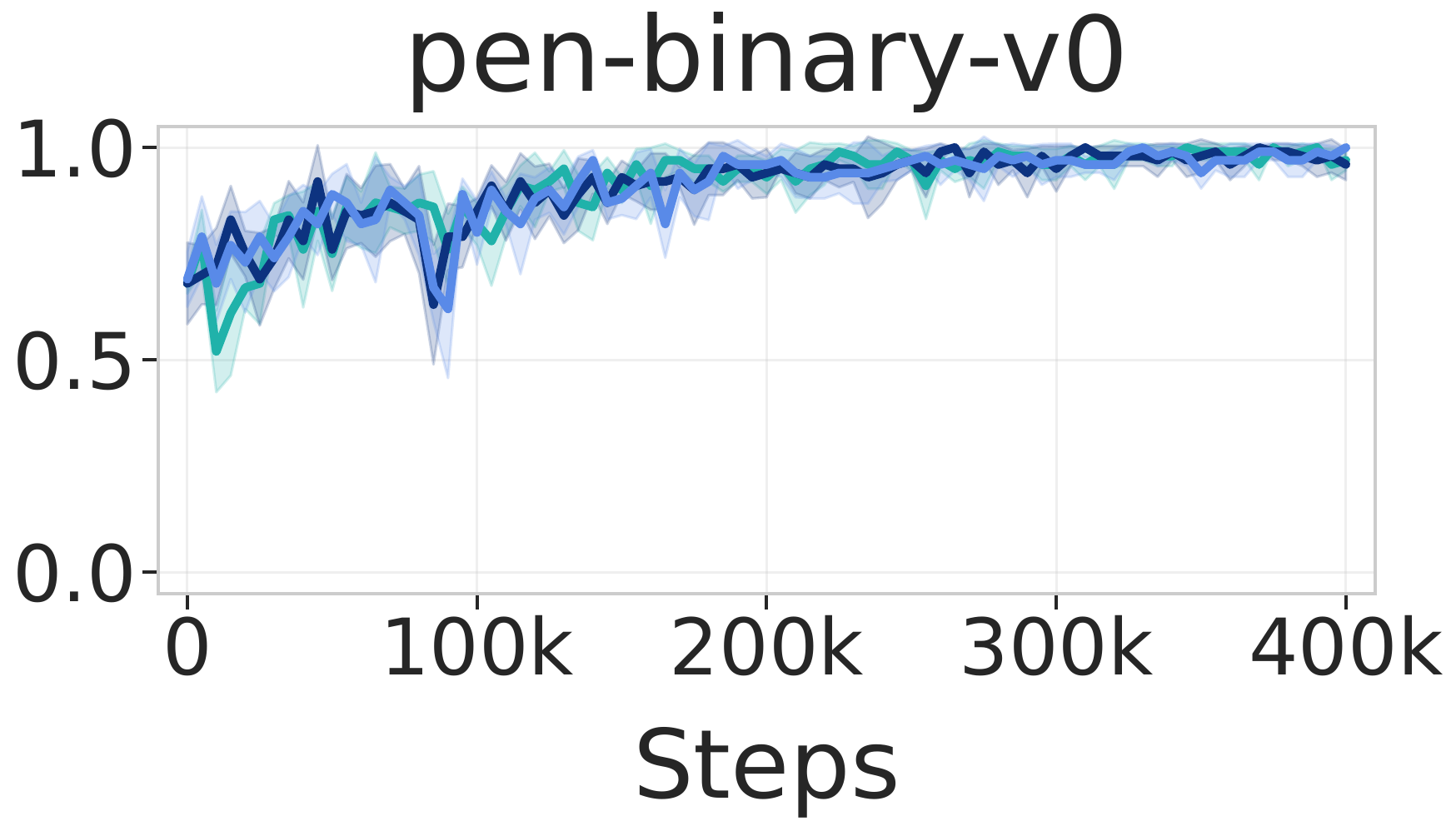}
    \end{subfigure}\\[1ex]
    \begin{subfigure}[c]{0.1\textwidth}
        \includegraphics[width=1.0\textwidth]{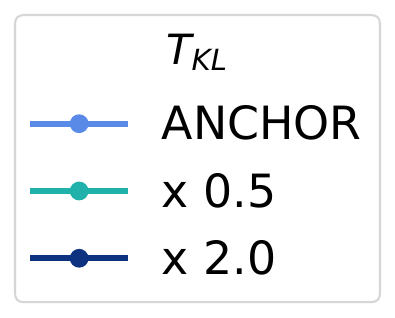}
    \end{subfigure}\hspace{0.1em}
    \begin{subfigure}[c]{0.26\textwidth}
        \includegraphics[width=1.0\linewidth]{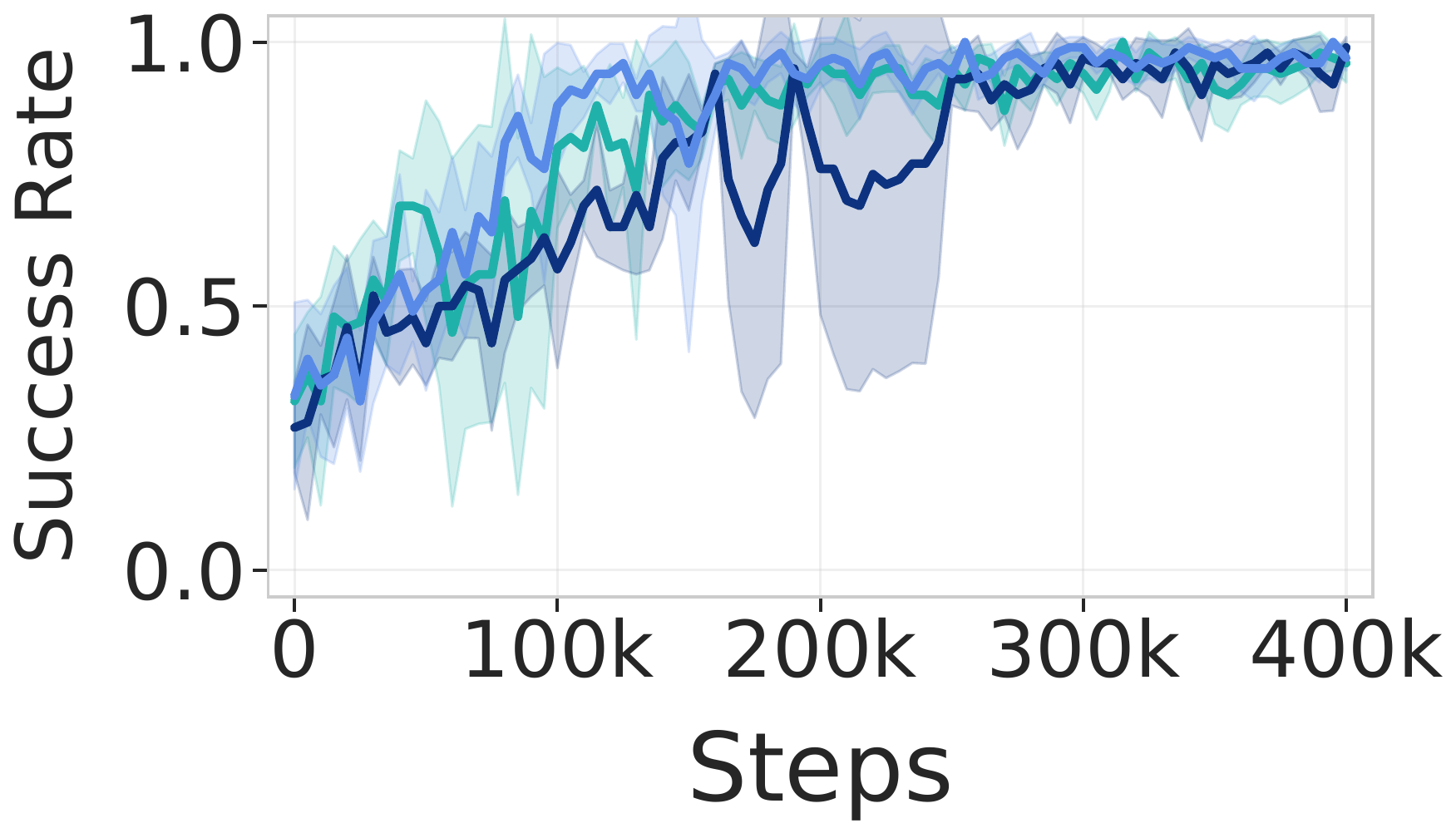}
    \end{subfigure}\hspace{0.1em}
    \begin{subfigure}[c]{0.26\textwidth}
        \includegraphics[width=1.0\linewidth]{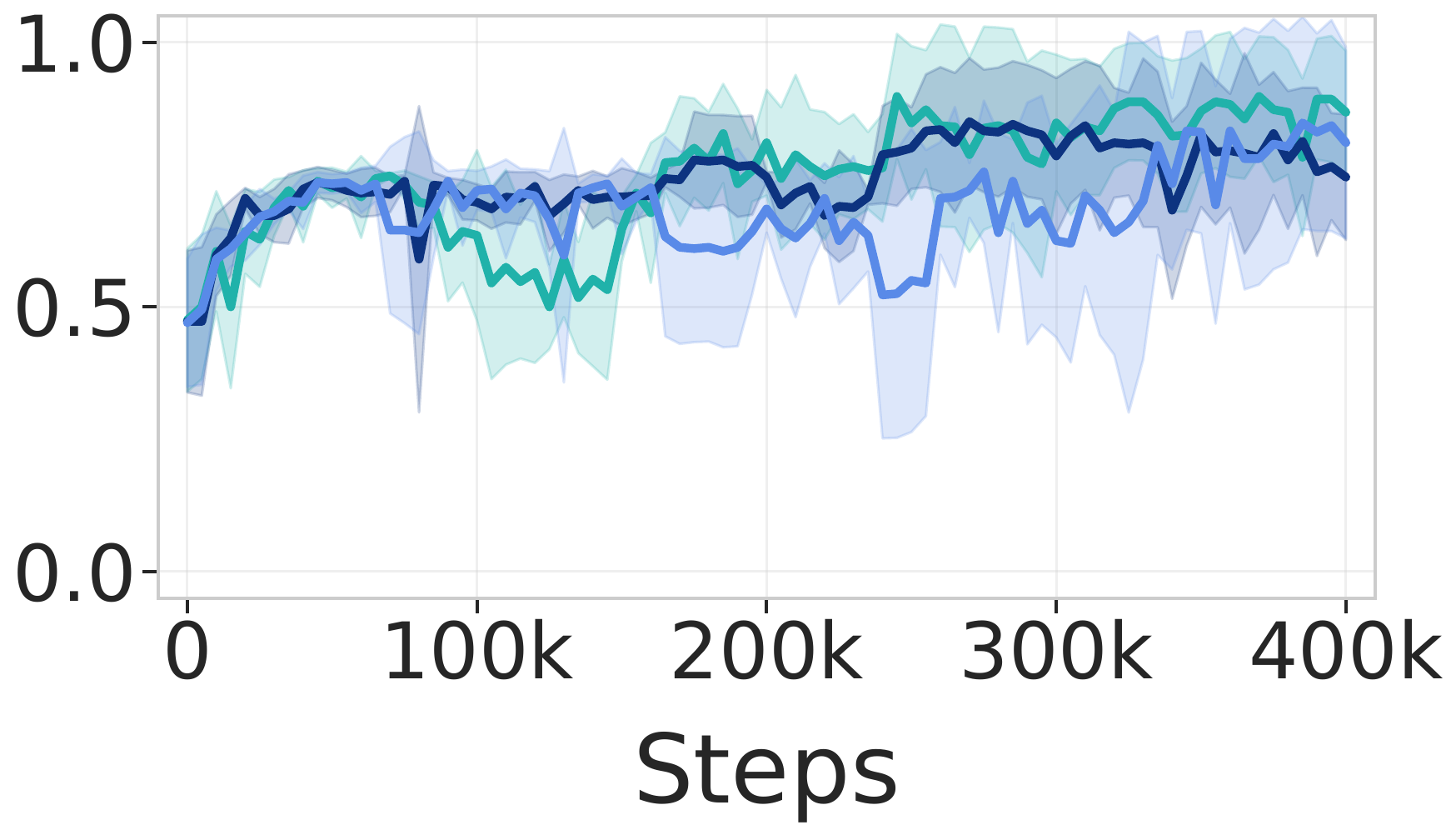}
    \end{subfigure}\hspace{0.1em}
    \begin{subfigure}[c]{0.26\textwidth}
        \includegraphics[width=1.0\linewidth]{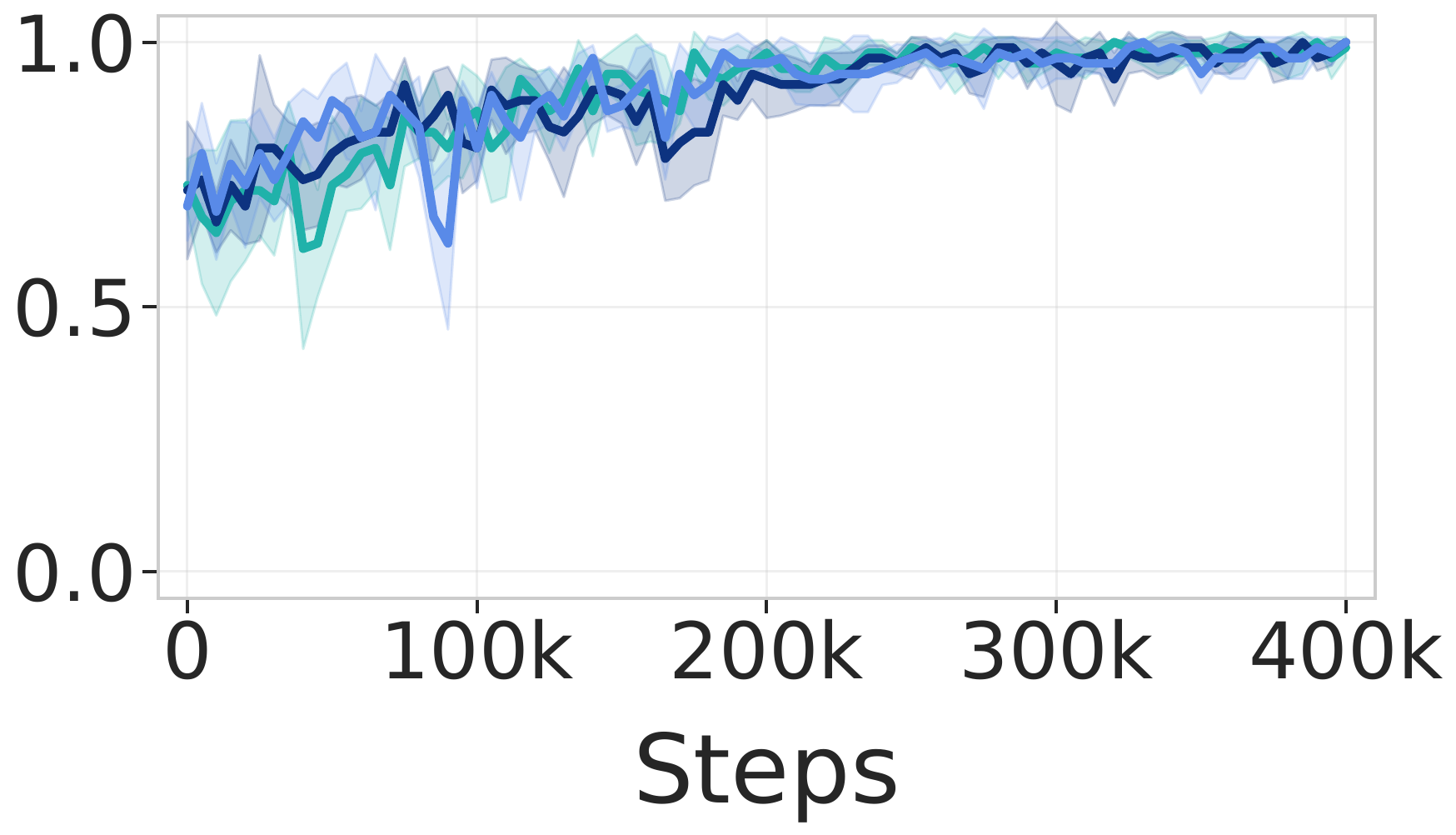}
    \end{subfigure}\\[1ex]
    \begin{subfigure}[c]{0.1\textwidth}
        \includegraphics[width=0.8\textwidth]{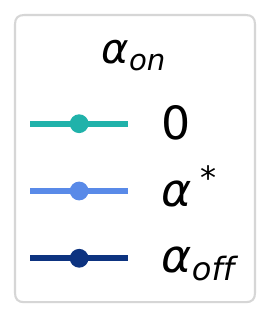}
    \end{subfigure}\hspace{0.1em}
    \begin{subfigure}[c]{0.26\textwidth}
        \includegraphics[width=1.0\linewidth]{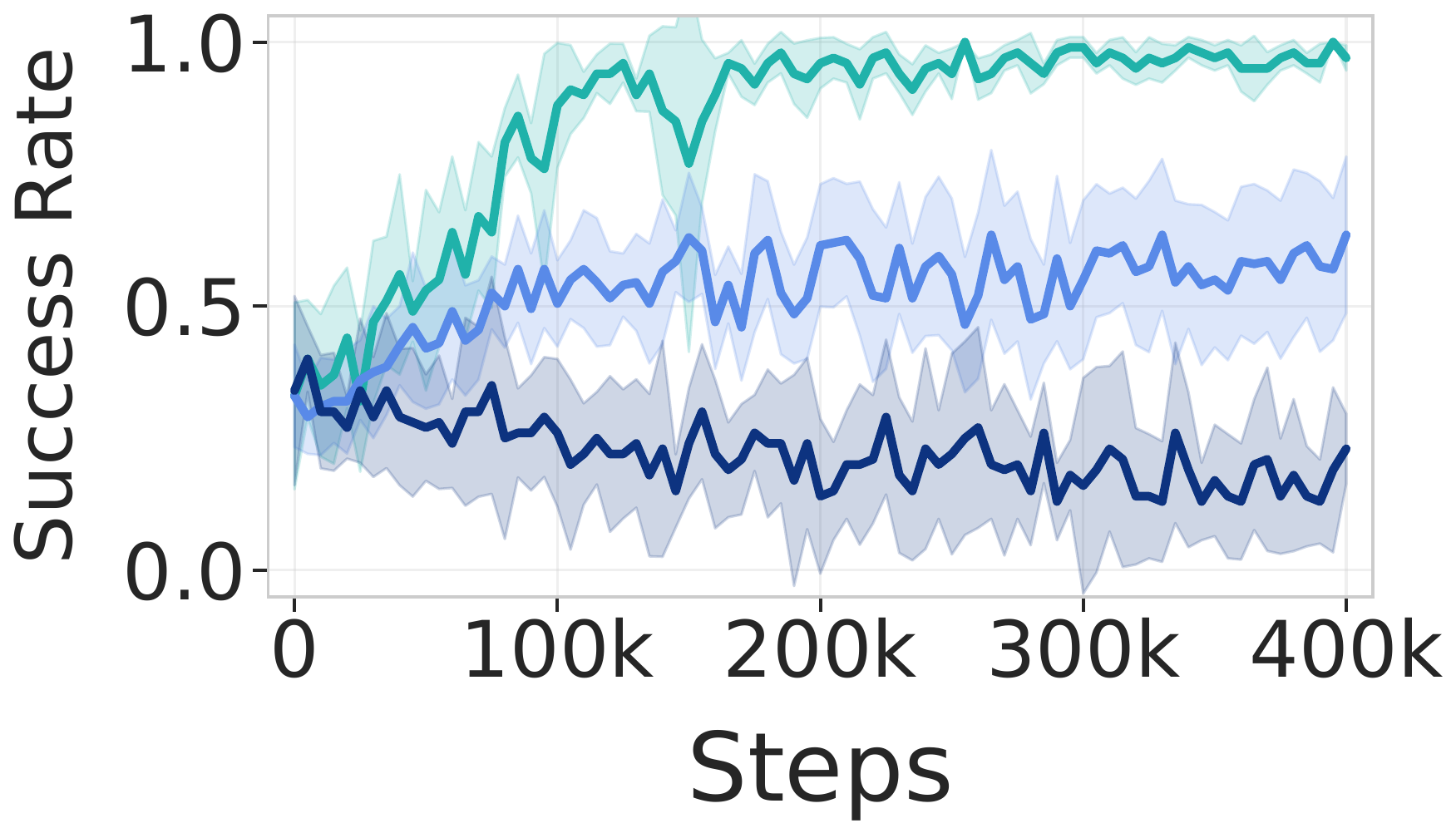}
    \end{subfigure}\hspace{0.1em}
    \begin{subfigure}[c]{0.26\textwidth}
        \includegraphics[width=1.0\linewidth]{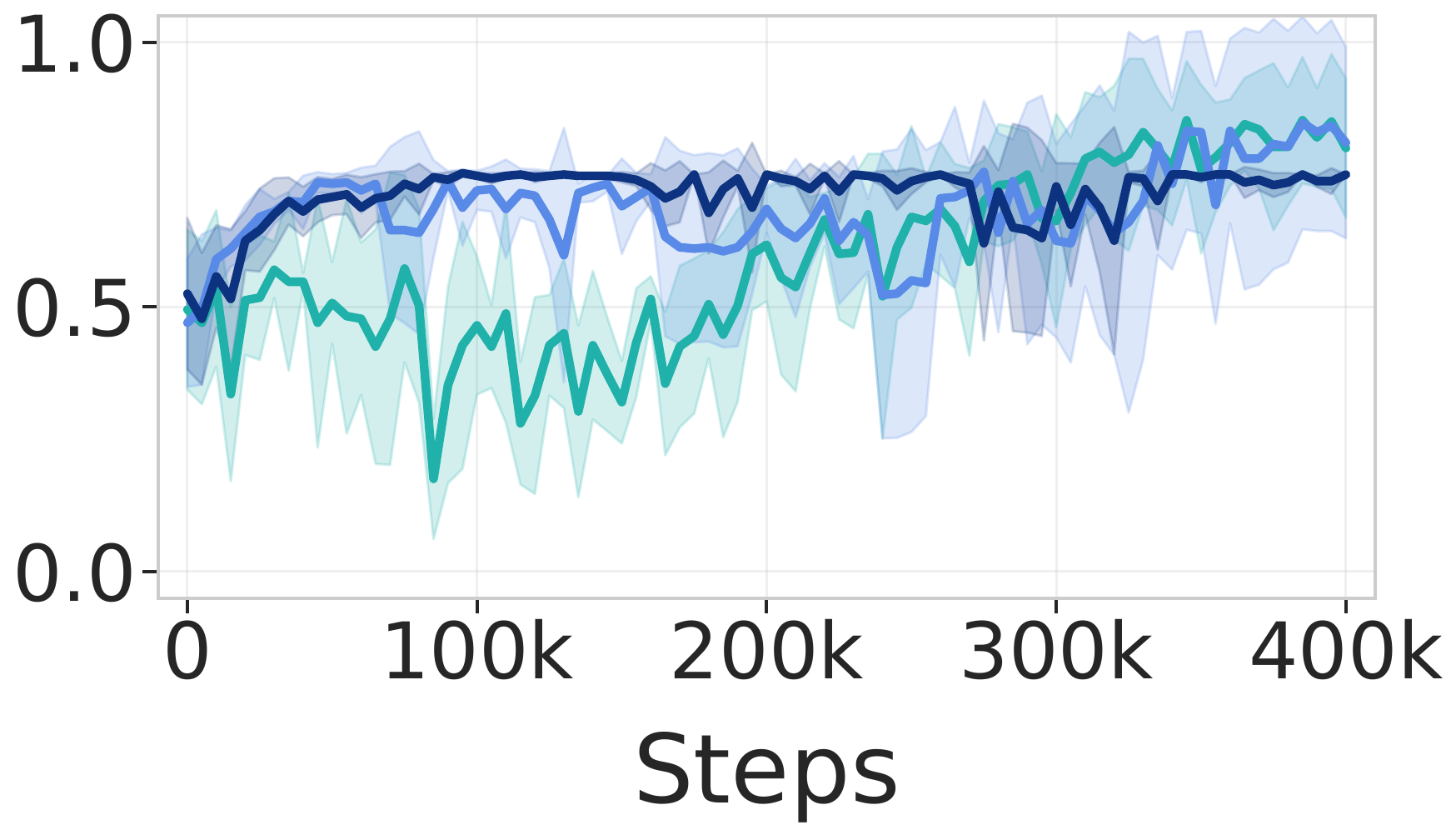}
    \end{subfigure}\hspace{0.1em}
    \begin{subfigure}[c]{0.26\textwidth}
        \includegraphics[width=1.0\linewidth]{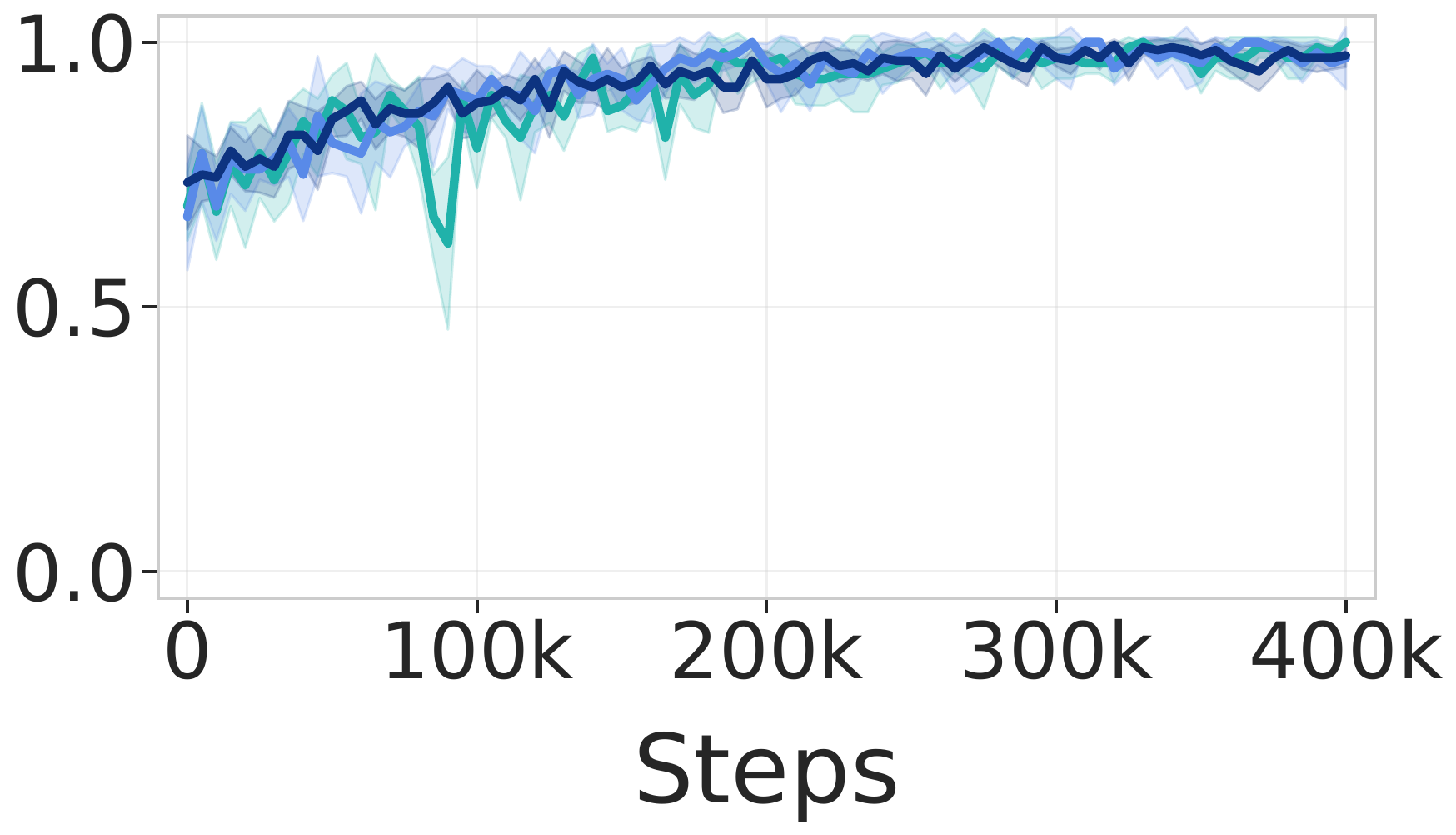}
    \end{subfigure}\\[1ex]

    \caption{\textbf{Hyperparameter sensitivity of ANCHOR.}
    We vary the initial KL weight $\beta_0$ (top), the KL annealing interval
    $T_{\mathrm{KL}}$ (middle), and the residual online conservative weight
    $\alpha_{\mathrm{on}}$ (bottom). Here, $\alpha^*$ denotes an intermediate value between
    $0$ and $\alpha_{\mathrm{off}}$, where $\alpha_{\mathrm{off}}$ is the conservative
    regularization weight used during offline pretraining. ANCHOR remains stable under
    moderate changes to KL-related hyperparameters, while $\alpha_{\mathrm{on}}$ controls the
    balance between early stability and final improvement.}
    \label{fig:anchor_ablation}
\end{figure}

Figure~\ref{fig:anchor_ablation} shows that ANCHOR is robust to moderate changes in its
KL-related hyperparameters. Although different choices of $\beta_0$ and
$T_{\mathrm{KL}}$ can affect learning stability, all tested variants continue to reduce
catastrophic failure across the three tasks and achieve similar asymptotic performance
within confidence intervals. Thus, ANCHOR's gains are not driven by a single carefully
tuned configuration, but reflect a robust effect of KL-anchored adaptation.

The residual conservative weight $\alpha_{\mathrm{on}}$ controls a different aspect of the
method. As shown in Figure~\ref{fig:anchor_ablation}, we compare three choices: removing
the conservative penalty, using an intermediate residual value between zero and the
offline-pretraining weight, and retaining the full offline-pretraining weight. Retaining the full
offline conservative weight leads to the lowest asymptotic performance in
antmaze-large-diverse and kitchen-mixed, consistent with the analysis in
Section~\ref{sec:motivation}. Removing the conservative penalty entirely achieves strong
final performance across tasks, but causes larger catastrophic failure in kitchen-mixed.
These results support the design choice of ANCHOR: reducing conservative regularization can
improve online adaptability, while retaining a small residual penalty can help stabilize
the early online phase depending on the task. Additional ablation studies on other
hyperparameters and an analysis of how the initial KL weight affects policy distribution
shift are provided in Appendix~\ref{sec:appendix:ablation_studies}.

\clearpage
\section{Additional Experiments and Protocol Details}
\label{sec:appendix:additional_evidence}
The additional experiments below use the final five-seed summaries where available, superseding earlier smaller-seed summaries. Values are mean $\pm$ SE unless explicitly labeled as bootstrap confidence intervals. The shared-duration sensitivity check remains a two-seed study. We distinguish these additional protocols from the original six-task benchmark: the available implementations do not justify treating every comparison as a change to the actor regularizer alone.

\subsection{Implementation and Interpretation of the Main Objective}
\label{sec:appendix:objective_details}
Equations~\ref{eq:anchor-actor-loss}--\ref{eq:anchor-actor-opt} present the entropy-free actor objective to expose the role of the anchor. The CQL/SAC implementation also retains its learned entropy temperature $\tau_t$, so its actor maximizes
\[
\mathbb E_{s\sim\mathcal D_{\mathrm{on}},\,a\sim\pi_\phi}
[Q_t(s,a)-\tau_t\log\pi_\phi(a|s)]
-\beta_t\mathbb E_{s\sim\mathcal D_{\mathrm{on}}}
D_{\mathrm{KL}}(\pi_\phi(\cdot|s)\|\pi_{\mathrm{off}}(\cdot|s)).
\]
The KL is computed analytically between the underlying diagonal Gaussian policies; the common invertible action squashing preserves it. The reference actor remains frozen. Schedule time $t$ counts adaptation steps after the collection-only warmup. The main critic retains a fixed residual CQL coefficient, whereas some additional diagnostic protocols below anneal it.

The lemmas in Appendix~\ref{sec:appendix:lemma} concern local occupancy control and a KL budget under the offline policy's state marginal. A soft penalty on online replay states does not guarantee that budget. Reliability under offline-policy occupancy is a separate premise; that occupancy need not equal the offline dataset distribution. The explicit constants in Equations~\ref{eq:thm1_offsupport} and~\ref{eq:thm2_main} bound additional error exposure and regret under the stated assumptions, rather than establishing a convergence guarantee or an optimal annealing schedule for the neural implementation.

\subsection{Related Policy-Side Adaptations}
\label{sec:appendix:policy_controls}
Table~\ref{tab:policy_controls} evaluates the defining mechanisms of PEX, PROTO, and Improving TD3-BC with CQL initialization. All three use online-only replay, a 5K collection-only warmup, batch size 512, UTD 1, and a 400K interaction budget. The implementation defaults disable the online CQL penalty, while the original ANCHOR configuration uses $\alpha_{\mathrm{on}}=1$ for kitchen-partial and zero for the other two tasks. Without run-level configuration records establishing an override, this difference prevents attributing every performance gap solely to fixed versus moving policy references.

\textbf{PEX adaptation.}
We freeze the offline CQL actor, initialize a separate online actor, and transfer the critic. Their action proposals are selected using a Q-based softmax with inverse temperature 10 on all three tasks. The online actor uses the PEX-SAC composite-action update. Original PEX uses offline and online replay; our adaptation changes data access and initialization~\citep{zhang2023pex}.

\textbf{PROTO adaptation.}
The current-policy versus delayed-reference log ratio enters both the actor loss and Bellman target. The reference follows a Polyak update after each actor update, with rate $5\times10^{-5}$. The initial regularization coefficient and linear annealing duration match ANCHOR's task-specific KL settings. Unlike the original mixed-replay pipeline~\citep{li2023proto}, this adaptation uses online-only replay.

\textbf{Improving TD3-BC adaptation.}
The actor uses a Q objective normalized by the minibatch mean absolute Q-value and an action-level mean squared error against actions from online replay. Its coefficient decreases geometrically from 0.4 to 0.2 over 245K updates~\citep{beeson2022improving}. We retain the CQL/SAC network and critic instead of replacing the backbone with deterministic TD3. These experiments compare complete adapted configurations and should not be interpreted as reproductions of the original methods' published results.

\begin{table}[ht]
\centering\small
\caption{\textbf{Policy-side adaptations under CQL initialization.} Five-seed mean $\pm$ SE; final success is higher-is-better and 0--100K CF is lower-is-better. ITD3-BC abbreviates the Improving TD3-BC-style adaptation.}
\label{tab:policy_controls}
\setlength{\tabcolsep}{5pt}
\begin{tabular}{llcccc}
\toprule
Task & Metric & PEX & PROTO & ITD3-BC & ANCHOR\\
\midrule
antmaze-large-play & Final & $0.91\pm.02$ & $0.97\pm.02$ & $0.52\pm.11$ & $0.96\pm.02$\\
 & CF & $0.13\pm.03$ & $0.07\pm.04$ & $0.08\pm.04$ & $0.11\pm.09$\\
kitchen-partial & Final & $0.58\pm.11$ & $0.97\pm.02$ & $0.89\pm.06$ & $1.00\pm.00$\\
 & CF & $0.71\pm.07$ & $0.47\pm.11$ & $0.69\pm.07$ & $0.06\pm.02$\\
pen-binary & Final & $0.94\pm.04$ & $0.95\pm.03$ & $0.86\pm.06$ & $1.00\pm.00$\\
 & CF & $0.49\pm.01$ & $0.46\pm.06$ & $0.15\pm.03$ & $0.14\pm.04$\\
\bottomrule
\end{tabular}
\end{table}

\subsection{Fixed and Annealed Behavior-Cloning Controls}
\label{sec:appendix:bc}
The additional BC results use weights 1 and 10. Annealed weights decrease linearly to zero over ANCHOR's task-specific interval; fixed weights remain constant. Table~\ref{tab:bc_controls} reports the final five-seed summaries. In the available implementation, the default actor penalty is
\[
\lambda_t\mathbb E_{(s,a)\sim\mathcal D_{\mathrm{off}}}[-\log\pi_\phi(a|s)],
\]
with a separate offline reference minibatch, whereas critic training uses newly collected data. An optional alternative imitates actions in the current online batch. Neither is action matching to a frozen policy checkpoint by default. The launchers also enable CQL-coefficient annealing to zero, rather than the main method's fixed residual penalty. We therefore identify these as offline-data-assisted BC controls and avoid a claim that Table~\ref{tab:bc_controls} isolates exact KL versus action-level BC under identical replay and critic conditions.

A larger annealed BC coefficient substantially improves Kitchen and pen stability, but degrades both reported metrics on antmaze-large-play. Fixed BC 10 is more stable than ANCHOR on pen-binary but reaches lower final success. The pattern illustrates a retention--adaptation tradeoff; it does not demonstrate that stronger regularization is always inferior.

\begin{table}[ht]
\centering\small
\caption{\textbf{Five-seed BC controls.} Each cell gives final success / 0--100K CF, both as mean $\pm$ SE. Data access and critic schedules are detailed in Appendix~\ref{sec:appendix:bc}.}
\label{tab:bc_controls}
\setlength{\tabcolsep}{5pt}
\begin{tabular}{lccc}
\toprule
Task & Annealed BC $1\to0$ & Annealed BC $10\to0$ & ANCHOR\\
\midrule
antmaze-large-play & $.94\pm.02\,/\,.23\pm.04$ & $.85\pm.05\,/\,.31\pm.05$ & $.96\pm.02\,/\,.11\pm.09$\\
kitchen-partial & $.99\pm.01\,/\,.69\pm.07$ & $.99\pm.01\,/\,.11\pm.06$ & $1.00\pm.00\,/\,.06\pm.02$\\
pen-binary & $.95\pm.02\,/\,.62\pm.07$ & $.98\pm.01\,/\,.15\pm.06$ & $1.00\pm.00\,/\,.14\pm.04$\\
\midrule
Task & Fixed BC 1 & Fixed BC 10 & ANCHOR\\
\midrule
kitchen-partial & $.96\pm.02\,/\,.68\pm.10$ & $.57\pm.08\,/\,.22\pm.08$ & $1.00\pm.00\,/\,.06\pm.02$\\
pen-binary & $.97\pm.02\,/\,.57\pm.03$ & $.93\pm.02\,/\,.07\pm.04$ & $1.00\pm.00\,/\,.14\pm.04$\\
\bottomrule
\end{tabular}
\end{table}

\subsection{Aggregate Statistical Reliability}
\label{sec:appendix:reliability}
Following~\citet{agarwal2021deep}, we report the IQM over task-run scores and stratified-bootstrap 95\% confidence intervals (50,000 replications), and PoI with 2,000 bootstrap replications. Final-performance statistics use the 400K endpoint of the original six-task benchmark. For CF, the values are sign-flipped when computing PoI, so larger PoI always favors ANCHOR. SAC has no reported CF comparison because it has no offline-pretrained policy to retain.

\begin{table}[ht]
\centering\small
\caption{\textbf{Aggregate IQM and 95\% bootstrap confidence intervals.} The baseline in each row is the strongest by that aggregate metric.}
\label{tab:iqm}
\begin{tabular}{lcc}
\toprule
Metric & ANCHOR & Strongest baseline\\
\midrule
Final success at 400K & 0.993 [0.981, 1.000] & WSRL: 0.944 [0.879, 0.972]\\
CF over 0--100K & 0.081 [0.048, 0.123] & IQL: 0.135 [0.104, 0.171]\\
\bottomrule
\end{tabular}
\end{table}

\begin{table}[ht]
\centering\small
\caption{\textbf{Probability of improvement of ANCHOR, with 95\% bootstrap confidence intervals.} These compare aggregate distributions across the six tasks.}
\label{tab:poi}
\begin{tabular}{lcc}
\toprule
Baseline & Final success at 400K & CF over 0--100K\\
\midrule
WSRL & 0.710 [0.607, 0.813] & 0.767 [0.653, 0.880]\\
PORL & 0.657 [0.553, 0.763] & 0.857 [0.763, 0.940]\\
Cal-QL & 0.790 [0.693, 0.880] & 0.747 [0.627, 0.853]\\
CQL & 0.853 [0.763, 0.937] & 0.757 [0.653, 0.860]\\
IQL & 1.000 [1.000, 1.000] & 0.590 [0.513, 0.667]\\
SAC & 0.983 [0.950, 1.000] & ---\\
\bottomrule
\end{tabular}
\end{table}

Every reported PoI interval has a lower bound above 0.5. This supports aggregate improvement within the evaluated task set; overlapping per-task uncertainty bands are not themselves hypothesis tests, and the aggregate analysis does not establish per-environment significance. The confidence intervals and PoI values are retained from the experimental summaries, not reconstructed from rounded task means and SEs.

\subsection{RND State-Novelty Diagnostics}
\label{sec:appendix:novelty}
The monitor follows Random Network Distillation~\citep{burda2019rnd}. States are normalized using offline dataset statistics and clipped to $[-5,5]$. Five independently initialized two-layer predictors (width 128) fit a fixed 64-dimensional random target for 10K updates with batch size 256 and learning rate $10^{-4}$. The mean feature prediction error, averaged across predictors, defines the novelty score. A snapshot of the offline-fitted predictors is frozen for evaluation. The high-novelty threshold is the 95th percentile on up to 10K offline calibration states; the reported rate is the fraction of measured online states exceeding that threshold. States are measured every 10 steps and logged in 1K-step windows. The monitor is diagnostic only and does not affect the actor or critic objective.

These runs use 100K online steps for kitchen-mixed and 60K for door-binary, batch size 512, UTD 1, and a 5K warmup. The diagnostic launcher linearly reduces the conservative weight from 5 to 0 over 160K on kitchen-mixed and from 1 to 0 over 40K on door-binary. The $\alpha$-only control uses the same schedule without actor anchoring; WSRL disables both penalties. BC uses a separate offline reference batch for its actor penalty. Thus the ANCHOR versus $\alpha$-only comparison isolates the actor anchor within this diagnostic protocol, whereas the BC comparison also differs in data access. These are distinct runs from the main 400K benchmark.

On kitchen-mixed, the frozen-checkpoint anchor accompanies lower early novelty and smaller failure. On door-binary, the weak initial policy has little performance to lose and all methods show similarly high novelty. The differing novelty and CF windows do not establish temporal precedence. These summaries support an association, not a causal claim that novelty causes collapse or that the critic is irrelevant.

\begin{table}[ht]
\centering\small
\caption{\textbf{Five-seed novelty and failure diagnostics.} Novelty is measured over 0--30K; CF over 0--100K for kitchen-mixed and 0--60K for door-binary. Novelty is a visitation statistic, not an objective to minimize.}
\label{tab:novelty}
\setlength{\tabcolsep}{5pt}
\begin{tabular}{llcccc}
\toprule
Task & Metric & WSRL & $\alpha$-only & Annealed BC & ANCHOR\\
\midrule
kitchen-mixed & Novelty & $.85\pm.01$ & $.79\pm.02$ & $.77\pm.01$ & $.44\pm.01$\\
 & CF & $.61\pm.02$ & $.57\pm.03$ & $.58\pm.03$ & $.03\pm.01$\\
door-binary & Novelty & $.98\pm.02$ & $.96\pm.03$ & $.96\pm.03$ & $.95\pm.04$\\
 & CF & $.06\pm.06$ & $.05\pm.05$ & $.08\pm.08$ & $.07\pm.07$\\
\bottomrule
\end{tabular}
\end{table}

\subsection{Implicit-Policy Extension}
\label{sec:appendix:flow}
We extend temporary anchoring to FQL and IFQL~\citep{park2025flow} on four OGBench tasks~\citep{park2025ogbench}. Their implicit policies lack tractable action densities, so we use a coupled action-distance proxy, not an exact KL between flow-policy distributions. For the same latent noise $z$, current and frozen actors produce actions $a_\phi(s,z)$ and $a_{\mathrm{off}}(s,z)$, and we penalize
\begin{equation}
\label{eq:flow_proxy}
\mathbb E_{s,z}\frac{\|a_\phi(s,z)-a_{\mathrm{off}}(s,z)\|_2^2}{2\sigma^2},\qquad \sigma=1.
\end{equation}
This is the KL between equal-variance Gaussian kernels centered on the coupled actions; it is not the original policies' KL. The shared coefficient starts at 1 and decreases linearly to zero over 400K steps. Neither backbone receives a CQL penalty or other critic modification.

The task names in Table~\ref{tab:ogbench} are \texttt{scene-play-singletask}, \texttt{puzzle-4x4-play-singletask}, \texttt{humanoidmaze-medium-navigate-singletask}, and \texttt{antsoccer-arena-navigate-singletask}. The reported extension uses the same initial anchor weight 1 and linear 400K annealing interval for every task and backbone, without task-specific anchor tuning. FQL and IFQL keep their base critic objectives.

The implementation freezes the one-step flow actor for FQL and the iterative flow actor for IFQL at the start of the online phase. It draws shared standard-normal latent noise for the current and frozen actors, clips both resulting actions to $[-1,1]$, and computes Equation~\ref{eq:flow_proxy} with smoothing scale 1. For IFQL the current-policy gradient passes through its flow integration. This is a coupled action-space penalty with a Gaussian-kernel interpretation; it does not evaluate the intractable action-density KL of either implicit policy. Accordingly, the occupancy lemmas for exact policy KL cannot be applied directly to these runs.

The supplied OGBench launchers default to 1M offline and 1M online steps, batch size 256, UTD 1, 5K online warmup, and evaluation every 10K steps over 50 episodes. The aggregate record reports five seeds and final scores but does not preserve their exact evaluation-window selection or all runtime overrides. We therefore retain them as reported final scores and do not combine them with the six-task 400K endpoint IQM. In particular, antsoccer's final improvement should not be described as a reduction in early failure, and neither FQL task shows a final-performance gain in the five-seed results.

\begin{table}[ht]
\centering\small
\caption{\textbf{OGBench extension, five seeds (mean $\pm$ SE).} ``+A'' uses the implicit-policy proxy in Equation~\ref{eq:flow_proxy}. Short task names are expanded in Appendix~\ref{sec:appendix:flow}. CF is measured over 0--100K.}
\label{tab:ogbench}
\setlength{\tabcolsep}{5pt}
\begin{tabular}{llcccc}
\toprule
Backbone & Task & Base final & +A final & Base CF & +A CF\\
\midrule
FQL & scene & $1.00\pm.00$ & $1.00\pm.00$ & $0.26\pm.06$ & $0.16\pm.07$\\
FQL & puzzle-4x4 & $0.40\pm.22$ & $0.40\pm.22$ & $0.07\pm.01$ & $0.09\pm.01$\\
IFQL & humanoidmaze-medium & $0.25\pm.07$ & $0.92\pm.01$ & $0.71\pm.02$ & $0.19\pm.08$\\
IFQL & antsoccer-arena & $0.08\pm.05$ & $0.40\pm.04$ & $0.18\pm.03$ & $0.21\pm.06$\\
\bottomrule
\end{tabular}
\end{table}

\subsection{Additional Offline-Replay Control}
\label{sec:appendix:replay_control}
The original diagnostic figure makes two one-factor comparisons. Table~\ref{tab:replay_control} adds the previously unreported replay-enabled, no-conservative-penalty condition alongside two controls. It is an additional three-condition summary, not a numerical table of all four factorial cells; the remaining original condition is shown in Figure~\ref{fig:offline_data_alpha_analysis}.

\begin{table}[ht]
\centering\small
\caption{\textbf{Replay/conservatism control, five seeds.} Cells give CF / final success (mean $\pm$ SE). The average row is the reported aggregate.}
\label{tab:replay_control}
\setlength{\tabcolsep}{5pt}
\begin{tabular}{lccc}
\toprule
Task & Replay, no $\alpha$ & No replay, no $\alpha$ & Replay, with $\alpha$\\
\midrule
kitchen-partial & $.69\pm.07\,/\,.76\pm.08$ & $.68\pm.06\,/\,.60\pm.06$ & $.58\pm.10\,/\,.62\pm.13$\\
kitchen-mixed & $.49\pm.06\,/\,.70\pm.02$ & $.49\pm.07\,/\,.41\pm.13$ & $.48\pm.07\,/\,.70\pm.05$\\
door-binary & $.15\pm.04\,/\,1.00\pm.00$ & $.25\pm.03\,/\,.79\pm.20$ & $.28\pm.03\,/\,.74\pm.19$\\
pen-binary & $.65\pm.06\,/\,.93\pm.03$ & $.61\pm.04\,/\,.89\pm.03$ & $.28\pm.09\,/\,.94\pm.03$\\
\midrule
Average & $.50\pm.06\,/\,.85\pm.03$ & $.51\pm.04\,/\,.67\pm.07$ & $.40\pm.05\,/\,.75\pm.06$\\
\bottomrule
\end{tabular}
\end{table}

Replay without conservatism does not reduce mean CF relative to removing both. With replay retained, conservatism reduces average CF but also lowers average final success. This qualifies any blanket claim that retaining replay alone prevents collapse.

\subsection{Exploratory Shared-Weight Duration Check}
\label{sec:appendix:sensitivity}
An additional two-seed check held the initial KL weight shared across tasks, kept the residual critic settings unchanged, and varied only the duration between 40K and 160K. Table~\ref{tab:shared_duration} retains the reported values. The experiment summary does not retain the numerical shared initial weight, and the available launcher grid contains multiple candidates; we do not infer a value from the outcomes. This table is exploratory evidence about duration, not a fully specified tuning-free configuration or a statistical-significance claim.

\begin{table}[ht]
\centering\small
\caption{\textbf{Two-seed shared-weight duration check.} Cells give final success / 0--100K CF, mean $\pm$ SE. The initial-weight value is unresolved in the retained experiment record.}
\label{tab:shared_duration}
\begin{tabular}{lcc}
\toprule
Task & $T_{\mathrm{KL}}=40\mathrm K$ & $T_{\mathrm{KL}}=160\mathrm K$\\
\midrule
antmaze-large-play & $.90\pm.05\,/\,.05\pm.00$ & $1.00\pm.00\,/\,.12\pm.07$\\
kitchen-partial & $1.00\pm.00\,/\,.03\pm.01$ & $.99\pm.01\,/\,.04\pm.03$\\
pen-binary & $1.00\pm.00\,/\,.35\pm.10$ & $1.00\pm.00\,/\,.18\pm.12$\\
\bottomrule
\end{tabular}
\end{table}

\end{document}